%% file: main.tex
\documentclass{article}

\usepackage{microtype}
\usepackage{graphicx}
\usepackage{subcaption}
\usepackage{booktabs}
\usepackage{hyperref}
\usepackage[utf8]{inputenc}
\usepackage{csquotes}
\usepackage[accepted]{icml2026}
\usepackage{xcolor}

\definecolor{myblue}{RGB}{0,0,255}
\colorlet{myred}{red}
\input{macros}

\input{proj-macros}
\input{acronyms}

\graphicspath{{figs/}}
 
\icmltitlerunning{\pname: Optimal Transport-Guided Adversarial Attacks on Graph Neural Network-Based Bot Detection}

\begin{document}

\twocolumn[
  \icmltitle{\pname: Optimal Transport-Guided Adversarial Attacks on Graph Neural Network-Based Bot Detection}

  \icmlsetsymbol{equal}{*}

  \begin{icmlauthorlist}
    \icmlauthor{Kunal Mukherjee}{vt}
    \icmlauthor{Zulfikar Alom}{tol}
    \icmlauthor{Tran Gia Bao Ngo}{ma}
    \icmlauthor{Cuneyt Gurcan Akcora}{uf}
    \icmlauthor{Murat Kantarcioglu}{vt}
  \end{icmlauthorlist}

    \icmlaffiliation{vt}{Department of Computer Science, Virginia Tech, Virginia, USA}
    \icmlaffiliation{ma}{Department of Computer Science, Manitoba, Canada}
   \icmlaffiliation{tol}{Department of Computer Science, University of Toledo, Ohio, USA}
   \icmlaffiliation{uf}{AI Initiative, University of Central Florida, Florida, USA}

  \icmlcorrespondingauthor{Kunal Mukherjee}{mkunal@vt.edu}
  \icmlcorrespondingauthor{Murat Kantarcioglu}{muratk@vt.edu}

  \icmlkeywords{Social Bot Detection, Adversarial Attacks, Graph Neural Networks, Optimal Transport, Robustness}

  \vskip 0.3in
]

% this must go after the closing bracket ] following \twocolumn[ ...

% This command actually creates the footnote in the first column listing the
% affiliations and the copyright notice. The command takes one argument, which
% is text to display at the start of the footnote. The \icmlEqualContribution
% command is standard text for equal contribution. Remove it (just {}) if you
% do not need this facility.

% Use ONE of the following lines. DO NOT remove the command.
% If you have no special notice, KEEP empty braces:
\printAffiliationsAndNotice{}  % no special notice (required even if empty)
% Or, if applicable, use the standard equal contribution text:
% \printAffiliationsAndNotice{\icmlEqualContribution}

\input{sections/abstract}
\input{sections/intro}
\input{sections/related}
\input{sections/preliminaries}
\input{sections/Problem_Formulation}
\input{sections/method}

\input{sections/eval}
\input{sections/conc}
\input{sections/icml}

\bibliographystyle{icml2026}
\bibliography{refs/prov}

%%%%%%%%%%%%%%%%%%%%%%%%%%%%%%%%%%%%%%%%%%%%%%%%%%%%%%%%%%%%%%%%%%%%%%%%%%%%%%%
%%%%%%%%%%%%%%%%%%%%%%%%%%%%%%%%%%%%%%%%%%%%%%%%%%%%%%%%%%%%%%%%%%%%%%%%%%%%%%%
% APPENDIX
%%%%%%%%%%%%%%%%%%%%%%%%%%%%%%%%%%%%%%%%%%%%%%%%%%%%%%%%%%%%%%%%%%%%%%%%%%%%%%%
%%%%%%%%%%%%%%%%%%%%%%%%%%%%%%%%%%%%%%%%%%%%%%%%%%%%%%%%%%%%%%%%%%%%%%%%%%%%%%%
\newpage
\appendix
\onecolumn
\input{sections/appendix.tex}
%%%%%%%%%%%%%%%%%%%%%%%%%%%%%%%%%%%%%%%%%%%%%%%%%%%%%%%%%%%%%%%%%%%%%%%%%%%%%%%
%%%%%%%%%%%%%%%%%%%%%%%%%%%%%%%%%%%%%%%%%%%%%%%%%%%%%%%%%%%%%%%%%%%%%%%%%%%%%%%

\end{document}

%% file: macros.tex
\usepackage{caption}

\usepackage[framemethod=tikz]{mdframed}
\usepackage[most]{tcolorbox}
\usepackage[normalem]{ulem}
\usepackage{amsfonts}
\usepackage{amsmath}
\usepackage{amssymb}
\usepackage{bbm}
\usepackage{arydshln}
\usepackage{balance}
\usepackage{blindtext}
\usepackage{cancel}
\usepackage{caption}
\usepackage{color,soul}
\usepackage{comment}
\usepackage{enumitem}
\usepackage{fontawesome}

\usepackage{longtable}
\usepackage{array}
\usepackage{pdflscape}  % or lscape
\usepackage{makecell}

\usepackage{listings}
\usepackage{mathtools}
\usepackage{multirow}
\usepackage{pifont}
\usepackage{tabularx}
\usepackage{threeparttable}
\usepackage{tikz}
\usepackage{xcolor,colortbl}
\usepackage{xspace}

\usepackage{makecell}
\usepackage{svg}

\usepackage{etoolbox}
\usepackage{varwidth}

\usetikzlibrary{arrows.meta,positioning,fit,calc,shapes.misc}
\usepackage{rotating} 

\definecolor{winered}{rgb}{0.7,0,0}
\definecolor{gray}{gray}{0.7}
\definecolor{darkpastelgreen}{rgb}{0.01, 0.75, 0.24}
\definecolor{cadmiumgreen}{rgb}{0.0, 0.42, 0.24}
\definecolor{brickred}{rgb}{0.8, 0.25, 0.33}
\definecolor{cornellred}{rgb}{0.7, 0.11, 0.11}
\definecolor{burgundy}{rgb}{0.5, 0.0, 0.13}
\definecolor{frenchblue}{rgb}{0.0, 0.45, 0.73}
\definecolor{light-gray}{gray}{0.92}
\definecolor{lightlight-gray}{gray}{0.97}
\definecolor{codegray}{gray}{0.90}
\definecolor{inputgray}{gray}{0.90}
\definecolor{darkgreen}{RGB}{40,125,40}

\newcommand{\cmt}[1]{}

\newcommand{\homedir}{\raise.17ex\hbox{$\scriptstyle\sim$}}

\global\mdfdefinestyle{rtboxstyle}{%
linecolor=black,%
leftmargin=0cm,rightmargin=0cm,linewidth=0.5pt,
roundcorner=3,
skipbelow=0pt,backgroundcolor=lightlight-gray
}

\newcommand{\code}[1]{\texttt{#1}}

\mdfdefinestyle{MyFrame}{%
     roundcorner=1pt,
     innertopmargin=6pt,
     innerbottommargin=6pt,
     innerrightmargin=6pt,
     innerleftmargin=6pt,
    }

\graphicspath{{figure/}{figures/}}
\DeclareGraphicsExtensions{.pdf,.eps,.png,.jpg,.jpeg}

\long\def\comment#1{}

\renewcommand{\paragraph}[1]{\smallskip\noindent\emph{#1}\quad}
\def\Snospace~{\S{}}

\newcommand{\heading}[1]{{\vspace{2pt}\noindent\bf{#1}}} % inside section
\newcommand{\headinggi}[1]{{\em{#1}}} % at beginning of section -- no extra
\newcommand{\ballnumber}[1]{
    \tikz[baseline=(myanchor.base)]
    \node[circle,fill=black,inner sep=1.5pt] (myanchor)
    {\color{-.}\bfseries\footnotesize #1};}

\newcommand{\eg}{{\it e.g.,~}}
\newcommand{\ie}{{\it i.e.,~}}

\newcommand{\optional}[1]{}

\definecolor{paneTitle}{HTML}{0B3C5D}
\definecolor{paneBack}{HTML}{F7F9FC}
\definecolor{logBack}{HTML}{F3F7FF}
\definecolor{edgeBack}{HTML}{FFF7F0}
\definecolor{okGreen}{HTML}{2E7D32}
\definecolor{warnAmber}{HTML}{EF6C00}
\definecolor{badRed}{HTML}{C62828}

\tcbset{
  pane/.style={
    enhanced,
    colback=paneBack,
    colframe=paneTitle,
    coltitle=white,
    fonttitle=\bfseries\small,
    boxrule=0.6pt,
    arc=2pt,
    left=4pt,right=4pt,top=4pt,bottom=4pt,
    title filled,
  },
  logpane/.style={
    pane,
    colback=logBack,
    fonttitle=\bfseries\small\ttfamily,
  },
  edgepane/.style={
    pane,
    colback=edgeBack,
    fonttitle=\bfseries\small\ttfamily,
  }
}

%% file: proj-macros.tex
\def\pname{{\textsc{BoCloak}}\xspace}
\usepackage{fp}

\newif\ifbiblatex
\biblatexfalse

\newif\ifcomments
\commentstrue
\ifcomments
\usepackage{todonotes}
	\newcommand{\murat}[1]{\textcolor{blue}{[MK: #1]}}
	\newcommand{\kunal}[1]{\textcolor{orange}{[{\bf KUNAL:} #1]}}
    \newcommand{\zulfikar}[1]{{{\textcolor{blue}{\textbf{Zulfikar: }}}{\textcolor{blue}{#1}}}}
	\newcommand{\fix}[1]{\textcolor{red}{\hl{#1}}}
	\newcommand{\corr}[2]{\sout{#1} \hl{#2}}
\else
	\newcommand{\murat}[1]{}
	\newcommand{\wc}[1]{}
	\newcommand{\kunal}[1]{}
    \newcommand{\zulfikar}[1]{}
	\newcommand{\fix}[1]{}
	\newcommand{\corr}[2]{}
\fi

\RequirePackage[normalem]{ulem} %DIF PREAMBLE
\RequirePackage{color}\definecolor{RED}{rgb}{1,0,0}\definecolor{BLUE}{rgb}{0,0,1} %DIF PREAMBLE
\providecommand{\DIFdel}[1]{{\protect\color{red}\sout{#1}}}                      %DIF PREAMBLE
\providecommand{\DIFdel}[1]{}                      %DIF PREAMBLE
\newcommand{\bluehl}[1]{%
  \colorbox{blue!15}{\textcolor{blue}{\textbf{#1}}}%
}

%% file: acronyms.tex
\usepackage[acronym]{glossaries}

\newacronym{hdl}{HDL}{High-level Dynamic Language}

\newacronym{ml}{ML}{Machine Learning}

\newacronym{ai}{AI}{Artificial Intelligence}

\newacronym{nn}{NN}{Neural Network}

\newacronym{gnn}{GNN}{Graph Neural Network}
\newcommand{\gnn}{\gls*{gnn}\xspace}

\newacronym{dnn}{DNN}{Deep Neural Network}

\newacronym{rnn}{RNN}{Recurrent Neural Network}

\newacronym{llm}{LLM}{Large Language Model}

\newacronym{rag}{RAG}{Retrieval-Augmented Generation}

\newacronym{mlm}{MLM}{Masked Language Model}

\newacronym{gan}{GAN}{Generative Adversarial Network}

\newacronym{nlp}{NLP}{Natural Language Processing}

\newacronym{pl}{PL}{Programming Language}

\newacronym{vm}{VM}{Virtual Machine}

\newacronym{bert}{BERT}{Bidirectional Encoder Representations from Transformers}

\newacronym{csn}{CSN}{Code Search Net}

\newacronym{sota}{SOTA}{State-Of-The-Art}
\newcommand{\sota}{\gls*{sota}\xspace}

\newacronym{pypi}{PyPI}{Python Package Index}

\newacronym{cdm}{CDM}{Common Data Model}

\newacronym{tc}{TC}{Transparent Computing}

\newacronym{dt}{DT}{Decision Tree}

\newacronym{apt}{APT}{Advanced Persistent Threat}

\newacronym{cve}{CVE}{Common Vulnerabilities and Exposures}

\newacronym{ids}{IDS}{Intrusion Detection System}

\newacronym{pids}{PIDS}{\emph{Provenance-based Intrusion Detection System}}

\newacronym{av}{AV}{Anti-Virus}

\newacronym{edr}{EDR}{Endpoint Detection and Response}

\newacronym{poi}{POI}{Point-Of-Interest}

\newacronym{vlan}{VLAN}{Virtual Local Area Network}

\newacronym{ctwo}{C2}{Command and Control}

\newacronym{csg}{CSG}{Computer Security Group}

\newacronym{wicys}{WiCyS}{Women In Cyber Security}

\newacronym{nsa}{NSA}{National Security Agency}

\newacronym{dod}{DoD}{Department of Defense}

\newacronym{lsm}{LSM}{Log Structured Merge}

\newacronym{me}{ME}{Microelectronics}

\newacronym{ci}{CI}{CyberInfrastructure}

\newacronym{soc}{SoC}{System-on-Chip}

\newacronym{tle}{TLE}{Two-line element set}

\newacronym{wal}{WAL}{Write Ahead Log}

\newacronym{obc}{OBC}{On-Board Computer}

\newacronym{dos}{DoS}{Denial-of-Services}

\newacronym{ttp}{TTP}{Tactics, techniques, and Procedures}

\newacronym{cot}{CoT}{chain-of-
thought}

\newacronym{cti}{CTI}{Cyber Threat Intelligence}

\newacronym{ioc}{IOC}{Indicators Of Compromise}

\newacronym{jspoc}{JSpOC}{The Joint Space Operations Center}

\newacronym{cfg}{CFG}{Control Flow Graph}

\newacronym{cdg}{CDG}{Control Dependency Graph}

\newacronym{dgl}{DGL}{Deep Graph Library}

%% file: sections/abstract.tex
\begin{abstract}
 
%The rise of human-mimicking bot accounts on social media poses serious risks to public discourse.  
%To counter this, bot detectors increasingly rely on Graph Neural Networks (GNNs), which may perform poorly in adversarial settings where attackers evolve, and attacks must satisfy domain-specific and temporal constraints.  
%Existing graph attacks either ignore such constraints or rely on gradient access and global perturbations that are implausible in real social systems.
%MK: I tried to rewrote the paragraph. Please carefully check. KM: done
The rise of bot accounts on social media poses significant risks to public discourse. To address this threat, modern bot detectors increasingly rely on Graph Neural Networks (GNNs). However, the effectiveness of these GNN-based detectors in real-world settings remains poorly understood. In practice, attackers continuously adapt their strategies as well as must operate under domain-specific and temporal constraints, which can fundamentally limit the applicability of existing attack methods. %Many prior graph attack methods  ignore such constraints. 
As a result, there is a critical need for robust GNN-based bot detection methods under realistic, constraint-aware attack scenarios.
%MK: i do not think we want to undermine. We should say something like evaluate the robustness of  et. Please check my change. KM: done
We introduce \pname~\footnote{\url{https://github.com/kunmukh/bocloak}} to systematically evaluate the robustness of GNN-based bot detection via both edge editing and node injection adversarial attacks under realistic constraints. \pname constructs a probability measure over spatio-temporal neighbor features and learns an optimal transport (OT) geometry that separates human and bot behaviors. It then decodes transport plans into sparse, plausible edge edits that evade detection while obeying real-world constraints. We evaluate \pname across three social bot datasets, five state-of-the-art bot detectors, three adversarial defenses, and compare it against four leading graph adversarial attack baselines. \pname achieves up to 80.13\% higher attack success rates while using 99.80\% less GPU memory under real-world constraints. \pname shows that OT provides a lightweight, principled framework for bridging adversarial attacks and real-world bot detection.

\end{abstract}

%% file: sections/intro.tex
\section{Introduction}
Bot accounts that mimic human behavior pose a persistent threat on social platforms. They distort engagement, amplify misinformation, and manipulate public discourse at scale~\cite{liu2025evolution}. To mitigate this, GNN-based detectors are widely deployed, combining user profiles, content, and social connectivity~\citep{varol2017online,kudugunta2018deep,feng2021botrgcn,feng2021rgt}.
%ferrara2016rise,yang2024sebot,he2024botdgt

Despite strong performance in controlled settings, the robustness of these detectors under realistic adversarial conditions remains poorly understood. Existing graph attacks~\citep{zugner2018nettack,geisler2021_robustness_of_gnns_at_scale,alom2025gottack} typically assume an attacker can perturb node features or flip edges arbitrarily, subject only to a fixed budget. These assumptions collapse in real deployments. Bot cannot rewire arbitrary users or gain instant access to high-reputation connections. Instead, they must create new accounts, learn from the demise of previously flagged bots, and establish plausible follow patterns in order to evade detection.

This setting introduces core constraints. First, the social graph is large, and probing all possible nodes is infeasible due to computational costs, API limits, and detection risk. Second, the number of feasible edge configurations grows combinatorially with the size of the graph. Third, while bots must connect to amplify content, overt coordination exposes them to cluster-based detection. Once one bot account is flagged, its connections can trigger cascading bans. Effective attacks must therefore operate under partial observability, behavioral plausibility, and limited connectivity.  

\begin{figure}[!t]
\centering
\resizebox{0.9\columnwidth}{!}{%
    \includegraphics{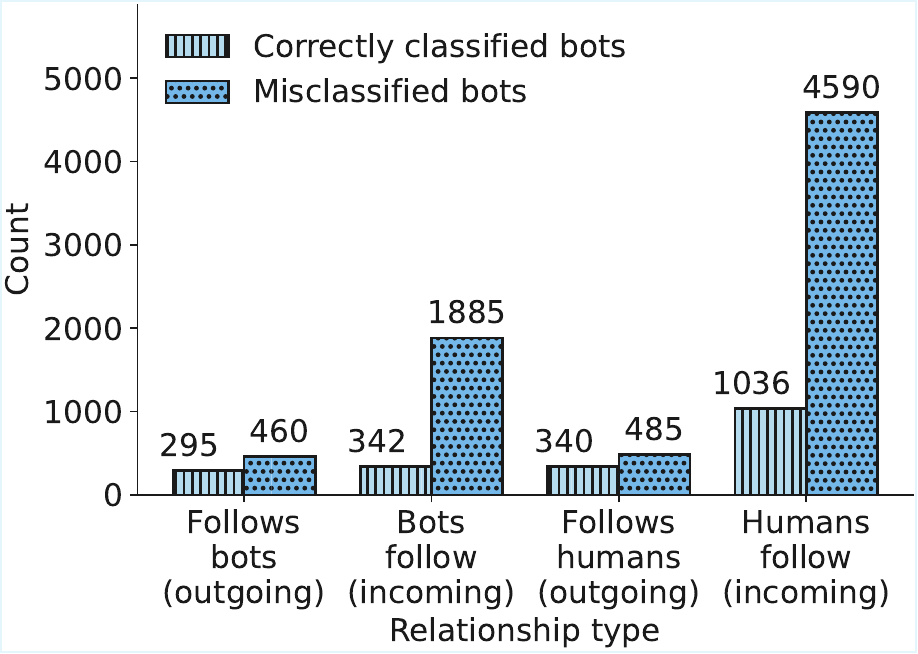}%
}
\caption{Relationship counts for classified bots in TwiBot-22 by BotRGCN. Counts are computed from the perspective of bot accounts. Each category indicates whether a bot has at least one outgoing or incoming relationship of the specified type. Categories are not mutually exclusive: a single bot may contribute to multiple bars. Misclassified bots are more likely to have edges incident to humans, especially incoming follows.}
\label{fig:rel-count}
\end{figure}

Prior work is fragmented: feature-space attacks~\citep{cresci2019better,cresci2021coming} ignore network structure, recent node-editing methods introduce helper bots for evasion~\citep{wang2023mybrother,wang2025nodeinjection,liu2024social}  but do not characterize vulnerable local neighborhoods or how to construct them under temporal and domain constraints.  As a result, \sota detection methods are fragile; our results in ~\autoref{fig:rel-count} show that even a single human follower can mislead a \sota bot detector in 81.50\% of the cases. 
%MK: not sure this next sentence flows well. Please check my improvement. KM: done
This striking fragility highlights a gap in existing defenses: despite these vulnerabilities, prior work has not connected bot evasion to a structured geometric attack formulation.

We argue that optimal transport (OT) ~\cite{petric2019got}, with its roots in measure theory and convex optimization, provides a principled and efficient way to reason in such constrained settings. Our proposed method, \pname, uses OT to compare probability distributions over spatio-temporal features of nodes and their neighbors, and finds the minimal cost of transforming one local neighborhood distribution (e.g., from a bot) into another (e.g., from a human account). OT plans are low-cost, interpretable transformations (example in App.~\ref{sec:case-study}) that can be computed efficiently with partial network access or model gradients. These properties make OT ideal for generating sparse, plausible edge edits for attacking under realistic constraints.

We evaluate \pname across three social bot datasets, five bot detectors, and three defense strategies, comparing it against four \sota graph adversarial attack baselines. \pname consistently achieves 80.13\% higher attack success rates while requiring negligible resource overhead and maintaining social plausibility. Notably, it performs equally well in unconstrained settings with no domain limitations. Our results push OT-based adversarial attacks to the forefront for efficient, model-agnostic bot detection.  

\textbf{Our contributions are as follows:}
\begin{itemize}[leftmargin=*, nosep]
    \item We present the first application of optimal transport theory in social bot detection and attacks on GNNs. 
    \item We develop a learnable OT geometry that aligns spatio-temporal graph transport costs with adversarial objectives and realistic domain constraints.
     \item We demonstrate that \pname evades \sota bot detectors under both constrained and unconstrained attacks, achieving 80.13\% higher attack success. \pname{} bypasses five prominent defenses and sustains high evasion rates under three adversarial hardening techniques.
     %\pname bypasses five recent and prominent defenses and maintains high evasion rates even against three adversarially hardened models.
    \item We demonstrate that \pname has significantly lower computational overhead than SOTA attacks. Across three datasets, it achieves up to 99.80\% lower GPU memory usage than PR-BCD and FGA, and runs up to $20\times$ faster than \sota adversarial attacks (\eg Nettack and GOttack), without gradient access or model-specific tuning.  
\end{itemize}

%% file: sections/related.tex
\section{Related Works}

\heading{Social Bot Detection.}
Early work on social bot detection relied on supervised classifiers~\cite{ferrara2016rise,varol2017online,kudugunta2018deep}. As bots evolved from simple spammers to coordinated, human-like accounts, research shifted toward graph-based detectors that more explicitly use social structure. The release of social graph datasets for bot detection, such as Cresci-2015, TwiBot-20, and TwiBot-22~\citep{cresci2015fakers, feng2021twibot20, feng2022twibot22} standardized evaluation across user attributes, node, and edge types. To address the growing gap in realism between benchmarks and emerging adversaries, the BotSim-24~\cite{qiao2025botsim} dataset demonstrates that LLM-driven bots substantially degrade the performance of existing detectors. These datasets catalyzed the development of \gnn-based bot detectors such as  BotRGCN~\cite{feng2021botrgcn}, S-HGN~\cite{lv2021we}, and RGT~\cite{feng2021rgt}. 
% Recent advancement, such as SEBot~\cite{yang2024sebot}, uses structural entropy and multi-view contrastive learning to capture hierarchical patterns in multi-relational graphs, while BotDGT~\cite{he2024botdgt} models the dynamic evolution of follow edges via dynamic graph transformers. 

We treat these detectors as victim models to be attacked and study how an adaptive adversary can inject new bots with carefully engineered neighborhoods that remain undetected.

% \heading{GNN-based Adversarial Attacks.}
% Adversarial vulnerabilities of GNNs have been widely studied in graph benchmarks. Nettack~\cite{zugner2018nettack} introduced targeted poisoning and evasion attacks by greedily perturbing features and edges around a victim node, while Fast Gradient Attack (FGA)~\cite{chen2018fga} used gradient saliency to select high-impact edge rewiring operations. Subsequent work GOttack~\cite{alom2025gottack} learned universal perturbation patterns by exploiting the orbit structure of nodes, achieved strong cross-architecture transfer, and PR-BCD~\cite{geisler2021_robustness_of_gnns_at_scale} formulated topology attacks as continuous optimization over an adjacency-perturbation matrix and solved it via projected randomized block coordinate descent, scaling to large graphs under a global edge budget.
% However, these attacks typically perturb existing nodes and edges rather than inject new ones and largely ignore temporal and social plausibility constraints (e.g., a bot cannot arbitrarily induce many human users to follow it). 
\heading{GNN-based Adversarial Attacks.}
Adversarial vulnerabilities of GNNs have been widely studied in graph benchmarks. Nettack~\cite{zugner2018nettack} introduced targeted poisoning and evasion attacks by greedily perturbing features and edges around a victim node, while Fast Gradient Attack (FGA)~\cite{chen2018fga} used gradient saliency to select high-impact edge rewiring operations. Subsequent work GOttack~\cite{alom2025gottack} learned universal perturbation patterns by exploiting the orbit structure of nodes, achieved strong cross-architecture transfer, and PR-BCD~\cite{geisler2021_robustness_of_gnns_at_scale} formulated topology attacks as continuous optimization over an adjacency-perturbation matrix and solved it via projected randomized block coordinate descent, scaling to large graphs under a global edge budget. More recent graph attacks include LR-BCD, which adds local node-wise constraints to randomized block-coordinate attacks~\citep{gosch2023adversarial}, partial graph attack (PGA), which allocates a global attack budget to selected vulnerable nodes~\citep{zhu2024partial}, and EvA, which directly optimizes discrete graph perturbations with an evolutionary black-box procedure~\citep{akhondzadeh2026eva}.

% Recent works~\cite{wang2023mybrother, wang2025nodeinjection} showed how a node-injection attack that adds a helper bot and its connections around a target bot to jointly evade GNN-based detectors, and build a combined attack-and-defense framework. Liu \etal~\cite{liu2024social} studied robustness and simple defenses for Twitter bot detectors.

% \pname differs from these works along two axes: (i) we formulate bot evasion directly in the space of domain and time-aware neighborhood distributions and learn an OT ground cost that shapes an explicit human-bot geometry; and (ii) our attack is optimized to mimic misclassified bots while respecting strict node-injection, degree, and temporal constraints; providing the first OT-based evaluation of node-injection attacks on social bot detectors.

\pname differs from these works along two axes: (i) we formulate bot evasion directly in the space of domain and time-aware neighborhood distributions and learn an OT ground cost that shapes an explicit human-bot geometry; and (ii) our attack is optimized to mimic misclassified bots while respecting strict node-injection, degree, and temporal constraints. LR-BCD and EvA are not designed around social-network feasibility constraints such as no forced human follow-backs, while PGA is a global budget-allocation attack rather than a per-target test-time evasion attack. These differences motivate a domain-specific attack design rather than a direct reuse of general graph attacks.

%% file: sections/preliminaries.tex
\section{Preliminaries}
\label{sec:otprelims}

We model a social network as a directed graph
$\mathcal{G} = (\mathcal{V}, \mathcal{E}, \mathbf{X}),$ where $\mathcal{V}$ is the set of nodes (accounts), $\mathcal{E} \subseteq \mathcal{V} \times \mathcal{V}$ is the set of directed edges (follow relations), and $\mathbf{X} \in \mathbb{R}^{|\mathcal{V}| \times d}$ denotes node features (profile, posts, tweets). For a node $v \in \mathcal{V}$, let $\mathcal{N}_k(v)\subseteq\mathcal{V}$ denote the set of nodes in its $k$-hop ego-neighborhood. We use $\mathcal{Z}_v=\{z_i\}_{i=1}^{m_v}$ for the corresponding multiset of neighbor feature vectors, where each $z_i$ is constructed from $\mathbf{X}$ and graph structure and encodes attribute, relational, and temporal information. This separates the node set $\mathcal{N}_k(v)$ from the feature vectors on which OT is computed.

We now summarize the optimal transport (OT) formulation~\cite{petric2019got,peyre2019computational} used to compare two empirical distributions over $\mathbb{R}^d$. 
Given two discrete distributions
$\mu_1 = \sum_{i=1}^m a_i \,\delta_{z_i}$ and
$\mu_2 = \sum_{j=1}^n b_j \,\delta_{\tilde z_j}$,
where $\delta_z$ denotes the Dirac measure concentrated at $z \in \mathbb{R}^d$~\citep{tao2011introduction},
with nonnegative weights $a_i$ and $b_j$ satisfying $\sum_i a_i = 1$ and $\sum_j b_j = 1$,
and a nonnegative ground cost $c(z_i, \tilde z_j)$ between feature vectors, we define the cost matrix $C \in \mathbb{R}^{m \times n}$ by $C_{ij} = c(z_i,\tilde z_j)$. The (unregularized) OT distance, $W_c(\mu_1,\mu_2)$ is:
$
\min_{P \in \mathbb{R}^{m \times n}_{\ge 0}} \langle P, C \rangle_F\text{ }
\text{s.t. } P \mathbf{1}_n = a,\;
P^\top \mathbf{1}_m = b
$, 
where $\langle P, C \rangle_F = \sum_{i,j} P_{ij} C_{ij}$ is the Frobenius inner product between $P$ and $C$, and $\mathbf{1}_m$ is the ones column vector in $\mathbb{R}^m$.

OT can be entropy-regularized for computational stability and scalability. Throughout the paper, $\log$ denotes the natural logarithm and we use the standard convention $0\log 0=0$. For $\varepsilon > 0$, the regularized problem adds the term $\varepsilon \sum_{i,j} P_{ij} (\log P_{ij} - 1)$ to the objective. This term may be negative because $P_{ij}\in[0,1]$ and because $\sum_{i,j}P_{ij}=1$; the ``$-1$'' contributes only the additive constant $-\varepsilon$, independent of $P$, but yields the convenient derivative $\varepsilon\log P_{ij}$. The resulting problem is strictly convex and has a unique solution $P^\star$ that can be efficiently computed via Sinkhorn~\cite{cuturi2013sinkhorn} iterations. The corresponding transport cost $D(\mu_1,\mu_2) = \sum_{i,j} P^\star_{ij} c(z_i, \tilde z_j)$ defines a distance between empirical distributions. The optimal plan admits a closed-form scaling structure $P^\star = \mathrm{diag}(u)\, K\, \mathrm{diag}(v)$, where $u \in \mathbb{R}^m$ and $v \in \mathbb{R}^n$ are scaling vectors that enforce the marginal constraints, and $K_{ij} = \exp(-c(z_i, \tilde z_j)/\varepsilon)$ is the Gibbs kernel derived from the cost matrix.

\heading{Semi-Supervised Node Classification.}
Let a node subset $\mathcal{V}_{\mathrm{train}} \subset \mathcal{V}$ be labeled with class labels $y_v \in \mathcal{Y}$ (e.g., human vs.\ bot), while the remaining nodes are unlabeled.
A GNN-based classifier $f_\Theta$ maps each node $v$ to a predictive distribution over labels: $f_\Theta : (\mathcal{G}, \mathbf{X}) \mapsto \bigl( p_\Theta(y \mid v) \bigr)_{v \in \mathcal{V}}$. 

\heading{Adversarial Attack.}
An adversarial attack perturbs the graph $\mathcal{G}$ to induce incorrect predictions by a trained classifier $f_\Theta$. We focus on test-time evasion attacks that modify graph inputs while keeping the model fixed. 

The attacker may change the graph structure and node features under a fixed perturbation budget $\Delta$. Let $\mathbf{A}$ and $\mathbf{A}'$ be the adjacency matrices of the original and perturbed graphs, and let $\mathbf{X}$ and $\mathbf{X}'$ be the corresponding node feature matrices. The total perturbation is constrained as:
\begin{equation}
\sum_{u,v} \left| \mathbf{A}_{uv} - \mathbf{A}'_{uv} \right| 
\;+\;
\|\mathbf{X} - \mathbf{X}'\|_0
\;\leq\; \Delta,
\end{equation}
where $\|\cdot\|_0$ denotes the number of feature entries changed. The resulting perturbed graph $\mathcal{G}' = (\mathcal{V}, \mathcal{E}', \mathbf{X}')$ is used at inference time, and the attacker’s goal is to cause target nodes to be misclassified with the modified input.
%under the modified inputs.
%MK: this formulization is standard. at least at a sentence saying how we relax it. KM: done

%% file: sections/Problem_Formulation.tex
\subsection{Problem Formulation}\label{subsec:problem-formulation}

Let $\mathcal{G}=(\mathcal{V},\mathcal{E},\mathbf{X})$ be a directed social graph, where $\mathcal{V}$ denotes user accounts, $\mathcal{E}\subseteq\mathcal{V}\times\mathcal{V}$ directed follow relations, and $\mathbf{X}\in\mathbb{R}^{|\mathcal{V}|\times d}$ node features. A subset $\mathcal{V}_L\subset\mathcal{V}$ is labeled for semi-supervised bot detection, with $y_v\in\{\textsf{human},\textsf{bot}\}$. A deployed bot detector $f$ is trained on $\mathcal{G}$ and outputs class probabilities $p_f(y\mid v;\mathcal{G})$ for any node $v$.

The attacker operates at test time and introduces a new bot node $v_{\mathrm{t}}\notin\mathcal{V}$ with features $\mathbf{x}_{\mathrm{t}}$. The attacker selects a set of incident edges, $
\Delta\mathcal{E}(v_{\mathrm{t}})\subseteq ({v_{\mathrm{t}}}\times\mathcal{V})\cup(\mathcal{V}\times{v_{\mathrm{t}}})$,
resulting in the modified graph, $
\mathcal{G}'=(\mathcal{V}\cup\{v_{\mathrm{t}}\},\ \mathcal{E}\cup\Delta\mathcal{E}(v_{\mathrm{t}}),\ \mathbf{X}\cup\{\mathbf{x}_{\mathrm{t}}\})$.
Outgoing follow edges from $v_{\mathrm{t}}$ are freely chosen. Incoming edges (follow-backs) are not directly controllable for humans and must respect domain constraints.

% \heading{Domain Constraints.}
% Edits are constrained to a feasible set $\mathcal{F}(B)$ defined by a fixed budget $B$ and plausibility criteria:
% {\small \[
% \mathcal{F}(B)=\Bigl\{\Delta\mathcal{E}:\ |\Delta\mathcal{E}|\le B,\ \Delta\mathcal{E}\text{ incident to }v_{\mathrm{t}},\ \Psi(\Delta\mathcal{E})\le 0\Bigr\}.
% \]}
% %MK: the formulization says  \Psi(\Delta\mathcal{E})\le 0\. Please clarify how this function captures edge plausibility conditions. KM: in the next section, it is described in details
% Here, $\Psi(\Delta\mathcal{E})$ captures edge plausibility conditions, including directionality and temporal alignment, discussed later.

\heading{Domain Constraints.}
All constrained results use the same admissible-edit set for every attack. Let $\mathcal{V}_{\mathrm{hum}}$ be the human nodes. Edits are constrained to a feasible set $\mathcal{F}(B)$ defined by a per-target budget $B$ and hard social-plausibility rules:
{\small
\[
\begin{aligned}
\mathcal{F}(B)=\{\Delta\mathcal{E}:&\ |\Delta\mathcal{E}|\le B,
\ \Delta\mathcal{E}\subseteq (\{v_{\mathrm{t}}\}\times\mathcal{V})\cup(\mathcal{V}\times\{v_{\mathrm{t}}\}),\\
& \Delta\mathcal{E}\cap\bigl(\mathcal{V}_{\mathrm{hum}}\times\{v_{\mathrm{t}}\}\bigr)=\emptyset,\\
&
\ \Psi_{\mathrm{dir}}(\Delta\mathcal{E})\le 0,
\ \Psi_{\mathrm{time}}(\Delta\mathcal{E})\le 0\}.
\end{aligned}
\]
}
The first condition enforces the edge budget; the second makes every edit incident to the target bot; the third prohibits forced human$\rightarrow$bot follow-backs; and $\Psi_{\mathrm{dir}}$ and $\Psi_{\mathrm{time}}$ reject directionally invalid or temporally inconsistent edits. Thus the attacker cannot rewire unrelated graph regions, create arbitrary human follow-backs, or modify features, labels, or metadata of any non-target node.

% \heading{Attacker's Capability.}
% The attacker performs targeted evasion by injecting or editing a bot $v_{\mathrm{t}}$ so that it is misclassified as \textsf{human}. Edits are restricted to edges incident to $v_{\mathrm{t}}$ and must respect a hard edge budget $B$. The attacker can freely choose outgoing follow edges from $v_{\mathrm{t}}$ to existing nodes, but cannot force incoming follow-backs from existing users (e.g, cannot make an arbitrary human follow its bot). No changes are permitted to the edges between any other pairs of nodes, nor to the features, labels, or metadata (\eg account age) of any node other than $v_{\mathrm{t}}$ 
% %MK: check my justification I added below KM: done
% since the bot creator may not control how a human follows other nodes.

\heading{Attacker's Capability.}
The attacker performs targeted evasion by injecting or editing a bot $v_{\mathrm{t}}$ so that it is misclassified as \textsf{human}. Edits are restricted to edges incident to $v_{\mathrm{t}}$ and must respect a hard edge budget $B$. The attacker can freely choose outgoing follow edges from $v_{\mathrm{t}}$ to existing nodes, but cannot force incoming follow-backs from existing users (e.g., cannot make an arbitrary human follow its bot). No changes are permitted to the edges between any other pairs of nodes, nor to the features, labels, or metadata (\eg account age) of any node other than $v_{\mathrm{t}}$.

\heading{Attacker’s Knowledge.}
We assume a black-box setting. The attacker has access to the training graph and its node labels, as well as the feature schema (profile, content, and edges), but \textit{no knowledge of the victim model’s architecture, parameters, gradients, or output scores}. The attacker cannot query the model or observe logits. Instead, the attacker observes which accounts are flagged or removed by the platform and which accounts remain active over time, as reflected through observable metadata such as account age. For injected bots, the attacker may assign non-temporal profile and content features (\eg username, bio).
%consistent with the feature schema.

We do not consider poisoning (training-time) or availability (global degradation) attacks. The \textit{focus is on inserting bots that evade detection under realistic deployment conditions.}

\begin{figure*}[!htb]
\centering
\resizebox{0.68\linewidth}{!}{%
\includegraphics{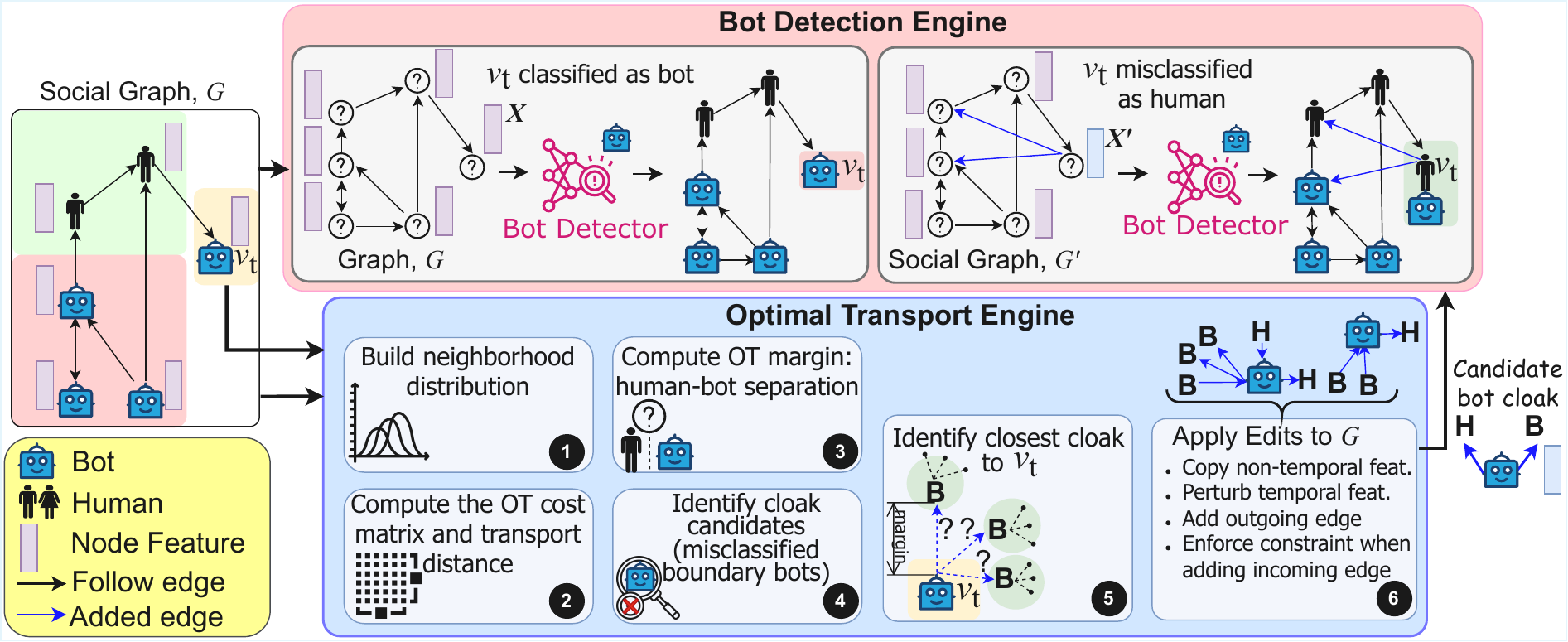}%
}
\caption{Overview of \pname. The method learns an optimal transport geometry over $k$-hop neighborhood distributions and uses transport plans to guide sparse, plausible local edge edits for evasion.}
\label{fig:overview}
\end{figure*}

\heading{Victim Models.}
We consider two categories of victim models~\citep{bacciu_gentle_2020}: vanilla detectors, which are trained without explicit adversarial robustness objectives, and defended detectors, which integrate adversarial defense. Attacks on vanilla models establish a baseline vulnerability level, whereas evaluations on defended models assess the effectiveness of robustness strategies. Notably, these defenses typically operate without prior knowledge (\ie non-adaptive) of the specific attack techniques employed.

%% file: sections/method.tex
\section{\pname: Constrained Graph Attacks via Optimal Transport}

We present \pname (shown in \autoref{fig:overview}), a framework for constrained adversarial attacks on social graphs that generate evasive bot accounts through node editing or injection.

\heading{Challenges.}
Issues that shape the attack design: search and feasibility. First, the neighborhood around a bot is combinatorial because a target bot can connect to a large candidate pool with many possible degree patterns, interaction roles, and temporal profiles. Direct search over edge sets $\Delta\mathcal{E}$ is not feasible at the scale of social graphs. Second, bot edits must respect behavioral and structural constraints. The attacker cannot rewire the rest of the graph or force follow-backs. \textit{These constraints exclude many perturbations assumed in standard adversarial settings}.

% \heading{\pname Philosophy.}
% \pname is a geometry-first method that treats local neighborhoods, rather than edges, as first-class objects of adversarial manipulation. Each node is represented by a probability measure over its $k$-hop ego neighborhood in a spatiotemporal feature space, and optimal transport compares these distributions under a learned ground cost. This induces a metric geometry in which bots and humans (potentially) occupy distinct regions, with misclassified bots near the decision boundary. \pname selects such boundary bots as cloak templates to derive sparse, feasible edge edits by decoding transport plans.

\heading{\pname Philosophy.}
\pname is a geometry-first method that treats local neighborhoods, rather than individual edges, as first-class objects of adversarial manipulation. Each node is represented by a probability measure over its ego-neighborhood in a spatio-temporal feature space, and optimal transport compares these distributions under a learned ground cost. This induces a metric geometry in which bots and humans occupy distinct regions, with misclassified bots near the decision boundary. \pname selects such boundary bots as cloak templates to derive sparse, feasible edge edits by decoding transport plans.

\heading{Why OT rather than scalar matching?}
A homophily or embedding similarity score can say that two neighborhoods are close, but it does not specify which neighbor mass should be aligned under a budget. OT is a mapping that returns both a scalar distance and a coupling $P^\star$: the distance selects a template region to imitate, while the coupling identifies the few high-mass neighbor correspondences that can be decoded into feasible target-incident edits. Thus \pname reconstructs a plausible local neighborhood rather than copying an entire ego-network or greedily searching over all edges.

\subsection{Neighborhood and Boundary Definitions}
\label{subsec:ot-neighborhood}

Let $\mathcal{N}_k(v)$ denote the node set in the $k$-hop ego-neighborhood of node $v$; in experiments we use $k=1$ unless otherwise stated. For each neighbor $\eta \in \mathcal{N}_k(v)$ we define a feature vector $\phi_v(\eta)\in\mathbb{R}^d$ that summarizes attributes of $\eta$ and its relation to $v$. The construction of $\phi_v(\eta)$ is domain dependent, but the method only requires that $\phi_v(\eta)$ be a fixed-dimensional representation that can encode static attributes, interaction role features, content signals, and temporal signals. 

\heading{Neighborhood Distribution and Cost Measures.}
We represent the neighborhood of $v$ as an empirical probability measure (as shown in \ballnumber{1} of \autoref{fig:overview} )
$$
\mu_v = \sum_{\eta \in \mathcal{N}_k(v)} w_v(\eta)\, \delta_{\phi_v(\eta)},
$$ where $\delta_x$ is a point mass at $x$ and $w_v(\eta)$ is a normalized importance weight. The weighting scheme lets the measure emphasize neighbors that carry a stronger behavioral signal for detection. We set $w_v(\eta)$ by normalizing an importance score $a_v(\eta)>0$ (details in Appendix \autoref{sec:prob-measure}, ~\autoref{eq:mu-weighted}):
$$w_v(\eta)=\frac{a_v(\eta)}{\sum_{\eta'\in\mathcal{N}_k(v)} a_v(\eta')}.
$$
The score $a_v(\eta)$ is defined as the product of a structural term and a temporal term:
$a_v(\eta)=g_{\mathrm{str}}\bigl(s(\eta)\bigr)\,
g_{\mathrm{temp}}\bigl(t(\eta)\bigr)$, where $s(\eta)$ is a scalar structural statistic and $t(\eta)$ is a scalar temporal statistic. The product acts as an ``and'' gate: a neighbor receives high mass only when it is both structurally informative and temporally plausible, rather than merely satisfying one criterion. The functions $g_{\mathrm{str}}$ and $g_{\mathrm{temp}}$ are monotone and chosen to emphasize informative neighbors while avoiding extreme weight concentration.

\heading{What topology is retained?}
\pname does not require the full platform graph at attack time. Instead, topology enters through the ego-neighborhood node set, directed edge-role features, in/out-degree statistics, and feasibility checks on candidate edits. This deliberately encodes the structural information available to an attacker with partial observability while avoiding assumptions that the attacker can inspect or rewire arbitrary distant graph regions.

We define a ground cost between neighbor feature vectors through a learnable embedding. Let $h_\theta: \mathbb{R}^d \to \mathbb{R}^{d_{\mathrm{emb}}}$ be a neural map, and let $\mathbf{M} \succeq 0$ be a learnable positive semidefinite matrix. For neighbor features $z = \phi_v(\eta)$ and $z' = \phi_\xi(\eta')$, with $\eta \in \mathcal{N}_k(v)$ and $\eta' \in \mathcal{N}(\xi)$, we define $c_\theta(z, z') = \| h_\theta(z) - h_\theta(z') \|_{\mathbf{M}}^2$ where $\|u\|_{\mathbf{M}}^2 \triangleq u^\top \mathbf{M} u$. This choice yields a smooth cost that correlates features and supports stable entropic OT computations.

\subsection{Optimal Transport Distance Between Nodes}
\label{subsec:ot-distance}

Let $c_\theta: \mathbb{R}^d \times \mathbb{R}^d \to \mathbb{R}_{\ge 0}$ be a learned ground cost between neighboring feature vectors. Using the entropy-regularized optimal transport formulation from Section~\ref{sec:otprelims}, we define the distance between two nodes $v$ and $\xi$ via their neighborhood measures, (as shown in \ballnumber{2} of \autoref{fig:overview}).
%MK: why there is blue and red in this formula? is it for checking something? KM: I think Cynuet did it to help understand how cost and transport plan is related

Given neighborhood measures $\mu_v = \sum_{i=1}^m a_i \delta_{z_i}$
  {and} $\mu_\xi = \sum_{j=1}^n b_j \delta_{z'_j}$,
we construct the cost matrix $\textcolor{myred}{C_{ij}} = c_\theta(z_i, z'_j)$ and solve the following entropic optimal transport problem:

{\small
\begin{equation}
\label{eq:ot-trans}
% $$
\textcolor{myblue}{P^\star_{v\xi}} \in \arg\min_{P \in \mathcal{U}(a,b)}
{
\underbrace{\sum_{i,j} P_{ij} \textcolor{myred}{C_{ij}}}_{\text{core}}
+
\underbrace{\varepsilon \sum_{i,j} P_{ij} (\log P_{ij} - 1)}_{\text{regularizer}}
},
% $$
\end{equation}
}

where the transport polytope is
$\mathcal{U}(a,b) = \left\{ P \in \mathbb{R}_{\ge 0}^{m \times n} \,:\, P \mathbf{1}_n = a,\ P^\top \mathbf{1}_m = b \right\}$. Here $\mathbf{1}_m$ and $\mathbf{1}_n$ enforce that $P$ has row sums $a$ and column sums $b$, further details in Appendix~\autoref{sec:ground-cost} and \autoref{sec:entropic-ot}.

We define the OT distance between nodes $v$ and $\xi$ as the transport cost induced by the optimal plan:

{\small
\begin{equation}
\label{eq:ot-cost}
D_\theta(v, \xi) \triangleq \langle \textcolor{myblue}{P^\star_{v\xi}}, C_{v\xi} \rangle.
\end{equation}
}

We compute the plan $\textcolor{myblue}{P^\star_{v\xi}}$ using Sinkhorn iterations, 
% detailed derivation in ~\autoref{app:en-ot-scomp}. The distance compares neighborhoods by transporting mass between their features under the learned cost $c_\theta$.
detailed derivation in~\autoref{app:en-ot-scomp}. With $K=\exp(-C/\varepsilon)$, Sinkhorn alternates $u\leftarrow a\oslash(Kv)$ and $v\leftarrow b\oslash(K^\top u)$ for $T_{\mathrm{sink}}$ iterations and returns $P^\star\approx\mathrm{diag}(u)K\mathrm{diag}(v)$. The distance compares neighborhoods by transporting mass between their features under the learned cost $c_\theta$.

\heading{Learning a Label-Aware Geometry.}
%MK: do we learn \theta or h_theta ? \kunal: \theta
% We train $\theta$ so that neighborhoods from the same class are close and different classes are separated. Let $\Pi^+$ and $\Pi^-$ denote sets of the same label and different label node pairs sampled from $\mathcal{V}\times\mathcal{V}$. We fit $\theta$ with a contrastive margin objective

% {\small
% \begin{equation}
% \label{eq:theta-learning}
% \min_{\theta}\
% \sum_{(i,j)\in\Pi^+} D_\theta(i,j)
% \;+\;
% \sum_{(i,k)\in\Pi^-}
% \max\left(0,\, \gamma - D_\theta(i,k)\right),
% \end{equation}
% }

% where $\gamma>0$ is a margin hyperparam. The first term contracts within class neighborhoods. The second term penalizes different class pairs whose OT distance is below $\gamma$.

% We train $\theta$ offline with a multi-term loss that shapes margins and favors sparse and plausible alignments. We define a margin surrogate loss that depends on $m(v)$, which is specified in the next subsection, and on whether $v$ is currently misclassified by $f_\Theta$. We further add a sparsity loss that penalizes diffuse transport plans (details in Appendix \autoref{app:full-obj}), then add a plausibility loss that penalizes transport mass assigned to feature mismatches that violate domain rules.  
We train $\theta$ offline with a multi-term loss that shapes margins and favors sparse and plausible alignments. For each training bot $v$, the compact objective is
{\small
\begin{equation}
\label{eq:main-compact-loss}
\begin{aligned}
\mathcal{L}_{\pname}(\theta)
&=
\mathbb{E}_{v\in\mathcal{V}_{\mathrm{train}}}
\left[
\lambda_{\mathrm{BCE}}L_{\mathrm{BCE}}(v;\theta)
+
\lambda_{\mathrm{sp}}L_{\mathrm{sp}}(v;\theta)
\right. \\
&\qquad\left.
+
\lambda_{\mathrm{pl}}L_{\mathrm{pl}}(v;\theta)
\right].
\end{aligned}
\end{equation}
}
Here $L_{\mathrm{BCE}}$ shapes the OT margin $m(v)$, $L_{\mathrm{sp}}$ penalizes diffuse transport plans so that decoded edits are sparse, and $L_{\mathrm{pl}}$ penalizes transport mass assigned to feature mismatches that violate degree or temporal plausibility rules. Appendix~\autoref{app:full-obj} gives the full definitions and ~\autoref{alg:train-ot-geometry} gives the training procedure.
 
\subsection{Boundary Cloak Templates}
\label{subsec:boundary-cloak}

% Let $\mathcal{V}_{\mathrm{bot}}$ and $\mathcal{V}_{\mathrm{hum}}$ denote the sets of bot and human nodes as identified by the ground truth. For $v \in \mathcal{V}_{\mathrm{bot}}$, define the closest human and bot distances by $d_{\mathrm{hum}}(v) = \min_{h \in \mathcal{V}_{\mathrm{hum}}} D_\theta(v, h)$, and $d_{\mathrm{bot}}(v) = \min_{\xi \in \mathcal{V}_{\mathrm{bot}} \setminus \{v\}} D_\theta(v, \xi),$ and let the OT margin be $m(v) = d_{\mathrm{hum}}(v) - d_{\mathrm{bot}}(v)$, (details in Appendix~\autoref{subsec:ot-margin-main} and \ballnumber{3} of \autoref{fig:overview}). Large $m(v)$ implies $v$ lies deeper in the bot region; small or negative values is proximity to the human.

Let $\mathcal{V}_{\mathrm{bot}}$ and $\mathcal{V}_{\mathrm{hum}}$ denote the sets of bot and human nodes as identified by the ground truth. For $v \in \mathcal{V}_{\mathrm{bot}}$, define the closest human and bot distances by $d_{\mathrm{hum}}(v) = \min_{h \in \mathcal{V}_{\mathrm{hum}}} D_\theta(v, h)$, and $d_{\mathrm{bot}}(v) = \min_{\xi \in \mathcal{V}_{\mathrm{bot}} \setminus \{v\}} D_\theta(v, \xi),$ and let the OT margin be $m(v) = d_{\mathrm{hum}}(v) - d_{\mathrm{bot}}(v)$, (details in Appendix~\autoref{subsec:ot-margin-main} and \ballnumber{3} of \autoref{fig:overview}). Large positive $m(v)$ means $v$ is closer to other bots than to humans in OT space; small or negative values indicate proximity to the human region.

We then enforce a margin threshold $m(v) \le \tau_{\mathrm{bdry}}$ to select bot nodes whose neighborhoods are closer to human nodes than to other bots under $D_\theta$, further details in Appendix~\autoref{app:train-op}. We define the boundary set as $\mathcal{B}_{\mathrm{bdry}} = \{ v \in \mathcal{V}_{\mathrm{bot}} : m(v) \le \tau_{\mathrm{bdry}}\},$ where $\tau_{\mathrm{bdry}}$ is a validation-set threshold.
%MK: it sounds like we control the f_Theta.  Please rewrite this as we are merely observing the results. this could confuse a reviewer. Please note we assume blackbox access.  KM: done
Using a pre-trained node classifier $f_\Theta$ as a black-box, we define the set of misclassified bots that the classifier predicts as human:
% tiny cannot be used with math mode
% \begin{equation}
% \label{eq:boundary-set}
% {\tiny \mathcal{B}_{\mathrm{mis}} \triangleq
% \left\{
% v \in \mathcal{V}:
% y_v = \textsf{bot}
% \land
% \arg\max_y p_\Theta(y \mid v; \mathcal{G}) = \textsf{human}
% \right\}.}
% \end{equation}
% \begin{equation}
% \label{eq:boundary-set}
% \scalebox{0.80}{$
$\mathcal{B}_{\mathrm{mis}} =
\left\{
v \in \mathcal{V}:
y_v = \textsf{bot}
\land
\arg\max_y p_\Theta(y \mid v; \mathcal{G}) = \textsf{human}
\right\}$.
% $}
% \end{equation}

The cloak candidate set in \autoref{eq:ot-relaxed-attack} can be instantiated as $\mathcal{B}_{\mathrm{cand}}=\mathcal{B}_{\mathrm{bdry}}\cap\mathcal{B}_{\mathrm{mis}}$  (as shown in \ballnumber{4} of \autoref{fig:overview}). 
% or as a small pool sampled from $\mathcal{B}_{\mathrm{bdry}}\cup\mathcal{B}_{\mathrm{mis}}$. % under a reuse cap 
These nodes serve as cloak templates because they already satisfy the detector decision rule on the current graph.  After training the OT geometry (details in Appendix~\autoref{app:full-obj} and algorithm in \autoref{alg:train-ot-geometry}), we compute $D_\theta(v,\xi)$ for the pairs needed to evaluate $d_{\mathrm{hum}}(v)$ and $d_{\mathrm{bot}}(v)$ on the bot set. We then form $\mathcal{B}_{\mathrm{bdry}}$ and $\mathcal{B}_{\mathrm{mis}}$ and select a small pool of cloak templates for each target. 
%MK: OT solves? Please clarify. KM: done
This restriction limits the number of OT iterations required during attack generation. 
%The full OT-guided attack procedure is given in ~\autoref{alg:bocloak-newbot}.

\subsection{Attack Objectives and Plans}
\label{subsec:attack-plans}

\begin{figure*}[!t]
\centering
\begin{subfigure}[t]{0.48\linewidth}
\centering
\resizebox{1.0\columnwidth}{!}{%
\includegraphics{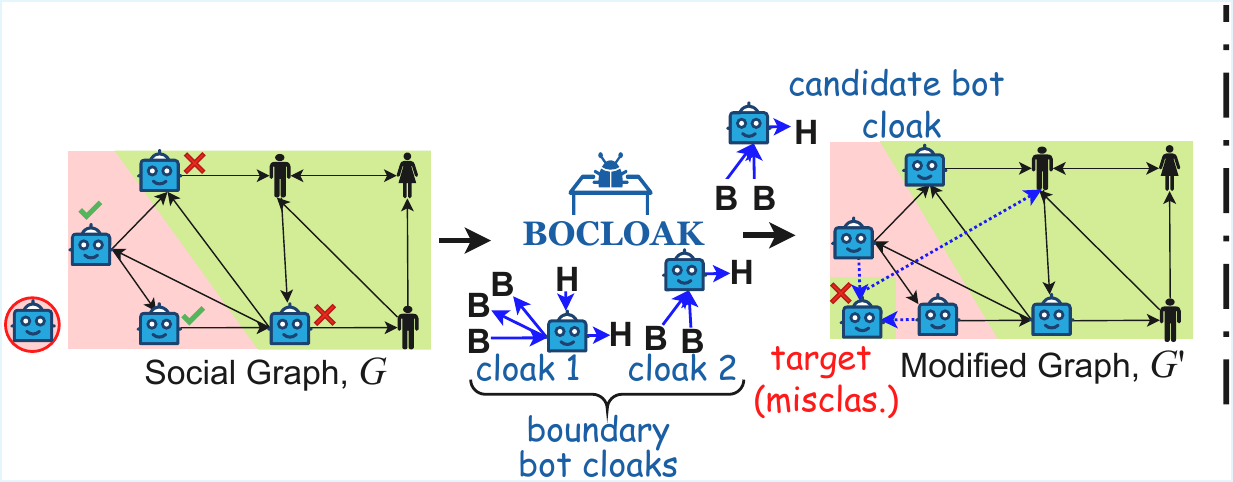}%
}
\caption{\pname node injection pipeline.}
\label{fig:newbot}
\end{subfigure}
\begin{subfigure}[t]{0.48\linewidth}
\centering
\resizebox{1.0\columnwidth}{!}{%
\includegraphics{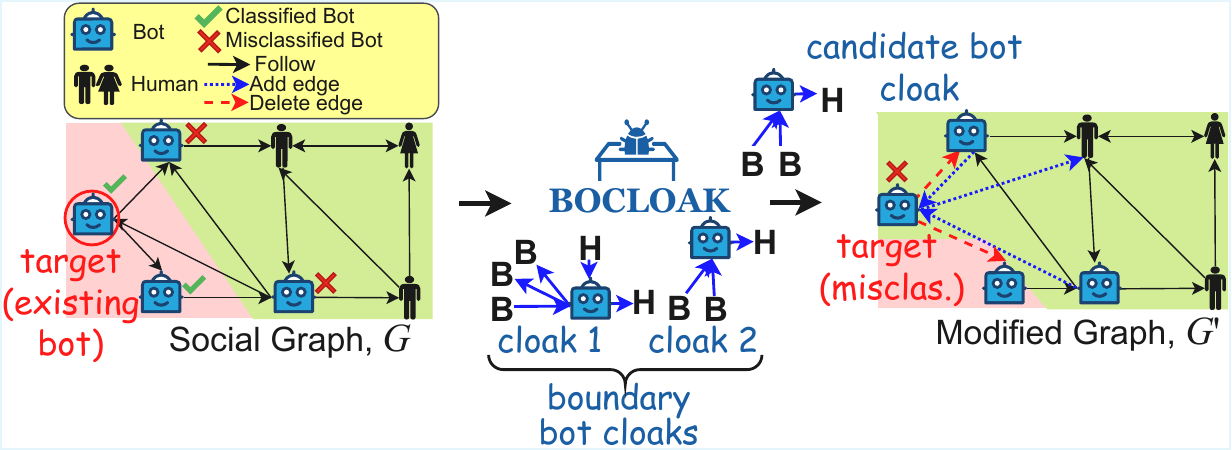}%
}
\caption{\pname node editing pipeline.}
\label{fig:oldbot}
\end{subfigure}
\caption{\pname pipeline for node injection and node editing. It operates on local neighborhoods rather than individual edges. Boundary bot cloaks are identified in the learned optimal transport geometry and used to guide sparse, plausible edge edits. (a) Injection constructs a new bot with a human-like neighborhood. (b) Editing transforms an existing bot’s neighborhood to induce misclassification.}
\label{fig:pipeline}
\end{figure*}

We consider two attack scenarios: node injection and editing. In node injection (\autoref{fig:newbot}), the attacker introduces a new bot node $v_{\mathrm{t}}\notin\mathcal{V}$ and adds a set of outgoing edges incident to $v_{\mathrm{t}}$. In node editing (\autoref{fig:oldbot}), the attacker selects an existing bot node $v_{\mathrm{t}}\in\mathcal{V}$ and modifies its incident edges through $\Delta\mathcal{E}$. Node injection can be viewed as a special case of node editing in which $v_{\mathrm{t}}$ initially has no incident edges.

\heading{Conceptual Objective.}
The attacker aims to construct a neighborhood around $v_{\mathrm{t}}$ that resembles a cloak template while respecting feasibility (details in Appendix~\autoref{app:op-newbot}). Let $\mu_{\mathrm{t}}$ denote the neighborhood measure of $v_{\mathrm{t}}$ after applying $\Delta\mathcal{E}$, where $\mu_{\mathrm{t}}(\Delta\mathcal{E})$ is explicit dependency.
% We write $\mu_{\mathrm{t}}(\Delta\mathcal{E})$ when we need to make this dependence explicit. 
\pname uses the following OT-guided objective as a target specification:

{\small
\begin{equation}
\label{eq:ot-relaxed-attack}
\min_{\Delta\mathcal{E} \in \mathcal{F}(B)} \;
\min_{v_c \in \mathcal{B}}
D_\theta\!\bigl(\mu_{\mathrm{t}}(\Delta\mathcal{E}), \mu_c\bigr)+\lambda_{\mathrm{pl}}  \Phi(\Delta\mathcal{E})+\lambda_{\mathrm{sp}}  |\Delta\mathcal{E}|
\end{equation}
}
where $\mathcal{F}(B)$ encodes the edge budget and hard feasibility constraints, $\Phi(\Delta\mathcal{E})$ penalizes implausible edits under domain rules, and $|\Delta\mathcal{E}|$ enforces sparsity. The inner minimization selects a cloak template $v_c$ from bot candidates~$\mathcal{B_{\mathrm{cand}}}$.

\subsection{Approximation Strategy}
\label{subsec:approx-strategy}

Direct optimization of~\autoref{eq:ot-relaxed-attack} is not tractable because $\Delta\mathcal{E}$ ranges over discrete edge sets and the objective depends on neighborhood measures induced by those edits. \pname first learns $D_\theta$ offline, then identifies a small candidate pool of cloak templates near the detector boundary, and finally converts a transport plan into concrete edge edits; full loss objective is described in Appendix~\autoref{app:full-obj}, ~\autoref{eq:overall-loss}. The decoding step maps an OT plan to $\Delta\mathcal{E}$ under the budget and plausibility constraints.

\subsection{Decoding OT Plans into Discrete Edges}
\label{subsec:decode-edges}

Given a target bot $v_{\mathrm{t}}$, \pname selects a cloak $v_c$ from the candidate set $\mathcal{B}_{\mathrm{bdry}}$ (\pname algorithm described in ~\autoref{alg:bocloak-newbot} and  shown in \ballnumber{5} of \autoref{fig:overview}). Although $v_c$ is itself a bot, the victim model predicts it as human, placing it near or within the human region of the learned OT geometry. To identify which parts of its neighborhood contribute most to this human-like appearance, we compare $v_c$ to a nearby human $v_h \in \arg\min_{h \in \mathcal{V}_{\mathrm{hum}}} D_\theta(v_c, h)$.%, where $\mathcal{V}_{\mathrm{hum}}$ is human set.

\pname computes the entropic OT plan $P^\star_{ch}$ between the neighborhood distributions $\mu_c$ and $\mu_h$ using Sinkhorn iterations under the learned cost $c_\theta$ (details in Appendix~\autoref{sec:entropic-ot}). This plan aligns neighbors of $v_c$ with those of $v_h$ according to structural, behavioral, and temporal similarity.

From this OT plan, \pname extracts a small set of high-mass cloak-side neighbors and converts them into the target bot's incident edges, subject to the budget $B$ and feasibility constraints. In node injection, these edges define the initial neighborhood of a new node. In node editing, they replace or augment a subset of existing incident edges (as shown in \ballnumber{6} of \autoref{fig:overview}). This decoding step yields a discrete set $\Delta\mathcal{E}$ that approximates the continuous objective in~\autoref{eq:ot-relaxed-attack}.

\heading{Why Bot-Cloaks?}
We prefer misclassified bots over true humans as cloak templates because existing bots are more likely to maintain connections aligned with bot objectives, such as promotion or coordination. Mimicking arbitrary humans may yield bots that evade detection but fail to generate meaningful engagement.
%MK: should we mention that finding such nodes are always feasible since the bot detector makes mistakes? KM: :)

%MK: never mind my previous comments. Special case explains it.
\heading{Special Case.}
If no bot cloaks are available, \pname falls back to selecting a nearby human in OT space and constructs an evading bot by mimicking that human’s neighborhood.

% \heading{Complexity.}
% Let $m$ and $n$ denote the neighborhood sizes of $\mu_v$ and $\mu_\xi$. One Sinkhorn OT solve between them costs $O(T_{\mathrm{sink}}mn)$ time and $O(mn)$ space, which dominates \pname's execution (details in Appendix~\autoref{app:complexity}).
\heading{Complexity.}
Let $m$ and $n$ denote the neighborhood sizes of $\mu_v$ and $\mu_\xi$. One Sinkhorn OT solve between them costs $O(T_{\mathrm{sink}}mn)$ time and $O(mn)$ space, which dominates \pname's execution (details in Appendix~\autoref{app:complexity}). The dependence is on local ego-neighborhood density rather than the full graph size. For very large social graphs or high-degree nodes, \pname can use the same formulation with two practical truncations: retain only the top-$K$ boundary templates and keep only the highest-weight neighbors when forming each empirical measure.

%% file: sections/eval.tex
\section{Experiments}

In this section, we present experimental results that demonstrate the effectiveness and practicality of the \pname.
% We demonstrate \pname's effectiveness and practicality.

\input{tables/degree-short}

\input{tables/old-node-50}
\heading{Datasets.} We evaluate \pname on three widely used social bot detection datasets: TwiBot-22~\citep{feng2022twibot22}, a large-scale graph-based benchmark featuring heterogeneous relations and high-quality annotations; Cresci-2015~\citep{cresci2015fakers}, which includes early social spambots, fake followers, and genuine users with explicit interaction graphs; and BotSim-24~\citep{qiao2025botsim}, a recent benchmark that simulates highly human-like social botnets. The statistics of the datasets are summarized in \autoref{tab:avg-node-degree-short}.

\heading{Bot Detection Models.} We employ the \gnn-based node classifiers, such as GCN~\citep{kipf2017gcn} and GAT~\citep{velickovic2017gat} as baseline methods. We adopt the prominent BotRGCN~\citep{feng2021botrgcn}, Simple-HGNN (S-HGN)~\citep{lv2021we}, and RGT~\citep{feng2021rgt} models as \sota bot detectors. 
%The detection performance is shown in App. \autoref{tab:f1-gnn-bots-defense}.

% \heading{GNN Attack Models.} We compare \pname{} with random baselines and SOTA adversarial attack frameworks, including Nettack~\citep{zugner2018nettack}, FGA~\citep{chen2018fga}, PR-BCD~\citep{geisler2021_robustness_of_gnns_at_scale}, and GOttack~\citep{alom2025gottack} across both vanilla bot detectors and their defended counterparts, including GNNGuard~\citep{zhang2020gnnguard}, GRAND~\citep{feng2020grand}, and RobustGCN~\citep{zhu2019robustgcn} for node editing attack. We compare \pname{} with random baselines for node injection attack since no node injection attack for social graph has been designed.
\heading{GNN Attack Models.} We compare \pname{} with random baselines and SOTA adversarial attack frameworks, including Nettack~\citep{zugner2018nettack}, FGA~\citep{chen2018fga}, PR-BCD~\citep{geisler2021_robustness_of_gnns_at_scale}, GOttack~\citep{alom2025gottack}, and the LR-BCD control~\citep{gosch2023adversarial} across both vanilla bot detectors and their defended counterparts, including GNNGuard~\citep{zhang2020gnnguard}, GRAND~\citep{feng2020grand}, and RobustGCN~\citep{zhu2019robustgcn} for node editing attack. We additionally evaluate NoisyGNN~\citep{ennadir2024noisygnn} as a robustness-control defense on the strongest BotRGCN setting. We compare \pname{} with random baselines for node injection attack since no prior node-injection attack for social graphs is directly comparable under our feasibility rules.

For each dataset, we follow the original training protocols for these detectors, including train/validation splits and early-stopping criteria. We further verify that our implementations match or exceed the reported clean accuracies before evaluating adversarial attacks, as summarized in \autoref{tab:f1-gnn-bots-defense}. We follow the official train/validation/test as described by dataset authors~\citep{feng2021twibot20,feng2022twibot22,qiao2025botsim}.
%, and otherwise adopt the standard $60/20/20$ split used in prior work~\cite{feng2021botrgcn} on the Cresci dataset.

\heading{Experimental Setup.}
We evaluate \pname{} under two scenarios: \emph{node editing} and \emph{node injection}. In node editing, we uniformly sample $50$ correctly classified bot nodes, reset the graph to the original (unperturbed) graph before attacking each target, and allow both edge additions and deletions, but no human follow-back. 
This follows the standard protocol used in prior works (both Nettack and GOttack), but we use $50$ targets to improve statistical stability.
In node injection, we follow the same regime as editing, but we introduce a new bot into the graph and connect it either randomly (since there are no competitors) or using a bot cloak from \pname{}; only edge additions are allowed, and no human-follow-back edge can be created.
%cuneyt:we need to say that existing attack methods do not have node injection

\heading{Common constrained protocol.}
For the constrained columns, every attack receives the same per-target budget $\Delta$ and the same feasible set $\mathcal{F}(\Delta)$ from \autoref{subsec:problem-formulation}. Baseline attacks produce their ranked perturbation proposals using their original optimization procedures and dataset loaders adapted to our graphs. We then apply a deterministic feasibility wrapper: invalid edits are rejected, and the baseline continues down its own ranked proposal list until either $\Delta$ admissible target-incident edits are applied or the list is exhausted. We do not rerank, hand-pick, or add \pname-specific information to baseline proposals. For FGA and PR-BCD, we also evaluate in-loop projection controls that project candidate perturbations onto the bot$\rightarrow$human admissible edge set during optimization.

\heading{Hyper-parameters.}
Hyperparameters of bot detectors and their defense variants are provided in App. Tables~\ref{tab:hparams-defense} and \ref{tab:hparams-defense-variants}; \sota adversarial attack framework are provided in App. Table~\ref{tab:hparams-attack}. \pname{} uses 1-hop ego-neighborhoods, standard in other domains for capturing behavioral signals~\cite{dong2024pure, mukherjee2025zrex}. 

\heading{Computational Environment.}
Our implementation release: \url{https://github.com/kunmukh/bocloak}. Experiments were run on a Linux cluster with an AMD EPYC 7313 16-Core CPU, 1 TiB RAM, and 8 NVIDIA L40S 46 GB GPUs.

\heading{Evaluation Criteria.}
For a fixed black-box detector $f$, per-bot edge budget $\Delta$, and attack $\mathcal{A}$, we focus on a set of target bots $\mathcal{V}_t$ that are correctly classified on the social graph $G$, \ie $f(G,v_t)=\text{bot } 
\forall v_t\in\mathcal{V}_t$. 
In node editing, $v_t$ is an \emph{existing bot} 
% that was sampled from the correctly classified bots
and in node injection, $v_t$ is a \emph{new bot} to be injected by \pname.
The perturbed graph $\mathcal{G}'$ contains both edge additions/deletions for bot editing but only edge addition for bot injection.  
Given a target $v_t\in\mathcal{V}_t$, the attack $\mathcal{A}$ constructs a perturbed graph $\mathcal{G}'$ by applying at most $\Delta$ edge edits incident to $v_t$. $v_t$ is \emph{flipped} if $f(\mathcal{G}',v_t)=\text{human}$ while $f(\mathcal{G},v_t)=\text{bot}$, and define the indicator: $\code{BotFlip}(v_t)= \mathbb{I}\bigl[ f(\mathcal{G},v)=\text{bot} \ \wedge\ f(\mathcal{G}',v_t)=\text{human} \bigr]$. 
We use the misclassification rate: 
$\frac{1}{|\mathcal{V}_{\mathrm{t}}|}
  \sum_{v\in\mathcal{V}_{\mathrm{t}}} \texttt{BotFlip}(v),
$ as the primary performance metric for evaluation. Higher misclassification values indicate poorer detector performance and greater attack effectiveness. 
In all the results, the best performance is in bold, and the second best is underlined.

% \subsection{Attack Analysis}

\subsection{Node Editing Results}

We give the budget $\Delta=1$ node editing results against GAT and two \sota bot detectors in \autoref{tab:old-node-50-budget-1}. For results with all budgets and other bot detectors, see App.~\ref{tab:old-node-50-rest}, and for results with reuse of the same bot cloak, see App.~\ref{tab:old-node-50-samtemplate}. %For results without domain constraints, where any edge may be edited, see \autoref{tab:old-node-50-nodomainrules}.%cuneyt: we give the unconstrained results here

% \autoref{tab:old-node-50-budget-1} shows, \textit{unconstrained \sota attacks are effective, but once we enforce social-domain constraints, their performance drops drastically}. 
\autoref{tab:old-node-50-budget-1} shows that unconstrained \sota attacks can be effective, but their performance drops sharply once we enforce social-domain constraints. 
This drop is expected because generic graph attacks often rely on perturbations that are high-impact mathematically but infeasible operationally, such as rewiring unrelated users, violating edge directionality, or inducing arbitrary human$\rightarrow$bot follow-backs.

To verify that this gap is not simply caused by a weak post-hoc filtering implementation, we additionally evaluate constrained control variants in Appendix~\ref{app:constraint-controls}. 
In particular, FGA-mod and PR-BCD-mod add an in-loop projection step onto the admissible bot$\rightarrow$human edit set, LR-BCD adds a newer local-budget attack baseline, and Homophily replaces OT-guided coupling with embedding/homophily-based matching. 
As shown in \autoref{tab:constraint-controls}, these controls improve some constrained baselines only modestly and remain far below \pname, indicating that the performance gap is driven by the mismatch between generic graph perturbation objectives and domain-faithful social-bot constraints.
We further test NoisyGNN as an additional robustness-control defense on TwiBot-22/BotRGCN; results are reported in App.~\autoref{tab:noisygnn-results}. NoisyGNN reduces constrained generic attacks to below $6\%$ misclassification, but \pname{} still achieves $85.12\%$, showing that hidden-state noise alone does not remove the domain-feasible neighborhood vulnerability exploited by \pname{}.

Within the constrained setting, increasing budget improves \sota only modestly (\eg on Cresci-15 their rates rise from 3--9\% at $\Delta{=}1$ to 12--23\% at $\Delta{=}5$), but they remain far below \pname. By contrast, \pname consistently achieves \textbf{80.13\%} higher success attack rates than the best-performing constrained \sota{}.
%MK: I wonder whether we want to add a sentence on how we evaluated these SOTA techniques under constraints before ??
% Within the constrained setting, increasing budget improves \sota only modestly (\eg on Cresci-15 their rates rise from 3--9\% at $\Delta{=}1$ to 12--23\% at $\Delta{=}5$), but they remain far below \pname. By contrast, \pname consistently achieves \textbf{80.13\%} higher success attack rates than the best-performing constrained \sota{}.

An insight is that detector choice matters more for \sota attacks than for \pname: \sota can reach the low-20\% range on BotRGCN for some settings at $\Delta{=}5$, but still fails to approach \pname, suggesting \pname learns transferable ``bot cloaks'' rather than brittle, model-specific perturbations. Standard defenses only weakly affect \pname{} on Cresci-15 and TwiBot-22, with scores clustering tightly across vanilla and defended detectors, whereas constrained \sota attacks stay uniformly low because the feasible edit space removes many of their exploitable directions.
\pname{} is consistently the strongest attack model and remains highly effective even when constraints are active.%, achieving high attack success on Cresci-15 and TwiBot-22 (roughly $\sim$80--99\%+ across detectors/defenses). %BotSim-24 is the most challenging dataset where \pname{} is notably lower at small budgets (e.g., $\sim58\%$ at $\Delta{=}1$ for BotRGCN) and exhibits larger drops under defenses than other datasets, indicating that the dataset is comparatively more resistant to feasible, domain-valid cloaking.

\begin{figure}[!htb] 
\centering 
\resizebox{0.90\columnwidth}{!}{% 
    \includegraphics{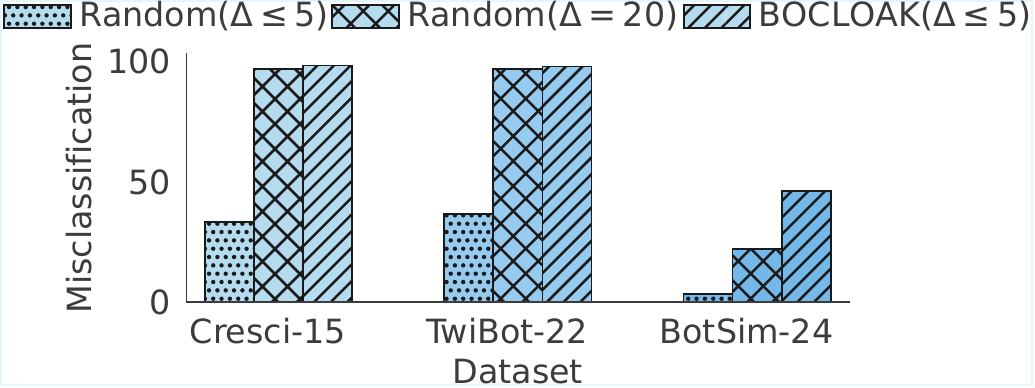}% 
} 
\caption{\textbf{Node Injection Attack:} Misclassification rate ($\uparrow$, in \%) for fifty target nodes by \pname against best \sota bot detector, BotRGCN. Extended results in App. \autoref{tab:new-node-1} and \autoref{tab:new-node-def-rest}.}
\label{fig:node-inject-plot} 
\end{figure}

\heading{Node Injection.}
Node injection results are shown in \autoref{fig:node-inject-plot}. For extended results, see App.~\ref{tab:new-node-1} and App.~\ref{tab:new-node-def-rest}, and for results when the same bot cloak is reused, see App.~\ref{tab:new-node-1-samtemplate} and App.~\ref{tab:new-node-def-samtemplate}. 
%MK: what is the bocloak budget in figure 4. please specify ?? Cuneyt: yes what is that? KM budget=5 figs updated as well
% For node injection, we compare \pname{} against a \emph{random} procedure with the domain constraints.
On Cresci-15 and TwiBot-22, \pname{} achieves near-perfect injection success with very small budgets (high-90\% with 1-3 edges), and additional edges yield only marginal gains, suggesting the decision boundary is easy to cross once a minimally human-like ego-neighbourhood is formed. %In the large-budget procedure, the gap to random largely vanishes: with a 20-edge budget, random injection reaches similarly high misclassification (mid-90\%), so \pname{} is at par with random when the attacker can connect broadly. cuneyt: this would work against us.

%For the BotSim-24 dataset, the success is low at tiny budgets ($14\%$ at $B{=}1$) and rises steadily into the high-60\% range by $B{=}20$, indicating a sharper, harder-to-hit margin where a few arbitrary bot cloaks are insufficient. Consistently, random remains weak on BotSim-24 even at $B{=}20$ (still around the mid-teens), highlighting that OT-guided template selection contributes most when the boundary is sparse and naive linking rarely lands in the ``human manifold.'' 
Overall, injection robustness can look deceptively ``solved'' on easier benchmarks (where even random succeeds at high budget), whereas \pname{}'s OT geometry- and plausibility- aware construction is most revealing on harder datasets like BotSim-24 where random search fails and success scales smoothly with budget.

\input{tables/ablation-small}
\heading{Ablation Studies.}
\autoref{tab:ablation-small} (extended in App.~\ref{tab:ablation}) shows that the OT-margin BCE surrogate is the primary contributor: removing it leads to a pronounced success drop, confirming its role in enforcing a separable OT margin that keeps misclassified boundary cloaks close to the human region.
% The plausibility term is the next most impactful; removing $\lambda_{\mathrm{pl}}$ yields moderate success because it prevents unrealistic transport alignments (\eg matching neighbors with large degree/age gaps) that do not translate into effective edits. Within plausibility, degree consistency dominates temporal calibration: disabling $\alpha_{\deg}$ degrades more than disabling $\alpha_{\mathrm{age}}$, indicating that preserving local connectivity is the stronger constraint for realistic cloak transfer.

% \begin{figure}[!htb]
%     \centering
%     \resizebox{0.80\columnwidth}{!}{%
%         \includegraphics{figs/sensitivity.pdf}%
%     }
%     \caption{\textbf{Sensitivity to OT regularizer ($\varepsilon$).} Misclassification rate (in \%) for flipping fifty existing correctly classified bots as the OT regularizer ($\varepsilon$) is varied, impacting the human-bot decision boundary on vanilla BotRGCN. For extended results refer to \autoref{tab:sensitivity}}
%     \label{fig:sensitivity-small}
% \end{figure}

\begin{figure}[!htb]
    \centering
    % \begin{subfigure}[t]{\columnwidth}
    %     \centering
    %     \resizebox{0.80\columnwidth}{!}{%
    %         \includegraphics{figs/sensitivity.pdf}%
    %     }
    %     \caption{OT regularizer ($\varepsilon$).}
    %     \label{fig:eps}
    % \end{subfigure}
    % \begin{subfigure}[t]{\columnwidth}
    %     \centering
    %     \resizebox{0.80\columnwidth}{!}{%
    %         \includegraphics{figs/sensitivity2.pdf}%
    %     }
    %     \caption{BCE threshold ($\tau_{\text{bdry}}$).}
    %     \label{fig:tau}
    % \end{subfigure}
    \centering 
    \resizebox{0.95\columnwidth}{!}{% 
        \includegraphics{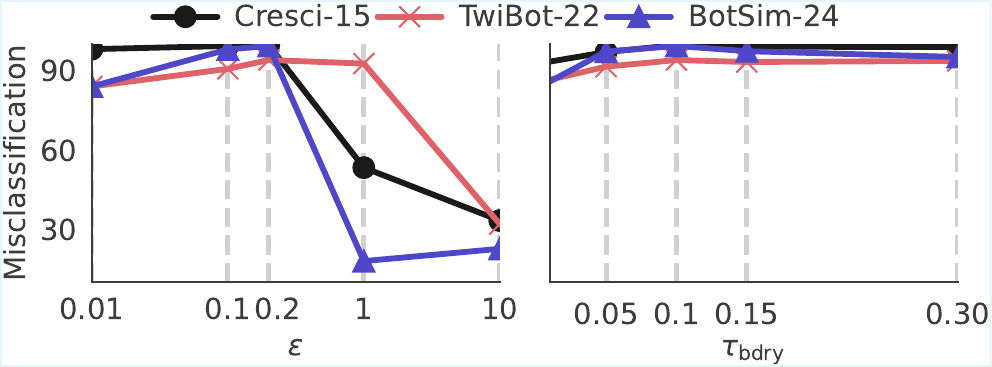}% 
    } 

    \caption{\textbf{Sensitivity Analysis of OT regularizer ($\varepsilon$) and margin threshold ($\tau_{\mathrm{bdry}}$)}
    Misclassification rate ($\uparrow$, in \%) for flipping fifty existing correctly classified bots as $\varepsilon$ and $\tau_{\mathrm{bdry}}$ is varied. 
    % $\varepsilon$ impacting the human-bot decision boundary on vanilla BotRGCN with budget $\Delta$ of 5 and $\tau_{\text{bdry}}$ controls the strength of the OT margin constraint required for a successful cloak.
    Extended results in App.~\ref{tab:sensitivity}.}
    \label{fig:sensitivity-small}
\end{figure}

\heading{Sensitivity Analysis.}
\autoref{fig:sensitivity-small} illustrates that \pname{} is most effective at intermediate OT regularization and margin threshold. With very small $\varepsilon$ (e.g., $0.01$), transport plans become sharp and unstable, leading to higher variance and moderate flips (about $83$–$85\%$), and large ($\ge 10$), the plan becomes diffuse, blurring the OT-induced margin. 
% In contrast, moderate $\varepsilon$ preserves multiple plausible correspondences while maintaining margin structure, yielding the strongest and most consistent boundary crossings ($98$–$100\%$). When $\varepsilon$ is large ($\ge 10$), the plan becomes overly diffuse, blurring the OT-induced margin and sharply reducing flip rates (\eg 53\% to 14\% on Cresci-15 as $\varepsilon$ increases from $1$ to $100$). 
Varying the margin threshold $\tau_{\mathrm{bdry}}$ shows a similar optimal value: around $0.10$, where it maximizes flip rates by enforcing a meaningful OT-margin without over-constraining the search.
% Overly large $\tau_{\text{bdry}}$ (\eg $\ge 0.30$) can cause performance to plateau or drop (notably on TwiBot-22 and BotSim-24), consistent with a stricter margin requirement that leaves many trials persistently penalized and increases run-to-run variance.

\input{tables/time-gpu-small}
\heading{Overhead Analysis.}
\autoref{tab:time-gpu-small} describes the resource usage. \pname{} incurs \textbf{99.80\%} less GPU footprint (13MB) because it does not require GPU computation and runs up to \textbf{$20\times$} faster than \sota attacks, extended in App \ref{tab:time-gpu}.

\heading{Limitation.}
We assume fixed feature vectors throughout the study. In practice, social network content, such as tweets, changes over time and can affect both detection and evasion. However, modeling these effects requires temporally resolved content and metadata, which are not yet available.

%% file: tables/degree-short.tex
\begin{table}[!htb]
\centering
\resizebox{0.90\columnwidth}{!}{%
\begin{tabular}{lcccc}
\toprule
\textbf{Dataset} 
& \textbf{\#Nodes} 
& \textbf{\#Edges} 
& \textbf{\#Bots} 
& \textbf{\#Humans} \\
\midrule
Cresci-15   & 5{,}301       & 14{,}220        & 3{,}351   & 1{,}950 \\
TwiBot-22   & 1{,}000{,}000 & 170{,}185{,}937 & 139{,}943 & 860{,}057 \\
BotSim-24   & 2{,}907       & 46{,}518        & 1{,}000   & 1{,}907 \\
\bottomrule
\end{tabular}%
}
\caption{\textbf{Dataset Statistics:} Summary of bot detection datasets. Detailed statistics are in Appendix ~\autoref{tab:avg-node-degree}.}
\label{tab:avg-node-degree-short}
\end{table}

%% file: tables/old-node-50.tex
\begin{table*}[!htb]
\centering
\setlength{\tabcolsep}{2.2pt}
\renewcommand{\arraystretch}{1.05}

\resizebox{\textwidth}{!}{%
\begin{tabular}{ll ccccc ccccc ccccc}
\toprule
\multirow{3}{*}{\textbf{Dataset}} &
\multirow{3}{*}{\makecell[c]{\textbf{Attack}}} &
\multicolumn{5}{c}{\textbf{GAT}} &
\multicolumn{5}{c}{\textbf{BotRGCN}} &
\multicolumn{5}{c}{\textbf{S-HGN}} \\
\cmidrule(lr){3-7}
\cmidrule(lr){8-12}
\cmidrule(lr){13-17}
& & \textbf{Unconstrained} & \multicolumn{4}{c}{\textbf{Constrained}} & \textbf{Unconstrained} & \multicolumn{4}{c}{\textbf{Constrained}} & \textbf{Unconstrained} & \multicolumn{4}{c}{\textbf{Constrained}} \\
\cmidrule(lr){4-7}
\cmidrule(lr){9-12}
\cmidrule(lr){14-17}
&
& \textbf{Vanilla} & \textbf{Vanilla}  & \textbf{+GNNGuard}  & \textbf{+GRAND}  & \textbf{+RobustGCN}
& \textbf{Vanilla} & \textbf{Vanilla}  & \textbf{+GNNGuard}  & \textbf{+GRAND}  & \textbf{+RobustGCN}
& \textbf{Vanilla} & \textbf{Vanilla}  & \textbf{+GNNGuard}  & \textbf{+GRAND}  & \textbf{+RobustGCN} \\
\midrule

% -------- Cresci-15 --------
\multirow{5}{*}{Cresci-15}
& Nettack
& \bluehl{95.45\scriptsize{$\pm$3.22}} & 8.17\scriptsize{$\pm$3.23} & 6.59\scriptsize{$\pm$3.40} & \underline{7.54\scriptsize{$\pm$2.07}} & 7.28\scriptsize{$\pm$3.19}
& 94.67\scriptsize{$\pm$6.55} & \underline{4.32\scriptsize{$\pm$2.10}} & \underline{3.78\scriptsize{$\pm$1.95}} & \underline{3.21\scriptsize{$\pm$1.88}} & \underline{2.67\scriptsize{$\pm$1.72}}
& \underline{94.11\scriptsize{$\pm$2.45}} & 6.04\scriptsize{$\pm$3.00} & 4.81\scriptsize{$\pm$3.18} & 4.51\scriptsize{$\pm$3.60} & 4.88\scriptsize{$\pm$2.71} \\
& FGA
& 94.75\scriptsize{$\pm$1.50} & 8.52\scriptsize{$\pm$1.92} & 5.74\scriptsize{$\pm$2.39} & 4.73\scriptsize{$\pm$3.67} & \underline{7.93\scriptsize{$\pm$2.91}}
& 96.89\scriptsize{$\pm$4.81} & 3.95\scriptsize{$\pm$2.44} & 3.40\scriptsize{$\pm$2.12} & 2.98\scriptsize{$\pm$1.76} & 2.15\scriptsize{$\pm$1.60}
& 92.54\scriptsize{$\pm$1.23} & 5.22\scriptsize{$\pm$3.17} & 4.96\scriptsize{$\pm$3.09} & 4.31\scriptsize{$\pm$2.07} & 3.44\scriptsize{$\pm$2.01} \\
& PR-BCD
& 95.02\scriptsize{$\pm$1.38} & \underline{8.79\scriptsize{$\pm$1.91}} & 6.85\scriptsize{$\pm$2.35} & 7.48\scriptsize{$\pm$2.37} & 7.92\scriptsize{$\pm$3.24}
& 96.36\scriptsize{$\pm$8.36} & 4.10\scriptsize{$\pm$3.05} & 3.60\scriptsize{$\pm$2.80} & 2.85\scriptsize{$\pm$2.20} & 1.90\scriptsize{$\pm$1.45}
& 94.12\scriptsize{$\pm$1.23} & 7.57\scriptsize{$\pm$3.84} & 6.91\scriptsize{$\pm$2.43} & 6.70\scriptsize{$\pm$2.84} & 5.58\scriptsize{$\pm$2.45} \\
& GOttack
& 94.98\scriptsize{$\pm$5.11} & 7.61\scriptsize{$\pm$2.91} & \underline{7.14\scriptsize{$\pm$3.65}} & 5.08\scriptsize{$\pm$3.06} & 6.20\scriptsize{$\pm$1.94}
& \underline{98.54\scriptsize{$\pm$12.40}} & 3.55\scriptsize{$\pm$2.34} & 3.41\scriptsize{$\pm$2.52} & 2.28\scriptsize{$\pm$1.22} & 2.66\scriptsize{$\pm$1.67}
& 94.12\scriptsize{$\pm$1.21} & \underline{8.10\scriptsize{$\pm$2.83}} & \underline{7.57\scriptsize{$\pm$2.04}} & \underline{7.92\scriptsize{$\pm$1.87}} & \underline{6.39\scriptsize{$\pm$2.12}} \\
& \textbf{\pname{}} (ours)
& \underline{95.25\scriptsize{$\pm$1.11}} & \bluehl{93.10\scriptsize{$\pm$2.34}} & \bluehl{94.20\scriptsize{$\pm$1.89}} & \bluehl{92.45\scriptsize{$\pm$3.11}} & \bluehl{91.88\scriptsize{$\pm$4.56}}
& \bluehl{99.34\scriptsize{$\pm$0.29}} & \bluehl{99.34\scriptsize{$\pm$0.29}} & \bluehl{98.91\scriptsize{$\pm$0.54}} & \bluehl{99.06\scriptsize{$\pm$0.41}} & \bluehl{99.72\scriptsize{$\pm$0.18}}
& \bluehl{95.80\scriptsize{$\pm$2.50}} & \bluehl{92.18\scriptsize{$\pm$4.73}} & \bluehl{91.60\scriptsize{$\pm$2.15}} & \bluehl{91.92\scriptsize{$\pm$3.08}} & \bluehl{91.40\scriptsize{$\pm$1.67}} \\
\midrule

% -------- TwiBot-22 --------
\multirow{5}{*}{TwiBot-22}
& Nettack
& 81.22\scriptsize{$\pm$1.23} & 13.51\scriptsize{$\pm$8.15} & 11.92\scriptsize{$\pm$9.38} & 8.59\scriptsize{$\pm$7.73} & \underline{12.91\scriptsize{$\pm$8.91}}
& 79.77\scriptsize{$\pm$6.77} & \underline{9.33\scriptsize{$\pm$3.06}} & \underline{8.67\scriptsize{$\pm$4.62}} & 4.67\scriptsize{$\pm$5.03} & \underline{8.00\scriptsize{$\pm$5.29}}
& 83.51\scriptsize{$\pm$1.22} & 5.03\scriptsize{$\pm$4.72} & 3.24\scriptsize{$\pm$9.20} & 3.72\scriptsize{$\pm$10.81} & 4.73\scriptsize{$\pm$7.85} \\
& FGA
& 73.75\scriptsize{$\pm$1.22} & 12.32\scriptsize{$\pm$8.10} & 8.59\scriptsize{$\pm$4.52} & 9.94\scriptsize{$\pm$6.41} & 5.57\scriptsize{$\pm$10.78}
& 71.90\scriptsize{$\pm$4.23} & 4.00\scriptsize{$\pm$2.00} & 7.33\scriptsize{$\pm$2.31} & \underline{6.67\scriptsize{$\pm$4.16}} & 1.33\scriptsize{$\pm$6.43}
& 75.85\scriptsize{$\pm$2.54} & \underline{5.79\scriptsize{$\pm$6.52}} & \underline{4.57\scriptsize{$\pm$4.96}} & 3.41\scriptsize{$\pm$6.87} & 3.19\scriptsize{$\pm$9.89} \\
& PR-BCD
& 73.86\scriptsize{$\pm$1.50} & \underline{13.51\scriptsize{$\pm$4.52}} & \underline{12.52\scriptsize{$\pm$9.38}} & \underline{11.32\scriptsize{$\pm$6.41}} & 12.24\scriptsize{$\pm$4.52}
& 75.77\scriptsize{$\pm$6.05} & 8.67\scriptsize{$\pm$2.31} & 8.67\scriptsize{$\pm$5.03} & 6.23\scriptsize{$\pm$4.16} & 7.33\scriptsize{$\pm$2.31}
& 78.12\scriptsize{$\pm$1.98} & 4.58\scriptsize{$\pm$5.29} & 3.74\scriptsize{$\pm$3.34} & 3.61\scriptsize{$\pm$7.20} & 3.20\scriptsize{$\pm$7.92} \\
& GOttack
& \underline{82.12\scriptsize{$\pm$2.56}} & 11.41\scriptsize{$\pm$7.59} & 10.50\scriptsize{$\pm$6.38} & 9.78\scriptsize{$\pm$3.93} & 10.33\scriptsize{$\pm$5.34}
& \underline{84.29\scriptsize{$\pm$11.63}} & 6.89\scriptsize{$\pm$3.33} & 7.15\scriptsize{$\pm$4.25} & 6.01\scriptsize{$\pm$2.48} & 5.32\scriptsize{$\pm$2.58}
& 80.85\scriptsize{$\pm$2.30} & 5.55\scriptsize{$\pm$7.98} & 4.04\scriptsize{$\pm$8.93} & \underline{4.44\scriptsize{$\pm$7.82}} & \underline{4.95\scriptsize{$\pm$8.39}} \\
& \textbf{\pname{}} (ours)
& \bluehl{95.25\scriptsize{$\pm$2.50}} & \bluehl{94.38\scriptsize{$\pm$3.41}} & \bluehl{93.72\scriptsize{$\pm$1.66}} & \bluehl{93.90\scriptsize{$\pm$4.28}} & \bluehl{95.12\scriptsize{$\pm$3.33}}
& \bluehl{86.67\scriptsize{$\pm$2.31}} & \bluehl{86.67\scriptsize{$\pm$2.31}} & \bluehl{84.67\scriptsize{$\pm$11.02}} & \bluehl{83.67\scriptsize{$\pm$3.06}} & \bluehl{87.33\scriptsize{$\pm$1.15}}
& \bluehl{98.12\scriptsize{$\pm$1.75}} & \bluehl{97.42\scriptsize{$\pm$1.39}} & \bluehl{96.15\scriptsize{$\pm$3.74}} & \bluehl{95.90\scriptsize{$\pm$2.06}} & \bluehl{95.11\scriptsize{$\pm$4.28}} \\
\midrule

% -------- BotSim-24 --------
\multirow{5}{*}{BotSim-24}
& Nettack
& 53.77\scriptsize{$\pm$4.12} & 5.52\scriptsize{$\pm$1.92} & 3.74\scriptsize{$\pm$2.39} & 4.73\scriptsize{$\pm$3.67} & 4.93\scriptsize{$\pm$8.10}
& 56.57\scriptsize{$\pm$6.63} & 0.00\scriptsize{$\pm$0.00} & 0.00\scriptsize{$\pm$0.00} & 0.00\scriptsize{$\pm$0.00} & \underline{3.00\scriptsize{$\pm$2.00}}
& 55.54\scriptsize{$\pm$5.55} & 4.71\scriptsize{$\pm$2.19} & 3.44\scriptsize{$\pm$3.09} & 3.35\scriptsize{$\pm$2.67} & 3.65\scriptsize{$\pm$7.44} \\
& FGA
& 52.09\scriptsize{$\pm$1.59} & 5.79\scriptsize{$\pm$1.91} & 4.85\scriptsize{$\pm$2.35} & 4.48\scriptsize{$\pm$2.37} & 4.92\scriptsize{$\pm$3.24}
& 55.67\scriptsize{$\pm$6.96} & 0.00\scriptsize{$\pm$0.00} & 0.00\scriptsize{$\pm$0.00} & 54.98\scriptsize{$\pm$1.33} & 0.00\scriptsize{$\pm$0.00}
& 50.41\scriptsize{$\pm$2.50} & \underline{5.72\scriptsize{$\pm$2.56}} & \underline{4.63\scriptsize{$\pm$2.12}} & \underline{5.06\scriptsize{$\pm$3.15}} & \underline{4.54\scriptsize{$\pm$3.73}} \\
& PR-BCD
& 51.26\scriptsize{$\pm$4.11} & \underline{7.32\scriptsize{$\pm$8.10}} & \underline{6.59\scriptsize{$\pm$4.52}} & 6.65\scriptsize{$\pm$7.59} & \underline{7.14\scriptsize{$\pm$3.65}}
& 50.62\scriptsize{$\pm$8.04} & \underline{2.00\scriptsize{$\pm$2.00}} & \underline{5.33\scriptsize{$\pm$2.31}} & \underline{0.67\scriptsize{$\pm$1.15}} & 0.00\scriptsize{$\pm$0.00}
& 52.11\scriptsize{$\pm$2.50} & 5.20\scriptsize{$\pm$8.04} & 4.24\scriptsize{$\pm$4.63} & 4.43\scriptsize{$\pm$4.99} & 4.26\scriptsize{$\pm$1.88} \\
& GOttack
& \bluehl{91.50\scriptsize{$\pm$1.50}} & 7.08\scriptsize{$\pm$3.06} & 6.20\scriptsize{$\pm$1.94} & \underline{6.67\scriptsize{$\pm$3.09}} & 5.87\scriptsize{$\pm$2.27}
& \underline{57.18\scriptsize{$\pm$2.32}} & 0.00\scriptsize{$\pm$0.00} & 0.00\scriptsize{$\pm$0.00} & 0.00\scriptsize{$\pm$0.00} & 0.00\scriptsize{$\pm$0.00}
& \underline{88.33\scriptsize{$\pm$1.33}} & 4.42\scriptsize{$\pm$1.83} & 3.25\scriptsize{$\pm$3.38} & 3.65\scriptsize{$\pm$3.12} & 3.21\scriptsize{$\pm$2.45} \\
& \textbf{\pname{}} (ours)
& \underline{90.19\scriptsize{$\pm$1.50}} & \bluehl{88.74\scriptsize{$\pm$3.12}} & \bluehl{90.05\scriptsize{$\pm$1.47}} & \bluehl{89.30\scriptsize{$\pm$4.66}} & \bluehl{88.40\scriptsize{$\pm$2.29}}
& \bluehl{58.22\scriptsize{$\pm$7.20}} & \bluehl{58.22\scriptsize{$\pm$7.20}} & \bluehl{52.68\scriptsize{$\pm$1.15}} & \bluehl{54.98\scriptsize{$\pm$2.45}} & \bluehl{55.10\scriptsize{$\pm$3.98}}
& \bluehl{90.75\scriptsize{$\pm$2.33}} & \bluehl{89.16\scriptsize{$\pm$4.25}} & \bluehl{90.44\scriptsize{$\pm$2.11}} & \bluehl{89.78\scriptsize{$\pm$1.63}} & \bluehl{88.90\scriptsize{$\pm$3.74}} \\
\bottomrule
\end{tabular}%
}

\caption{\textbf{Node Editing Attack with Budget $\Delta=1$:} Misclassification rate ($\uparrow$, in \%) for flipping fifty correctly classified bots by \pname{} and SOTA adversarial attacks against best SOTA bot detectors: GAT, BotRGCN, and S-HGN, with adversarial defenses. For results against other bot detectors, see App. \ref{tab:old-node-50-rest}, see App. \ref{tab:old-node-50-samtemplate} when the same bot cloak is reused, see App. \ref{tab:old-node-50-nodomainrules} when no domain constraints are enforced. The best performance is shown in bold, and the second best is underlined.}
\label{tab:old-node-50-budget-1}
\end{table*}

%% file: tables/ablation-small.tex
\begin{table}[!htb]
\centering
\resizebox{1.\columnwidth}{!}{%
\begin{tabular}{lcccccc}
\toprule
\textbf{Dataset} 
& $-\lambda_{\mathrm{BCE}}$
& $-\lambda_{\mathrm{sp}}$
& $-\lambda_{\mathrm{pl}}$
& $-\alpha_{\deg}$
& $-\alpha_{\mathrm{age}}$
& \textbf{\pname{}} \\
\midrule
\textbf{Cresci-17} 
& 38.21\scriptsize{$\pm$5.12}
& 96.88\scriptsize{$\pm$5.89}
& 57.99\scriptsize{$\pm$3.38}
& 73.23\scriptsize{$\pm$0.23}
& 87.23\scriptsize{$\pm$1.78}
& \bluehl{99.67\scriptsize{$\pm$0.20}} \\
\textbf{TwiBot-22} 
& 22.45\scriptsize{$\pm$6.44}
& 92.12\scriptsize{$\pm$7.65}
& 55.23\scriptsize{$\pm$6.77}
& 70.98\scriptsize{$\pm$4.29}
& 83.88\scriptsize{$\pm$6.63}
& \bluehl{94.00\scriptsize{$\pm$2.00}} \\
\textbf{BotSim-24} 
& 32.45\scriptsize{$\pm$5.64}
& 96.55\scriptsize{$\pm$0.66}
& 60.34\scriptsize{$\pm$5.09}
& 69.22\scriptsize{$\pm$2.15}
& 85.24\scriptsize{$\pm$1.38}
& \bluehl{99.33\scriptsize{$\pm$1.15}} \\
\bottomrule
\end{tabular}%
}
\caption{\textbf{Ablation Study:} Misclassification rate ($\uparrow$, in \%) for node editing attack against vanilla BotRGCN with budget $\Delta$ of $5$. Extended result in App. \ref{tab:ablation}. }
\label{tab:ablation-small}
\end{table}

%% file: tables/time-gpu-small.tex
\begin{table}[!htb]
\centering
\resizebox{0.7\columnwidth}{!}{%
\begin{tabular}{lccc}
\toprule
\multirow{2}{*}{\textbf{Attack}}
& \multicolumn{3}{c}{\textbf{BotSim-24}} \\
\cmidrule(lr){2-4}
& \textbf{Time (sec)} ($\downarrow$)
& \textbf{RAM (MB)} ($\downarrow$)
& \textbf{GPU (MB)} ($\downarrow$) \\
\midrule
% Random   & 0.18\scriptsize{$\pm$0.02} & 8237\scriptsize{$\pm$11} & 20812\scriptsize{$\pm$2} \\
% Nettack  & 1.4\scriptsize{$\pm$1.33} & 21102\scriptsize{$\pm$12133} & \underline{5982\scriptsize{$\pm$4}} \\
% FGA      & \underline{1.1\scriptsize{$\pm$0.06}} & \underline{9036\scriptsize{$\pm$72}} & 9222\scriptsize{$\pm$28} \\
% PR-BCD   & 2.9\scriptsize{$\pm$0.47} & 9037\scriptsize{$\pm$72} & 6178\scriptsize{$\pm$4} \\
% GOttack  & 853.9\scriptsize{$\pm$19.95} & 10121\scriptsize{$\pm$214} & 5982\scriptsize{$\pm$15} \\
% \textbf{\pname{}} & \bluehl{0.5\scriptsize{$\pm$0.14}} & \bluehl{7974\scriptsize{$\pm$25}} & \bluehl{13\scriptsize{$\pm$2}} \\
Nettack  & 19.7\scriptsize{$\pm$1.65} & 2973\scriptsize{$\pm$628} & \underline{156\scriptsize{$\pm$40}} \\
FGA      & \underline{0.3\scriptsize{$\pm$1.57}} & \underline{1720\scriptsize{$\pm$13}} & 286\scriptsize{$\pm$2} \\
PR-BCD   & 2.1\scriptsize{$\pm$0.75} & 1857\scriptsize{$\pm$21} & 286\scriptsize{$\pm$1} \\
GOttack  & 7.5\scriptsize{$\pm$1.75} & 2069\scriptsize{$\pm$623} & 240\scriptsize{$\pm$2} \\
\textbf{\pname{}} & \bluehl{0.1\scriptsize{$\pm$0.13}} & \bluehl{1450\scriptsize{$\pm$2}} & \bluehl{11\scriptsize{$\pm$1}} \\
\bottomrule
\end{tabular}%
}
\caption{\textbf{System Overhead:} System resource usage by \pname{} and \sota adversarial attacks against vanilla BotRGCN with budget $\Delta$ of 1. Extended result in App.~\ref{tab:time-gpu}.}
\label{tab:time-gpu-small}
\end{table}

%% file: sections/conc.tex
\section{Conclusion}

We introduced \pname, which constructs realistic $k$-hop neighbourhood under directional, social, and temporal constraints. \pname is a geometry-first attack framework that learns a label-aware optimal transport ground cost that explicitly separates human and bot behavioral regions. Across three datasets, five victim detectors with three defenses, and compared against four \sota adversarial attacks, \pname consistently achieves 80.13\% higher evasion success under realistic constraints and using 99.80\% less GPU memory. This work closes a key realism gap in the robustness evaluation of social bot detectors. 

%% file: sections/icml.tex
\section*{Acknowledgements}

We thank the anonymous reviewers for their helpful feedback. The research reported here in were supported in part by NSF awards DMS-2204795, OAC-2115094, CNS-2331424, ITE-2452833, ARL/Army Research Office awards W911NF-24-1-0202 and W911NF-24-2-0114, and Virginia Commonwealth Cyber Initiative grants.

\section*{Impact Statement}

This paper presents research whose goal is to advance the field of machine learning, with a focus on understanding the robustness and limitations of graph-based bot detection systems. Our work studies constrained adversarial attacks in order to characterize failure modes of existing detectors and to motivate the design of more reliable and resilient defenses.

While the techniques introduced here could, in principle, be misused to evade deployed detection systems, the intent of this work is diagnostic rather than operational. By revealing how bot accounts can exploit neighborhood-level vulnerabilities, our results provide insights that can help platform designers and researchers strengthen detection models and improve trustworthiness. We do not release tools for large-scale misuse, and the attacks studied operate under restrictive assumptions intended to mirror realistic constraints.

Overall, we believe the primary societal impact of this work is to contribute to safer and more robust machine learning systems for online platforms. We do not foresee immediate negative societal consequences beyond those already well understood in the study of adversarial machine learning.

% \heading{LLM Usage.}
% We acknowledge the use of LLMs to help improve the writing of this article. All content, ideas, and results are our own. The LLM helped improve clarity, grammar, style, and LaTeX formatting.

%% file: sections/appendix.tex
\appendix

\input{tables/notation}

\section{Appendix}

\section{Summary of Notations Used in This Paper}

Table~\ref{tab:compact_notations} summarizes the key notations used throughout the paper, covering graph structure, neighborhood representations, optimal transport quantities, and attack-specific objects.

% \section{\pname Philosophy.}
% \pname is a geometry-first method that treats local neighborhoods, rather than individual edges, as first-class objects of adversarial manipulation. Each node is represented by a probability measure over its ego-neighborhood in a spatio-temporal feature space, and optimal transport compares these distributions under a learned ground cost. This induces a metric geometry in which bots and humans occupy distinct regions, with misclassified bots near the decision boundary. \pname selects such boundary bots as cloak templates to derive sparse, feasible edge edits by decoding transport plans.

% \section{Steps}

\section{Optimal Transport \pname-specific Extended Discussion}

In this section, we present a detailed discussion of the Optimal Transport–based components of \pname, including formulation, and algorithmic steps.

\subsection{Step 1: Spatio-Temporal Feature Extraction}
For each node $v$, let $\mathcal{N}(v)$ denote its $1$-hop neighborhood (all in- or out-neighbors that share at least one directed edge with $v$). For every neighbor $u \in \mathcal{N}(v)$, we build a structural and temporal feature vector $\phi_v(u) \in \mathbb{R}^d$, that captures node type, relational properties with respect to $v$, degree information, and temporal behavior.

Concretely, we organize $\phi_v(u)$ into four feature types: \textbf{static}, \textbf{behavioral}, \textbf{content}, and \textbf{temporal}:

\begin{itemize}[noitemsep,leftmargin=*,topsep=0pt]
    \item \textbf{Static features} $\mathbf{X}_{\mathrm{stat}}(u)$ (intrinsic, time-invariant or slowly varying):
    \(
        t(u)\in\{0,1\}
    \)
    (human/bot indicator) and degree statistics
    \(
        \deg_{\mathrm{in}}(u),\deg_{\mathrm{out}}(u)
    \)
    (raw and/or normalized).

    \item \textbf{Behavioral features} $\mathbf{X}_{\mathrm{beh}}(u;v)$ (how $u$ interacts with $v$ in the ego graph):
    the directed edge role w.r.t.\ $v$, \break
       $ r_v(u)\in\{0,1,2\}=
        \begin{cases}
            1 & \text{if } u \to v \text{ (follower)}, \\
            2 & \text{if } v \to u \text{ (followee)}, \\
            0 & \text{if } u \leftrightarrow v \text{ (mutual follow)}.
        \end{cases}$ 
        %additional interaction and activity counts (\eg retweets, shares) appended here if available.

    \item \textbf{Content features} $\mathbf{X}_{\mathrm{cont}}(u)$ (what the account posts or describes):
    a fixed-length content vector (\eg username, user bio, text/profile embeddings and content-derived attributes),
    denoted $\mathrm{cont\_feat}(u)$.

    \item \textbf{Temporal features} $\mathbf{X}_{\mathrm{temp}}(u;v)$ (when the account exists/acts):
    normalized account age $\mathrm{age\_norm}(u)\in[0,1]$ and relative age
    \(
        \mathrm{age\_diff\_norm}(u;v)=\mathrm{age\_norm}(u)-\mathrm{age\_norm}(v)
    \),
    capturing temporal alignment between $u$ and $v$.
\end{itemize}

We collect these into a single spatio-temporal neighbor vector
\begin{equation}
\label{eq:feature-vector}
\phi_v(u)
=
\bigl[
\underbrace{\mathbf{X}_{\mathrm{stat}}(u)}_{\text{Static}},
\underbrace{\mathbf{X}_{\mathrm{beh}}(u;v)}_{\text{Behavioral}},
\underbrace{\mathbf{X}_{\mathrm{cont}}(u)}_{\text{Content}},
\underbrace{\mathbf{X}_{\mathrm{temp}}(u;v)}_{\text{Temporal}}
\bigr]\in\mathbb{R}^d,
\end{equation}
where, in our implementation,
$\mathbf{X}_{\mathrm{stat}}(u)=[t(u),\deg_{\mathrm{in}}(u),\deg_{\mathrm{out}}(u)]$,
$\mathbf{X}_{\mathrm{beh}}(u;v)=[r_v(u)]$,
$\mathbf{X}_{\mathrm{cont}}(u)=[\mathrm{cont\_feat}(u)]$,
and
$\mathbf{X}_{\mathrm{temp}}(u;v)=[\mathrm{age\_norm}(u),\mathrm{age\_diff\_norm}(u;v)]$.
In this spatio-temporal feature space, neighbors with similar labels, structural roles, degrees, and temporal behavior lie close to each other.

%\subsubsection{Step 2: From Neighbor Features to Local Neighborhood Probability Measures}
\subsection{Step 2: Local Neighborhood Probability Measures}\label{sec:prob-measure}
We represent the $1$-hop neighborhood of $v$ as an empirical probability measure on the feature space. Let $\delta_z$ denote a point mass (Dirac measure~\cite{tao2011introduction}) at location $z \in \mathbb{R}^d$, \ie for any bounded measurable test function $f$: $\int f(x)\, d\delta_z(x) = f(z)$.
%zulfikar; revised the paragraph without equations number.

\heading{Non-uniform, importance-weighted neighborhoods.}
Some neighbors are more informative about bot vs. human behavior than others. We therefore use a non-uniform weighting scheme. Let $a_v(u) > 0$ denote an unnormalized importance score for neighbor $u$ and define the normalized weights:
\begin{equation}
    w_v(u)
    =
    \frac{a_v(u)}{\sum_{u' \in \mathcal{N}(v)} a_v(u')}.
    \label{eq:weights-normalized}
\end{equation}

The importance score combines degree-based and time-based priorities:
\begin{equation}
    a_v(u)
    =
    g_{\mathrm{deg}}\bigl(\deg_{\mathrm{raw}}(u)\bigr)\;
    g_{\mathrm{time}}\bigl(\mathrm{age\_norm}(u)\bigr),
    \label{eq:importance-score}
\end{equation}
where, $ g_{\mathrm{deg}}\bigl(\deg_{\mathrm{raw}}(u)\bigr)=
        1 + \alpha_{\mathrm{deg}} \log\bigl(1 + \deg_{\mathrm{raw}}(u)\bigr), \text{with } \alpha_{\mathrm{deg}} \ge 0,$  to give higher but saturating importance to high-degree neighbors; likewise, 
        $g_{\mathrm{time}}\bigl(\mathrm{age\_norm}(u)\bigr)
        = 1 + \alpha_{\mathrm{time}}\,
        \mathrm{age\_norm}(u)$,  with $\alpha_{\mathrm{time}} \ge 0$ to emphasize long-lived accounts
    that have persisted over the observation window.

We then define the neighborhood probability measure:
\begin{equation}
    \mu_v
    =
    \sum_{u \in \mathcal{N}(v)}
    w_v(u)\, \delta_{\phi_v(u)}
    \label{eq:mu-weighted}
\end{equation}
% where $\delta_{\phi}$ denotes the Dirac measure (unit point mass)~\cite{tao2011introduction} at location $\phi$.

%\begin{itemize}[noitemsep,leftmargin=*, topsep=0pt]
%   \item \emph{degree-dependent weighting:} a monotone function:
%  \begin{equation}
%    \begin{aligned}
%        g_{\mathrm{deg}}\bigl(\deg_{\mathrm{raw}}(u)\bigr)
%        &=
%        1 + \alpha_{\mathrm{deg}} \log\bigl(1 + \deg_{\mathrm{raw}}(u)\bigr), \\
%        &\text{with } \alpha_{\mathrm{deg}} \ge 0,
%    \end{aligned}
%    \label{eq:g-deg}
%    \end{equation}
%    to give higher but saturating importance to high-degree neighbors;
%
%    \item \emph{time-dependent weighting:} a function:
 %   \begin{equation}
%        g_{\mathrm{time}}\bigl(\mathrm{age\_norm}(u)\bigr)
%        =
%        1 + \alpha_{\mathrm{time}}\,
%        \mathrm{age\_norm}(u),
%        \label{eq:g-time}
%    \end{equation}
%    with $\alpha_{\mathrm{time}} \ge 0$ to emphasize long-lived accounts
%    that have persisted over the observation window.
%\end{itemize}
This yields neighborhood distributions that concentrate more mass on structurally central and temporally persistent neighbors.

%\subsubsection{Step 3: Neural Ground Cost Between Neighbor Features}
\subsection{Step 3: Feature-based Neighbor Cost Computation}\label{sec:ground-cost}
A \emph{ground cost} $c_\theta$ between neighbor features in order to compare neighborhood distributions. Given two neighbor feature vectors $z, z' \in \mathbb{R}^d$, we first map each through a neural embedding network $h_\theta : \mathbb{R}^d \to \mathbb{R}^{d_{\mathrm{emb}}}$ and then compute a \emph{squared Mahalanobis distance} in the embedding space. 
 
Specifically, we let $\tilde{z} = h_\theta(z)$ and $\tilde{z}' = h_\theta(z')$ and define ground cost $c_\theta$
\begin{equation}
c_\theta(z,z')
    =
    \bigl(\tilde{z} - \tilde{z}'\bigr)^\top
    \mathbf{M}
    \bigl(\tilde{z} - \tilde{z}'\bigr),
\end{equation}
where $\mathbf{M} = \mathbf{L}^\top \mathbf{L} \succeq 0$ is a positive-semidefinite matrix parameterized via a learnable matrix $\mathbf{L}$. Since this is the Mahalanobis distance~\cite{mahalanobis1936} applied in a learned embedding space. It is expressive enough to reweight and correlate features, while remaining smooth, differentiable, and computationally efficient for entropic OT and Sinkhorn solvers~\cite{flamary2021pot, peyre2019computational}.

\subsection{Step 4: Entropic OT Between Neighborhoods}\label{sec:entropic-ot}

Given two neighborhoods $\mu_v = \sum_{i=1}^m a_i \delta_{z_i}$ and $\mu_w = \sum_{j=1}^n b_j \delta_{z'_j}$, we assemble a cost matrix $C \in \mathbb{R}^{m \times n}$ with entries $C_{ij} = c_\theta(z_i, z'_j)$, where $z_i$ and $z'_j$ are the neighbor feature vectors from $v$ and $w$, respectively, and $a_i, b_j \ge 0$ are their corresponding normalized importance weights ($\sum_i a_i = \sum_j b_j = 1$). The discrete OT problem seeks a transport plan $P \in \mathbb{R}^{m \times n}_{\ge 0}$ with given marginals: $\mathcal{U}(a,b)=\Bigl\{P \in \mathbb{R}^{m \times n}_{\ge 0}:P \mathbf{1}_n = a,\;P^\top \mathbf{1}_m = b\Bigr\}$, where $\mathbf{1}_k$ denotes an all-ones vector of length $k$.

For $\varepsilon > 0$, the entropic-regularized OT problem is

\[    \operatorname{OT}_\varepsilon(\mu_v, \mu_w)
    =
    \min_{P \in \mathcal{U}(a,b)}\;
    \Biggl[
        \sum_{i,j} P_{ij} C_{ij}
        +
        \varepsilon
        \sum_{i,j} P_{ij}
        \bigl( \log P_{ij} - 1 \bigr)
    \Biggr]
\]
Let $P^\star_{vw}$ denote an optimal entropic transport plan, i.e.,
\[ 
P^\star_{vw} \in
\arg\min_{P \in \mathcal{U}(a,b)}\;
\Biggl[
\sum_{i,j} P_{ij} C_{ij}
+
\varepsilon \sum_{i,j} P_{ij}(\log P_{ij} - 1)
\Biggr]
\]
 
The entropic regularizer, $\varepsilon$, encourages diffuse plans and yields a unique minimizer $P^\star$; moreover, after introducing the Gibbs kernel $K=\exp(-C/\varepsilon)$, the objective can be expressed (up to an additive constant) as a KL divergence, making $\varepsilon$ a \textit{temperature} that controls plan sharpness (small $\varepsilon$: near-permutation, large $\varepsilon$: diffuse). The ideal plan lies between a completely diffuse plan and a brittle near-permutation plan, stabilizing the OT geometry by providing smooth, well-conditioned transport maps that capture multiple plausible alignments among similar neighbors, rather than completely diffuse or a brittle, permutational matching plan.

As is standard in entropic OT, the objective can be rewritten as a Kullback-Leibler (KL) divergence between $P$ and a Gibbs kernel constructed from the cost matrix $C$; see Appendix~\ref{app:ot-lagrangian} for a detailed derivation. In particular, introducing the Gibbs kernel $K = \exp(-C/\varepsilon)$, the optimal plan $P^\star$ can be written in a scaling form $P^\star = \mathrm{diag}(u)\, K\, \mathrm{diag}(v)$ where the scaling vectors $u$ and $v$ are obtained via Sinkhorn iterations that enforce the marginal constraints $P^\star \mathbf{1}_n = a$ and $(P^\star)^\top \mathbf{1}_m = b$. 
 
We define OT distance as the transport cost induced by the entropic plan:
\begin{equation}
D_\theta(v,w)
\triangleq
\bigl\langle P^\star_{vw}, C_{vw}\bigr\rangle
=
\sum_{i,j} (P^\star_{vw})_{ij}\, (C_{vw})_{ij}.
\label{eq:ot-cost-p}
\end{equation}
Entropic regularization is used only to obtain a stable and unique plan $P^\star_{vw}$ via Sinkhorn.

\subsection{Step 5: OT-Margin Between Human And Bot}
\label{subsec:ot-margin-main}

Let $\mathcal{B}$ denote the set of bot accounts and $\mathcal{H}$ the set of human accounts. 
Using the OT distance $D_\theta(\cdot,\cdot)$ from \autoref{eq:ot-cost-p}, for each bot 
$v \in \mathcal{B}$, the minimum distance to any human account is
$d_H(v) = \min_{h \in \mathcal{H}} D_\theta(v,h)$,  while the minimum distance to another bot account is $d_B(v) = \min_{b \in \mathcal{B} \setminus \{v\}} D_\theta(v,b)$. The OT-based margin is then given by $m(v) = d_H(v) - d_B(v)$. 

This OT-based margin ensures:
\begin{itemize}[noitemsep,leftmargin=*,topsep=0pt]
    \item if $m(v)$ is large and positive, then $v$ is much closer (in OT space)
          to other bots than to any human, indicating a robust bot-like neighborhood;
    \item if $m(v)$ is small or negative, $v$ lies close to (or inside) the
          ``human region'' of OT space, suggesting that small structural
          perturbations could push $v$ across the classifier's decision boundary.
\end{itemize}

We also introduce a margin threshold $\tau_{\mathrm{bdry}}$ and label a bot $v$ as a boundary bot when $m(v)\le \tau_{\mathrm{bdry}}$, meaning its neighborhood is closer to the human manifold than to the bot manifold under $D_\theta$.

This OT-based margin with margin threshold, $\tau_{\mathrm{bdry}}$ ensures:
\begin{itemize}[noitemsep,leftmargin=*,topsep=0pt]
    \item if $m(v) > \tau_{\mathrm{bdry}}$, then $v$ is well inside the bot manifold (robustly bot-like), since it is closer to bots than to humans by a comfortable margin;
    \item if $m(v) \le \tau_{\mathrm{bdry}}$, then $v$ is a boundary bot whose neighborhood is comparatively human-close in OT space, so small structural perturbations may push it across the decision boundary (especially when $m(v)\le 0$).
\end{itemize}

\heading{Adversarial Interpretation.}
Viewing OT space as the \emph{attack space}, the adversary seeks a neighborhood distribution $\mu'_v$ for the target bot such that the classifier labels it as human:
\begin{equation}
    \min_{\mu'_v:\, f(\mu'_v) = \text{human}} D_\theta(\mu_v, \mu'_v).
    \label{eq:min-perturbation}
\end{equation}
The optimal value of \autoref{eq:min-perturbation} reflects the robustness of the bot detector around $v$: a large minimal perturbation implies that the current neighborhood is robust to realistic edits, while a small one indicates vulnerability.

In \pname, we do not optimize \autoref{eq:min-perturbation} directly. Instead, we use existing bots as \emph{cloak} templates for $\mu'_v$. Choosing a mimic bot with small $d_H(\cdot)$ or negative margin $m(\cdot)$ yields a blueprint for the \emph{minimal} structural changes (edge edits) needed to make the target bot's neighborhood distribution human-like.

Under this formulation, misclassified bots naturally reveal vulnerable regions of the decision boundary in OT space, and the OT distance quantifies how far a correctly classified bot lies from such regions. Bots with small positive margin are our candidate \emph{boundary bots}, whereas bots with negative or near-zero margin are already evasive.

\subsection{Step 6: OT Geometry Training Objective}\label{app:full-obj}

The OT geometry (parameters $\theta$ and $\mathbf{M} = \mathbf{L}^\top \mathbf{L}$) is trained offline using a multi-tasked optimization loss (refer to ~\autoref{eq:overall-loss} and ~\autoref{alg:train-ot-geometry}). We use three complementary loss components accounting for margin, sparsity and plausibility.
 
\heading{Binary Cross-Entropy (BCE) Loss via OT-margin surrogate.}
Rather than differentiating through discrete edge edits and the classifier, we directly shape the OT margin. For each bot $v$, let $y_v \in \{0,1\}$ denote a label that encodes whether $v$ is currently misclassified ($y_v=1$) or correctly classified ($y_v=0$). We apply a logistic surrogate to the margin:
 
\begin{equation}
    L_{\mathrm{BCE}}(v;\theta)
    =
    \mathrm{BCE}\Bigl(
        \sigma\bigl(- m(v) / \tau_{\mathrm{BCE}}\bigr),
        y_v
    \Bigr),
\label{eq:loss-bce}
\end{equation}
 
where $\sigma$ is the logistic sigmoid, $\tau_{\mathrm{BCE}} > 0$ is a sigmoid temperature hyperparameter, and $\mathrm{BCE}$ is the standard binary cross-entropy. This encourages misclassified bots to have a small or negative margin and correctly classified bots to have a large positive margin, thereby sharpening the decision boundary in OT space.

\heading{Sparsity loss via OT plan entropy.}
To encourage attack sparsity (few edge changes for the target bot), we regularize the entropy of the OT plan between a bot $v$ and its nearest human $h^\star(v)$.
For each bot $v$, the nearest human is:
\begin{equation}
    h^\star(v) = \arg\min_{h \in \mathcal{H}} D_\theta(v,h),
\end{equation}
where $D_\theta(v,h)$ is the OT distance between their neighborhood measures.
 
Let $P_{vh}^\star \in \mathbb{R}_{\ge 0}^{m \times n}$ be the Sinkhorn plan between $\mu_v$ and $\mu_{h^\star(v)}$, where rows $i = 1,\dots,m$ index neighbors $u_i \in \mathcal{N}(v)$ of the bot $v$ and columns $j = 1,\dots,n$ index neighbors $u'_j \in \mathcal{N}(h^\star(v))$ of its closest human. Thus $P_{ij}$ denotes how much probability mass from bot-side neighbor $u_i$ is transported to human-side neighbor $u'_j$.

We define row-wise and column-wise conditional entropies that measure how concentrated the transport mass is within each row/column of $P$ (\ie whether each neighbor aligns to only a few counterparts versus spreading its mass broadly), where $p_{j\mid i}= P_{ij}/a_i$ denotes the row-conditional and $q_{i\mid j}= P_{ij}/b_j$ denotes the column-conditional distribution:
\begin{equation}
H_{\mathrm{row}}(P)
 = \sum_{i=1}^{m} a_i\Bigl(-\sum_{j=1}^{n} p_{j\mid i}\log p_{j\mid i}\Bigr)
\notag\\
= -\sum_{i=1}^{m}\sum_{j=1}^{n} P_{ij}\log\frac{P_{ij}}{a_i},
\label{eq:row-cond-entropy}
\end{equation}
\begin{equation}
H_{\mathrm{col}}(P)
= \sum_{j=1}^{n} b_j\Bigl(-\sum_{i=1}^{m} q_{i\mid j}\log q_{i\mid j}\Bigr)
\notag\\
= -\sum_{i=1}^{m}\sum_{j=1}^{n} P_{ij}\log\frac{P_{ij}}{b_j},
\label{eq:col-cond-entropy}
\end{equation}

Here, $p_{j\mid i}=P_{ij}/a_i$ describes how the mass of bot-side neighbor $u_i$ is distributed across human-side neighbors, and $q_{i\mid j}=P_{ij}/b_j$ describes how the mass received by human-side neighbor $u'_j$ is distributed across bot-side neighbors. When $H_{\mathrm{row}}(P)$ is small, each bot-side neighbor concentrates its transport on only a few human-side neighbors; when $H_{\mathrm{col}}(P)$ is small, each human-side neighbor receives mass from only a few bot-side neighbors. 

The sparsity loss for bot $v$ as the sum of these two entropies:
 
\begin{equation}
L_{\mathrm{sp}}(v;\theta)
=
H_{\mathrm{row}}\!\bigl(P_{vh}^{\star}\bigr)
+
H_{\mathrm{col}}\!\bigl(P_{vh}^{\star}\bigr).
\label{eq:loss-sparsity-entropy}
\end{equation}
 
By construction, $L_{\mathrm{sp}}(v;\theta)$ is large when the transport mass in $P_{vh}^\star$ is distributed broadly across many bot-side and human-side neighbors, resulting in high row- and column-wise conditional entropies. Minimizing $L_{\mathrm{sp}}(v;\theta)$ therefore encourages OT plans in which the transport concentrates on a small set of high-confidence template matches between bot-side neighbors and human-side neighbors (i.e., a localized, few-to-few alignment). In the context of \pname, these template matches can be interpreted as a small set of ``bot templates'' (human-like local patterns) that the target bot must mimic, which in turn corresponds to attack strategies requiring only a few edge edits around the target bot.

\headinggi{Derivation and Properties.}
Using \autoref{eq:row-cond-entropy} and the marginal constraint $P\mathbf 1_n=a$,
\begin{align}
H_{\mathrm{row}}(P)
&= -\sum_{i,j} P_{ij}\log P_{ij} + \sum_{i} a_i\log a_i.
\label{eq:row-expand}
\end{align}
Similarly, using $P^\top\mathbf 1_m=b$,
\begin{align}
H_{\mathrm{col}}(P)
&= -\sum_{i,j} P_{ij}\log P_{ij} + \sum_{j} b_j\log b_j.
\label{eq:col-expand}
\end{align}

Therefore,
\begin{align}
H_{\mathrm{row}}(P)+H_{\mathrm{col}}(P)
&=
2\Bigl(-\sum_{i,j} P_{ij}\log P_{ij}\Bigr)
+\sum_i a_i\log a_i+\sum_j b_j\log b_j,
\label{eq:sum-expand}
\end{align}
\ie $L_{\mathrm{sp}}$ is equivalent to minimizing the entropy of the transport plan and is thus non-constant in $P$.

\headinggi{Why this encourages sparsity?}
For each fixed row $i$, the Shannon entropy satisfies $H(p_{\cdot\mid i})\ge 0$ with equality iff $p_{\cdot\mid i}$ is a point mass. Hence minimizing $H_{\mathrm{row}}(P)=\sum_i a_i H(p_{\cdot\mid i})$ encourages each row to concentrate mass on few columns; analogously minimizing $H_{\mathrm{col}}(P)$ encourages concentration within each column.

Moreover, the feasible set $\mathcal U(a,b)=\{P\ge 0:\;P\mathbf 1_n=a,\;P^\top\mathbf 1_m=b\}$ is a convex polytope, and $H(P)=-\sum_{i,j}P_{ij}\log P_{ij}$ is concave in $P$. Therefore, minimizing $H(P)$ over $\mathcal U(a,b)$ attains a minimum at an extreme point of $\mathcal U(a,b)$, which is a feasible solution with at most $m+n-1$ nonzero entries, \ie a sparse transport plan.

\heading{Plausibility loss via degree and account-age alignment.}
We also regularize the OT plan to preserve social and temporal  plausibility. Large numbers of high-degree, long-lived humans following very new, low-degree bots are unlikely in realistic networks. 
 
We define a per-pair plausibility cost that penalizes degree and account-age mismatches between matched neighbors:

\[ 
\phi_{\mathrm{pl}}(i,j)
    = \Bigl(
    \alpha_{\deg}\, \bigl|\deg(i) - \deg(j)\bigr|
    \\
    \quad
    +
    \alpha_{\mathrm{age}}\,
    \bigl|\mathrm{age}(i) - \mathrm{age}(j)\bigr|
    \Bigr)
\]
 
with $\alpha_{\deg}, \alpha_{\mathrm{age}} \ge 0$, where $\deg(i)$ and $\deg(j)$ are the degrees of neighbors $u_i$ and $u'_j$, and $\mathrm{age}(\cdot)$ is their normalized account age. Larger $\alpha_{\deg}$ (resp.\ $\alpha_{\mathrm{age}}$) increases the penalty for degree (resp.\ account-age) mismatches, biasing $P_{vh}^\star$ toward structurally and temporally aligned neighbor matches.

The plausibility loss aggregates these per-pair costs under the OT plan:
\begin{equation}
    L_{\mathrm{pl}}(v;\theta)
    =
    \sum_{i,j} P_{vh,ij}^\star\, \phi_{\mathrm{pl}}(i,j).
    \label{eq:pl-loss}
\end{equation}
$L_{\mathrm{pl}}(v;\theta)$ measures how much transport mass is placed on socially implausible matches where high-degree, long-lived humans are aligned with very new, low-degree bots (or vice versa). $\phi_{\mathrm{pl}}(i,j)$ increases as the degree/age mismatch increases, and $L_{\mathrm{pl}}(v;\theta)$ is large when the OT plan puts mass on implausible matches. Under this loss, neighbor pairs that both (i) receive substantial transport mass $P_{vh,ij}^\star$ and (ii) exhibit large degree or age gaps contribute more strongly to the penalty, encouraging OT plans that align neighbors with similar degree and temporal profiles.

\heading{Multi-task Objective.}
Putting the pieces together, we train the OT geometry by minimizing

\begin{equation}
    \mathcal{L}_{\pname}(\theta) =
    \mathbb{E}_{v \in \mathcal{V}_{\mathrm{train}}}
    \bigl[
        \lambda_{\mathrm{BCE}}\, L_{\mathrm{BCE}}(v;\theta)
        +
        \lambda_{\mathrm{sp}}\, L_{\mathrm{sp}}(v;\theta)
        +
        \lambda_{\mathrm{pl}}\, L_{\mathrm{pl}}(v;\theta)
    \bigr]
\label{eq:overall-loss}
\end{equation}
%\begin{equation}
%\begin{split}
%  \mathcal{L}_{\pname}(\theta)
 %   &=
 %   \mathbb{E}_{v \in \mathcal{V}_{\mathrm{train}}}
  %  \bigl[
   %     \lambda_{\mathrm{BCE}}\, L_{\mathrm{BCE}}(v;\theta)
    %    +
     %   \lambda_{\mathrm{sp}}\, L_{\mathrm{sp}}(v;\theta)
      %  \\
       % &\qquad\qquad
       % +
        %\lambda_{\mathrm{pl}}\, L_{\mathrm{pl}}(v;\theta)
  %  \bigr],
%\end{split}
%\label{eq:overall-loss}
%\end{equation}
with non-negative weights $\lambda_{\mathrm{BCE}}, \lambda_{\mathrm{sp}}$, and $\lambda_{\mathrm{pl}}$.

We treat $(\varepsilon, TopK)$ as OT hyperparameters: $\varepsilon$ is the entropic regularization used in Sinkhorn during both training and inference, while $K$ is used only at attack time to select the Top $K$ OT-salient neighbors in \autoref{alg:helper-otguidedneighbors}. ~\autoref{alg:train-ot-geometry} details the offline optimization procedure used to learn the OT geometry $(\theta,\mathbf M)$.

\input{algos/train-ot}
\input{algos/newbot-algo}

\subsection{Step 7: OT Margins and Boundary Bots in \pname}\label{app:train-op}

Once the OT geometry is trained (using ~\autoref{alg:train-ot-geometry}), we precompute $D_\theta(v,w)$ for bot-human pairs, obtain margins $m(v)$ for all bots, and identify boundary bots ($\mathcal{B}_{\mathrm{bdry}}$) and misclassified bots ($\mathcal{B}_{\mathrm{mis}}$), as described in ~\autoref{subsec:ot-margin-main}. Training learns a neural ground cost $c_\theta$ (with Mahalanobis metric $\mathbf M$) by differentiating through a Sinkhorn OT solver with entropic regularization $\varepsilon$. For each training bot, the procedure mines its nearest human and nearest-bot neighbors under the current OT geometry, and forms an OT margin $m(v)=d_H(v)-d_B(v)$. It then updates $(\theta,\mathbf M)$ to shape this margin via the margin threshold ($\tau_{\mathrm{bdry}}$) and sigmoid temperature  ($\tau_{\mathrm{BCE}}$) while regularizing the transport plan with sparsity and plausibility losses. Given the learned OT boundary, \autoref{alg:bocloak-newbot} performs OT-guided cloaking that pushes correctly classified bots across the decision boundary, causing them to be misclassified as human and thus evading the bot detector.

\subsection{Step 8: OT-Guided \pname Attack}\label{app:op-newbot}

\autoref{alg:bocloak-newbot} summarizes the \pname OT-guided pipeline for bot editing attacks (with helper routines detailed in \autoref{app:helper-bdry-bots}, \autoref{app:helper-getImportanceWeights}, \autoref{app:helper-sampletemp},  \autoref{app:helper-otguidedneighbors} and \autoref{alg:helper-CloneCloak}). For a given target bot $v_{\mathrm{tar}}$, once OT margins and boundary bots are precomputed, \pname selects a small set of OT bot cloak templates $b \in \mathcal{B}_{\mathrm{bdry}} \cap \mathcal{B}_{\mathrm{mis}}$, as described in~\autoref{app:helper-bdry-bots}. Rather than choosing templates uniformly, \pname samples each template bot $b$ from an importance-weighted distribution $p(b)$ that favors (i) structurally cheaper templates (fewer required edges) and (ii) templates closer to the OT boundary (higher boundary priority). To avoid overusing a single blueprint, we enforce a reuse cap: each bot template can be selected at most three times while any template remains under the cap; once all templates reach this cap, we reset the counters and continue sampling under the same policy. This captures a practical constraint: if a successful template bot is reused too many times and later detected, then its cloaked bots may be flagged as well. Therefore, we cap template reuse to reduce correlated detection risk across generated bots.

%\zulfikar{Kunal: Please add a brief explanation of Algorithm 1, referring to the key steps / major line.  Please justify the choice of setting the cap to 3.  Reviewers may question why 3 was selected instead of 5 or another value.}

Before applying any edge edits, we initialize the non-temporal node features of $v_{\mathrm{tar}}$ to match those of the current template bot $b$, while keeping temporal features such as account age consistent with a newly created account. For each template $b$, we then retrieve the OT plan between $b$ and its closest human, interpret high-mass entries in the plan as \emph{critical neighbor pairs}, and translate those pairs into concrete edge edits between $v_{\mathrm{tar}}$ and the neighbors of $b$, subject to a strict edge budget and plausibility constraints. In other words, the OT plan tells us \emph{which edges matter most} for making $b$ appear human-like, and \pname reuses this blueprint for $v_{\mathrm{tar}}$ to construct sparse, realistic neighborhoods that cross the decision boundary.

In summary, OT provides two key signals for \pname:
(i) OT margins identify boundary bots whose neighborhoods are already close to the human manifold and hence are promising candidate cloaks; and
(ii) the OT plan between a template and its nearest human identifies which neighbors are most critical to clone, enabling sparse, plausible edge edits that still effectively push new or existing correctly classified bots across the decision boundary.

\heading{Special Case: Perfect Bot-Detector.}
In rare cases, the underlying bot detector may be ``perfect'' on the current graph, so that no bots are misclassified, and the OT margin test fails to identify any boundary or misclassified templates, i.e., $\mathcal{B}_{\mathrm{bdry}} \cup \mathcal{B}_{\mathrm{mis}} = \emptyset$. In this scenario, \pname falls back to a purely human behavior-guided strategy that still exploits the learned OT geometry. For each bot $v_{\mathrm{tar}}$ in an allowed structural category, we first identify its most human-like account:
$h^\star(v_{\mathrm{tar}}) = \arg\min_{h \in \mathcal{H}} D_\theta(v_{\mathrm{tar}}, h)$,
and treat $h^\star(v_{\mathrm{tar}})$ as a \emph{human template}. We then instantiate a new synthetic bot node $v_{\mathrm{new}}$ for node injection or update the target node for node editing by perturbing the non-temporal features and mimicking follow edges of $h^\star(v_{\mathrm{tar}})$, subject to the same budget and plausibility constraints. 

Human templates whose cloned neighborhood would exceed the budget are skipped. This human-template fallback creates highly human-like synthetic bots whose neighborhoods lie close to the human manifold in OT space, and can still yield evading examples even when no misclassified or boundary bot templates are available. This method greedily pulls the neighborhood mass of $v_{\mathrm{tar}}$ toward the human manifold in OT space using only edge additions that progressively shrinking $D_\theta(v_{\mathrm{tar}}, h)$ eventually pushes the target across the decision boundary and yields an evading bot even when no misclassified bot cloak templates are available.

\section{Entropic Optimal Transport, Sinkhorn Scaling, and Complexity}\label{app:en-ot-scomp}

This section presents the mathematical derivation of entropic Optimal Transport in \pname, the associated Sinkhorn scaling method, and its computational complexity.

\label{app:ot-lagrangian}

% \subsection{Why Mahalanobis and not Wasserstein as ground cost?}
% Note that $c_\theta$ is a \emph{pointwise} cost between individual neighbor features, whereas the OT distance $D_\theta$ already provides a \emph{Wasserstein-type} distance between entire neighborhood distributions. Using a Mahalanobis metric for $c_\theta$ has two advantages: (i) it yields a closed-form, PSD quadratic form that is cheap to evaluate and easy to differentiate through; (ii) it lets OT focus on aligning neighborhoods at the distribution level, instead of nesting an additional Wasserstein problem inside the ground cost. In contrast, using a Wasserstein distance as the ground cost between points would require solving a second OT problem for each pair of neighbors, which is computationally prohibitive and unnecessary given that $D_\theta$ already measures Wasserstein distance between the local distributions.

\subsection{Entropic OT objective and KL form}\label{sec:entropic-kl}

Recall the entropic OT problem in \autoref{sec:entropic-ot}:
\begin{equation}
\begin{aligned}
    \operatorname{OT}_\varepsilon(\mu_v, \mu_w)
    =
    \min_{P \in \mathcal{U}(a,b)}\;
    &\Biggl[
        \sum_{i,j} P_{ij} C_{ij}
        +
        \varepsilon
        \sum_{i,j} P_{ij}
        \bigl( \log P_{ij} - 1 \bigr)
    \Biggr],
\end{aligned}
\label{eq:app-entropic-ot}
\end{equation}
where $\mathcal{U}(a,b)$ is the transportation polytope
$P \mathbf{1}_n = a$, $P^\top \mathbf{1}_m = b$, $P_{ij} \ge 0$.
We introduce the Gibbs kernel
\begin{equation}
    K_{ij} = \exp(-C_{ij}/\varepsilon).
    \label{eq:app-gibbs}
\end{equation}

Using the identity
\begin{equation}
    \sum_{i,j} P_{ij} C_{ij}
    +
    \varepsilon \sum_{i,j} P_{ij}(\log P_{ij} - 1)
    =
    \varepsilon \sum_{i,j} P_{ij} \log \frac{P_{ij}}{\tilde K_{ij}}
    + \text{const},
\end{equation}
where $\tilde K_{ij} = \exp(-C_{ij}/\varepsilon)$ and the constant does not
depend on $P$, we can rewrite \autoref{eq:app-entropic-ot} as
\begin{equation}
    \operatorname{OT}_\varepsilon(\mu_v, \mu_w)
    =
    \varepsilon \min_{P \in \mathcal{U}(a,b)}
    \mathrm{KL}\!\bigl(P \,\|\, K\bigr) + \text{const},
    \label{eq:app-kl-form}
\end{equation}
where
$\mathrm{KL}(P \,\|\, K) = \sum_{i,j} P_{ij} \log(P_{ij}/K_{ij})$. Thus entropic OT selects, among all matrices with the desired marginals, the plan $P$ that is closest (in KL divergence) to the Gibbs kernel $K$.

Previously, we defined the OT distance between nodes $v$ and $w$ as the transport term under the entropic optimal plan:
\[
D_\theta(v,w) \triangleq \langle P^\star_{vw}, C_{vw}\rangle.
\]
The full entropic objective value in \autoref{eq:app-entropic-ot} satisfies, for $\varepsilon>0$,
\[
\operatorname{OT}_\varepsilon(\mu_v,\mu_w)
=
D_\theta(v,w)
+
\varepsilon \sum_{i,j} (P^\star_{vw})_{ij}\bigl(\log (P^\star_{vw})_{ij}-1\bigr),
\]
so $\operatorname{OT}_\varepsilon(\mu_v,\mu_w)$ and $D_\theta(v,w)$ are generally not equal except when $\varepsilon=0$ or when the entropic regularizer evaluates to zero at $P^\star_{vw}$.

\subsection{Lagrangian and scaling form}\label{sec:lang}

The constrained problem \autoref{eq:app-entropic-ot} admits a dual. Consider the Lagrangian
\begin{equation}
\begin{aligned}
    \mathcal{L}(P,\alpha,\beta)=
    \sum_{i,j} P_{ij} C_{ij}
    +
    \varepsilon
    \sum_{i,j} P_{ij}(\log P_{ij} - 1)
    %&\quad
    + \sum_i \alpha_i \Bigl( a_i - \sum_j P_{ij} \Bigr)
    + \sum_j \beta_j \Bigl( b_j - \sum_i P_{ij} \Bigr)
\end{aligned}
\label{eq:app-lagrangian}
\end{equation}
where $\alpha \in \mathbb{R}^m$, $\beta \in \mathbb{R}^n$ are dual
variables for the marginal constraints. Differentiating w.r.t.\ $P_{ij}$
and setting the derivative to zero yields
\[
    C_{ij}
    +
    \varepsilon (\log P_{ij})
    - \alpha_i - \beta_j = 0,
\]
so that
\begin{equation}
    P_{ij}
    =
    \exp\!\Bigl(\tfrac{\alpha_i}{\varepsilon}\Bigr)\,
    \exp\!\Bigl(-\tfrac{C_{ij}}{\varepsilon}\Bigr)\,
    \exp\!\Bigl(\tfrac{\beta_j}{\varepsilon}\Bigr)
    =
    u_i K_{ij} v_j,
    \label{eq:app-scaling}
\end{equation}
where we define the scaling vectors
$u_i = \exp(\alpha_i/\varepsilon)$ and $v_j = \exp(\beta_j/\varepsilon)$.
In matrix form,
\begin{equation}
    P^\star = \mathrm{diag}(u)\, K\, \mathrm{diag}(v).
    \label{eq:app-sinkhorn-form}
\end{equation}
The marginal constraints $P^\star \mathbf{1}_n = a$ and
$(P^\star)^\top \mathbf{1}_m = b$ translate to
\begin{equation}
    u \odot (K v) = a,
    \qquad
    v \odot (K^\top u) = b,
    \label{eq:app-scaling-eqs}
\end{equation}
where $\odot$ denotes element-wise multiplication.

\subsection{Sinkhorn iterations}

The Sinkhorn algorithm~\cite{cuturi2013sinkhorn} iteratively rescales $u$ and $v$ to satisfy
\autoref{eq:app-scaling-eqs}. Starting from $v^{(0)} = \mathbf{1}_n$,
a typical scheme is:
\begin{align}
    u^{(t+1)} &= a \oslash (K v^{(t)}), \label{eq:app-u-update} \\
    v^{(t+1)} &= b \oslash (K^\top u^{(t+1)}), \label{eq:app-v-update}
\end{align}
where $\oslash$ is element-wise division. After $T_{\mathrm{sink}}$ iterations,
we obtain approximate scalings $u^{(T_{\mathrm{sink}})}$ and
$v^{(T_{\mathrm{sink}})}$ and hence an approximate optimal plan, $P^{(T_{\mathrm{sink}})}$,
\[
    P^{(T_{\mathrm{sink}})} =
    \mathrm{diag}\bigl(u^{(T_{\mathrm{sink}})}\bigr)
    K
    \mathrm{diag}\bigl(v^{(T_{\mathrm{sink}})}\bigr).
\]
Under mild conditions, the iterates converge geometrically and the resulting
plan satisfies the marginal constraints up to a small numerical tolerance.

\subsection{Computational and Time Complexity}
\label{app:complexity}

Let $m = |\mathcal{N}(v)|$ and $n = |\mathcal{N}(w)|$ denote the neighborhood
sizes of $\mu_v$ and $\mu_w$, respectively. Constructing the cost matrix
$C \in \mathbb{R}^{m \times n}$ with entries $C_{ij}=c_\theta(z_i,z'_j)$ and the
Gibbs kernel $K=\exp(-C/\varepsilon)$ requires $O(mn)$ ground-cost evaluations
(\autoref{sec:ground-cost}) and exponentials, and stores $O(mn)$ numbers.

Each Sinkhorn iteration updates the scaling vectors via the matrix--vector
products $K v$ and $K^\top u$, each costing $O(mn)$ operations. After
$T_{\mathrm{sink}}$ iterations, one entropic OT solve between $\mu_v$ and $\mu_w$
therefore costs
\[
\text{time } = O(T_{\mathrm{sink}}mn),
\qquad
\text{space } = O(mn),
\]
where the space accounts for storing $C$ and $K$ (and an additional $O(m+n)$ for
the scaling vectors). Consequently, the OT computation that dominates one
iteration of \pname for finding one cloak is
$O(T_{\mathrm{sink}}mn)$ in time and $O(mn)$ in space complexity.

Computing $d_H(v)$ and $d_B(v)$ for all bots requires OT distances between each
$v \in \mathcal{B}$ and all $h \in \mathcal{H}$, as well as all
$b \in \mathcal{B}\setminus\{v\}$. If $|\mathcal{B}| = N_B$ and
$|\mathcal{H}| = N_H$ and typical neighborhood sizes are $m$ and $n$, then the
number of Sinkhorn solves is $N_B N_H + N_B(N_B-1)$, and the total time for a
na\"ive full pairwise computation is $O\bigl(T_{\mathrm{sink}}mn\,(N_B N_H + N_B^2)\bigr)$ with per-solve memory $O(mn)$.

% \heading{Caching.}
% In the \pname setting, we must evaluate OT distances between many pairs of
% bots and humans when computing margins \autoref{subsec:ot-margin-main}.
% To make this tractable, we (i) restrict OT computation to a candidate set of humans and bots selected via only calculating OT for bots with node degree less than or equal to budget provided (\eg $\Delta=1$), (ii) reuse the same kernel $K$ across multiple solves whenever $C$ depends only on distance in the learned embedding space, and (iii) cache $D_\theta(v,w)$ for frequently queried pairs. These approximations preserve the structure of the OT margin while keeping the precomputation stage within our computational budget.
\heading{Caching.}
In the \pname setting, we must evaluate OT distances between many pairs of
bots and humans when computing margins \autoref{subsec:ot-margin-main}.
To make this tractable, we (i) restrict OT computation to a candidate set of humans and bots selected via only calculating OT for bots with node degree less than or equal to budget provided (\eg $\Delta=1$), (ii) reuse the same kernel $K$ across multiple solves whenever $C$ depends only on distance in the learned embedding space, and (iii) cache $D_\theta(v,w)$ for frequently queried pairs. For higher-degree nodes, we can additionally truncate each empirical measure to its top-weighted neighbors before solving OT, and for larger graphs we restrict cloak search to the top-$K$ boundary candidates. These approximations preserve the structure of the OT margin while keeping the precomputation stage within our computational budget.

% \section{Sparsity Loss: Extended Discussion}\label{sec:sparse-detail}

\input{sections/bocloak_helper_function}

\section{Bot Detectors (Victim Models)}

The underlying bot detectors attacked by \pname are treated as frozen black-box classifiers. Our victim set includes:

\begin{itemize}[leftmargin=*,noitemsep]
  % \item \textbf{RF}: a random forest classifier~\citep{breiman2001randomforest} on hand-crafted user and content features (no graph), as a non-graph baseline.

  \item \textbf{GCN}~\citep{kipf2017gcn} standard homogeneous GNNs that operate on the social graph with node features.
  \item \textbf{GAT}~\citep{velickovic2017gat} an attention-based model for heterogeneous graphs that operate on the social graph with node features.
  \item \textbf{BotRGCN}~\citep{feng2021botrgcn}, a heterogeneous GNN where a standard GCN is enhanced using relation-aware message propagation for bot detection.
  \item \textbf{Simple-HGNN (S-HGN)}~\citep{lv2021we}, a heterogeneous GNN similar to BotRGCN, but complex relation-specific modules are replaced with lightweight type-wise attention and relation-aware message propagation.
  \item \textbf{RGT} (Relational Graph Transformer)~\citep{feng2021rgt}, a heterogeneity-aware bot detector that uses relation-specific transformer blocks.

    % \item \textbf{HGT} (Heterogeneous Graph Transformer)~\citep{hu2020hgt}, a transformer-based model for large heterogeneous graphs that we instantiate on the multi-relational TwiBot-22 schema.

  % \item \textbf{SEBot}~\citep{yang2024sebot}, which leverages structural entropy and multi-view contrastive learning to capture hierarchical neighborhood patterns for bot detection.

  % \item \textbf{BotDGT}~\citep{he2024botdgt}, a dynamic graph transformer that models the temporal evolution of the social network and has state-of-the-art performance on time-resolved benchmarks.
\end{itemize}

\section{\gnn Adversarial Defense}

Although our primary focus is on evaluating \pname{} against \emph{vanilla} bot detectors, we also study how GNN robustness interventions affect \pname's performance (more details in ~\autoref{sec:victim-defense}):

\begin{itemize}[leftmargin=*,noitemsep]

  \item \textbf{GNNGuard}~\citep{zhang2020gnnguard}, which reweights and prunes edges based on feature similarity scores, suppressing adversarial shortcuts while preserving informative neighbors.

  \item \textbf{GRAND}~\citep{feng2020grand}, a graph random neural network that uses random propagation and consistency regularization across stochastic augmentations to improve both generalization and robustness.

  \item \textbf{RobustGCN}~\citep{zhu2019robustgcn}, which replaces deterministic neighborhood aggregation with a Gaussian-based formulation to downweight noisy or adversarial neighbors.

  \item \textbf{NoisyGNN}~\citep{ennadir2024noisygnn}, an additional adversarial-training/architecture-noise control that injects stochasticity into hidden representations while adding minimal complexity.

  % \item \textbf{ElasticGNN}~\citep{liu2021elasticgnn}, which introduces elastic message passing and residual connections to stabilize training and dampen the influence of perturbed edges and nodes.

  % \item \textbf{RUNG}~\citep{hou2024robustgnn}, which proposes an unbiased aggregation scheme derived from robust optimization principles, achieving state-of-the-art robustness on large, attack-prone graphs.
  % \item \textbf{NoisyGNN}
  % \item \textbf{GCORN}
  % \item \textbf{GNN-Jaccard}~\citep{wu2019adversarial}, a preprocessing-based defense that prunes edges between nodes with low Jaccard similarity in the feature space, thereby removing many adversarial or semantically inconsistent connections before GNN training.

\end{itemize}

\section{Adversarial Attacks}

We compare \pname against four \sota adversarial attack frameworks:

\begin{itemize}[leftmargin=*,noitemsep]
  \item \textbf{Nettack}~\citep{zugner2018nettack}, a targeted attack that greedily perturbs edges and features to maximally decrease the victim model's logit margin while respecting a budget.

  \item \textbf{FGA} (Fast Gradient Attack)~\citep{chen2018fga}, which uses a first‑order (locally linear) approximation of the victim model and gradient‑based importance score of each edge to rank candidate edge perturbations.

  \item \textbf{PR-BCD} (Projected Randomized Block Coordinate Descent)~\citep{geisler2021_robustness_of_gnns_at_scale}, a scalable white-box topology attack that optimizes a sparse adjacency-perturbation matrix over randomized edge blocks while enforcing a global edge budget.

  \item \textbf{GOttack}~\citep{alom2025gottack}, a universal attack that learns graph orbit representations and optimizes a single perturbation pattern that generalizes across many target nodes and even across different victim architectures.   

  % \item \textbf{PGD}~\citep{madry2018pgd}, where we treat the adjacency as a continuous variable, run projected gradient descent to obtain an adversarial adjacency, and then discretize the top-$B$ edge changes around the target node.
  
  % \item \textbf{SGA} (Simplified Gradient-based Attack)~\citep{li2021sga}, a scalable multi-stage gradient attack designed for large-scale graphs.
\end{itemize}

\section{Implementation}\label{sec:imp}
% We implement \pname{}, all baselines, and victim models in PyTorch and PyTorch Geometric. 

% \heading{Computational Environment.}
% Experiments were run using Python 3.8.19 and PyTorch 2.3.0 on a Linux cluster with an AMD EPYC 7313 16-Core Processor (providing 64 logical CPU(s)), 1.0 TiB of RAM, and 8 NVIDIA L40S GPUs, each with 46 GB of dedicated memory. GNNs were implemented using PyTorch Geometric 2.5.3, and author-provided defense methods were implemented in the author-provided repositories not in those libraries.

The bot detector code was adapted from an up-to-date codebase~\footnote{\url{https://github.com/LuoUndergradXJTU/TwiBot-22}} maintained by the authors of~\cite{feng2022twibot22}. This codebase has been used by numerous studies and recent work~\cite{qiao2025botsim}. We implement \pname{} in PyTorch and PyTorch Geometric. \sota adversarial attack code bases~\footnote{\url{https://github.com/danielzuegner/nettack}, \url{https://deeprobust.readthedocs.io/en/latest/_modules/deeprobust/graph/targeted_attack/fga.html}, \url{https://github.com/sigeisler/robustness_of_gnns_at_scale}, \url{https://github.com/cakcora/GOttack}} have been adopted by running the original author's code from their shared repository and only modifying the dataset loaders to execute on large social graphs such as TwiBot-22, which contains 1M nodes.

\subsection{\gnn Adversarial Defenses: Extended Discussion}\label{sec:victim-defense}

% We integrated three \gnn defenses~\footnote{\url{https://github.com/mims-harvard/GNNGuard}, \url{https://github.com/THUDM/GRAND}, \url{https://github.com/ZW-ZHANG/RobustGCN}}: GNNGuard~\citep{zhang2020gnnguard}, GRAND~\citep{feng2020grand}, and RobustGCN~\citep{zhu2019robustgcn}, with our victim \gnn bot detector models. We faithfully reused the original defense implementation and extended it to a heterogeneous bot detector without violating any defense's modeling assumptions.
We integrated three primary \gnn defenses~\footnote{\url{https://github.com/mims-harvard/GNNGuard}, \url{https://github.com/THUDM/GRAND}, \url{https://github.com/ZW-ZHANG/RobustGCN},\url{https://github.com/sennadir/noisygnn}}: GNNGuard~\citep{zhang2020gnnguard}, GRAND~\citep{feng2020grand}, and RobustGCN~\citep{zhu2019robustgcn}, with our victim \gnn bot detector models. We also include NoisyGNN~\citep{ennadir2024noisygnn} as an additional robustness control on the strongest TwiBot-22/BotRGCN setting. We faithfully reused the original defense implementations and extended them to a heterogeneous bot detector without violating any defense's modeling assumptions.

\heading{Common interface and invariants.}
All victim models share a unified forward signature, so we map the raw node features into a shared hidden space using the same projection as the vanilla models. Each \gnn defense then modifies either the adjacency used by the detector (GNNGuard) or appends a defense head that operates on the detector's hidden representation (GRAND, RobustGCN). This design ensures: (i) the defended and vanilla models are comparable (same input projection, same label space), and (ii) the defense-specific changes are localized and auditable.

\heading{Summary:}
The \emph{only} changes relative to the original implementation when extending to heterogeneous \gnn-based bot detectors:
(i) GNNGuard: relation-wise edge pruning;
(ii) GRAND: a post-detector propagation head; and
(iii) RobustGCN: a post-detector Gaussian aggregation head.

\subsubsection{GNNGuard}

GNNGuard computes an attention score per edge via feature cosine similarity, removes edges below a threshold, and returns the pruned edges along with a renormalized edge weight vector \({\alpha}\)~\citep{zhang2020gnnguard}. Crucially, the output edge list is a \emph{subset} of the input edges, and the returned weights align one-to-one with the kept edges.

\headinggi{Why a relational adaptation is necessary.}
Victim models operate on a typed edge set, with \(\texttt{edge type}\in\{0,\dots,R-1\}^E\) indicating relation identity. The original GNNGuard implementation is defined for a single (homogeneous) adjacency. Applying it naively to a concatenated multi-relation edge list would (a) mix relation semantics when pruning and normalization are computed, and (b) potentially break the invariant that \(\texttt{edge type}\) aligns with the filtered edges.

\headinggi{Our extension: per-relation filtering with type-safe reconstruction.}
We implement a helper that:
(1) partitions edges by relation \(r\) using \(\texttt{mask}(\texttt{edge type}=r)\);
(2) runs the original GNNGuard primitive \emph{independently} on each relation-specific edges; and
(3) concatenates the surviving edges across relations.
This preserves two correctness properties: (i) no cross-relation edges are introduced (only deletions), and (ii) \(\texttt{edge type}\) remains perfectly aligned after filtering. As in the original GNNGuard setting, the defense is realized through edge pruning and neighborhood renormalization.

\headinggi{Implementation Details.}
\begin{itemize}[leftmargin=*,noitemsep, topsep=0pt]
  \item \textbf{GCN} and \textbf{BotRGCN.}
  We apply GNNGuard filtering per relation, then run the standard BotRGCN on the filtered edges. 

    \item \textbf{GAT.}
  We apply GNNGuard filtering per relation and then run the GAT on the filtered adjacency matrix.

  \item \textbf{S-HGN.}
  S-HGN uses \texttt{pre\_alpha} to pass attention information from layer 1 to layer 2. This creates an important constraint: the two layers must see a \emph{fixed} edge set so that the attention tensor dimensions match.  Therefore, we \emph{guard once per forward pass} and reuse the same filtered adjacency in both S-HGN layers. This is a correctness-critical choice: guarding separately per layer could change \(E\) between layers and invalidate the \texttt{pre\_alpha} alignment.

\end{itemize}

\headinggi{GNNGuard Correctness.}
Across all three detectors, our implementation satisfies:
(i) \emph{soundness of edits}: the defense only removes edges, never fabricating new inter-node connections;
(ii) \emph{type preservation}: for typed graphs, the relation id of each retained edge is preserved and reattached consistently;
(iii) \emph{shape invariants}: in S-HGN, the edge set is fixed across layers to ensure \texttt{pre\_alpha} tensor compatibility.

\subsubsection{GRAND}
GRAND improves robustness through stochastic random propagation (e.g., node dropout) and consistency regularization across multiple augmented views of the graph~\citep{feng2020grand}. The GRAND implementations expose an inference/encoder path that returns (log-)softmax probabilities rather than raw logits. We expect our detector models to return logits, from which we compute \texttt{CrossEntropy} (CE) loss. Passing normalized probabilities (or \texttt{log\_softmax} outputs) into this logits-based loss would be inconsistent with the intended objective. We therefore introduce a lightweight wrapper, \texttt{FixedGRAND}, a wrapper around the original GRAND that preserves GRAND's hyperparameters (drop-node rate, propagation order, number of samples), implements GRAND's normalized propagation, and returns raw \texttt{logits}.

\headinggi{Implementation Details.}
GRAND's random propagation is defined on an adjacency; it does not require typed message functions. Our design keeps relation awareness in the detectors and applies GRAND as a \emph{regularizing smoothing-and-dropout head} on the resulting representation. Our implementation reproduces the practice of applying GRAND-style consistency regularization on top of learned embeddings while avoiding the redefinition of GRAND for every relation type~\citep{zhang2021scr}. The three victim models: GCN, GAT, BotRGCN, and S-HGN produce hidden states \(\mathbf{H}\), then \texttt{FixedGRAND} performs GRAND propagation on \(\mathbf{H}\) and outputs logits.

\headinggi{GRAND Correctness.}
Our integration ensures that:
(i) the implementation preserves GRAND's propagation mechanism (normalized adjacency + order-\(K\) averaging + node dropout),
(ii) outputs are logits compatible with a CE loss, and
(iii) the detector computations are unchanged, ensuring the defense layer is isolated to the intended GRAND regularization layer.

\subsubsection{RobustGCN}

RobustGCN replaces deterministic aggregation with Gaussian-distributed hidden states that propagate both the mean and variance, and samples node representations for classification~\citep{zhu2019robustgcn}. It additionally provides a KL regularizer to constrain the latent distribution.

\headinggi{Implementation Details.}
Similar to GRAND, the three victim models: GCN, GAT, BotRGCN, and S-HGN produce a hidden representation \(\mathbf{H}\), then apply a \texttt{RobustGCN} head on \(\mathbf{H}\).

\headinggi{RobustGCN Correctness.}
Our implementation ensures that we reuses the original RobustGCN forward pass (mean/variance propagation + sampling), and integrates as a modular head without changing the victim's architecture.

\input{tables/hyperparameter-defense}
\input{tables/hyperparameter-attack}

\section{Hyper-parameters}\label{sec:hyperparameter}
% \autoref{tab:hparams-defense}, \autoref{tab:hparams-defense-variants}, and \autoref{tab:hparams-attack} describe the hyperparameters of bot detectors, their adversarial defended variants, and the adversarial attack methods.

% \autoref{tab:hparams-defense} and \autoref{tab:hparams-defense-variants} report detector and defense hyperparameters, across GCN/GAT/BotRGCN/S-HGN based bot detectors, we largely standardize optimization (Adam, dropout 0.5, weight decay \(3\times10^{-5}\), neighbor sampling 256, patience 50) and train to convergence with large epochs (typically 1000 epochs), so robustness comparisons are not driven by under-training. Architectural differences then reflect model capacity rather than tuning artifacts (e.g., GAT uses 4 layers with 2 heads; RGT increases hidden size to 256 and batch size to 512 while using fewer epochs, 500, to balance compute). 
\autoref{tab:hparams-defense} and \autoref{tab:hparams-defense-variants} report detector and defense hyperparameters, across GCN/GAT/BotRGCN/S-HGN based bot detectors, we largely standardize optimization (Adam, dropout 0.5, weight decay \(3\times10^{-5}\), neighbor sampling 256, patience 50) and train to convergence with large epochs (typically 1000 epochs), so robustness comparisons are not driven by under-training. Architectural differences then reflect model capacity rather than tuning artifacts (e.g., GAT uses 4 layers with 2 heads; RGT increases hidden size to 256 and batch size to 512 while using fewer epochs, 500, to balance compute). We selected \(\varepsilon=0.2\) and \(\tau_{\mathrm{bdry}}=0.1\) from the validation sensitivity analysis in \autoref{tab:sensitivity}; the same settings are then fixed across datasets unless otherwise stated.

The aggregation choice also aligns with the defense mechanism: GRAND variants use \texttt{sum} aggregation (consistent with stochastic node dropping and consistency regularization), whereas most other settings keep \texttt{mean} for stability under neighbor sampling. Defense-specific hyperparameters are intentionally lightweight, GNNGuard's similarity threshold (0.1) without self-loops, GRAND's node drop rate (0.1) with \(\lambda=0.5\) and 5 Monte Carlo samples, RobustGCN's \(\gamma=1.0\) with KL regularization \(5\times10^{-4}\), and NoisyGNN injects additive Gaussian hidden-state noise with $\texttt{noise\_ratio\_1}=0.05$ after the first GCN layer during training only; so any remaining vulnerability is less attributable to fragile tuning and more to the underlying constraints of the adversarial setting.

% \autoref{tab:hparams-attack} summarizes the hyperparameters used to instantiate each adversarial attack baseline under a common evaluation protocol: we attack 50 target nodes per run with a small per-target edit budget \(B\in\{1,3,5\}\). For scalability, all attacks fall back to a localized \(k\)-hop subgraph (with \(k=2\) and at most 50k nodes) when \(|V|\ge 10^5\), otherwise operating on the full graph, which keeps runtime and memory comparable across datasets. Nettack uses a conservative learning-rate cutoff (0.004) and an early-stop rule when no admissible structural edits remain, while FGA is configured for edge rewiring only; PR-BCD follows its standard randomized block-coordinate optimization with 125 epochs, 100 resamples, and a probability-margin loss. In contrast, \pname replaces surrogate gradients entirely with a learned OT guidance signal (no surrogate model), using moderate entropic regularization \(\epsilon=0.2\) with 30 Sinkhorn iterations and a compact cost network \((128,256)\). The additional domain constraint guided edit-policy constraints (e.g., max reuse 3, and prohibiting human\(\rightarrow\)bot follow-backs) bias perturbations toward sparse and operationally plausible edits under social-graph constraints.
\autoref{tab:hparams-attack} summarizes the hyperparameters used to instantiate each adversarial attack baseline under a common evaluation protocol: we attack 50 target nodes per run with a small per-target edit budget \(B\in\{1,3,5\}\). For scalability, all attacks fall back to a localized \(k\)-hop subgraph (with \(k=2\) and at most 50k nodes) when \(|V|\ge 10^5\), otherwise operating on the full graph, which keeps runtime and memory comparable across datasets. Nettack uses a conservative learning-rate cutoff (0.004) and an early-stop rule when no admissible structural edits remain, while FGA is configured for edge rewiring only; PR-BCD follows its standard randomized block-coordinate optimization with 125 epochs, 100 resamples, and a probability-margin loss. LR-BCD follows the local-budget attack of~\citet{gosch2023adversarial}; in our targeted setting, its node-wise local budget is set equal to the per-target budget. FGA-mod and PR-BCD-mod use the same hyperparameters as FGA and PR-BCD but project each candidate update onto the admissible bot$\rightarrow$human edge set before the update is committed. In contrast, \pname replaces surrogate gradients entirely with a learned OT guidance signal (no surrogate model), using moderate entropic regularization \(\epsilon=0.2\) with 30 Sinkhorn iterations and a compact cost network \((128,256)\). The additional domain constraint guided edit-policy constraints (e.g., max reuse 3, and prohibiting human\(\rightarrow\)bot follow-backs) bias perturbations toward sparse and operationally plausible edits under social-graph constraints.

% For OT, we use entropic regularization with $\varepsilon = 0.2$ and Sinkhorn iterations $T_{\mathrm{sink}} = 30$. Hyperparameters $(\lambda_{\mathrm{BCE}}, \lambda_{\mathrm{sp}}, \lambda_{\mathrm{pl}})$ are tuned on the validation split of each dataset.

\section{Dataset}\label{sec:dataset-app}
We evaluated \pname on three widely used social bot detection datasets: TwiBot-22~\citep{feng2022twibot22}, Cresci-2015~\citep{cresci2015fakers}, and BotSim-24~\citep{qiao2025botsim}, with summary statistics reported in \autoref{tab:avg-node-degree}. TwiBot-22 is by far the largest benchmark in our study (\(1\)M nodes and \(170\)M edges), and its scale comes with heterogeneous social signals (multiple relation types and rich user metadata) that more closely resemble real-world detection settings. Consistent with this realism, the degree gap between bots and humans is present (Avg Deg.\ 3.56 vs.\ 7.00), which reduces the effectiveness of trivial heuristics and incentivizes methods that integrate both attribute and relational evidence. Overall, the three datasets span (i) large-scale heterogeneous graphs (TwiBot-22), (ii) sparse early-era spam graphs with sharp structural cues (Cresci-2015), and (iii) modern, dense, human-like interaction simulations where structural context is essential (BotSim-24), enabling a more comprehensive assessment of robustness across realistic operating conditions.

\input{tables/degree}

In contrast, Cresci-2015 is substantially smaller and markedly sparse (Avg Deg.\ 2.05), with bots exhibiting very low connectivity on average (0.22) compared to humans (5.18), reflecting early-generation spam/fake-follower behaviors and producing a clearer structural separation. Together, these two datasets provide complementary regimes: a large, heterogeneous and sparse graph (TwiBot-22) versus a small, sparse graph with strong degree-based signals (Cresci-2015), for stress-testing both effectiveness and scalability.

BotSim-24 is the newest dataset and is designed to simulate highly human-like bot interactions, making the \emph{node features} alone more challenging to distinguish from genuine users; in such cases, leveraging graph structure becomes crucial. This dataset is also notably dense despite its small size (Avg Deg.\ 45.39), and the connectivity patterns are different across classes: humans have very high average degree (59.12) while bots remain substantially connected (19.23), showcasing that bots are embedded in interaction neighborhoods rather than being isolated. Therefore, it increases redundancy in local neighborhoods (potentially stabilizing message passing) while simultaneously reducing the space of plausible structural perturbations available to an adversary.

% \input{tables/victim}
% \section{Bot Detection Performance}
% \autoref{tab:f1-gnn-bots-victim} shows the detection performance of different \gnn-based bot detectors and shows that BotRGCN is the best detector, followed by GAT and S-HGN. 

% \input{tables/defense}
% \section{Bot Detection with GNN Defense Performance}

%For each dataset, we follow the original training recipes for the bot detectors~\citep{feng2021twibot20,feng2022twibot22,feng2021botrgcn,feng2021rgt,yang2024sebot}, including train/validation splits and early-stopping criteria, and verify that our implementation matches or exceeds the reported clean accuracies before evaluating adversarial attacks (as shown in \autoref{tab:f1-gnn-bots-victim} and \autoref{tab:f1-gnn-bots-defense}).

 \begin{figure}[!h]
    \centering
    \resizebox{0.5\linewidth}{!}{%
        \includegraphics{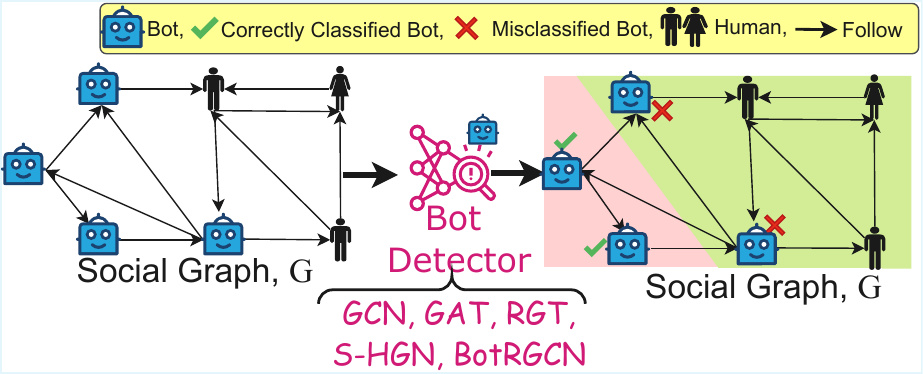}%
    }
    \caption{Overview of bot detector detection pipeline.}
    \label{fig:detect}
\end{figure}

\input{tables/defense}
\section{Bot Detection Performance}
% \autoref{tab:f1-gnn-bots-defense} shows the detection performance of the vanilla \gnn-based bot detectors and their defense architectures. It shows the detection performance of different \gnn-based bot detectors and shows that BotRGCN is the best detector for vanilla, followed by GAT and S-HGN.

\autoref{tab:f1-gnn-bots-defense} reports the F1 scores of vanilla bot detectors and their adversarially defended variants, detection pipeline illustrated in \autoref{fig:detect}. Across datasets, \textbf{BotRGCN} is the strongest overall detector: it attains the top vanilla performance on Cresci-15 (0.98) and BotSim-24 (0.96), and remains among the top performers on TwiBot-22 (0.58, tied for best), suggesting that explicitly modeling heterogeneous relation types is consistently beneficial when interaction semantics matter. GAT and S-HGN form a close second and third best bot detector, frequently appearing as runner-up (e.g., S-HGN is second on Cresci-15 vanilla and GAT is second on BotSim-24 vanilla), indicating that attention or heterophily-aware designs can match BotRGCN when the relational signal is strong.
A consistent pattern as noted by other works~\cite{qiao2025botsim} also, is that absolute F1 scores on TwiBot-22 are lower for all models than on Cresci-15 and BotSim-24, reflecting the larger scale and higher ambiguity of TwiBot-22 where graph size, diverse behaviors, and heterogeneous relations makes behavioral modeling challenging.

Defense variants exhibit mixed behaviors such as on Cresci-15, defenses largely preserve already vanilla performance, with small gains such as GAT+GRAND, consistent with defenses acting as mild regularizers when the classification boundary is easy. In contrast, on TwiBot-22 the best defended results are achieved by vanilla backbones rather than by defenses alone (e.g., BotRGCN remains best), suggesting that robustness mechanisms cannot compensate for a backbone that under-utilizes heterogeneous structure.

But, several defenses \emph{decrease} performance on BotSim-24 for the strongest vanilla models, which is showcases that aggressive perturbation filtering or stochastic regularization can remove informative dense-neighborhood signals in human-like interaction graphs. Overall, \autoref{tab:f1-gnn-bots-defense} suggests that (i) choosing a relation-aware backbone (BotRGCN) provides the most reliable gains, while (ii) defenses behave like regularizers whose benefit depends on its defense procedure.

\input{tables/noisygnn-results}

\section{NoisyGNN Adversarial Robustness}
\label{app:noisygnn-control}

\autoref{tab:noisygnn-results} reports an additional robustness-control experiment using NoisyGNN~\cite{ennadir2024noisygnn} on the strongest BotRGCN detector setting. 
Unlike GNNGuard, GRAND, and RobustGCN, which are evaluated as general defense variants, NoisyGNN is instantiated here as a targeted control on the relation-aware BotRGCN backbone. 
The defense injects Gaussian noise into the hidden representation after the first GCN layer during training and disables this noise during inference.

The results show that NoisyGNN further reduces the success of generic constrained attacks, with Nettack, FGA, PR-BCD, and GOttack remaining below $6\%$ misclassification. In contrast, \pname{} maintains an $85.12\%$ misclassification rate under the same constrained setting.

This suggests that stochastic hidden-state regularization alone does not remove the vulnerability exploited by \pname{}, because \pname{} operates by constructing domain-feasible neighborhood distributions rather than relying on fragile gradient perturbations.

\input{tables/old-node-50-rest}
\section{Bot Editing Results against \sota Bot Detectors}
\autoref{tab:old-node-50-rest} shows the bot editing results using \pname against the second- and third- best \sota bot detectors and their adversarial defense variants. \pname{} remains the strongest attacker across datasets, budgets, and defenses, achieving consistently $\textbf{80.13\%}$ high misclassification rate and substantially outperforming \sota adversarial attacks.

A consistent trend in \autoref{tab:old-node-50-rest} is that domain-constrained \sota adversarial attacks remain in the single-digit to low-teen misclassification range across both GAT and S-HGN and their defenses, even as the edit budget increases. In contrast, \pname{} produces large-scale flips under the same constraints, staying in the high-success regime on Cresci-15 and TwiBot-22 (typically 90--99\%+) across vanilla and defended detector variants. This gap indicates that \pname{} is not merely exploiting unconstrained rewiring freedom; it is effective even when edits must remain operationally feasible under domain constraints.

BotSim-24 consistently behaves as the most difficult dataset. Unlike Cresci-15 and TwiBot-22 (where \pname{} saturates quickly), BotSim-24 shows noticeably lower misclassification at smaller budgets and exhibits larger sensitivity to defense choice. This suggests that, in BotSim-24, producing a human-like neighborhood with sparse feasible edits is intrinsically harder, and robust training/denoising defenses can partially suppress boundary transfer. Importantly, however, the same setting also highlights how limited constrained \sota attacks are: they frequently remain near-zero to low-single-digit success, while \pname{} still achieves substantially higher flip rates, preserving a clear best performer even in the most resistant dataset.

\input{tables/new-node-1}

\input{tables/new-node-def-rest}

\input{tables/old-node-50-samtemplate}
\section{Bot Injection Results against \sota Bot Detectors}
\autoref{tab:new-node-1} illustrates the bot injection results using \pname against \sota bot detectors. \autoref{tab:new-node-def-rest} shows the bot injection results using \pname against the best three \sota bot detectors and their adversarial defense variants. Unlike node editing, there is no established \emph{node-injection} attack for social bot detection under realistic constraints; accordingly, we compare \pname{} to a constrained random injection baseline.

Across Cresci-15 and TwiBot-22, \pname{} achieves near-ceiling injection success even at very small budgets, consistently reaching close to $99\%$ misclassification across \emph{all} vanilla detectors in \autoref{tab:new-node-1}. We demonstrate that \pname{} is (to our knowledge) the first framework to explicitly evaluate and succeed at \emph{domain-constrained node injection} on social bot detectors, rather than restricting attention to perturbing existing nodes. Injection here is constrained to \emph{only} add edges for a new bot, without forcing human follow-backs, so success depends on constructing a locally human-like ego-neighborhood under a tight feasibility budget. 

\autoref{tab:new-node-def-rest} highlights a consistent efficiency gap. On BotRGCN (and its defense variants), random injection remains modest at $\Delta{\le}5$ (e.g., $\approx$31-37\% on Cresci-15/TwiBot-22), but climbs to the mid-90\% range only when allowed a large budget of 20 edges. In contrast, \pname{} reaches 100\% with $\Delta{\le}5$ across all BotRGCN defense variants on both Cresci-15 and TwiBot-22. This is similar for GAT and S-HGN based bot detectors, describing, when the attacker is constrained to realistic edits, \pname{} provides a \emph{high-success, low-budget} injection mechanism that random baselines cannot replicate without substantially more connectivity.

BotSim-24 does not saturate at small budgets. In \autoref{tab:new-node-1}, injection success increases steadily with budget (from $\sim$10--16\% at $\Delta{=}1$ to $\sim$50--56\% at $\Delta{=}5$), and only approaches high success at very large budgets for some detectors. Under defended settings (\autoref{tab:new-node-def-rest}), BotSim-24 remains the most resistant: \pname{} is still far above random at $\Delta{\le}5$, but absolute rates are lower and defenses hurt more than on the other datasets. BotSim-24 has a sharper or sparser ``human manifold'' for injected neighborhoods, so a few arbitrary edges rarely land in the right region, whereas \pname{}’s OT-guided template selection yields meaningful gains that scale with budget.

On the easier benchmarks (Cresci-15/TwiBot-22), the boundary appears relatively easy to cross once a minimally human-like neighborhood is formed, so \pname{} reaches ceiling quickly and additional edges give diminishing returns. Meanwhile, when the attacker is allowed to connect broadly (random with $\Delta{=}20$), the gap to \pname{} can shrink, because brute-force linking can eventually approximate the needed neighborhood structure. On BotSim-24, random remains weak even at large budgets in defended settings, indicating that naive linking often fails to reach the human region; here, OT-guided selection provides the largest advantage by systematically steering injections toward plausible, boundary-adjacent neighborhoods. 

\section{Bot Editing and Injection Results When the Best Bot Template is Reused Every Time}
\autoref{tab:new-node-1-samtemplate}, \autoref{tab:new-node-def-samtemplate}, and \autoref{tab:old-node-50-samtemplate} illustrate the experimentation results when the best bot cloak template is selected every time for node editing and injection. The template-reuse setting effectively approximates an attacker who \emph{cherry-picks} the most human-like bot and mass-produces clones of its neighborhood signature. However, this comes with an important realism caveat: reuse creates a single point of failure. If the best template (or its characteristic neighborhood ``fingerprint'') is ever detected, the same signature can be blacklisted, causing all its clones (and any targets edited toward it) to become detectable as well.

Because the reused template is intentionally chosen from the most human-like region of the detector’s decision boundary, it acts as a highly transferable cloak: once we decode its neighborhood structure into a feasible set of edits, the resulting perturbations repeatedly push many different targets into the human region. This effect is especially pronounced on Cresci-15 and TwiBot-22, where injection becomes essentially solved in this setting: \autoref{tab:new-node-1-samtemplate} shows 100\% misclassification across all vanilla detectors even at small budgets, and \autoref{tab:new-node-def-samtemplate} shows the same near-perfect behavior persists against the best detectors equipped with adversarial defenses. 

A similar trend holds for node editing when the cloak is reused: \autoref{tab:old-node-50-samtemplate} indicates that selecting a single strong cloak can sustain very high flip rates, often remaining best even under standard defenses, consistent with the idea that \pname{} learns a transferable ``bot cloak'' rather than a target-specific perturbation. 

Unlike Cresci-15/TwiBot-22, BotSim-24 does \emph{not} saturate at tiny budgets when reusing one template. In \autoref{tab:new-node-1-samtemplate}, BotSim-24 injection rises gradually with budget (from low success at $\Delta{=}1$ to higher success at larger budgets), and only reaches ceiling at very large budget for some detectors. For node editing, \autoref{tab:old-node-50-samtemplate} similarly reveals that while reuse is generally powerful, some defense combinations can still suppress performance (an example is a pronounced drop under RobustGCN in at least one configuration).

\input{tables/old-node-50-nodomainrules}
\section{Bot Editing Results When No Domain Constraints are Enforced}
\autoref{tab:old-node-50-nodomainrules} reports node-editing performance when we \emph{remove} all domain constraints and allow \sota attacks to freely rewire incident edges around the target bot under the same per-subgraph budgets. This setting is intentionally unrealistic for social platforms, but it is useful as it reveals how much of the previously reported \textit{attack strength} is driven by infeasible edits. In our threat model, \sota attacks collapse to single-digit/low-teen flip rates once we enforce social-domain constraints, which the main paper highlights as a core realism gap in prior graph-attack evaluations. 

Across all datasets and defenses, unconstrained variants of adversarial attacks become highly effective, often approaching ceiling misclassification. Suggesting that if an attacker could violate domain rules, the detector is extremely brittle. This aligns with the paper’s observation that unconstrained attacks can appear strong, but their effectiveness does not survive once feasibility constraints are imposed. 

While unconstrained \sota attacks inflate substantially, \pname{} stays consistently high and is frequently the top performer across datasets/defenses. On BotSim-24 at $\Delta{=}1$, \pname{} is best on vanilla and competitive elsewhere (58.22\% on vanilla), indicating that our cloak-transfer mechanism is not relying on \textit{illegal} rewiring to be effective; it remains potent even when we \emph{allow} that extra flexibility. The unconstrained setting can make \sota attacks seem overwhelmingly strong, but this success is largely driven by edits that are not feasible in real social systems. In contrast, \pname{} is designed around domain-faithful constraints, yet still remains highly effective even when those constraints are lifted and, crucially, the gap between constrained vs.\ unconstrained \sota performance underscores why evaluations that ignore domain rules can substantially mischaracterize real-world robustness.

\section{Constrained Baseline Controls}
\label{app:constraint-controls}

\input{tables/constrained-baseline-controls}

\autoref{tab:constraint-controls} reports additional controls  to showcase that the constrained baseline comparison is fair. 
All attacks are evaluated on TwiBot-22 against BotRGCN in the node-editing setting with budget $\Delta=1$, using the same target set and the same constrained feasible-edit set as the main experiments.

The mod variants test whether stronger constraint integration inside the baseline optimization closes the gap to \pname. 
Specifically, FGA-mod and PR-BCD-mod add an in-loop projection step so that candidate perturbations are restricted to admissible bot$\rightarrow$human edits before they are committed. 
This improves the constrained baselines modestly: FGA increases from $4.00\%$ to $7.11\%$, and PR-BCD increases from $8.67\%$ to $13.21\%$. 
However, both remain far below \pname, which achieves $86.67\%$ under the same budget and feasibility rules.

We also include LR-BCD as a newer local-budget graph attack. 
Although LR-BCD is stronger than PR-BCD in the unconstrained setting, its constrained performance remains low because it is still not designed around social-bot feasibility rules such as target-incident-only edits, no unrelated rewiring, temporal plausibility, and no forced human$\rightarrow$bot follow-backs.

Finally, the Homophily control tests whether the observed gain can be explained by simple neighborhood similarity or embedding matching. 
Although the Homophily baseline performs better than several constrained generic attacks, it still reaches only $13.12\%$ in the constrained setting. 
This supports our claim that OT is useful not merely as a scalar similarity score, but because its transport plan identifies which local neighbor correspondences should be decoded into sparse, feasible edits.

\input{tables/ablation}
\section{Ablation Study}
\autoref{tab:ablation} describes the ablation study of \pname{} against BotRGCN with adversarial defenses on three bot datasets, with a budget $\Delta$ of 5. We measure the impact of the different loss (\eg $L_{\mathrm{BCE}}$, $L_{\mathrm{sp}}$, $L_{\mathrm{pl}}$) in \pname's performance. We selectively deactivate each loss by setting $\lambda_{\mathrm{BCE}}$, $\lambda_{\mathrm{sp}}$, $\lambda_{\mathrm{pl}}$ to zero. We also measure the impact of node degree and node age on $L_{\mathrm{pl}}$ by setting $\alpha_{\deg}$, $\alpha_{\mathrm{age}}$ to zero.

The OT-margin BCE surrogate is the key driver: setting $\lambda_{\mathrm{BCE}}$ to zero causes a sharp collapse in flip rates across all datasets and defense combinations, consistent with its role in explicitly shaping the OT margin so boundary/misclassified templates remain near (or inside) the human region in OT space. The plausibility loss is the next most important, dropping $\lambda_{\mathrm{pl}}$ yields only mid-range success, because it discourages transport mass on socially implausible neighbor matches (\eg large degree/age gaps), which otherwise produces unrealistic alignments that do not transfer into effective edge edits. Finally, within plausibility, degree alignment matters more than age alignment (ablating $\alpha_{\deg}$ deteriorates more than ablating $\alpha_{\mathrm{age}}$), indicating that preserving local connectivity is a stronger constraint on realistic template transfer than temporal calibration alone. This is reflected in the real world, where new accounts tend to follow trusted old accounts, but new accounts do not always have diverse interests, so their follow accounts differ considerably.

\input{tables/sensitivity}

\section{Sensitivity Study}
\autoref{tab:sensitivity} describes the sensitivity of \pname{} against vanilla BotRGCN on three bot datasets with budget $\Delta$ of 5. We measure the impact of the entropic regularizer, $\varepsilon$ that encourages OT decision boundary plans. It controls the plan sharpness: smaller $\varepsilon$ yields near-permutation matchings, whereas larger $\varepsilon$ produces more diffuse couplings.  We also vary the margin threshold ($\tau_{\mathrm{bdry}}$), which sets the required OT margin for a successful cloak: increasing $\tau_{\mathrm{bdry}}$ imposes a stronger constraint (requiring bots to be closer to human templates by a larger margin), which can improve boundary crossing up to a point but may eventually over-constrain optimization.

When $\varepsilon$ is too small (\eg $0.01$), small neighborhood mismatches can cause unstable, high-variance boundary behavior despite occasional strong flips, yielding only $83$-$85\%$ flip rates on TwiBot-22 dataset against BotRGCN, transport remains sharp and diffuse enough to capture multiple plausible neighbor correspondences, producing the strongest and most consistent boundary crossings, with near-ceiling $98$-$100\%$ flip rates on Cresci-15 dataset against BotRGCN. For large $\varepsilon$ ($\geq 10$), the plan washes out toward diffuseness, blurring class structure in the OT-induced margin, so flip rates collapse (\eg $53\%$ to $14\%$ on Cresci-15 as $\varepsilon$ increases $1$ to $100$), indicating a weakened ability to target the boundary.

Across $\tau_{\mathrm{bdry}}$, Cresci-15 benefits from moderate margins: moving from $\tau_{\mathrm{bdry}}{=}0.00$ to $0.10$ lifts flips from $92.50\%$ to $99\%$ while tightening variance, suggesting reduced ``lucky flips'' and more consistent boundary crossings. In contrast, TwiBot-22 and BotSim-24 exhibit diminishing returns at large margins: after peaking around $\tau_{\mathrm{bdry}}{=}0.10$, further increasing $\tau_{\mathrm{bdry}}$ to $0.30$-$0.50$ reduces success (TwiBot-22 down to $90.11\%$, BotSim-24 down to $90.67\%$) and can increase each trail variability, consistent with a harder constraint that leaves optimization persistently penalized and yields plateauing or dropping flip rates.

\input{tables/time-gpu}

\section{System Overhead}\label{app:overhead}
We summarize the system-level cost of running \pname{} and competing \sota attacks in \autoref{tab:time-gpu}. The results show that \pname{} consistently operates with near-negligible GPU usage (9--13MB; \textbf{99.8\%} lower) and achieves up to \textbf{$20\times$} lower runtime across datasets, primarily because it does not invoke GPU-intensive optimization. Moreover, \pname{} remains RAM-efficient by restricting computation to loading the graph and querying the precomputed OT transport plan, whereas several baselines allocate tens of GBs to perform iterative perturbations over graphs and features. This yields predictable, near-constant overhead for \pname{}, unlike gradient/optimization-based methods that scale poorly as graphs grow.

Further, we observe a consistent pattern: optimization-driven attacks (\eg Nettack/PR-BCD/GOttack) incur multi-second runtimes and elevated RAM usage even at a fixed budget, indicating that perturbation search (not mere inference) dominates the end-to-end overhead. This trend becomes more pronounced as dataset scale increases: TwiBot-22 triggers the highest GPU reservations (on the order of multiple GBs) and the largest runtime variability, highlighting strong sensitivity to graph size and iterative optimization dynamics. We noticed that even attacks that appear fast in wall-clock time (\eg FGA) can still reserve substantial GPU memory, underscoring that runtime alone can understate operational cost; in contrast, \pname{} reduces attack generation to lightweight OT-based lookups with predictable, bounded memory.

\input{tables/struc-category}

\begin{figure}[!htb]
  \centering
  \begin{subfigure}[t]{0.40\linewidth}
    \centering
    \resizebox{\linewidth}{!}{\includegraphics{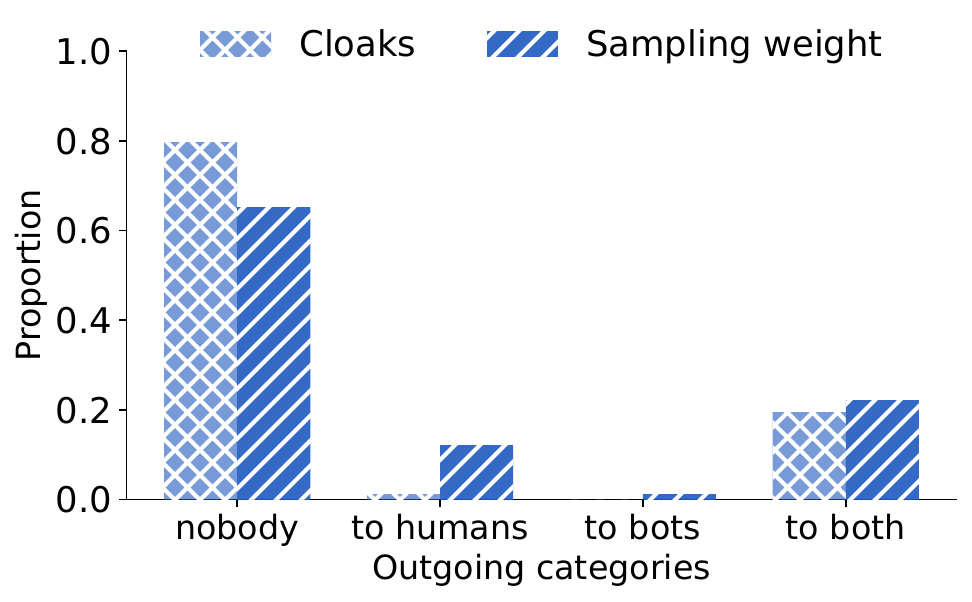}}
    % \caption{Outgoing categories.}
    \label{fig:case-edits-outgoing}
  \end{subfigure}\hfill
  \begin{subfigure}[t]{0.40\linewidth}
    \centering
    \resizebox{\linewidth}{!}{\includegraphics{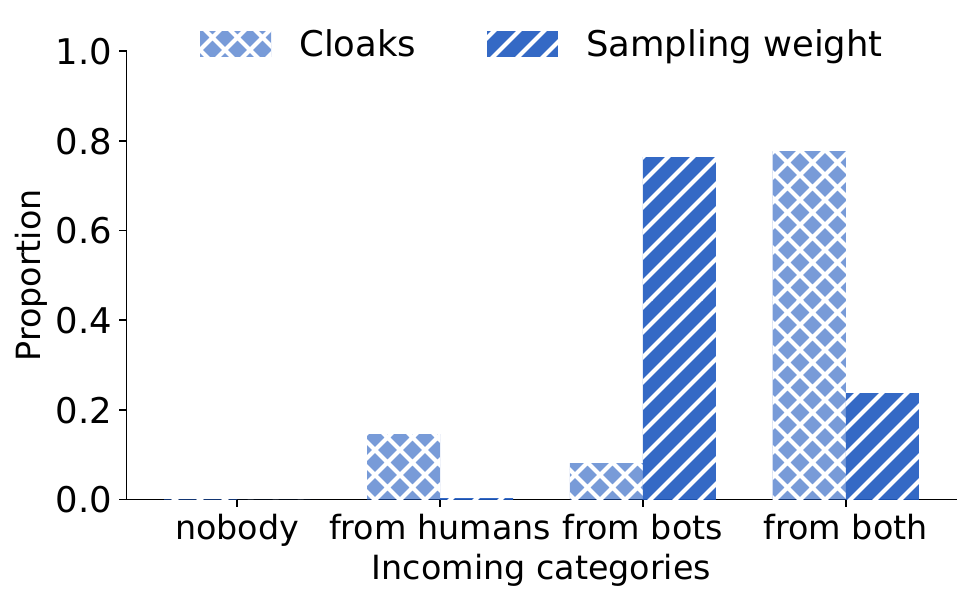}}
    % \caption{Incoming categories.}
    \label{fig:case-edits-incoming}
  \end{subfigure}

  \caption{\textbf{\pname induced edits in practice (Twibot-22).}
  Bars compare (i) the distribution of candidate cloaks across structural categories and  (ii) the normalized sampling mass assigned to those categories (\emph{Sampling weight}).
  (Left) Outgoing follow categories (target bot $\rightarrow$ others).
  (Right) Incoming follower categories (others $\rightarrow$ target bot).}
  \label{fig:case-edits-twibot}
\end{figure}
\section{\pname in Practice}

\heading{Characterizing the edits made by \pname (Twibot-22).}
To validate plausibility beyond the constraint-function definition, we summarize what graph edits \pname makes in a Twibot-22 against BotRGCN for node editing attack using a budget $\Delta{=}5$ over 50 trials. The attack flips 48/50 targets (96\% misclassifications rate) within budget. 

\heading{Template selection and constraint-induced pruning.}
\pname first selects OT-boundary accounts for templating (OT-boundary bots with incident edges $\leq 5$ are computed) and then filters misclassified-bot templates. 3,638 templates remain after filtering, while 11,709 are rejected by constraints and filters (\eg only 23.70\% retained; 76.30\% pruned). This demonstrates that the constraints substantially narrow the feasible set of candidate cloaks in practice.

\heading{Outgoing edges (to humans vs.\ bots).}
Using the structural categories~\autoref{tab:struc-category}, we characterize the different combinations (\emph{type}) of neighborhood structure \pname can mimic. ~\autoref{fig:case-edits-twibot} shows that the candidate cloaks are dominated by categories with \emph{no outgoing follows} (``Outgoing: nobody''). Moreover, the sampler's category weights concentrate mostly on templates with no outgoing edges, implying that \pname typically avoids creating new outgoing follow edges. When outgoing edges are present in the weighted sampler, they are overwhelmingly ``to humans'' rather than ``to bots''.

\heading{Incoming edges (from helper bots).}
The same category decomposition also describes who follows the edited account (incoming neighborhood). In the cloak pool, 85.50\% of templates fall in categories where the account is followed by bots or by a mix of bots and humans (``Incoming: from bots'' or ``Incoming: both''). After applying the sampler's category weights, 95.80\% of the sampling mass lies in these bot-involving incoming categories, indicating that \pname's most common successful mechanism is to realize (or match) \emph{bot-driven incoming followers}; assuming helper-bot follows under the constraint that human$\rightarrow$bot follow creation is disallowed.

\heading{How constraints shape which accounts are chosen.}
\autoref{fig:case-edits-twibot} indicates the feasible cloak accounts concentrate in \emph{low-degree} structural regimes. On the outgoing side (left), the vast majority of selected cloaks fall into the \emph{no-outgoing-follows} category (\texttt{to nobody}), with the remainder largely in \texttt{to both} and \texttt{to humans} , and essentially none in the \texttt{to bots} category. This indicates that, under our constraints, \pname prefers (and is often forced) to avoid creating new outgoing follows and instead rely on edits that minimally perturb the target's outgoing neighborhood.

On the incoming side (right), the sampling distribution places most of its mass on categories where the target is followed by bots (\texttt{from bots}), yet many candidate cloaks are dominated by \texttt{from both}. This mismatch is consistent with constraint-induced filtering: because we disallow creating human$\rightarrow$target bot follows, achieving a \texttt{from both} structure typically requires selecting cloaks that already have some human followers, while the remaining required signal is supplied via helper-bot incoming follows. In other words, the constraints steer \pname toward accounts whose local structure can be matched with a small number of bot-driven incoming edges without introducing implausible outgoing rewiring.

\begin{figure}[t]
  \centering
  \resizebox{0.7\linewidth}{!}{%
    \includegraphics{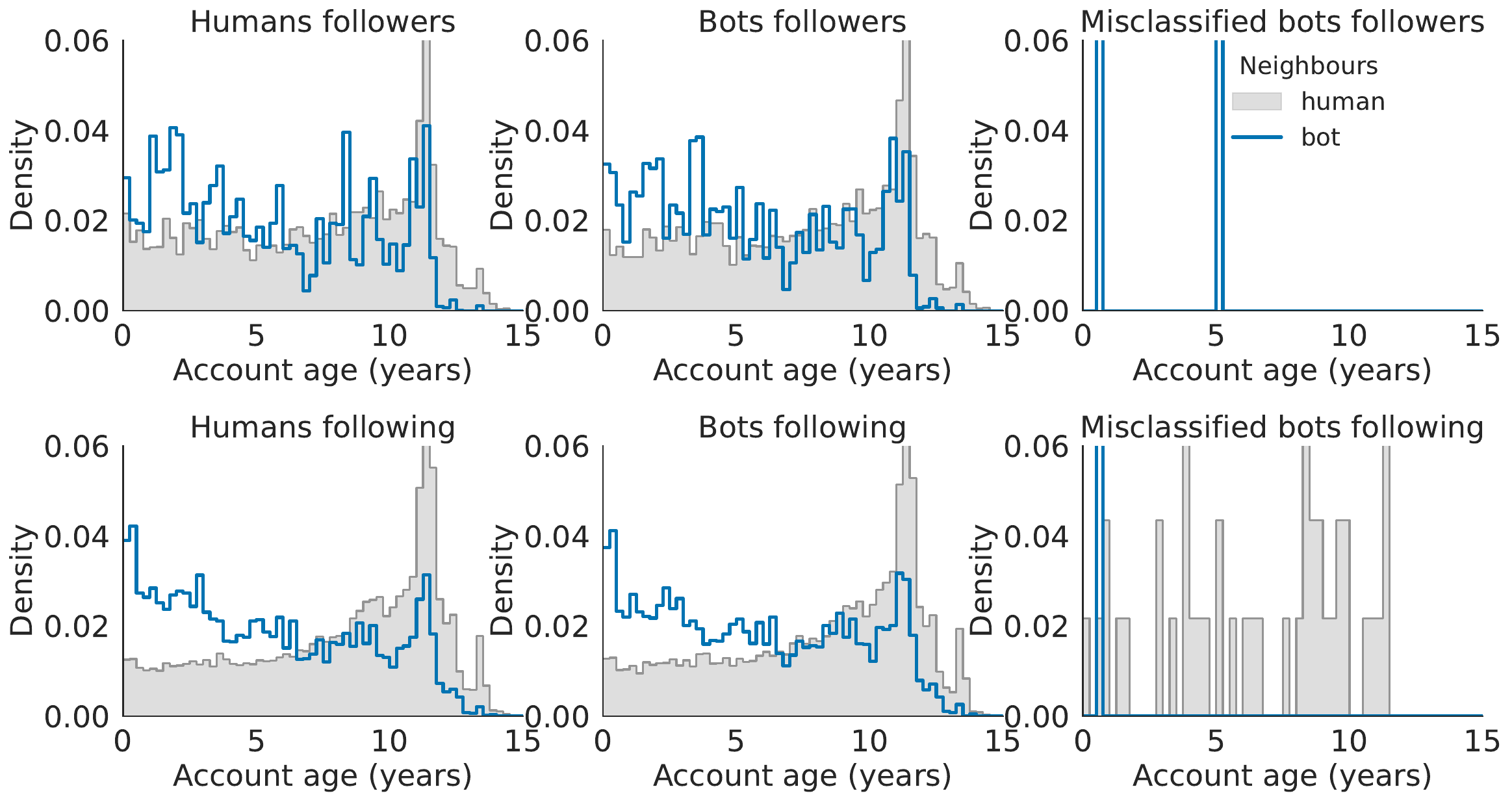}%
  }
  \caption{\textbf{Temporal distribution of neighbor accounts for TwiBot-22}. Top row shows followers; bottom row shows following. Shaded histograms represent human neighbors and outlined curves represent bot neighbors.}
  \label{fig:temp-hist}
\end{figure}

\heading{\pname temporal alignment for cloak selection}.
\autoref{fig:temp-hist} compares neighbor account-age distributions (followers and following) for (i) true humans, (ii) true bots, and (iii) misclassified bots after \pname attack. The temporal alignment constraint does not only shrink the candidate pool; it also biases \emph{which ages} are preferred when constructing human-looking neighborhoods. In particular, for misclassified bots we observe a consistent directional shift: selected \emph{human} neighbors skew toward \emph{older} accounts, whereas selected \emph{bot} neighbors skew toward \emph{newer} accounts. This effect is most visible in the following distributions, where human-neighbor mass for misclassified bots concentrates at higher ages, while bot neighbors appear closer to the low-age end.

This behavior reflects the interaction between temporal plausibility and the attack objective. Because \pname can only add edges that satisfy temporal alignment, it leverages temporally consistent accounts to move the target's neighborhood statistics toward those of humans. Older human accounts contribute stable, human-characteristic temporal signatures that help pull the target toward the human region of the decision boundary, while newer bot accounts are comparatively easier to match under temporal constraints and can satisfy connectivity requirements without introducing implausible temporal patterns. 

\begin{figure}[!htb]
  \centering
  \resizebox{0.7\linewidth}{!}{%
    \includegraphics{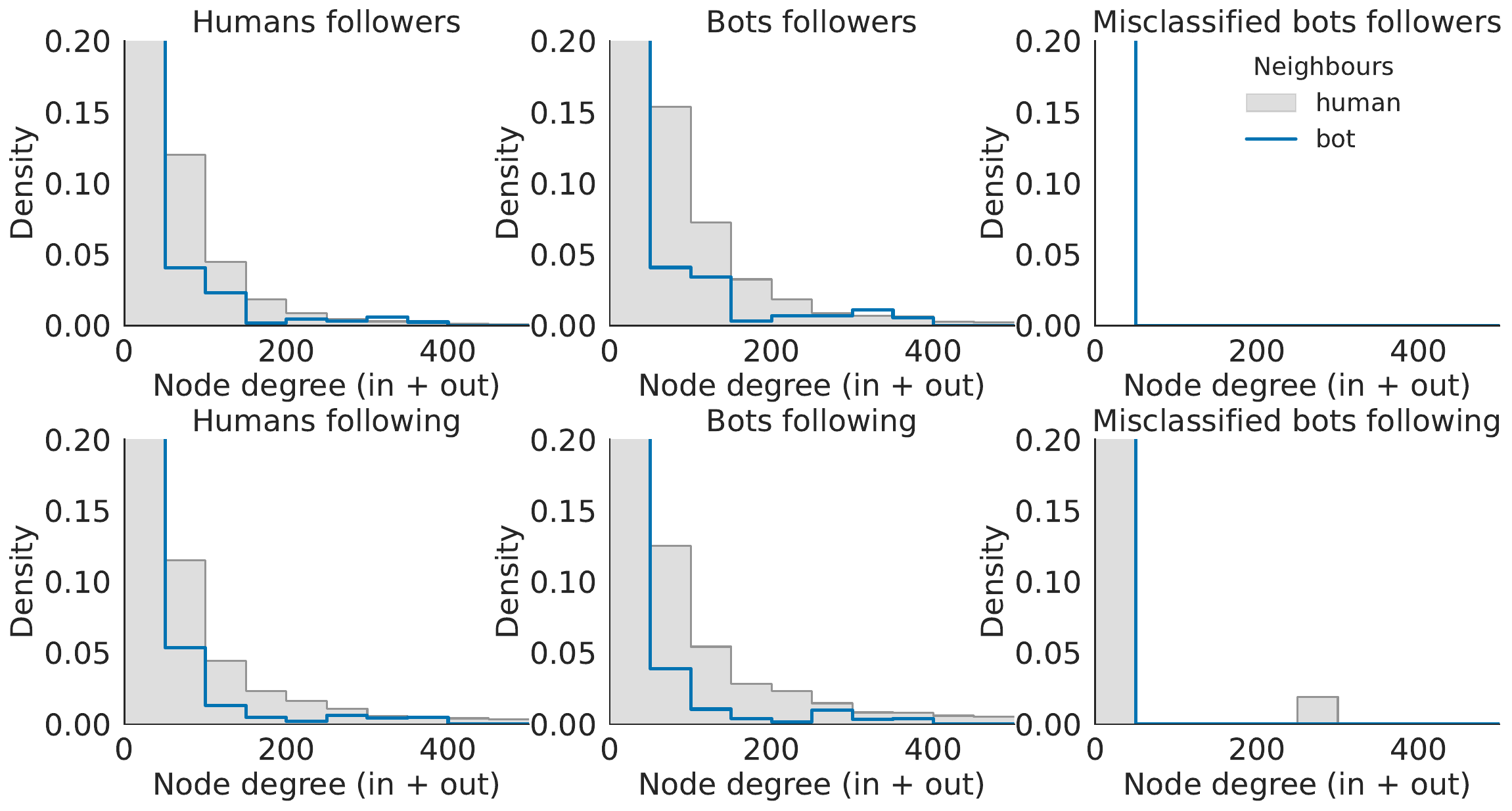}%
  }
  \caption{\textbf{Degree distribution of neighbor accounts for TwiBot-22}. Top row shows followers; bottom row shows following. Shaded histograms represent human neighbors and outlined curves represent bot neighbors.}
  \label{fig:deg-hist}
\end{figure}

\heading{\pname degree alignment for cloak selection}.
\autoref{fig:deg-hist} compares neighbor degree distributions (followers and following) for (i) true humans, (ii) true bots, and (iii) misclassified bots after the \pname attack. A consistent separation is that true bots are associated with heavier-tailed neighbor degrees than true humans, suggesting that bots are embedded in denser local structures (more high-degree neighbors). For misclassified bots, the distributions shift toward \emph{lower-degree} neighbors, especially in the followers neighbors, indicating that successful cloaking tends to attach the target to accounts with low connectivity that better match the human neighborhood in terms of degree.

We also suggests an asymmetry between followers and following in \autoref{fig:deg-hist}. The follower-side distributions show a clearer move away from the high-degree tail, whereas the following-side distributions are comparatively less shifted, implying that follower cloaks are a more effective (or less constrained) lever for altering degree-based signals at the target. Additionally, the misclassified-bot appear sparser than the true-class panels, consistent with the fact that only a small number of edges are added per target by \pname; with limited budget, \pname concentrates its additions on the most degree-informative connections, rather than attempting to reproduce the full human degree distribution. Overall, \pname succeeds by selectively using neighborhood degree composition, primarily by favoring low-degree follower connections to reduce bot-like structural signatures while maintaining plausible connectivity.

\input{figs/test-fig}
\section{\pname Case Study}\label{sec:case-study}

\heading{Target bot 81779 and (cloak 41575).}
Case study in \autoref{fig:case-study} illustrated that target bot 81779 is a low-degree node: it has total degree $5$ (in-degree $2$, out-degree $3$), and its in-neighbors are labeled as bot and two out-neighbors are bots and one is human. Temporally, the target account is relatively old (about $5.59$ years), and its average 1-hop neighbors are also mature (about $4.96$ years). In contrast, the chosen cloak 41575 1-hop neighbor is sparse (degree $4$, dominated by outgoing edges) and younger (about $2.64$ years).

\heading{Attack outcome and cloak signature.}
The detector initially predicts the target as bot, but flips to human after the cloak is applied. The cloaking constraints signature shows four edges are added, with the label breakdown indicating two newly added incoming edges from bots and two outgoing edge to a bot and a human. The target's incident edges are removed upon cloaking, these four edited edges effectively dominate the target's observable neighborhood, a regime where small absolute edge edits become large relative perturbations resulting in misclassification.

\heading{Neighborhood interpretation.}
Despite the target's sparsity, it is embedded in a large 2-hop neighborhood: target 81779 hop-2 nodes, which are predominantly human (about $942$ humans, $78$ bots; \ie $\approx 92\%$ human and $\approx 7.6\%$ bot). The 2-hop neighbors are also temporally mature (median account age $\approx 7.86$ years), suggesting the attack is not exploiting a locally bot-dense region. Instead, this case highlights a brittle operating point for low-degree nodes: when the pre-attack ego signal is effectively a single edge, a small number of constraint-matched additions (combined with an edge wipe) can overwhelm the evidence the classifier uses and induce a label flip.

\section{Defense against \pname}

\pname-inspired  style evasion is effective because it reshapes a node's \emph{neighborhood distribution} (rather than only a few individual edges) to look human-like while keeping edits sparse and socially plausible. A practical defense is therefore to (i) add an explicit \emph{cloak detector} that compares each account's $k$-hop ego-neighborhood distribution to human/bot reference distributions, (ii) adversarially harden the detector using \emph{feasible} perturbations that match the attacker constraints (incident-edge edits, budgeted, direction/temporal plausibility), and (iii) for detection complement graph structure with temporally constant behavioral/content signals that are harder to keep consistent under long-lived evasion. 

%% file: tables/notation.tex
\begin{landscape}
\thispagestyle{plain}
\centering
\small
\setlength{\tabcolsep}{4pt}
\renewcommand{\arraystretch}{0.01}
\captionsetup{type=table}
\captionof{table}{Summary of notations used in this paper.}
\begin{tabular}{@{}p{0.49\linewidth}@{\hspace{0.02\linewidth}}p{0.49\linewidth}@{}}

% ---------------- LEFT BLOCK ----------------
\begin{minipage}[t]{\linewidth}
\label{tab:compact_notations}
\vspace{0.4em}

\begin{tabularx}{\linewidth}{@{}lX@{}}
\toprule
\textbf{Symbol} & \textbf{Description} \\
\midrule
\multicolumn{2}{c}{\textbf{Graph and Nodes}} \\
\midrule
$\mathcal{G}=(\mathcal{V},\mathcal{E}, \mathbf{X})$ & Directed social graph with node set $\mathcal{V}$, edge set $\mathcal{E}$, and node features $\mathbf{X}$ \\
$v$ & Generic node \\
$v_{\mathrm{t}}$ & Target bot node \\
$\xi$ & Generic comparison/anchor node (e.g., template or human) \\
$\mathcal{V}_{\mathrm{bot}}, \mathcal{V}_{\mathrm{hum}}$ & Sets of bot and human nodes \\
$\mathcal{N}(v)$ & $k$-hop ego neighborhood of node $v$ \\
$N^{\mathrm{out}}(v)$ & Out-neighborhood (1-hop outgoing neighbors) of node $v$ \\
$u_i$ & The $i$-th neighbor (used in helper selection over $N^{\mathrm{out}}(v_{\mathrm{t}})$) \\
\midrule

\multicolumn{2}{c}{\textbf{Neighbor Features and Measures}} \\
\midrule
$\phi_v(\eta)$ & Spatio-temporal feature vector of neighbor $\eta$ relative to $v$ \\
$z_i$ & Shorthand for a neighbor feature vector (e.g., $z_i=\phi_v(u_i)$) \\
$d$ & Dimensionality of neighbor feature vectors \\
$z_i \in \mathbb{R}^d$ & Neighbor feature space and dimension \\
$w_v(\eta)$ & Normalized importance weight of neighbor $\eta$ \\
$\mu_v$ & Neighborhood probability measure of node $v$ (weighted empirical distribution) \\
$a,b$ & Discrete marginal weight vectors for two neighborhoods (entries $a_i$, $b_j$) \\
$m,n$ & Neighborhood sizes for $\mu_v$ and $\mu_w$ (numbers of atoms) \\
\midrule

\multicolumn{2}{c}{\textbf{Optimal Transport Geometry (Entropic OT)}} \\
\midrule
$h_\theta(\cdot)$ & Neural embedding used to define the ground cost \\
$c_\theta(z,z')$ & Learned ground cost between neighbor feature vectors \\
$C_{ij}$ & Ground-cost matrix entry, typically $C_{ij}=c_\theta(z_i,z'_j)$ \\
$\mathbf{M}$ & Positive semidefinite matrix defining Mahalanobis distance \\
$\mathbf{L}$ & Learnable factor to parameterize $\mathbf{M}$ (e.g., $\mathbf{M}=\mathbf{L}^\top\mathbf{L}$) \\
$\operatorname{OT}_\varepsilon(\mu_v,\mu_w)$ & Entropic-regularized OT objective between neighborhood measures \\
$\mathcal{U}(a,b)$ & Transportation polytope: couplings with marginals $a$ and $b$ \\
$P^\star_{vw}$ & Entropic optimal transport plan between $\mu_v$ and $\mu_w$ \\
$P_{ij}$ & $(i,j)$ entry of a coupling/plan matrix $P$ \\
$D_\theta(v,w)$ & OT-based distance between nodes $v$ and $w$ \\
$\varepsilon$ & Entropic regularization parameter \\
$K_{ij}$ & Gibbs kernel entry: $K_{ij}=\exp(-C_{ij}/\varepsilon)$ \\
$u,v$ & Sinkhorn scaling vectors; $P \approx \mathrm{diag}(u)\,K\,\mathrm{diag}(v)$ \\
$u^{(t)}, v^{(t)}$ & Sinkhorn iterates at iteration $t$ \\
$T_{\mathrm{sink}}$ & Number of Sinkhorn iterations \\
$\odot$ & Element-wise (Hadamard) product \\
$\mathrm{diag}(\cdot)$ & Diagonal matrix formed from a vector \\
$\mathbf{1}_k$ & All-ones vector of length $k$ \\
$\mathrm{KL}(P\|K)$ & KL divergence used in the entropic OT reformulation \\
% \midrule

\bottomrule
\end{tabularx}
\end{minipage}

&

% ---------------- RIGHT BLOCK ----------------
\begin{minipage}[t]{\linewidth}
\label{tab:compact_notations_right}
\vspace{0.4em}

\begin{tabularx}{\linewidth}{@{}lX@{}}
\toprule
\textbf{Symbol} & \textbf{Description} \\
\midrule
\multicolumn{2}{c}{\textbf{OT Margins and Cloak Templates}} \\
\midrule
$d_{\mathrm{hum}}(v)$, $d_H(v)$ & Distance from bot $v$ to nearest human in OT space \\
$d_{\mathrm{bot}}(v)$, $d_B(v)$ & Distance from bot $v$ to nearest bot in OT space \\
$m(v)$ & OT margin between human and bot regions for node $v$ \\
$\mathcal{B}_{\mathrm{bdry}}$ & Boundary bot set under OT geometry \\
$\tau_{\mathrm{bdry}}$ & Margin threshold used to define boundary bots \\
$\mathcal{B}_{\mathrm{mis}}$ & Bots misclassified as human by the detector \\
$\mathcal{B}$ & Candidate cloak template pool \\
\midrule

\multicolumn{2}{c}{\textbf{Attack Objectives and Constraints}} \\
\midrule
$\Delta\mathcal{E}$ & Set of edge edits incident to the target node \\
$B$ & Edge budget for the attack \\
$\mathcal{F}(B)$ & Feasible set of edge edits under budget and domain constraints \\
$\Phi(\Delta\mathcal{E})$ & Plausibility penalty for edge edits \\
$\lambda_{\mathrm{BCE}},
\lambda_{\mathrm{pl}}, \lambda_{\mathrm{sp}}$ & BCE, plausibility and sparsity weights \\
\midrule

\multicolumn{2}{c}{\textbf{Feature-Weighting Functions (Appendix)}} \\
\midrule
$\deg_{\mathrm{in}}(u), \deg_{\mathrm{out}}(u)$ & In-/out-degree statistics for node $u$ \\
$\deg_{\mathrm{raw}}(u)$ & Raw degree used in weighting functions \\
$\mathrm{age\_norm}(u)$ & Normalized account age feature used in weighting \\
$g_{\mathrm{deg}}(\cdot), g_{\mathrm{time}}(\cdot)$ & Degree-based and time-based importance functions \\
$\alpha_{\mathrm{deg}}, \alpha_{\mathrm{time}}$ & Hyperparameters controlling degree/time weighting strength \\
\midrule
\multicolumn{2}{c}{\textbf{Decoding, Construction, and Helper Selection}} \\
\midrule
$v_c$ & Selected cloak template bot \\
$v_h$ & Nearest human anchor in OT space \\
$h^\star(v)$ & Nearest-human selector in OT space \\
$P^\star_{ch}$ & OT plan between template bot and nearest human \\
$\mu_{\mathrm{t}}(\Delta\mathcal{E})$ & Neighborhood measure of $v_{\mathrm{t}}$ after edits $\Delta\mathcal{E}$ \\
$\rho_i$ & Row mass in $P^\star$ used for guided neighbor selection \\
$\rho$ & Vector of row masses $(\rho_i)_i$ \\
$\mathcal{U}$ & Selected neighbor subset returned by helper routine \\
$\mathrm{TopK}(\rho,K)$ & Operator returning indices of the largest $K$ entries of $\rho$ \\
$K$ & Top-$K$ selection size \\
\midrule

\multicolumn{2}{c}{\textbf{Classifier, Training Pairs, and Surrogates}} \\
\midrule
$f_\Theta$ & Trained node classifier/detector (parameters $\Theta$) \\
$p_\Theta(y \mid v; \mathcal{G})$ & Predicted class probability for label $y$ at node $v$ in graph $\mathcal{G}$ \\
$y_v$ & Ground-truth node label (e.g., \textsf{bot}/\textsf{human}) \\
$\Pi^+$, $\Pi^-$ & Positive and negative (different-class) training pair sets \\
$\gamma$ & Margin hyperparameter used in OT-geometry training loss \\
$\sigma(\cdot)$ & Logistic sigmoid function \\
$\tau_{\text{BCE}}$ & Temperature hyperparameter to regulate sigmoid function \\
$\Delta_{e, subg}$, $\Delta$ & Edge budget, max number of modified edges by \pname \\
\bottomrule
\end{tabularx}
\end{minipage}

\\
\end{tabular}

\end{landscape}

%% file: algos/train-ot.tex
% ------------------------------------------------------------
\begin{algorithm}[t]
\caption{\textsc{TrainOTGeometry}$(G,\mathcal V_{\mathrm{train}},\mathcal V,\mathcal V',\varepsilon,T_{\mathrm{sink}},\tau_{\mathrm{BCE}},, \tau_{\mathrm{bdry}},
\lambda_{\mathrm{BCE}},\lambda_{\mathrm{sp}},\lambda_{\mathrm{pl}})$}
\label{alg:train-ot-geometry}
\begin{algorithmic}[1]
\STATE {\bfseries Input:}
Graph $G=(\mathcal V,\mathcal E)$;
ground-truth labels $\mathcal V=\mathcal H\cup\mathcal B$;
predicted labels $\mathcal V=\mathcal H'\cup\mathcal B'$;
training node set $\mathcal V_{\mathrm{train}}\subseteq\mathcal V$ (typically bots);
entropic OT parameter $\varepsilon>0$;
Sigmoid temperature $\tau_{\mathrm{BCE}}>0$;
Boundary margin $\tau_{\mathrm{bdry}}>0$;
Sinkhorn iterations $T_{\mathrm{sink}}$;
loss weights $\lambda_{\mathrm{BCE}},\lambda_{\mathrm{sp}},\lambda_{\mathrm{pl}}\ge 0$.

\STATE {\bfseries Output:} trained OT geometry $(\theta,\mathbf M)$ where $\mathbf M=\mathbf L^\top \mathbf L\succeq 0$.

\vspace{2pt}
\STATE Initialize neural network parameters $\theta$ and matrix factor $\mathbf L$ (so $\mathbf M=\mathbf L^\top\mathbf L$).
\STATE Precompute neighbor features $\phi_v(u)$ and importance weights $w_v(u)$ for all $v\in\mathcal V_{\mathrm{train}}$.
\FOR{each training epoch}
    \FOR{each minibatch $\mathcal S\subseteq \mathcal V_{\mathrm{train}}\cap\mathcal B$}
        \STATE $L \leftarrow 0$
        \FOR{each bot $v\in\mathcal S$}
            \STATE Build weighted neighborhood measure $\mu_v=\sum_{u\in\mathcal N(v)} w_v(u)\,\delta_{\phi_v(u)}$ \hfill // \autoref{eq:mu-weighted}
            % \STATE (Optional) Restrict $\mathcal N(v)$ to Top-$K_{\mathrm{nbr}}$ by $w_v(\cdot)$ for efficiency.

            \STATE Set misclassification label
            $y_v \leftarrow \mathbbm 1[v\in\mathcal H']$ \hfill // true bot predicted human

            \STATE Find nearest human:
            $h^\star(v)\leftarrow \arg\min_{h\in\mathcal H} D_\theta(v,h)$ \hfill // \autoref{eq:ot-cost-p}
            \STATE Find nearest other bot:
            $b^\star(v)\leftarrow \arg\min_{b\in\mathcal B\setminus\{v\}} D_\theta(v,b)$

            \STATE Compute margin:
            $m(v)\leftarrow D_\theta(v,h^\star(v)) - D_\theta(v,b^\star(v))$ \hfill // \autoref{subsec:ot-margin-main}

            \STATE Compute $L_{\mathrm{BCE}}(v;\theta)$ from margin and $\tau_{\mathrm{bdry}}$ and $\tau_{\mathrm{BCE}}$ \hfill // \autoref{eq:loss-bce}
            \STATE Compute entropic OT plan $P^\star_{v,h^\star(v)}$ via Sinkhorn under $(\theta,\mathbf M,\varepsilon)$
            \STATE Compute sparsity loss $L_{\mathrm{sp}}(v;\theta)$ from plan entropies $P^\star_{v,h^\star(v)}$ \hfill // \autoref{eq:loss-sparsity-entropy}
            \STATE Compute plausibility loss $L_{\mathrm{pl}}(v;\theta)$ from plan entropies $P^\star_{v,h^\star(v)}$ \hfill // \autoref{eq:pl-loss}

            for $T_{\mathrm{sink}}$ iterations \hfill // \autoref{sec:entropic-ot}, \autoref{eq:app-u-update}--\autoref{eq:app-v-update}

            \STATE Accumulate:
            $L \leftarrow L + \lambda_{\mathrm{BCE}}L_{\mathrm{BCE}}(v;\theta)
                      + \lambda_{\mathrm{sp}}L_{\mathrm{sp}}(v;\theta)
                      + \lambda_{\mathrm{pl}}L_{\mathrm{pl}}(v;\theta)$
        \ENDFOR

        \STATE Gradient step on $(\theta,\mathbf L)$ using $L$ // backprop through Sinkhorn (\autoref{sec:entropic-ot}).
    \ENDFOR
\ENDFOR
\STATE \textbf{return} $(\theta,\mathbf M=\mathbf L^\top\mathbf L)$.
\end{algorithmic}
\end{algorithm}

%% file: algos/newbot-algo.tex
\begin{algorithm}[t]
%\caption{\pname Node Injection Attack. $(G, \mathcal V, \mathcal V', \theta, M, \tau_{\mathrm{bdry}},degree_{node})$}
\caption{\pname $(G, \mathcal V, \mathcal V', \theta, M, \varepsilon, TopK, \Delta, R, \texttt{flag\_hb},  \tau_{\mathrm{bdry}},degree_{node}, T)$}

\label{alg:bocloak-newbot}
\begin{algorithmic}[1]
\STATE {\bfseries Input:}
Graph $G=(\mathcal V,\mathcal E)$;
ground-truth labels $\mathcal V=\mathcal H\cup\mathcal B$;
predictions labels $\mathcal V'=\mathcal H'\cup\mathcal B'$;
trained OT geometry $(\theta, M)$ (obtained via \autoref{alg:train-ot-geometry}),
target node $v_{tar}$
OT params $(\varepsilon,TopK)$;
budget $\Delta$;
max reuse bot $R$;
flag \texttt{flag\_hb} to not allow human follow back;
boundary threshold $\tau_{\mathrm{bdry}}$;
max cloak bot node degree $degree_{node}$; and 
number of trials $T$.
\STATE {\bfseries Output:} attack statistics and successful bot cloaks.

\vspace{2pt}
% \STATE $\mathcal{T} \leftarrow \textsc{GetBoundaryTemp}(\theta, M, \varepsilon, K, V, V')$
\STATE $\mathcal{B}_{\mathrm{candidate}} \leftarrow \textsc{getBoundaryBotCandidates}(G, \mathcal V, \mathcal V', \theta, M, \tau_{\mathrm{bdry}},degree_{node})$ // \texttt{using} \autoref{alg:ot-margins}
\STATE $w_{candidate} \leftarrow \textsc{getImportanceWights}(\mathcal \mathcal{B}_{\mathrm{candidate}})$  
// \texttt{using} \autoref{alg:helper-getImportanceWeights}
\STATE Initialize template use-counts $u(t)\leftarrow 0:$ $\forall$ $t\in\mathcal \mathcal{B}_{\mathrm{candidate}}$
\STATE Initialize successful misclassified bot cloaks $B_{succesful}  \leftarrow \emptyset$
\FOR{$\_ \gets 1$ \textbf{to} $T$}
\STATE $c \leftarrow \textsc{SampleCloak}(\mathcal{B}_{\mathrm{candidate}}, w_{candidate},u, R)$ // \texttt{using} \autoref{alg:helper-sampletemp}
% \STATE $h^\star \leftarrow \textsc{NearestHuman}(\theta, M, t)$
\STATE $\mathcal U \leftarrow \textsc{OTGuidedNeighbors}(\theta, M, \varepsilon, TopK, G,c)$  // \texttt{using} \autoref{alg:helper-otguidedneighbors}
% \COMMENT{$\mathcal U=\varnothing$ means ``no restriction''}
\STATE $G', E_{\mathrm{add}}, v_{\mathrm{new}} \leftarrow \textsc{CloneCloak}(G, v_{tar}, c,\mathcal U, \texttt{flag\_hb})$ // \texttt{using} \autoref{alg:helper-CloneCloak}
\IF{$|E_{\mathrm{add}}| > \Delta$}
 \STATE record \code{budget exceeded} 
 \STATE \textbf{continue}
\ENDIF
% \STATE predict $\hat y(v_{\mathrm{new}})$ on $G'$.
\IF{$v_{\mathrm{new}}\neq\mathcal Bot$} %\COMMENT{misclassified as human}
 \STATE record \code{success}, $c$, $E_{\mathrm{add}}$, and $v_{\mathrm{new}}$
 \STATE $B_{succesful} \leftarrow B_{succesful} \cup \{[c, E_{\mathrm{add}}, v_{\mathrm{new}}]\}$
 \STATE $u(t)\leftarrow u(t)+1$
\ENDIF
\ENDFOR
\STATE \textbf{return} $B_{successful}$.
\end{algorithmic}
\end{algorithm}

%% file: sections/bocloak_helper_function.tex
\section{Algorithm~\ref{alg:bocloak-newbot}: Extended Discussion}
\label{sec:newbot-helpers}

\heading{Helper Functions.}
We define the helper routines invoked by \autoref{alg:bocloak-newbot} (Node Editing Attack). We (i) construct a boundary-ranked cloak set via OT geometry (\autoref{app:helper-bdry-bots}), (ii) derive importance sampling distributions (\autoref{app:helper-getImportanceWeights}), (iii) sample cloaks under a reuse cap (\autoref{app:helper-sampletemp}), (iv) compute OT-guided neighbor restrictions (\autoref{app:helper-otguidedneighbors}), including the nearest-human, and (v) clone the selected cloak (\autoref{app:helper-CloneCloak}). Underlying OT definitions (neighborhood measures, ground cost, entropic OT, and OT-margin) are detailed in ~\autoref{sec:ground-cost}--\autoref{subsec:ot-margin-main} and \autoref{app:ot-lagrangian}.

Let $\mathcal G=(\mathcal V,\mathcal E)$ be a directed social graph. Ground-truth labels induce a partition $\mathcal V=\mathcal H\cup\mathcal B$ (humans/bots), predicted labels induce $\mathcal V=\mathcal H'\cup\mathcal B'$ and target node $v_{\mathrm{tar}}$. The trained OT geometry is denoted by $(\theta,M)$, and OT hyperparameters are $(\varepsilon,K)$ where $K$ is the neighbor budget used for OT-guided restriction.
% ------------------------------------------------------------
% \input{algos/ot-margin-algo}
\begin{algorithm}[t]
  \caption{\textsc{getBoundaryBotCandidates}$(G, \mathcal V, \mathcal V', \theta, M, \tau_{\mathrm{bdry}},degree_{node})$}
  \label{alg:ot-margins}
  \begin{algorithmic}[1]
     \STATE {\bfseries Input:}
        Graph $\mathcal{G} = (\mathcal{V},\mathcal{E})$;
        ground-truth labels $\mathcal{V}=\mathcal{H}\cup\mathcal{B}$ (humans and bots);
        predicted labels $\mathcal{V}=\mathcal{H'}\cup\mathcal{B'}$ (predicted humans and bots);
        trained OT geometry $(\theta, M)$ (trained via \autoref{alg:train-ot-geometry}); its boundary threshold $\tau_{\mathrm{bdry}}$ and max boundary bot node degree $degree_{node}$.
    \STATE {\bfseries Output:}
        % OT margins $\{m(b)\}_{b\in\mathcal{B}}$,
        boundary bot candidates $\mathcal{B}_{\mathrm{candidate}}$,
        % misclassified bot pool $\mathcal{B}_{\mathrm{mis}}$
    \STATE $\mathcal{B}_{\mathrm{bdry}} \leftarrow \emptyset$
    \STATE $\mathcal{B}_{\mathrm{mis}} \leftarrow \emptyset$
    \FOR{each bot $b \in \mathcal{B}$}
        \IF{\textsc{GetNodeDegree}(b) $\leq degree_{node}$}
            \STATE build neighborhood measure $\mu_b$
            \STATE compute $d_H(b)$ and $d_B(b)$
            \STATE set $m(b) \leftarrow d_H(b) - d_B(b)$
            \IF{$m(b) \le \tau_{\mathrm{bdry}}$}
                \STATE $\mathcal{B}_{\mathrm{bdry}} \leftarrow \mathcal{B}_{\mathrm{bdry}} \cup \{b\}$
            \ENDIF
            \IF{$b \notin \mathcal{B'}$}
                \STATE $\mathcal{B}_{\mathrm{mis}} \leftarrow \mathcal{B}_{\mathrm{mis}} \cup \{b\}$
            \ENDIF
      \ENDIF
    \ENDFOR
    \STATE \textbf{return} \textsc{Sorted}$\left(\mathcal{B}_{\mathrm{bdry}} \cap \mathcal{B}_{\mathrm{mis}},\ \text{key}=-m(\cdot),\ \text{order}=\text{descending}\right)$
  \end{algorithmic}
\end{algorithm}

\subsection{\textsc{getBoundaryBotCandidates}$(G, \mathcal V, \mathcal V', \theta, M, \tau_{\mathrm{bdry}},degree_{node})$}
\label{app:helper-bdry-bots}

\autoref{alg:bocloak-newbot} calls \textsc{getBoundaryBotCandidates}~\autoref{alg:ot-margins} to identify the candidate bot cloaks, which is the intersection set of (i) \emph{boundary bots} (true bots that lie close to the human/bot decision boundary under the OT geometry) and (ii) \emph{misclassified bots} (true bots predicted as human), while also computing the OT margin $m(b)$ for each $b\in\mathcal B$, so that the candidate bots can be sorted by margin. Algorithm~\ref{alg:ot-margins} gives the exact procedure.

\heading{Inputs and outputs.}
It takes the socail graph $\mathcal G=(\mathcal V,\mathcal E)$; the ground-truth label $\mathcal V=\mathcal H\cup\mathcal B$; the predicted label $\mathcal V=\mathcal H'\cup\mathcal B'$; the trained OT geometry $(\theta,M)$ (trained via \autoref{alg:train-ot-geometry}, and defines neighborhood measures and OT-based distances); thresholds $\tau_{\mathrm{bdry}}$ (boundary proximity); and max boundary bot degree $degree_{node}$. It outputs the intersection set between the boundary candidate set and the misclassified-bot set, $\mathcal B_{\mathrm{bdry}} \cap \mathcal B_{\mathrm{mis}}$, sorted by margin $\{-m(b)\}_{b\in\mathcal B}$ in descending order.

\heading{Neighborhood measure and OT distances.}
For each bot $b\in\mathcal B$, we form a neighborhood measure $\mu_b$ (as in \autoref{subsec:ot-neighborhood}), using the OT geometry $(\theta,M)$ to embed and compare neighborhoods. We then compute two OT-geometry distances: (i) $d_H(b)$, the distance from $b$ to the human side (e.g., to a reference human neighborhood/aggregate or nearest-human prototype under $(\theta,M)$), and (ii) $d_B(b)$, the distance from $b$ to the bot side (analogously defined within the bot class). These quantities are exactly those referenced in Algorithm~\ref{alg:ot-margins}.
The routine iterates over true bots $b\in\mathcal B$ that have degree less than $degree_{node}$, builds the neighborhood measure $\mu_b$ (per \autoref{subsec:ot-neighborhood}), computes OT-geometry distances $d_H(b)$ and $d_B(b)$ under $(\theta,M)$, and calculates the OT margin, $m(b)\;=\; d_H(b)-d_B(b)$ (refer to ~\autoref{subsec:ot-margin-main} for the formal definition and interpretation).

It then forms two sets: boundary bots $\mathcal B_{\mathrm{bdry}}=\{b:\ 0<m(b)\le\tau_{\mathrm{bdry}}\}$ and misclassified bots $\mathcal B_{\mathrm{mis}}=\{b:\ b \notin \mathcal{B'} \}$. Finally, it returns the candidate bot cloaks, $\mathcal T \;=\; \mathcal B_{\mathrm{bdry}}\cap \mathcal B_{\mathrm{mis}}$,
sorted by margin $-m(\cdot)$ in descending order (Algorithm~\ref{alg:ot-margins}) to rank candidate bots.

% \heading{Ranking Candidates Bots}
% We use the same margin values to define a boundary proximity rank for weighting:
% \begin{equation}
% \mathrm{rank}(v)\;=\;\text{position of }v\text{ in the ascending sort of }m(\cdot),
% \label{eq:helper-boundary-rank}
% \end{equation}
% so smaller margins indicate nodes closer to the decision boundary.

% ------------------------------------------------------------
\begin{algorithm}[t]
\caption{\textsc{getImportanceWeights}$(\mathcal T)$}
\label{alg:helper-getImportanceWeights}
\begin{algorithmic}[1]
\STATE {\bfseries Input:} cloak records $\mathcal T$ (containing $\mathrm{rank}(t)$ and access to $\mathrm{category}(t), in_h,in_b,out_h,out_b$)
\STATE {\bfseries Output:} $p_{\mathrm{category}}(\cdot)$ and $\{p_{\mathrm{cloak}}(\cdot\mid c)\}_{c}$
\STATE For each $t\in\mathcal T$, compute $e(t)$ and $\eta(t)$ // \texttt{using} \autoref{eq:helper-in-eff-a}-\autoref{eq:helper-eta}
\STATE Group cloaks into buckets $\mathcal T_c$ by $c=\mathrm{category}(t)$
\STATE Compute $p_{\mathrm{category}}(\cdot)$ from // \texttt{using} \autoref{eq:helper-cat-avg-e}-\autoref{eq:helper-pcat}
\STATE For each category $c$, compute $p_{\mathrm{cloak}}(\cdot\mid c)$ from // \texttt{using} \autoref{eq:helper-wtmpl-edges}-\autoref{eq:helper-ptmpl}
\STATE \textbf{return} ($p_{\mathrm{category}}(\cdot)$ and $\{p_{\mathrm{cloak}}(\cdot\mid c)\}_c$)
\end{algorithmic}
\end{algorithm}

\subsection{\textsc{getImportanceWeights}$(\mathcal T)$}
\label{app:helper-getImportanceWeights}

\autoref{alg:bocloak-newbot} calls \textsc{getImportanceWeights}~\autoref{alg:helper-getImportanceWeights} with candidate bot cloaks $\mathcal{B}_{\mathrm{candidate}}$ to build a category distribution $p_{\mathrm{category}}(\cdot)$ and per-category cloak distributions $p_{\mathrm{cloak}}(\cdot\mid c)$.

\heading{Per-cloak structural quantities.}
For each cloak $t\in\mathcal T$, we can obtain directed neighbor-counts: $in_h(t),\; in_b(t),\; out_h(t),\; out_b(t)$, and a node category label from neighbor signature (details refer to \autoref{tab:struc-category}), $\mathrm{category}(t)\in\mathcal C$ (\eg \code{follow\_bot\_followed\_by\_bot\_and\_human}, \code{follow\_node\_followed\_by\_bot}). There can be sixteen categories, as bots and humans can follow and be followed by each other separately, together, or not at all. 

We define the edge-cost proxy used for weighting:
\begin{align}
\mathrm{in\_eff}(t) &= in_h(t)+in_b(t), \label{eq:helper-in-eff-a}\\
e(t) &= out_h(t)+out_b(t)+\mathrm{in\_eff}(t), \label{eq:helper-e-total-a}\\
\eta(t) &= \mathbbm 1[in_h(t)>0]. \label{eq:helper-eta}
\end{align}

\heading{Category-level weights.}
Let $\mathcal T_c = \{t\in\mathcal T:\mathrm{category}(t)=c\}$ and $M=|\mathcal T|$.

Define per-category averages:
\begin{align}
\overline e_c &= \frac{1}{|\mathcal T_c|}\sum_{t\in\mathcal T_c} e(t), \label{eq:helper-cat-avg-e}\\
\overline r_c &= \frac{1}{|\mathcal T_c|}\sum_{t\in\mathcal T_c} \mathrm{rank}(t). \label{eq:helper-cat-avg-r}
\end{align}
Let $e_{\min}= \min_{c:\,|\mathcal T_c|>0}\overline e_c$.

Define two factors and the category weight:
\begin{align}
w_{\mathrm{edges}}(c) &= \left(\frac{e_{\min}}{\max(\overline e_c,10^{-6})}\right)^2, \label{eq:helper-wcat-edges}\\
w_{\mathrm{rank}}(c) &= \frac{1}{1+\overline r_c/M}, \label{eq:helper-wcat-rank}\\
w_{\mathrm{category}}(c) &= w_{\mathrm{edges}}(c)\,w_{\mathrm{rank}}(c). \label{eq:helper-wcat}
\end{align}

Normalize to obtain:
\begin{equation}
p_{\mathrm{category}}(c)= \frac{w_{\mathrm{category}}(c)}{\sum_{c'} w_{\mathrm{category}}(c')}.
\label{eq:helper-pcat}
\end{equation}

\heading{Cloak-level weights within a category.}
Let $r_{\max}= \max_{t\in\mathcal T}\mathrm{rank}(t)$ with $r_{\max}\ge 1$.
Define:
\begin{align}
w_{\mathrm{edges}}(t) &= \frac{1}{1+e(t)}, \label{eq:helper-wtmpl-edges}\\
w_{\mathrm{rank}}(t) &= \frac{1}{1+\mathrm{rank}(t)/r_{\max}}, \label{eq:helper-wtmpl-rank}\\
w_{\mathrm{human}}(t) &=
\begin{cases}
0.5, & \eta(t)=1,\\
1, & \eta(t)=0,
\end{cases}
\label{eq:helper-wtmpl-hum}
\end{align}
and combine:
\begin{equation}
w_{\mathrm{cloak}}(t)= w_{\mathrm{edges}}(t)\,w_{\mathrm{rank}}(t)\,w_{\mathrm{human}}(t).
\label{eq:helper-wtmpl}
\end{equation}
Normalize within category $c$:
\begin{equation}
p_{\mathrm{cloak}}(t\mid c)= \frac{w_{\mathrm{cloak}}(t)}{\sum_{t'\in\mathcal T_c} w_{\mathrm{cloak}}(t')}.
\label{eq:helper-ptmpl}
\end{equation}

% ------------------------------------------------------------
\subsection{\textsc{SampleCloak}$(\mathcal{B}_{\mathrm{candidate}},w_{\mathrm{cadidate}},u,R)$}
\label{app:helper-sampletemp}

\autoref{alg:bocloak-newbot} calls \textsc{SampleCloak}~\autoref{alg:helper-sampletemp} to sample a cloak while enforcing a reuse cap $R$ whenever possible. Here $u(t)$ counts cloak uses (e.g., successful uses).

\heading{Cap enforcement.}
Let $\mathcal T_c$ be the cloaks in category $c$ and
\begin{equation}
\mathcal T_c^{(<R)} = \{t\in\mathcal T_c:\ u(t)<R \land t \in \mathcal{B}_{\mathrm{bdry}} \}.
\label{eq:helper-under-cap}
\end{equation}
If any category has $\mathcal T_c^{(<R)}\neq\emptyset$, we sample only among under-cap cloaks; otherwise we fall back to the original distributions.

\begin{algorithm}[t]
\caption{\textsc{SampleCloak}$(\mathcal{B}_{\mathrm{candidate}},w_{\mathrm{cadidate}},u,R)$}
\label{alg:helper-sampletemp}
\begin{algorithmic}[1]
\STATE {\bfseries Input:} Boundary bots $\mathcal{B}_{\mathrm{candidate}}$; $w_{\mathrm{cadidate}}=$ (category distribution $p_{\mathrm{category}}$; conditional cloak distributions $p_{\mathrm{cloak}}(\cdot\mid c)$), use-counts $u(\cdot)$ and reuse cap $R$.
\STATE {\bfseries Output:} sampled cloak $t$
\STATE For each category $c$, form $\mathcal T_c^{(<R)}$ using $\mathcal{B}_{\mathrm{candidate}}$ // \texttt{using} \autoref{eq:helper-under-cap}
\STATE Compute $\mathcal{B}_{\mathrm{candidate}}$ mass
$\alpha(c)= \sum_{t\in \mathcal T_c^{(<R)}} p_{\mathrm{cloak}}(t\mid c)$
\IF{$\sum_c p_{\mathrm{category}}(c)\alpha(c) > 0$}
  \STATE Sample $c$ with probability proportional to $p_{\mathrm{category}}(c)\alpha(c)$
  \STATE Sample $t$ from $p_{\mathrm{cloak}}(\cdot\mid c)$ restricted to $\mathcal T_c^{(<R)}$
\ELSE
  \STATE Sample $c\sim p_{\mathrm{category}}$ and then $t\sim p_{\mathrm{cloak}}(\cdot\mid c)$
\ENDIF
\STATE \textbf{return} $t$
\end{algorithmic}
\end{algorithm}

% ------------------------------------------------------------
\begin{algorithm}[t]
\caption{\textsc{OTGuidedNeighbors}$(\theta,M,\varepsilon,TopK,G,t)$}
\label{alg:helper-otguidedneighbors}
\begin{algorithmic}[1]
\STATE {\bfseries Input:} OT geometry $(\theta,M)$; OT params $(\varepsilon, TopK)$; graph $G$ and cloak $t$
\STATE {\bfseries Output:} restriction set $\mathcal U$ (or $\varnothing$ meaning no restriction)
\IF{$\varepsilon \le 0$ \OR $K \le 0$}
  \STATE \textbf{return} $\varnothing$
\ENDIF
\STATE Compute nearest human $h^\star(t)$ via // \texttt{using} \autoref{eq:helper-nearest-human-inside}
\STATE Compute entropic OT plan $P^\star_{t,h^\star(t)}$ under $(\theta,M,\varepsilon)$ between neighborhoods of $t$ and $h^\star(t)$ // \texttt{using} \autoref{sec:entropic-ot}
\STATE For each row $i$ (neighbor $u_i\in N^{\mathrm{out}}(t)$), compute row mass $\rho_i$ // \texttt{using} \autoref{eq:helper-row-mass}
\STATE Let $\mathcal U \leftarrow \{u_i:\ i\in \mathrm{TopK}(\rho,K)\}$
\STATE \textbf{return} $\mathcal U$
\end{algorithmic}
\end{algorithm}

\subsection{\textsc{OTGuidedNeighbors}$(\theta,M,\varepsilon,TopK,G,t)$}
\label{app:helper-otguidedneighbors}

\autoref{alg:bocloak-newbot} calls \textsc{OTGuidedNeighbors} \autoref{alg:helper-otguidedneighbors} to restrict which \emph{outgoing} neighbors of cloak $t$ are cloned. In the updated design, this helper also computes the nearest human reference.

\heading{Nearest-human reference.}
Let $d_{\theta,M}(\cdot,\cdot)$ be the OT-geometry distance (or embedding-induced distance) between nodes. The nearest human to cloak $t$ is
\begin{equation}
h^\star(t)= \arg\min_{h\in\mathcal H} d_{\theta,M}(t,h).
\label{eq:helper-nearest-human-inside}
\end{equation}

\heading{OT plan and row-mass scoring.}
Let $N^{\mathrm{out}}(t)$ be the outgoing neighbors of $t$ in $G$. When $\varepsilon>0$, we compute the Sinkhorn plan between the neighborhood measures of $t$ and $h^\star(t)$ as described in the \autoref{sec:entropic-ot}. Let rows of $P^\star_{t,h^\star(t)}$ correspond to neighbors $u_i\in N^{\mathrm{out}}(t)$. 

Define row masses:
\begin{equation}
\rho_i = \sum_j \bigl(P^\star_{t,h^\star(t)}\bigr)_{ij}.
\label{eq:helper-row-mass}
\end{equation}
Then select the OT-guided neighbor restriction set as, $\mathcal U = \{\,u_i:\ i\in \mathrm{TopK}(\rho,K)\,\}$.

% If OT guidance is disabled/unavailable (e.g., $\varepsilon=0$), we return $\mathcal U=\varnothing$ to denote “no restriction”.

% ------------------------------------------------------------
\begin{algorithm}[t]
\caption{\textsc{CloneCloak}$(G,v_{tar}, t,\mathcal U,\texttt{flag\_hb})$}
\label{alg:helper-CloneCloak}
\begin{algorithmic}[1]
\STATE {\bfseries Input:} graph $G=(\mathcal V,\mathcal E)$; target node $v_{tar}$; cloak $t$; restriction set $\mathcal U$ and flag \texttt{flag\_hb}
\STATE {\bfseries Output:} updated graph $G'$, added edges $E_{\mathrm{add}}$, and target node $v_{\mathrm{tar}}$
% \STATE Create a new node $v_{\mathrm{tar}}$ and set $E_{\mathrm{add}}\leftarrow\emptyset$
\STATE Copy the \emph{non-temporal} node features of $t$ to $v_{\mathrm{tar}}$.
\STATE Perturb the \emph{temporal} features of $v_{\mathrm{tar}}$ (e.g., age/timestamp-related features) for plausibility
\STATE Determine allowed outgoing neighbors $\widetilde N^{\mathrm{out}}(t)$ via // \texttt{using} \autoref{eq:helper-out-allowed}
\FOR{each outgoing edge $(t\rightarrow x)\in\mathcal E$ with $x\in \widetilde N^{\mathrm{out}}(t)$}
  \STATE Add $(v_{\mathrm{tar}}\rightarrow x)$ to $G'$ and to $E_{\mathrm{add}}$
\ENDFOR
\FOR{each incoming edge $(x\rightarrow t)\in\mathcal E$}
  \IF{\texttt{flag\_hb} is enabled \AND $x\in\mathcal H$}
    \STATE skip (do not add human follow-back)
  \ELSE
    \STATE Add $(x\rightarrow v_{\mathrm{tar}})$ to $G'$ and to $E_{\mathrm{add}}$
  \ENDIF
\ENDFOR
\STATE \textbf{return} $G',E_{\mathrm{add}},v_{\mathrm{tar}}$
\end{algorithmic}
\end{algorithm}

\subsection{\textsc{CloneCloak}$(G,v_{\mathrm{tar}},t,\mathcal U,\texttt{flag\_hb})$}
\label{app:helper-CloneCloak}

\autoref{alg:bocloak-newbot} calls \textsc{CloneCloak}~\autoref{alg:helper-CloneCloak} to edit a target node $v_{\mathrm{tar}}$ by cloning the cloak’s directed edges, perturbing the cloak's features and restricting outgoing edges to the OT-guided subset $\mathcal U$. The flag \texttt{flag\_hb} disables human follow-back (incoming edges from humans).

\heading{Outgoing cloning with restriction.}
Let $N^{\mathrm{out}}(t)$ be outgoing neighbors of $t$. Define the allowed outgoing set:
\begin{equation}
\widetilde N^{\mathrm{out}}(t)=
\begin{cases}
N^{\mathrm{out}}(t), & \mathcal U=\varnothing,\\
N^{\mathrm{out}}(t)\cap \mathcal U, & \text{otherwise}.
\end{cases}
\label{eq:helper-out-allowed}
\end{equation}

\heading{Directed edge cloning rule.}
Let $v_{\mathrm{tar}}$ be the target node. For each outgoing edge $(t\rightarrow x)\in\mathcal E$ with $x\in \widetilde N^{\mathrm{out}}(t)$, add
$(t\rightarrow x)\in\mathcal E \;\Longrightarrow\; (v_{\mathrm{tar}}\rightarrow x)\in\mathcal E'$.
For incoming edges $(x\rightarrow t)\in\mathcal E$, we add
$(x\rightarrow t)\in\mathcal E \;\Longrightarrow\; (x\rightarrow v_{\mathrm{tar}})\in\mathcal E'$, but if $x\in\mathcal H$ and \texttt{flag\_hb} is enabled, we \emph{skip} cloning that incoming human edge.

%% file: tables/hyperparameter-defense.tex
\begin{landscape}
\scriptsize
\renewcommand{\arraystretch}{1.15}
\setlength{\LTleft}{0pt}
\setlength{\LTright}{0pt}
\setlength{\tabcolsep}{4pt}

% ========================= Main table (training-focused) =========================
\begin{longtable}{@{}llcccccccccccccc@{}}
\caption{\textbf{Hyperparameters of Bot Detectors:} Hyperparameters for vanilla bot detectors and their defense variants.}\label{tab:hparams-defense}\\
\toprule
\textbf{Model} & \textbf{Variant} &
\textbf{Layer} &
\textbf{\makecell{Number\\Heads}} &
\textbf{\makecell{Hidden\\ Units}} &
\textbf{\makecell{Learning\\ Rate}} &
\textbf{\makecell{Batch\\ Size}} &
\textbf{Dropout} &
\textbf{\makecell{Weight\\ Decay}} &
\textbf{Epochs} &
\textbf{Patience} &
\textbf{Optimizer} &
\textbf{\makecell{Neighbour\\ Sampling}} &
\textbf{\makecell{Neighbour\\ Aggregation}} & 
\textbf{\makecell{Num Edge\\ Types}} &
\textbf{\makecell{Temporal\\ Smoothing}} \\
\midrule
\endfirsthead

\toprule
\textbf{Model} & \textbf{Variant} &
\textbf{Layer} &
\textbf{\makecell{Num.\\Heads}} &
\textbf{\makecell{Hidden\\ Units}} &
\textbf{\makecell{Learning\\ Rate}} &
\textbf{\makecell{Batch\\ Size}} &
\textbf{Dropout} &
\textbf{\makecell{Weight\\ Decay}} &
\textbf{Epochs} &
\textbf{Patience} &
\textbf{Optimizer} &
\textbf{\makecell{Neighbour\\ Sampling}} &
\textbf{\makecell{Neighbour\\ Aggregation}} &
\textbf{\makecell{Num Edge\\ Types}} &
\textbf{\makecell{Temporal\\ Smoothing}} \\
\midrule
\endhead

\midrule
\multicolumn{16}{r}{\textit{Continued on next page}}\\
\endfoot

\bottomrule
\endlastfoot

% ========================= GCN =========================
\multirow{4}{*}{GCN}
& Vanilla     & 2 & - & 128 & $1\times 10^{-3}$ & 128 & 0.5 & $3\times 10^{-5}$ & 1000 & 50 & Adam & 256 & mean & -&-   \\
& +GNNGuard   & 2 & - & 128 & $1\times 10^{-3}$ & 128 & 0.5 & $3\times 10^{-5}$ & 1000 & 50 & Adam & 256 & mean & -&-   \\
& +GRAND     & 2 & - & 128 & $1\times 10^{-3}$ & 128 & 0.5 & $3\times 10^{-5}$ & 1000 & 50 & Adam & 256 & sum & -&-   \\
& +RobustGCN & 2 & - & 128 & $1\times 10^{-3}$ & 128 & 0.5 & $3\times 10^{-5}$ & 1000 & 50 & Adam & 256 & mean & -&-   \\
& +NoisyGNN   & 2 & - & 128 & $1\times 10^{-3}$ & 256 & 0.5 & $3\times 10^{-5}$ & 1000 & 50 & Adam & 256 & mean & 2 & - \\

\cmidrule(l){1-16}

% ========================= GAT =========================
\multirow{4}{*}{GAT}
& Vanilla     & 4 & 2 & 128 & $1\times 10^{-3}$ & 256 & 0.5 & $3\times 10^{-5}$ & 1000 & 50 & Adam & 256 & mean & - & -   \\
& +GNNGuard   & 4 & 2 & 128 & $1\times 10^{-3}$ & 256 & 0.5 & $3\times 10^{-5}$ & 1000 & 50 & Adam & 256 & mean & - & -   \\
& +GRAND     & 4 & 2 & 128 & $1\times 10^{-3}$ & 256 & 0.5 & $3\times 10^{-5}$ & 1000 & 50 & Adam & 256 & sum & - & -   \\
& +RobustGCN & 4 & 2 & 128 & $1\times 10^{-3}$ & 256 & 0.5 & $3\times 10^{-5}$ & 1000 & 50 & Adam & 256 & mean & - & -   \\
\cmidrule(l){1-16}

% ========================= BotRGCN =========================
\multirow{4}{*}{BotRGCN}
& Vanilla     & 2 & - & 128 & $1\times 10^{-3}$ & 256 & 0.5 & $3\times 10^{-5}$ & 1000 & 50 & Adam & 256 & mean & 2 &-   \\
& +GNNGuard   & 2 & - & 128 & $1\times 10^{-3}$ & 256 & 0.5 & $3\times 10^{-5}$ & 1000 & 50 & Adam & 256 & mean &2  &-   \\
& +GRAND     & 2 & - & 128 & $1\times 10^{-3}$ & 256 & 0.5 & $3\times 10^{-5}$ & 1000 & 50 & Adam & 256 & sum & 2 &-   \\
& +RobustGCN & 2 & - & 128 & $1\times 10^{-3}$ & 256 & 0.5 & $3\times 10^{-5}$ & 1000 & 50 & Adam & 256 & mean & 2 &-   \\
\cmidrule(l){1-16}

% ========================= S-HGN =========================
% Note: For S-HGN, "Linear ch." is treated as Hidden Units (=128).
\multirow{4}{*}{S-HGN}
& Vanilla     & 2 & - & 128 & $1\times 10^{-3}$ & 256 & 0.5 & $3\times 10^{-5}$ & 1000 & 50 & Adam & 256 & mean & 2 & 0.05 \\
& +GNNGuard   & 2 & - & 128 & $1\times 10^{-2}$ & 256 & 0.5 & $3\times 10^{-5}$ & 1000 & 50 & Adam & 256 & mean & 2 & 0.05 \\
& +GRAND     & 2 & - & 128 & $1\times 10^{-3}$ & 256 & 0.5 & $3\times 10^{-5}$ & 1000 & 50 & Adam & 256 & sum & 2 & 0.05 \\
& +RobustGCN & 2 & - & 128 & $1\times 10^{-2}$ & 256 & 0.5 & $3\times 10^{-5}$ & 1000 & 50 & Adam & 256 & mean & 2 & 0.05  \\
\cmidrule(l){1-16}

% ========================= RGT =========================
\multirow{1}{*}{RGT}
& Vanilla     & 4 & 2 & 256 & $1\times 10^{-3}$ & 512 & 0.5 & $3\times 10^{-5}$ & 500 & 50 & Adam & 256 & - & 2 &-   \\

\end{longtable}

\vspace{0.6em}

% ========================= Stacked defense table =========================
\begin{longtable}{@{}lll@{}}
\caption{\textbf{Hyperparameters of \gnn-based Defenses.}}\label{tab:hparams-defense-variants}\\
\toprule
\textbf{GNN Defense} & \textbf{Hyperparameter} & \textbf{Value} \\
\midrule
\endfirsthead

\toprule
\textbf{Defense} & \textbf{Hyperparameter} & \textbf{Value} \\
\midrule
\endhead

\bottomrule
\endlastfoot

\multirow{2}{*}{GNNGuard}
& Threshold & 0.1 \\
& Add self-loops & False \\
\midrule

\multirow{5}{*}{GRAND}
& Node drop rate & 0.1 \\
& Propagation order & 1 \\
& Temperature & 0.1 \\
& Consistency Weight ($\lambda$) & 0.5 \\
& Monte Carlo Samples & 5 \\
\midrule

\multirow{2}{*}{RobustGCN}
& Robustness Parameter ($\gamma$) & 1.0 \\
& KL regularization & $5\times 10^{-4}$ \\

\midrule

\multirow{4}{*}{NoisyGNN}
& Backbone & BotRGCN \\
& Noise type & Additive Gaussian hidden-state noise \\
& Noise scale ($\texttt{noise\_ratio\_1}$) & 0.05; training only, disabled during inference \\

\end{longtable}

\end{landscape}

%% file: tables/hyperparameter-attack.tex
\begin{landscape}
\begin{table*}[!htb]
\centering
\caption{\textbf{Hyperparameters of Adversarial Attack Frameworks}.}
\label{tab:hparams-attack}

\scriptsize
\setlength{\tabcolsep}{3.5pt}
\renewcommand{\arraystretch}{1.10}

\begin{tabular}{@{}p{0.10\linewidth} p{0.16\linewidth} p{0.22\linewidth} p{0.24\linewidth} p{0.28\linewidth}@{}}
\toprule
\textbf{Attack} &
\textbf{Setup} &
\textbf{Scaling Safeguards} &
\textbf{Attack Specific} &
\textbf{Surrogate Specifics} \\
\midrule

\textbf{Nettack} &
\multirow{26}{*}{%
\begin{tabular}[t]{@{}l@{}}
$B\in\{1,3,5\}$; targets=50.\\
Targeted a specific node prediction.
\end{tabular}%
}
&
\multirow{26}{*}{%
\begin{tabular}[t]{@{}l@{}}
If $|V|\ge 10^5$: local $k$-hop, $k{=}2$, max nodes=50k.\\
Else: full graph.
\end{tabular}%
}
&
\begin{tabular}[t]{@{}l@{\hspace{4pt}}l@{}}
LR cutoff & 0.004 \\
Stop rule & early stop if no admissible structural edits \\
\end{tabular}
&
\multirow{15}{*}{%
\begin{tabular}[t]{@{}l@{\hspace{4pt}}l@{}}
Architecture & 2-layer GCN \\
Hidden units & 64 \\
Epochs & 200 \\
Optimizer & Adam \\
Learning rate & $1\times 10^{-3}$ \\
Weight decay & $5{\times}10^{-4}$ \\
\end{tabular}%
}
\\
% \midrule
\cmidrule(l){1-1}
\cmidrule(l){4-4}

\textbf{FGA} &
% $B\in\{1,3,5\}$; targets=50.\newline
% Structural-only (no features).\newline
% Direct. 
&
% If $|V|\ge 10^5$: local $k$-hop, $k{=}2$, max nodes=50k.\newline
% Else: full graph. 
&
\begin{tabular}[t]{@{}l@{\hspace{4pt}}l@{}}
Structure edits & enabled \\
Feature edits & disabled \\
\end{tabular}
&
% \begin{tabular}[t]{@{}l@{\hspace{4pt}}l@{}}
% Architecture & 2-layer GCN \\
% Hidden units & 64 \\
% Epochs & 200 \\
% Optimizer & Adam \\
% Learning rate & $10^{-3}$ \\
% Weight decay & $5{\times}10^{-4}$ \\
% \end{tabular}
\\
% \midrule
\cmidrule(l){1-1}
\cmidrule(l){4-4}

\textbf{PR-BCD} &
% $B\in\{1,3,5\}$; targets=50.\newline
% Structural-only (no features).\newline
% Direct.
&
% If $|V|\ge 10^5$: local $k$-hop, $k{=}2$, max nodes=50k.\newline
% Else: full graph. 
&
\begin{tabular}[t]{@{}l@{\hspace{4pt}}l@{}}
Epochs & 125 \\
Resample & 100 \\
Loss & probability margin loss\\
Step size & 1000 \\
Directed Graph & false \\
Block size & $\max(B{+}1,50k)$ \\
\end{tabular}
&
% \begin{tabular}[t]{@{}l@{\hspace{4pt}}l@{}}
% Architecture & 2-layer GCN \\
% Hidden units & 64 \\
% Epochs & 200 \\
% Optimizer & Adam \\
% Learning rate & $10^{-3}$ \\
% Weight decay & $5{\times}10^{-4}$ \\
% \end{tabular}
\\
% \midrule
\cmidrule(l){1-1}
\cmidrule(l){4-4}

\textbf{GOttack} &
% $B\in\{1,3,5\}$; targets=50.\newline
% Structural-only (no features).\newline
% Direct. 
&
&
\begin{tabular}[t]{@{}l@{\hspace{4pt}}l@{}}
Seed & 720 \\
Orbit type & 1518 \\
Orbit table & precomputed per dataset.
\end{tabular}
&
% \begin{tabular}[t]{@{}l@{\hspace{4pt}}l@{}}
% Architecture & 2-layer GCN \\
% Hidden units & 64 \\
% Epochs & 200 \\
% Optimizer & Adam \\
% Learning rate & $10^{-3}$ \\
% Weight decay & $5{\times}10^{-4}$ \\
% \end{tabular}
\\
% \midrule
\cmidrule(l){1-1}
\cmidrule(l){4-5}

\textbf{\pname} 
&

&

&
\begin{tabular}[t]{@{}l@{\hspace{4pt}}l@{}}
OT Regularizer $\epsilon$ & 0.2 \\
Sinkhorn iters & 30 \\
Cost MLP (hid, emb) & (128, 256) \\
Top boundary bots & 50 \\
Max reuse & 3 \\
Edit policy & Reset per trial \\
 & Add/del only for target bot \\
 & Forbid human$\rightarrow$bot follow-back \\
Train $C_\theta$ & $\lambda_{\text{BCE}}{=}2$, $\lambda_{\text{sp}}{=}0.05$, $\lambda_{\text{pl}}{=}0.10$ \\
& $\alpha_{\text{degree}}{=}0.8, \alpha_{\text{time}}{=}0.2$ \\
& Boundary margin $\tau_{\mathrm{bdry}}$=0.1 \\
& Sigmoid temperature $\tau_{\mathrm{BCE}}$=0.01 \\
& $\gamma{=}1\times 10^{-3}$, Batch=128 \\
Neighbor Sampling & 256 \\
$k$-hop neighborhood & $k$=1 or 1-hop ego neighborhood\\
\end{tabular}
&
\multirow{13}{*}{%
\begin{tabular}[t]{@{}l@{\hspace{4pt}}l@{}}
Surrogate & none \\
Guidance signal & learned OT cost $C_\theta$ \\
\end{tabular}
}
\\
\bottomrule
\end{tabular}
\end{table*}
\end{landscape}

%% file: tables/degree.tex
\begin{table*}[!htb]
\centering
\resizebox{0.8\linewidth}{!}{%
\begin{tabular}{lccccccc}
\toprule
\textbf{Dataset} 
& \textbf{\# of Nodes} 
& \textbf{\# of Edges} 
& \textbf{Avg Deg. } 
& \textbf{\# of Bots} 
& \textbf{Avg Deg. (Bots)} 
& \textbf{\# of Humans} 
& \textbf{Avg Deg. (Humans)} \\
\midrule
Cresci-15 &     5{,}301 &     14{,}220 &  2.05 &  3{,}351 &  0.22 &  1{,}950 &  5.18 \\
TwiBot-22 & 1{,}000{,}000 & 170{,}185{,}937 &  6.67 & 139{,}943 &  3.56 & 860{,}057 &  7.00 \\
BotSim-24    &     2{,}907 &    46{,}518 & 45.39 &  1{,}000 & 19.23 &  1{,}907 & 59.12 \\
% \midrule
% \textbf{Average} & 336{,}069 &56{,}748{,}891 & 18.04 & 48{,}098 & 7.67 & 287{,}971 & 23.82 \\
\bottomrule
\end{tabular}%
}
\caption{\textbf{Dataset Statistics:} Detailed bot detection dataset summaries.}
\label{tab:avg-node-degree}
\end{table*}

%% file: tables/defense.tex
\begin{table*}[!htb]
\centering
\resizebox{0.5\linewidth}{!}{%
\begin{tabular}{llcccc}
\toprule
\multirow{2}{*}{\textbf{Dataset}} & \multirow{2}{*}{\makecell[c]{\textbf{Victim}}} & \multirow{2}{*}{\textbf{Vanilla} ($\uparrow$)} & \multicolumn{3}{c}{\textbf{\gnn Defense}} \\
\cmidrule(lr){4-6} 
& & & \textbf{+GNNGuard} ($\uparrow$) & \textbf{+GRAND} ($\uparrow$) & \textbf{+RobustGCN} ($\uparrow$) \\
\midrule
\multirow{5}{*}{Cresci-15}
& GCN     & 0.93\scriptsize{\emph{$\pm$0.01}} & 0.93\scriptsize{\emph{$\pm$0.05}} & 0.93\scriptsize{\emph{$\pm$0.01}} & 0.92\scriptsize{\emph{$\pm$0.06}} \\
& GAT     & 0.97\scriptsize{\emph{$\pm$0.03}} & 0.97\scriptsize{\emph{$\pm$0.05}} & \bluehl{0.98\scriptsize{\emph{$\pm$0.01}}} & 0.97\scriptsize{\emph{$\pm$0.07}} \\
& BotRGCN & \bluehl{0.98\scriptsize{\emph{$\pm$0.08}}} & \bluehl{0.98\scriptsize{\emph{$\pm$0.05}}} & \underline{0.98\scriptsize{\emph{$\pm$0.03}}} & \bluehl{0.98\scriptsize{\emph{$\pm$0.02}}} \\
& S-HGN   & \underline{0.97\scriptsize{\emph{$\pm$0.02}}} & \underline{0.98\scriptsize{\emph{$\pm$0.06}}} & 0.96\scriptsize{\emph{$\pm$0.02}} & 0.97\scriptsize{\emph{$\pm$0.07}} \\
& RGT     & 0.97\scriptsize{\emph{$\pm$0.02}} & 0.97\scriptsize{\emph{$\pm$0.01}} & 0.97\scriptsize{\emph{$\pm$0.03}} & \underline{0.98\scriptsize{\emph{$\pm$0.04}}} \\
\midrule
\multirow{5}{*}{TwiBot-22}
& GCN     & 0.52\scriptsize{\emph{$\pm$0.05}} & \underline{0.50\scriptsize{\emph{$\pm$0.05}}} & 0.48\scriptsize{\emph{$\pm$0.03}} & 0.51\scriptsize{\emph{$\pm$0.04}} \\
& GAT     & 0.52\scriptsize{\emph{$\pm$0.02}} & 0.50\scriptsize{\emph{$\pm$0.10}} & 0.53\scriptsize{\emph{$\pm$0.00}} & \underline{0.53\scriptsize{\emph{$\pm$0.01}}} \\
& BotRGCN & \bluehl{0.58\scriptsize{\emph{$\pm$0.08}}} & \bluehl{0.52\scriptsize{\emph{$\pm$0.05}}} & \bluehl{0.56\scriptsize{\emph{$\pm$0.01}}} & 0.51\scriptsize{\emph{$\pm$0.06}} \\
& S-HGN   & \underline{0.58\scriptsize{\emph{$\pm$0.02}}} & 0.50\scriptsize{\emph{$\pm$0.08}} & \underline{0.56\scriptsize{\emph{$\pm$0.02}}} & \bluehl{0.55\scriptsize{\emph{$\pm$0.03}}} \\
& RGT     & 0.42\scriptsize{\emph{$\pm$0.04}} & 0.40\scriptsize{\emph{$\pm$0.01}} & 0.41\scriptsize{\emph{$\pm$0.02}} & 0.38\scriptsize{\emph{$\pm$0.01}} \\
\midrule
\multirow{5}{*}{BotSim-24}
& GCN     & 0.90\scriptsize{\emph{$\pm$0.01}} & 0.92\scriptsize{\emph{$\pm$0.03}} & \bluehl{0.93\scriptsize{\emph{$\pm$0.01}}} & \underline{0.93\scriptsize{\emph{$\pm$0.02}}} \\
& GAT     & \underline{0.95\scriptsize{\emph{$\pm$0.05}}} & 0.90\scriptsize{\emph{$\pm$0.01}} & 0.92\scriptsize{\emph{$\pm$0.02}} & 0.89\scriptsize{\emph{$\pm$0.03}} \\
& BotRGCN & \bluehl{0.96\scriptsize{\emph{$\pm$0.03}}} & 0.91\scriptsize{\emph{$\pm$0.01}} & 0.90\scriptsize{\emph{$\pm$0.06}} & 0.91\scriptsize{\emph{$\pm$0.14}} \\
& S-HGN   & 0.92\scriptsize{\emph{$\pm$0.09}} & \bluehl{0.94\scriptsize{\emph{$\pm$0.01}}} & \underline{0.93\scriptsize{\emph{$\pm$0.03}}} & 0.92\scriptsize{\emph{$\pm$0.20}} \\
& RGT     & 0.93\scriptsize{\emph{$\pm$0.05}} & \underline{0.94\scriptsize{\emph{$\pm$0.03}}} & 0.92\scriptsize{\emph{$\pm$0.01}} & \bluehl{0.94\scriptsize{\emph{$\pm$0.03}}} \\
\bottomrule
\end{tabular}%
}
\caption{\textbf{Node Classification Performance (F1 score):} Bot detection performance of different \sota bot detector architectures with \gnn-based adversarial defenses. The best performance is shown in bold, and the second best is underlined.}
\label{tab:f1-gnn-bots-defense}
\end{table*}

%% file: tables/noisygnn-results.tex
\begin{table}[!t]
\centering
\scriptsize
\setlength{\tabcolsep}{4pt}
\renewcommand{\arraystretch}{1.08}
\begin{tabular}{lccc}
\toprule
\textbf{Attack} 
& \textbf{Unconstrained Vanilla} 
& \textbf{Constrained Vanilla} 
& \textbf{+NoisyGNN} \\
\midrule
Nettack & 79.77$\pm$6.77 & 9.33$\pm$3.06 & 4.34$\pm$8.09 \\
FGA & 71.90$\pm$4.23 & 4.00$\pm$2.00 & 3.65$\pm$1.11 \\
PR-BCD & 75.77$\pm$6.05 & 8.67$\pm$2.31 & 5.45$\pm$0.13 \\
GOttack & 84.29$\pm$11.63 & 6.89$\pm$3.33 & 4.11$\pm$1.11 \\
\textbf{\pname{} (ours)} & \bluehl{86.67$\pm$2.31} & \bluehl{86.67$\pm$2.31} & \bluehl{85.12$\pm$2.23} \\
\bottomrule
\end{tabular}
\caption{
\textbf{NoisyGNN robustness results on TwiBot-22 against the strongest BotRGCN detector with node-editing budget $\Delta=1$}: 
NoisyGNN injects Gaussian noise into the hidden state after the first relational message-passing layer during training only. 
Even with this stochastic hidden-state defense, \pname{} remains substantially stronger than constrained generic graph attacks.
}
\label{tab:noisygnn-results}
\end{table}

%% file: tables/old-node-50-rest.tex
\begin{landscape}
\begin{table*}[!htb]
\centering
\setlength{\tabcolsep}{2.2pt}
\renewcommand{\arraystretch}{1.05}

\resizebox{\textwidth}{!}{%
\begin{tabular}{ll cccc cccc cccc}
\toprule
\multirow{2}{*}{\textbf{Dataset}} &
\multirow{2}{*}{\makecell[c]{\textbf{Attack}}} &
\multicolumn{4}{c}{\textbf{GAT}} &
\multicolumn{4}{c}{\textbf{BotRGCN}} &
\multicolumn{4}{c}{\textbf{S-HGN}} \\
\cmidrule(lr){3-6}\cmidrule(lr){7-10}\cmidrule(lr){11-14}
&
& \textbf{Vanilla} ($\uparrow$) & \textbf{+GNNGuard} ($\uparrow$) & \textbf{+GRAND} ($\uparrow$) & \textbf{+RobustGCN} ($\uparrow$)
& \textbf{Vanilla} ($\uparrow$) & \textbf{+GNNGuard} ($\uparrow$) & \textbf{+GRAND} ($\uparrow$) & \textbf{+RobustGCN} ($\uparrow$)
& \textbf{Vanilla} ($\uparrow$) & \textbf{+GNNGuard} ($\uparrow$) & \textbf{+GRAND} ($\uparrow$) & \textbf{+RobustGCN} ($\uparrow$) \\
\midrule

% ========================= Δ = 1 =========================
\multicolumn{14}{c}{$\Delta = 1$} \\
\midrule

% -------- Cresci-15 --------
\multirow{5}{*}{Cresci-15}
& Random
& 5.34\scriptsize{$\pm$1.65} 
& 3.75\scriptsize{$\pm$2.12} 
& 3.80\scriptsize{$\pm$2.85} 
& 3.11\scriptsize{$\pm$2.57}
& \underline{8.67\scriptsize{$\pm$5.03}} & \underline{7.33\scriptsize{$\pm$2.33}} & \underline{6.67\scriptsize{$\pm$3.43}} & \underline{8.00\scriptsize{$\pm$0.20}}
& 5.11\scriptsize{$\pm$x.xx} 
& 4.46\scriptsize{$\pm$x.xx} 
& 4.18\scriptsize{$\pm$x.xx} 
& 3.56\scriptsize{$\pm$x.xx} \\
& Nettack
% GAT (4)
& 8.17\scriptsize{$\pm$3.23} & 6.59\scriptsize{$\pm$3.40} & \underline{7.54\scriptsize{$\pm$2.07}} & 7.28\scriptsize{$\pm$3.19}
% BotRGCN (4)
& 4.32\scriptsize{$\pm$2.10} & 3.78\scriptsize{$\pm$1.95} & 3.21\scriptsize{$\pm$1.88} & 2.67\scriptsize{$\pm$1.72}
% S-HGN (4)
& 6.04\scriptsize{$\pm$3.00} & 4.81\scriptsize{$\pm$3.18} & 4.51\scriptsize{$\pm$3.60} & 4.88\scriptsize{$\pm$2.71} \\
& FGA
& 8.52\scriptsize{$\pm$1.92} & 5.74\scriptsize{$\pm$2.39} & 4.73\scriptsize{$\pm$3.67} & \underline{7.93\scriptsize{$\pm$2.91}}
& 3.95\scriptsize{$\pm$2.44} & 3.40\scriptsize{$\pm$2.12} & 2.98\scriptsize{$\pm$1.76} & 2.15\scriptsize{$\pm$1.60}
& 5.22\scriptsize{$\pm$3.17} & 4.96\scriptsize{$\pm$3.09} & 4.31\scriptsize{$\pm$2.07} & 3.44\scriptsize{$\pm$2.01} \\
& PR-BCD
& \underline{8.79\scriptsize{$\pm$1.91}} & 6.85\scriptsize{$\pm$2.35} & 7.48\scriptsize{$\pm$2.37} & 7.92\scriptsize{$\pm$3.24}
& 4.10\scriptsize{$\pm$3.05} & 3.60\scriptsize{$\pm$2.80} & 2.85\scriptsize{$\pm$2.20} & 1.90\scriptsize{$\pm$1.45}
& 7.57\scriptsize{$\pm$3.84} & 6.91\scriptsize{$\pm$2.43} & 6.70\scriptsize{$\pm$2.84} & 5.58\scriptsize{$\pm$2.45} \\
& GOttack
& 7.61\scriptsize{$\pm$2.91} & \underline{7.14\scriptsize{$\pm$3.65}} & 5.08\scriptsize{$\pm$3.06} & 6.20\scriptsize{$\pm$1.94}
& 3.55\scriptsize{$\pm$2.34} & 3.41\scriptsize{$\pm$2.52} & 2.28\scriptsize{$\pm$1.22} & 2.66\scriptsize{$\pm$1.67}
& \underline{8.10\scriptsize{$\pm$2.83}} & \underline{7.57\scriptsize{$\pm$2.04}} & \underline{7.92\scriptsize{$\pm$1.87}} & \underline{6.39\scriptsize{$\pm$2.12}} \\
& \textbf{\pname{}} (ours)
& \bluehl{93.10\scriptsize{$\pm$2.34}} & \bluehl{94.20\scriptsize{$\pm$1.89}} & \bluehl{92.45\scriptsize{$\pm$3.11}} & \bluehl{91.88\scriptsize{$\pm$4.56}}
& \bluehl{99.34\scriptsize{$\pm$0.29}} & \bluehl{98.91\scriptsize{$\pm$0.54}} & \bluehl{99.06\scriptsize{$\pm$0.41}} & \bluehl{99.72\scriptsize{$\pm$0.18}}
& \bluehl{92.18\scriptsize{$\pm$4.73}} & \bluehl{91.60\scriptsize{$\pm$2.15}} & \bluehl{91.92\scriptsize{$\pm$3.08}} & \bluehl{91.40\scriptsize{$\pm$1.67}} \\
\midrule

% -------- TwiBot-22 --------
\multirow{5}{*}{TwiBot-22}
& Random
& 6.67\scriptsize{$\pm$2.45} 
& 3.12\scriptsize{$\pm$2.11} 
& 3.87\scriptsize{$\pm$2.84} 
& 4.48\scriptsize{$\pm$1.01}
& 5.12\scriptsize{$\pm$6.11} 
& 2.00\scriptsize{$\pm$3.46} 
& 5.33\scriptsize{$\pm$13.61} 
& 3.00\scriptsize{$\pm$6.00}
& 4.55\scriptsize{$\pm$1.69} 
& 3.50\scriptsize{$\pm$1.48} 
& 3.11\scriptsize{$\pm$2.83} 
& 3.76\scriptsize{$\pm$2.55} \\
& Nettack
& 13.51\scriptsize{$\pm$8.15} & 11.92\scriptsize{$\pm$9.38} & 8.59\scriptsize{$\pm$7.73} & \underline{12.91\scriptsize{$\pm$8.91}}
& \underline{9.33\scriptsize{$\pm$3.06}} & \underline{8.67\scriptsize{$\pm$4.62}} & 4.67\scriptsize{$\pm$5.03} & \underline{8.00\scriptsize{$\pm$5.29}}
& 5.03\scriptsize{$\pm$4.72} & 3.24\scriptsize{$\pm$9.20} & 3.72\scriptsize{$\pm$10.81} & 4.73\scriptsize{$\pm$7.85} \\
& FGA
& 12.32\scriptsize{$\pm$8.10} & 8.59\scriptsize{$\pm$4.52} & 9.94\scriptsize{$\pm$6.41} & 5.57\scriptsize{$\pm$10.78}
& 4.00\scriptsize{$\pm$2.00} & 7.33\scriptsize{$\pm$2.31} & \underline{6.67\scriptsize{$\pm$4.16}} & 1.33\scriptsize{$\pm$6.43}
& \underline{5.79\scriptsize{$\pm$6.52}} & \underline{4.57\scriptsize{$\pm$4.96}} & 3.41\scriptsize{$\pm$6.87} & 3.19\scriptsize{$\pm$9.89} \\
& PR-BCD
& \underline{13.51\scriptsize{$\pm$4.52}} & \underline{12.52\scriptsize{$\pm$9.38}} & \underline{11.32\scriptsize{$\pm$6.41}} & 12.24\scriptsize{$\pm$4.52}
& 8.67\scriptsize{$\pm$2.31} & 8.67\scriptsize{$\pm$5.03} & 6.23\scriptsize{$\pm$4.16} & 7.33\scriptsize{$\pm$2.31}
& 4.58\scriptsize{$\pm$5.29} & 3.74\scriptsize{$\pm$3.34} & 3.61\scriptsize{$\pm$7.20} & 3.20\scriptsize{$\pm$7.92} \\
& GOttack
& 11.41\scriptsize{$\pm$7.59} & 10.50\scriptsize{$\pm$6.38} & 9.78\scriptsize{$\pm$3.93} & 10.33\scriptsize{$\pm$5.34}
& 6.89\scriptsize{$\pm$3.33} & 7.15\scriptsize{$\pm$4.25} & 6.01\scriptsize{$\pm$2.48} & 5.32\scriptsize{$\pm$2.58}
& 5.55\scriptsize{$\pm$7.98} & 4.04\scriptsize{$\pm$8.93} & \underline{4.44\scriptsize{$\pm$7.82}} & \underline{4.95\scriptsize{$\pm$8.39}} \\
& \textbf{\pname{}} (ours)
& \bluehl{94.38\scriptsize{$\pm$3.41}} & \bluehl{93.72\scriptsize{$\pm$1.66}} & \bluehl{93.90\scriptsize{$\pm$4.28}} & \bluehl{92.26\scriptsize{$\pm$2.13}}
& \bluehl{86.67\scriptsize{$\pm$2.31}} & \bluehl{84.67\scriptsize{$\pm$11.02}} & \bluehl{83.67\scriptsize{$\pm$3.06}} & \bluehl{85.30\scriptsize{$\pm$2.00}}
& \bluehl{95.66\scriptsize{$\pm$2.27}} & \bluehl{94.88\scriptsize{$\pm$4.93}} & \bluehl{94.41\scriptsize{$\pm$1.62}} & \bluehl{94.30\scriptsize{$\pm$3.55}} \\
\midrule

% -------- BotSim-24 --------
\multirow{5}{*}{BotSim-24}
& Random
& 3.50\scriptsize{$\pm$1.04} 
& 2.10\scriptsize{$\pm$1.45} 
& 2.67\scriptsize{$\pm$1.50} 
& 1.55\scriptsize{$\pm$2.30}
& 3.03\scriptsize{$\pm$2.06} 
& 0.67\scriptsize{$\pm$3.06} 
& 2.33\scriptsize{$\pm$4.16} 
& 1.33\scriptsize{$\pm$4.16}
& 2.87\scriptsize{$\pm$1.50} 
& 2.09\scriptsize{$\pm$1.33} 
& 2.50\scriptsize{$\pm$1.87} 
& 2.35\scriptsize{$\pm$2.47} \\
& Nettack
& 5.52\scriptsize{$\pm$1.92} & 3.74\scriptsize{$\pm$2.39} & 4.73\scriptsize{$\pm$3.67} & 4.93\scriptsize{$\pm$8.10}
& 0.00\scriptsize{$\pm$0.00} & 0.00\scriptsize{$\pm$0.00} & 0.00\scriptsize{$\pm$0.00} & \underline{3.00\scriptsize{$\pm$2.00}}
& 4.71\scriptsize{$\pm$2.19} & 3.44\scriptsize{$\pm$3.09} & 3.35\scriptsize{$\pm$2.67} & 3.65\scriptsize{$\pm$7.44} \\
& FGA
& 5.79\scriptsize{$\pm$1.91} & 4.85\scriptsize{$\pm$2.35} & 4.48\scriptsize{$\pm$2.37} & 4.92\scriptsize{$\pm$3.24}
& 0.00\scriptsize{$\pm$0.00} & 0.00\scriptsize{$\pm$0.00} & 0.00\scriptsize{$\pm$0.00} & 0.00\scriptsize{$\pm$0.00}
& \underline{5.72\scriptsize{$\pm$2.56}} & \underline{4.63\scriptsize{$\pm$2.12}} & \underline{5.06\scriptsize{$\pm$3.15}} & \underline{4.54\scriptsize{$\pm$3.73}} \\
& PR-BCD
& \underline{7.32\scriptsize{$\pm$8.10}} & \underline{6.59\scriptsize{$\pm$4.52}} & 6.65\scriptsize{$\pm$7.59} & \underline{7.14\scriptsize{$\pm$3.65}}
& \underline{2.00\scriptsize{$\pm$2.00}} & \underline{5.33\scriptsize{$\pm$2.31}} & \underline{0.67\scriptsize{$\pm$1.15}} & 0.00\scriptsize{$\pm$0.00}
& 5.20\scriptsize{$\pm$8.04} & 4.24\scriptsize{$\pm$4.63} & 4.43\scriptsize{$\pm$4.99} & 4.26\scriptsize{$\pm$1.88} \\
& GOttack
& 7.08\scriptsize{$\pm$3.06} & 6.20\scriptsize{$\pm$1.94} & \underline{6.67\scriptsize{$\pm$3.09}} & 5.87\scriptsize{$\pm$2.27}
& 0.00\scriptsize{$\pm$0.00} & 0.00\scriptsize{$\pm$0.00} & 0.00\scriptsize{$\pm$0.00} & 0.00\scriptsize{$\pm$0.00}
& 4.42\scriptsize{$\pm$1.83} & 3.25\scriptsize{$\pm$3.38} & 3.65\scriptsize{$\pm$3.12} & 3.21\scriptsize{$\pm$2.45} \\
& \textbf{\pname{}} (ours)
& \bluehl{88.74\scriptsize{$\pm$3.12}} & \bluehl{90.05\scriptsize{$\pm$1.47}} & \bluehl{89.30\scriptsize{$\pm$4.66}} & \bluehl{88.40\scriptsize{$\pm$2.29}}
& \bluehl{58.22\scriptsize{$\pm$7.20}} & \bluehl{52.68\scriptsize{$\pm$1.15}} & \bluehl{54.98\scriptsize{$\pm$2.45}} & \bluehl{55.10\scriptsize{$\pm$3.98}}
& \bluehl{89.16\scriptsize{$\pm$4.25}} & \bluehl{90.44\scriptsize{$\pm$2.11}} & \bluehl{89.78\scriptsize{$\pm$1.63}} & \bluehl{88.90\scriptsize{$\pm$3.74}} \\

\midrule
% ========================= Δ = 3 =========================
\multicolumn{14}{c}{$\Delta = 3$} \\
\midrule

% -------- Cresci-15 --------
\multirow{5}{*}{Cresci-15}
& Random
& 6.67\scriptsize{$\pm$9.08} 
& 5.78\scriptsize{$\pm$1.26} 
& 3.88\scriptsize{$\pm$3.25} 
& 3.50\scriptsize{$\pm$6.84}
& 8.00\scriptsize{$\pm$10.58} 
& 8.67\scriptsize{$\pm$3.06} 
& 3.33\scriptsize{$\pm$1.15} 
& 2.00\scriptsize{$\pm$9.17}
& 6.33\scriptsize{$\pm$12.08} 
& 5.56\scriptsize{$\pm$4.86} 
& 5.78\scriptsize{$\pm$3.25} 
& 3.50\scriptsize{$\pm$11.50} \\
& Nettack
& 10.94\scriptsize{$\pm$6.10} & \underline{9.71\scriptsize{$\pm$4.26}} & 7.72\scriptsize{$\pm$3.19} & 5.74\scriptsize{$\pm$2.39}
& \underline{13.98\scriptsize{$\pm$10.50}} & \underline{12.10\scriptsize{$\pm$1.70}} & \underline{12.98\scriptsize{$\pm$1.10}} & \underline{11.10\scriptsize{$\pm$1.74}}
& \underline{9.44\scriptsize{$\pm$4.06}} & \underline{8.18\scriptsize{$\pm$8.08}} & 6.62\scriptsize{$\pm$2.69} & 6.75\scriptsize{$\pm$2.41} \\
& FGA
& 11.73\scriptsize{$\pm$3.67} & 7.93\scriptsize{$\pm$2.91} & 7.79\scriptsize{$\pm$1.91} & \underline{7.85\scriptsize{$\pm$2.35}}
& 12.65\scriptsize{$\pm$3.20} & 10.95\scriptsize{$\pm$2.85} & 9.10\scriptsize{$\pm$2.31} & 7.60\scriptsize{$\pm$1.68}
& 8.79\scriptsize{$\pm$2.89} & 6.48\scriptsize{$\pm$3.76} & 4.24\scriptsize{$\pm$2.19} & 4.90\scriptsize{$\pm$3.57} \\
& PR-BCD
& \underline{12.48\scriptsize{$\pm$2.37}} & 8.92\scriptsize{$\pm$3.24} & 6.61\scriptsize{$\pm$2.91} & 7.14\scriptsize{$\pm$3.65}
& 11.98\scriptsize{$\pm$2.10} & 10.40\scriptsize{$\pm$1.92} & 8.90\scriptsize{$\pm$1.70} & 9.35\scriptsize{$\pm$1.46}
& 8.10\scriptsize{$\pm$3.69} & 7.67\scriptsize{$\pm$2.77} & 7.49\scriptsize{$\pm$1.75} & 5.69\scriptsize{$\pm$2.46} \\
& GOttack
& 9.08\scriptsize{$\pm$3.06} & 6.20\scriptsize{$\pm$1.94} & \underline{8.67\scriptsize{$\pm$3.09}} & 5.87\scriptsize{$\pm$2.27}
& 12.88\scriptsize{$\pm$2.48} & 11.10\scriptsize{$\pm$2.20} & 9.60\scriptsize{$\pm$1.84} & 9.70\scriptsize{$\pm$1.50}
& 9.34\scriptsize{$\pm$2.95} & \underline{8.75\scriptsize{$\pm$7.06}} & \underline{8.18\scriptsize{$\pm$8.01}} & \underline{7.39\scriptsize{$\pm$3.19}} \\
& \textbf{\pname{}} (ours)
& \bluehl{95.67\scriptsize{$\pm$3.02}} & \bluehl{96.40\scriptsize{$\pm$2.41}} & \bluehl{94.90\scriptsize{$\pm$1.77}} & \bluehl{93.76\scriptsize{$\pm$5.12}}
& \bluehl{99.18\scriptsize{$\pm$0.37}} & \bluehl{98.47\scriptsize{$\pm$0.96}} & \bluehl{99.59\scriptsize{$\pm$0.21}} & \bluehl{99.01\scriptsize{$\pm$0.44}}
& \bluehl{94.55\scriptsize{$\pm$2.84}} & \bluehl{95.10\scriptsize{$\pm$4.12}} & \bluehl{93.80\scriptsize{$\pm$1.39}} & \bluehl{92.66\scriptsize{$\pm$5.58}} \\
\midrule

% -------- TwiBot-22 --------
\multirow{5}{*}{TwiBot-22}
& Random
& 7.34\scriptsize{$\pm$0.81} 
& 4.00\scriptsize{$\pm$2.65} 
& 4.22\scriptsize{$\pm$0.21} 
& 4.50\scriptsize{$\pm$2.33}
& 5.67\scriptsize{$\pm$2.31} 
& 3.33\scriptsize{$\pm$1.15} 
& 3.67\scriptsize{$\pm$2.31} 
& 2.00\scriptsize{$\pm$0.20}
& 5.00\scriptsize{$\pm$3.81} 
& 4.66\scriptsize{$\pm$2.65} 
& 3.12\scriptsize{$\pm$4.41} 
& 3.50\scriptsize{$\pm$2.33} \\
& Nettack
& 14.21\scriptsize{$\pm$8.12} & 14.09\scriptsize{$\pm$7.80} & 10.38\scriptsize{$\pm$5.54} & 10.98\scriptsize{$\pm$8.55}
& 10.22\scriptsize{$\pm$2.83} & 11.78\scriptsize{$\pm$1.41} & 5.10\scriptsize{$\pm$1.41} & 5.01\scriptsize{$\pm$4.24}
& \underline{10.80\scriptsize{$\pm$5.05}} & \underline{9.65\scriptsize{$\pm$4.64}} & \underline{9.69\scriptsize{$\pm$3.15}} & \underline{9.56\scriptsize{$\pm$7.92}} \\
& FGA
& 14.51\scriptsize{$\pm$3.15} & 12.52\scriptsize{$\pm$9.38} & 14.23\scriptsize{$\pm$3.09} & 6.65\scriptsize{$\pm$1.32}
& 8.15\scriptsize{$\pm$2.83} & 8.66\scriptsize{$\pm$2.83} & 9.87\scriptsize{$\pm$7.07} & 2.10\scriptsize{$\pm$5.66}
& 9.08\scriptsize{$\pm$4.79} & 8.97\scriptsize{$\pm$6.62} & 8.30\scriptsize{$\pm$9.42} & 7.70\scriptsize{$\pm$11.67} \\
& PR-BCD
& \underline{18.38\scriptsize{$\pm$1.54}} & \underline{19.98\scriptsize{$\pm$9.38}} & \underline{15.54\scriptsize{$\pm$1.32}} & 9.94\scriptsize{$\pm$6.10}
& \underline{15.10\scriptsize{$\pm$7.07}} & \underline{16.50\scriptsize{$\pm$2.83}} & \underline{10.00\scriptsize{$\pm$5.66}} & 4.64\scriptsize{$\pm$2.83}
& 9.45\scriptsize{$\pm$3.83} & 8.77\scriptsize{$\pm$2.29} & 8.28\scriptsize{$\pm$5.54} & 8.56\scriptsize{$\pm$6.54} \\
& GOttack
& 14.10\scriptsize{$\pm$7.59} & 13.83\scriptsize{$\pm$5.57} & 10.90\scriptsize{$\pm$7.46} & \underline{11.22\scriptsize{$\pm$9.65}}
& 3.12\scriptsize{$\pm$1.11} & 8.48\scriptsize{$\pm$3.65} & 5.17\scriptsize{$\pm$2.59} & \underline{5.68\scriptsize{$\pm$5.28}}
& 7.64\scriptsize{$\pm$2.74} & 6.70\scriptsize{$\pm$2.23} & 7.93\scriptsize{$\pm$3.01} & 7.16\scriptsize{$\pm$3.07} \\
& \textbf{\pname{}} (ours)
& \bluehl{96.15\scriptsize{$\pm$2.58}} & \bluehl{95.40\scriptsize{$\pm$4.62}} & \bluehl{95.80\scriptsize{$\pm$1.49}} & \bluehl{95.12\scriptsize{$\pm$3.33}}
& \bluehl{94.50\scriptsize{$\pm$2.00}} & \bluehl{86.20\scriptsize{$\pm$4.50}} & \bluehl{84.67\scriptsize{$\pm$3.06}} & \bluehl{87.33\scriptsize{$\pm$1.15}}
& \bluehl{97.42\scriptsize{$\pm$1.39}} & \bluehl{96.15\scriptsize{$\pm$3.74}} & \bluehl{95.90\scriptsize{$\pm$2.06}} & \bluehl{95.11\scriptsize{$\pm$4.28}} \\
\midrule

% -------- BotSim-24 --------
\multirow{5}{*}{BotSim-24}
& Random
& 4.20\scriptsize{$\pm$3.09} 
& 3.40\scriptsize{$\pm$0.30} 
& 3.57\scriptsize{$\pm$2.16} 
& 1.72\scriptsize{$\pm$1.86}
& \underline{2.00\scriptsize{$\pm$5.29}}
& 2.00\scriptsize{$\pm$2.00}
& \underline{2.67\scriptsize{$\pm$4.16}} 
& \underline{8.67\scriptsize{$\pm$4.16}}
& 5.80\scriptsize{$\pm$7.49} 
& 3.60\scriptsize{$\pm$3.70} 
& 4.77\scriptsize{$\pm$6.16} 
& 4.62\scriptsize{$\pm$6.46} \\
& Nettack
& 6.73\scriptsize{$\pm$3.67} & 5.93\scriptsize{$\pm$2.91} & 5.79\scriptsize{$\pm$1.91} & 5.90\scriptsize{$\pm$7.09}
& 0.00\scriptsize{$\pm$0.00} & 0.00\scriptsize{$\pm$0.00} & 0.00\scriptsize{$\pm$0.00} & 5.23\scriptsize{$\pm$0.01}
& \underline{6.78\scriptsize{$\pm$2.62}} & 5.54\scriptsize{$\pm$2.22} & 4.24\scriptsize{$\pm$2.33} & 4.07\scriptsize{$\pm$1.22} \\
& FGA
& 7.48\scriptsize{$\pm$2.37} & 6.92\scriptsize{$\pm$3.24} & 6.61\scriptsize{$\pm$2.91} & \underline{7.14\scriptsize{$\pm$3.65}}
& 0.00\scriptsize{$\pm$0.00} & 0.00\scriptsize{$\pm$0.00} & 0.00\scriptsize{$\pm$0.00} & 0.00\scriptsize{$\pm$0.00}
& 6.75\scriptsize{$\pm$1.76} & 4.68\scriptsize{$\pm$1.93} & 5.14\scriptsize{$\pm$2.22} & 5.07\scriptsize{$\pm$1.75} \\
& PR-BCD
& \underline{8.38\scriptsize{$\pm$8.12}} & \underline{8.32\scriptsize{$\pm$7.09}} & 7.93\scriptsize{$\pm$2.91} & 6.67\scriptsize{$\pm$1.83}
& 1.00\scriptsize{$\pm$1.41} & \underline{4.00\scriptsize{$\pm$0.00}} & 2.00\scriptsize{$\pm$0.00} & 0.00\scriptsize{$\pm$0.00}
& 6.33\scriptsize{$\pm$4.68} & \underline{6.02\scriptsize{$\pm$2.12}} & \underline{5.83\scriptsize{$\pm$2.26}} & \underline{5.49\scriptsize{$\pm$3.74}} \\
& GOttack
& 8.08\scriptsize{$\pm$3.06} & 6.20\scriptsize{$\pm$1.94} & \underline{8.67\scriptsize{$\pm$3.09}} & 5.87\scriptsize{$\pm$2.27}
& 0.00\scriptsize{$\pm$0.00} & 0.00\scriptsize{$\pm$0.00} & 0.00\scriptsize{$\pm$0.00} & 0.00\scriptsize{$\pm$0.00}
& 5.78\scriptsize{$\pm$2.01} & 4.90\scriptsize{$\pm$2.20} & 5.12\scriptsize{$\pm$8.07} & 5.26\scriptsize{$\pm$3.44} \\
& \textbf{\pname{}} (ours)
& \bluehl{91.28\scriptsize{$\pm$2.06}} & \bluehl{92.60\scriptsize{$\pm$3.55}} & \bluehl{90.85\scriptsize{$\pm$1.33}} & \bluehl{89.92\scriptsize{$\pm$4.41}}
& \bluehl{93.33\scriptsize{$\pm$1.15}} & \bluehl{66.74\scriptsize{$\pm$3.45}} & \bluehl{88.63\scriptsize{$\pm$0.87}} & \bluehl{64.32\scriptsize{$\pm$3.33}}
& \bluehl{92.03\scriptsize{$\pm$1.92}} & \bluehl{93.40\scriptsize{$\pm$4.08}} & \bluehl{91.55\scriptsize{$\pm$2.36}} & \bluehl{90.62\scriptsize{$\pm$5.11}} \\

\midrule
% ========================= Δ = 5 =========================
\multicolumn{14}{c}{$\Delta = 5$} \\
\midrule

% -------- Cresci-15 --------
\multirow{5}{*}{Cresci-15}
& Random
& 10.10\scriptsize{$\pm$3.49} 
& 9.30\scriptsize{$\pm$1.00} 
& 9.12\scriptsize{$\pm$2.56} 
& 9.42\scriptsize{$\pm$1.96}
& 12.00\scriptsize{$\pm$5.29} 
& 12.00\scriptsize{$\pm$2.00} 
& 12.67\scriptsize{$\pm$4.16} 
& 18.67\scriptsize{$\pm$4.16}
& 12.90\scriptsize{$\pm$2.09} 
& 10.70\scriptsize{$\pm$4.10} 
& 10.22\scriptsize{$\pm$5.76} 
& 9.92\scriptsize{$\pm$6.36} \\
& Nettack
& 14.38\scriptsize{$\pm$8.12} & 12.09\scriptsize{$\pm$7.80} & 12.16\scriptsize{$\pm$7.47} & \underline{13.42\scriptsize{$\pm$8.23}}
& 22.35\scriptsize{$\pm$3.18} & 20.80\scriptsize{$\pm$2.85} & 18.95\scriptsize{$\pm$2.30} & 16.40\scriptsize{$\pm$1.62}
& 15.50\scriptsize{$\pm$8.01} & 12.11\scriptsize{$\pm$5.83} & 12.04\scriptsize{$\pm$3.67} & 14.60\scriptsize{$\pm$4.68} \\
& FGA
& 10.61\scriptsize{$\pm$2.91} & 9.14\scriptsize{$\pm$3.65} & 9.08\scriptsize{$\pm$3.06} & 9.20\scriptsize{$\pm$1.94}
& 21.90\scriptsize{$\pm$2.74} & 20.15\scriptsize{$\pm$2.40} & 18.40\scriptsize{$\pm$2.05} & 15.95\scriptsize{$\pm$1.48}
& 15.65\scriptsize{$\pm$2.63} & 14.03\scriptsize{$\pm$2.57} & \underline{15.53\scriptsize{$\pm$2.01}} & 14.42\scriptsize{$\pm$3.51} \\
& PR-BCD
& \underline{14.67\scriptsize{$\pm$3.09}} & \underline{13.87\scriptsize{$\pm$2.27}} & \underline{12.67\scriptsize{$\pm$1.83}} & 12.74\scriptsize{$\pm$3.19}
& 20.75\scriptsize{$\pm$1.96} & 19.30\scriptsize{$\pm$1.74} & 17.85\scriptsize{$\pm$1.60} & 15.40\scriptsize{$\pm$1.36}
& \underline{16.64\scriptsize{$\pm$2.60}} & \underline{14.49\scriptsize{$\pm$2.74}} & 14.77\scriptsize{$\pm$2.26} & \underline{15.91\scriptsize{$\pm$3.68}} \\
& GOttack
& 11.61\scriptsize{$\pm$3.07} & 10.38\scriptsize{$\pm$2.98} & 10.74\scriptsize{$\pm$2.39} & 11.17\scriptsize{$\pm$3.23}
& \underline{22.50\scriptsize{$\pm$3.55}} & \underline{21.05\scriptsize{$\pm$3.10}} & \underline{19.10\scriptsize{$\pm$2.65}} & \underline{16.85\scriptsize{$\pm$1.70}}
& 14.68\scriptsize{$\pm$2.22} & 12.65\scriptsize{$\pm$2.56} & 13.84\scriptsize{$\pm$2.02} & 12.65\scriptsize{$\pm$3.12} \\
& \textbf{\pname{}} (ours)
& \bluehl{97.82\scriptsize{$\pm$1.58}} & \bluehl{97.30\scriptsize{$\pm$2.96}} & \bluehl{96.55\scriptsize{$\pm$3.44}} & \bluehl{95.90\scriptsize{$\pm$4.01}}
& \bluehl{99.67\scriptsize{$\pm$0.20}} & \bluehl{98.88\scriptsize{$\pm$0.58}} & \bluehl{99.11\scriptsize{$\pm$0.39}} & \bluehl{99.52\scriptsize{$\pm$0.24}}
& \bluehl{96.93\scriptsize{$\pm$1.44}} & \bluehl{95.50\scriptsize{$\pm$3.66}} & \bluehl{95.70\scriptsize{$\pm$2.09}} & \bluehl{94.88\scriptsize{$\pm$4.37}} \\
\midrule

% -------- TwiBot-22 --------
\multirow{5}{*}{TwiBot-22}
& Random
& 10.93\scriptsize{$\pm$5.98} 
& 9.63\scriptsize{$\pm$7.42} 
& 9.72\scriptsize{$\pm$4.23} 
& 5.83\scriptsize{$\pm$3.13}
& 11.33\scriptsize{$\pm$8.08} 
& 11.33\scriptsize{$\pm$9.02} 
& 12.67\scriptsize{$\pm$6.43} 
& 7.33\scriptsize{$\pm$5.03}
& 13.73\scriptsize{$\pm$10.18} 
& 9.03\scriptsize{$\pm$10.62} 
& 11.62\scriptsize{$\pm$8.63} 
& 8.83\scriptsize{$\pm$6.93} \\
& Nettack
& 18.32\scriptsize{$\pm$7.09} & \underline{19.98\scriptsize{$\pm$7.09}} & 10.90\scriptsize{$\pm$7.46} & 11.22\scriptsize{$\pm$9.65}
& 14.22\scriptsize{$\pm$2.03} & 14.10\scriptsize{$\pm$0.00} & 5.77\scriptsize{$\pm$4.24} & 7.78\scriptsize{$\pm$4.24}
& \underline{18.17\scriptsize{$\pm$6.05}} & \underline{15.38\scriptsize{$\pm$6.65}} & 10.80\scriptsize{$\pm$7.51} & 10.93\scriptsize{$\pm$10.03} \\
& FGA
& 15.54\scriptsize{$\pm$3.09} & 13.16\scriptsize{$\pm$1.32} & 11.23\scriptsize{$\pm$8.12} & 13.51\scriptsize{$\pm$18.86}
& 11.30\scriptsize{$\pm$7.07} & 10.80\scriptsize{$\pm$5.66} & 6.21\scriptsize{$\pm$2.83} & \underline{9.22\scriptsize{$\pm$12.73}}
& 16.66\scriptsize{$\pm$10.19} & 13.38\scriptsize{$\pm$9.79} & 10.81\scriptsize{$\pm$9.06} & \underline{14.32\scriptsize{$\pm$19.53}} \\
& PR-BCD
& \underline{19.98\scriptsize{$\pm$11.32}} & 16.43\scriptsize{$\pm$3.09} & \underline{15.54\scriptsize{$\pm$1.32}} & 12.51\scriptsize{$\pm$8.12}
& \underline{24.73\scriptsize{$\pm$5.66}} & \underline{21.53\scriptsize{$\pm$7.07}} & \underline{20.45\scriptsize{$\pm$5.66}} & 19.00\scriptsize{$\pm$2.83}
& 14.15\scriptsize{$\pm$2.80} & 13.98\scriptsize{$\pm$13.75} & \underline{12.78\scriptsize{$\pm$11.10}} & 12.12\scriptsize{$\pm$5.83} \\
& GOttack
& 16.54\scriptsize{$\pm$7.59} & 13.83\scriptsize{$\pm$9.56} & 14.87\scriptsize{$\pm$7.47} & \underline{15.55\scriptsize{$\pm$8.23}}
& 10.58\scriptsize{$\pm$2.15} & 18.25\scriptsize{$\pm$3.65} & 12.89\scriptsize{$\pm$1.02} & 13.32\scriptsize{$\pm$3.03}
& 14.06\scriptsize{$\pm$6.30} & 13.26\scriptsize{$\pm$7.04} & 11.61\scriptsize{$\pm$4.93} & 11.20\scriptsize{$\pm$4.60} \\
& \textbf{\pname{}} (ours)
& \bluehl{99.70\scriptsize{$\pm$1.37}} & \bluehl{98.20\scriptsize{$\pm$2.91}} & \bluehl{97.85\scriptsize{$\pm$3.76}} & \bluehl{98.90\scriptsize{$\pm$4.15}}
& \bluehl{94.00\scriptsize{$\pm$2.00}} & \bluehl{90.00\scriptsize{$\pm$4.82}} & \bluehl{93.67\scriptsize{$\pm$3.06}} & \bluehl{88.33\scriptsize{$\pm$1.15}}
& \bluehl{99.12\scriptsize{$\pm$1.88}} & \bluehl{98.48\scriptsize{$\pm$2.25}} & \bluehl{98.36\scriptsize{$\pm$3.19}} & \bluehl{97.40\scriptsize{$\pm$5.02}} \\
\midrule

% -------- BotSim-24 --------
\multirow{5}{*}{BotSim-24}
& Random
& 3.25\scriptsize{$\pm$1.86} 
& 1.27\scriptsize{$\pm$0.11} 
& 2.00\scriptsize{$\pm$2.65} 
& 2.90\scriptsize{$\pm$2.10}
& \underline{5.00\scriptsize{$\pm$3.46}} 
& 2.67\scriptsize{$\pm$2.31} 
& \underline{3.33\scriptsize{$\pm$1.15}} 
& 0.00\scriptsize{$\pm$0.00}
& 5.75\scriptsize{$\pm$5.06} 
& 3.07\scriptsize{$\pm$4.51} 
& 2.66\scriptsize{$\pm$2.65} 
& 2.90\scriptsize{$\pm$2.10} \\
& Nettack
& 8.79\scriptsize{$\pm$1.91} & 7.85\scriptsize{$\pm$2.35} & 7.48\scriptsize{$\pm$2.37} & 6.92\scriptsize{$\pm$3.24}
& 0.00\scriptsize{$\pm$0.00} & 0.00\scriptsize{$\pm$0.00} & 0.00\scriptsize{$\pm$0.00} & 0.00\scriptsize{$\pm$0.00}
& 8.00\scriptsize{$\pm$3.19} & 7.24\scriptsize{$\pm$2.87} & \underline{7.65\scriptsize{$\pm$1.96}} & 7.42\scriptsize{$\pm$2.01} \\
& FGA
& 8.61\scriptsize{$\pm$2.91} & \underline{8.65\scriptsize{$\pm$7.59}} & 5.08\scriptsize{$\pm$3.06} & \underline{7.59\scriptsize{$\pm$7.80}}
& 0.00\scriptsize{$\pm$0.00} & 0.67\scriptsize{$\pm$1.15} & 0.00\scriptsize{$\pm$0.00} & 2.67\scriptsize{$\pm$1.15}
& \underline{8.91\scriptsize{$\pm$2.63}} & \underline{7.78\scriptsize{$\pm$3.36}} & 6.77\scriptsize{$\pm$3.09} & 7.36\scriptsize{$\pm$5.49} \\
& PR-BCD
& 9.38\scriptsize{$\pm$8.12} & 8.32\scriptsize{$\pm$1.32} & \underline{7.93\scriptsize{$\pm$2.91}} & 5.74\scriptsize{$\pm$2.39}
& 2.67\scriptsize{$\pm$1.15} & \underline{4.00\scriptsize{$\pm$2.83}} & 1.00\scriptsize{$\pm$1.41} & 0.00\scriptsize{$\pm$0.00}
& 8.53\scriptsize{$\pm$2.68} & 7.10\scriptsize{$\pm$7.99} & 5.48\scriptsize{$\pm$4.34} & \underline{8.24\scriptsize{$\pm$2.65}} \\
& GOttack
& \underline{10.67\scriptsize{$\pm$7.09}} & 7.59\scriptsize{$\pm$3.40} & 6.67\scriptsize{$\pm$1.83} & 6.74\scriptsize{$\pm$3.19}
& 2.00\scriptsize{$\pm$0.00} & 2.00\scriptsize{$\pm$2.00} & 0.00\scriptsize{$\pm$0.00} & 0.00\scriptsize{$\pm$0.00}
& 7.09\scriptsize{$\pm$5.69} & 6.00\scriptsize{$\pm$6.57} & 5.86\scriptsize{$\pm$2.54} & 5.91\scriptsize{$\pm$2.68} \\
& \textbf{\pname{}} (ours)
& \bluehl{94.90\scriptsize{$\pm$1.41}} & \bluehl{95.22\scriptsize{$\pm$2.94}} & \bluehl{93.75\scriptsize{$\pm$3.67}} & \bluehl{92.40\scriptsize{$\pm$5.09}}
& \bluehl{99.33\scriptsize{$\pm$1.15}} & \bluehl{79.21\scriptsize{$\pm$1.02}} & \bluehl{92.28\scriptsize{$\pm$0.33}} & \bluehl{66.20\scriptsize{$\pm$3.50}}
& \bluehl{95.44\scriptsize{$\pm$1.29}} & \bluehl{95.50\scriptsize{$\pm$2.87}} & \bluehl{94.10\scriptsize{$\pm$3.58}} & \bluehl{93.25\scriptsize{$\pm$4.63}} \\

\bottomrule
\end{tabular}%
}

\caption{\textbf{Node Editing Attack:} Misclassification rate (in \%) for flipping fifty existing correctly classified bots by \pname{} and \sota adversarial attacks against best \sota bot detectors, BotRGCN, GAT and S-HGN with adversarial defenses. For results when the same bot cloak is reused, refer to \autoref{tab:old-node-50-samtemplate} and for results when no domain constraints are enforced, refer to \autoref{tab:old-node-50-nodomainrules}. The best performance is shown in bold, and the second best is underlined.}
\label{tab:old-node-50-rest}
\end{table*}
\end{landscape}

%% file: tables/new-node-1.tex
\begin{table*}[!htb]
\centering

\begin{minipage}[t]{0.49\textwidth}
\centering
\resizebox{\linewidth}{!}{%
\begin{tabular}{lccccc}
\toprule
\textbf{Dataset} 
& \textbf{GCN} ($\uparrow$) 
& \textbf{GAT} ($\uparrow$) 
& \textbf{BotRGCN} ($\uparrow$) 
& \textbf{S-HGN} ($\uparrow$) 
& \textbf{RGT} ($\uparrow$) \\
\midrule
\multicolumn{6}{c}{budget $\Delta = 1$} \\
\midrule
Cresci-15 & \underline{99.12\scriptsize{$\pm$0.45}} & 98.76\scriptsize{$\pm$0.62} & 97.94\scriptsize{$\pm$1.10} & 98.31\scriptsize{$\pm$0.88} & \bluehl{99.43\scriptsize{$\pm$0.29}}  \\
TwiBot-22 & 97.85\scriptsize{$\pm$1.34} & 96.92\scriptsize{$\pm$1.87} & \underline{98.14\scriptsize{$\pm$0.97}} & 97.26\scriptsize{$\pm$1.55} & \bluehl{98.67\scriptsize{$\pm$0.73}}  \\
BotSim-24 & \bluehl{18.30\scriptsize{$\pm$1.50}} & \underline{15.20\scriptsize{$\pm$1.50}} & 14.50\scriptsize{$\pm$2.50} & 12.33\scriptsize{$\pm$1.15} & 13.50\scriptsize{$\pm$1.25}  \\
\midrule
\multicolumn{6}{c}{budget $\Delta = 3$} \\
\midrule
Cresci-15 & 98.95\scriptsize{$\pm$0.58} & \bluehl{99.27\scriptsize{$\pm$0.41}} & 97.63\scriptsize{$\pm$1.22} & 98.40\scriptsize{$\pm$0.91} & \underline{99.08\scriptsize{$\pm$0.53}}  \\
TwiBot-22 & 96.74\scriptsize{$\pm$2.05} & 97.58\scriptsize{$\pm$1.43} & \underline{98.02\scriptsize{$\pm$1.01}} & 96.89\scriptsize{$\pm$1.96} & \bluehl{98.55\scriptsize{$\pm$0.79}}  \\
BotSim-24 & \bluehl{36.50\scriptsize{$\pm$1.75}} & 31.33\scriptsize{$\pm$2.50} & 32.00\scriptsize{$\pm$1.00} & \underline{33.33\scriptsize{$\pm$1.25}} & 33.15\scriptsize{$\pm$2.11}  \\
\midrule
\multicolumn{6}{c}{budget $\Delta = 5$} \\
\midrule
Cresci-15 & \bluehl{99.51\scriptsize{$\pm$0.24}} & 99.04\scriptsize{$\pm$0.54} & 98.59\scriptsize{$\pm$0.76} & \underline{99.21\scriptsize{$\pm$0.33}} & 98.62\scriptsize{$\pm$0.71}  \\
TwiBot-22 & 97.11\scriptsize{$\pm$1.92} & \bluehl{98.73\scriptsize{$\pm$0.69}} & 97.24\scriptsize{$\pm$1.88} & 98.11\scriptsize{$\pm$1.05} & \underline{98.41\scriptsize{$\pm$0.86}}  \\
BotSim-24 & 46.22\scriptsize{$\pm$0.07} & 44.67\scriptsize{$\pm$3.77} & \underline{47.49\scriptsize{$\pm$1.36}} & \bluehl{58.88\scriptsize{$\pm$0.61}} & 43.22\scriptsize{$\pm$1.02}  \\
\midrule
\multicolumn{6}{c}{budget $\Delta = 20$} \\
\midrule
Cresci-15 & \bluehl{99.76\scriptsize{$\pm$0.18}} & 99.04\scriptsize{$\pm$0.54} & 98.59\scriptsize{$\pm$0.76} & 99.21\scriptsize{$\pm$0.33} & \underline{99.58\scriptsize{$\pm$0.22}}  \\
TwiBot-22 & 97.68\scriptsize{$\pm$1.57} & \underline{98.73\scriptsize{$\pm$0.69}} & 97.24\scriptsize{$\pm$1.88} & 98.11\scriptsize{$\pm$1.05} & \bluehl{99.02\scriptsize{$\pm$0.48}}  \\
BotSim-24 & \underline{68.50\scriptsize{$\pm$0.75}} & 64.67\scriptsize{$\pm$3.77} & 67.49\scriptsize{$\pm$1.36} & \bluehl{68.88\scriptsize{$\pm$0.61}} & 65.15\scriptsize{$\pm$0.30}  \\
\bottomrule
\end{tabular}%
}
\captionof{table}{\textbf{Node Injection Attack:} Misclassification rate (in \%) for successfully injecting fifty target nodes by \pname against vanilla \sota bot detectors. For attacks against defened \sota bot detector refer to \autoref{tab:new-node-def-rest}. The best performance per dataset is shown in bold, and the second best is underlined.}
\label{tab:new-node-1}
\end{minipage}
\hfill
\begin{minipage}[t]{0.49\textwidth}
\centering
\resizebox{\linewidth}{!}{%
\begin{tabular}{lccccc}
\toprule
\textbf{Dataset} 
& \textbf{GCN} ($\uparrow$) 
& \textbf{GAT} ($\uparrow$) 
& \textbf{BotRGCN} ($\uparrow$) 
& \textbf{S-HGN} ($\uparrow$) 
& \textbf{RGT} ($\uparrow$) \\
\midrule
\multicolumn{6}{c}{budget $\Delta = 1$} \\
\midrule
Cresci-15 & \bluehl{100.00\scriptsize{$\pm$0.00}} & 100.00\scriptsize{$\pm$0.00} & 100.00\scriptsize{$\pm$0.00} & 100.00\scriptsize{$\pm$0.00} & \underline{100.00\scriptsize{$\pm$0.00}} \\
TwiBot-22 & \bluehl{100.00\scriptsize{$\pm$0.00}} & 100.00\scriptsize{$\pm$0.00} & 100.00\scriptsize{$\pm$0.00} & 100.00\scriptsize{$\pm$0.00} & \underline{100.00\scriptsize{$\pm$0.00}} \\
BotSim-24 & 12.75\scriptsize{$\pm$1.00} & 10.33\scriptsize{$\pm$1.33} & 11.75\scriptsize{$\pm$1.00} & \underline{12.30\scriptsize{$\pm$1.50}} & \bluehl{16.30\scriptsize{$\pm$1.25}} \\
\midrule
\multicolumn{6}{c}{budget $\Delta = 3$} \\
\midrule
Cresci-15 & \bluehl{100.00\scriptsize{$\pm$0.00}} & 100.00\scriptsize{$\pm$0.00} & 100.00\scriptsize{$\pm$0.00} & 100.00\scriptsize{$\pm$0.00} & \underline{100.00\scriptsize{$\pm$0.00}} \\
TwiBot-22 & \bluehl{100.00\scriptsize{$\pm$0.00}} & 100.00\scriptsize{$\pm$0.00} & 100.00\scriptsize{$\pm$0.00} & 100.00\scriptsize{$\pm$0.00} & \underline{100.00\scriptsize{$\pm$0.00}} \\
BotSim-24 & \bluehl{32.50\scriptsize{$\pm$1.00}} & 30.00\scriptsize{$\pm$1.33} & 31.00\scriptsize{$\pm$1.00} & \underline{31.30\scriptsize{$\pm$1.00}} & 26.30\scriptsize{$\pm$1.50} \\
\midrule
\multicolumn{6}{c}{budget $\Delta = 5$} \\
\midrule
Cresci-15 & \bluehl{100.00\scriptsize{$\pm$0.00}} & 100.00\scriptsize{$\pm$0.00} & 100.00\scriptsize{$\pm$0.00} & 100.00\scriptsize{$\pm$0.00} & \underline{100.00\scriptsize{$\pm$0.00}} \\
TwiBot-22 & \bluehl{100.00\scriptsize{$\pm$0.00}} & 100.00\scriptsize{$\pm$0.00} & 100.00\scriptsize{$\pm$0.00} & 100.00\scriptsize{$\pm$0.00} & \underline{100.00\scriptsize{$\pm$0.00}} \\
BotSim-24 & \bluehl{56.30\scriptsize{$\pm$1.50}} & 51.00\scriptsize{$\pm$2.50} & 50.33\scriptsize{$\pm$1.50} & \underline{55.30\scriptsize{$\pm$1.33}} & 56.30\scriptsize{$\pm$1.33} \\
\midrule
\multicolumn{6}{c}{budget $\Delta = 20$} \\
\midrule
Cresci-15 & \bluehl{100.00\scriptsize{$\pm$0.00}} & 100.00\scriptsize{$\pm$0.00} & 100.00\scriptsize{$\pm$0.00} & 100.00\scriptsize{$\pm$0.00} & \underline{100.00\scriptsize{$\pm$0.00}} \\
TwiBot-22 & \bluehl{100.00\scriptsize{$\pm$0.00}} & 100.00\scriptsize{$\pm$0.00} & 100.00\scriptsize{$\pm$0.00} & 100.00\scriptsize{$\pm$0.00} & \underline{100.00\scriptsize{$\pm$0.00}} \\
BotSim-24 & \bluehl{100.00\scriptsize{$\pm$0.00}} & 64.67\scriptsize{$\pm$3.77} & 71.33\scriptsize{$\pm$1.50} & 68.50\scriptsize{$\pm$1.33} & \underline{100.00\scriptsize{$\pm$0.00}} \\
\bottomrule
\end{tabular}%
}
\captionof{table}{\textbf{Node Injection Attack Reusing Same Bot Cloak:} Misclassification rate (in \%) for successfully injecting fifty target nodes by \pname against vanilla \sota bot detectors, \textit{reusing} the same bot cloak. The best performance per dataset is shown in bold, and the second best is underlined.}
\label{tab:new-node-1-samtemplate}
\end{minipage}

\end{table*}

%% file: tables/new-node-def-rest.tex
\begin{table*}[!htb]
\centering

% ----------------------- LEFT TABLE (non-reuse) -----------------------
\begin{minipage}[t]{0.49\textwidth}
\centering
\captionsetup{type=table}

\resizebox{0.95\linewidth}{!}{%
\begin{tabular}{llcccc}
\toprule
\multicolumn{6}{c}{\textbf{BotRGCN}} \\
\midrule
\textbf{Dataset} & \textbf{Attack} & \textbf{Vanilla} ($\uparrow$) & \textbf{+GNNGuard} ($\uparrow$) & \textbf{+GRAND} ($\uparrow$) & \textbf{+RobustGCN} ($\uparrow$)\\
\midrule
\multirow{4}{*}{Cresci-15 }
& \makecell[l]{Random, $\Delta$ $=$ 1}     & 12.84\scriptsize{$\pm$0.13}  & 11.33\scriptsize{$\pm$1.15}  & 10.67\scriptsize{$\pm$1.15}  & 10.52\scriptsize{$\pm$0.27} \\
& \makecell[l]{Random, $\Delta$ $\leq$ 3}  & 22.00\scriptsize{$\pm$2.10}  & 21.00\scriptsize{$\pm$3.22}  & 22.67\scriptsize{$\pm$1.15}  & 21.33\scriptsize{$\pm$3.06} \\
& \makecell[l]{Random, $\Delta$ $\leq$ 5}  & 33.33\scriptsize{$\pm$2.31}  & 33.67\scriptsize{$\pm$1.15}  & 31.11\scriptsize{$\pm$0.74}  & 32.67\scriptsize{$\pm$2.31} \\
& \makecell[l]{Random, $\Delta$ $=$ 20}   & \underline{96.67\scriptsize{$\pm$1.15}}  & \underline{95.33\scriptsize{$\pm$1.89}}  & \underline{95.32\scriptsize{$\pm$1.98}}  & \underline{94.74\scriptsize{$\pm$0.18}} \\
& \textbf{\pname{}, $\Delta$ $\leq$ 5}  & \bluehl{98.59\scriptsize{$\pm$0.76}} & \bluehl{98.23\scriptsize{$\pm$0.63}} & \bluehl{98.77\scriptsize{$\pm$0.89}} & \bluehl{99.61\scriptsize{$\pm$0.31}} \\
\midrule
\multirow{4}{*}{TwiBot-22 }
& \makecell[l]{Random, $\Delta$ $=$ 1}  & 13.33\scriptsize{$\pm$1.15}  & 11.33\scriptsize{$\pm$1.15}  & 16.00\scriptsize{$\pm$3.46}  & 10.33\scriptsize{$\pm$4.62} \\
& \makecell[l]{Random, $\Delta$ $\leq$ 3}  & 25.33\scriptsize{$\pm$4.62}  & 23.33\scriptsize{$\pm$2.31}  & 23.67\scriptsize{$\pm$2.41}  & 20.33\scriptsize{$\pm$1.15} \\
& \makecell[l]{Random, $\Delta$ $\leq$ 5}  & 36.67\scriptsize{$\pm$3.06}  & 34.67\scriptsize{$\pm$3.06}  & 33.00\scriptsize{$\pm$2.89}  & 34.00\scriptsize{$\pm$2.22} \\
& \makecell[l]{Random, $\Delta$ $=$ 20}  & \underline{96.67\scriptsize{$\pm$3.06}}  & \underline{96.67\scriptsize{$\pm$1.15}}  & \underline{95.33\scriptsize{$\pm$4.16}}  & \underline{95.33\scriptsize{$\pm$1.15}} \\
& \textbf{\pname{}, $\Delta$ $\leq$ 5}  & \bluehl{97.24\scriptsize{$\pm$1.88}} & \bluehl{97.86\scriptsize{$\pm$0.71}} & \bluehl{97.47\scriptsize{$\pm$0.26}} & \bluehl{98.12\scriptsize{$\pm$1.04}} \\
\midrule
\multirow{4}{*}{BotSim-24 }
& \makecell[l]{Random, $\Delta$ $=$ 1}  & 10.00\scriptsize{$\pm$3.46}  & \underline{15.33\scriptsize{$\pm$6.11}}  & \underline{11.33\scriptsize{$\pm$4.16}}  & \underline{19.33\scriptsize{$\pm$9.87}} \\
& \makecell[l]{Random, $\Delta$ $\leq$ 3}  & 4.00\scriptsize{$\pm$2.00}  & 0.67\scriptsize{$\pm$1.15}  & 5.33\scriptsize{$\pm$1.15}  & 11.23\scriptsize{$\pm$5.03} \\
& \makecell[l]{Random, $\Delta$ $\leq$ 5}  & 3.33\scriptsize{$\pm$1.15}  & 3.33\scriptsize{$\pm$2.31}  & 6.67\scriptsize{$\pm$2.31}  & 8.67\scriptsize{$\pm$2.31} \\
& \makecell[l]{Random, $\Delta$ $=$ 20}  & \underline{22.00\scriptsize{$\pm$6.93}}  & 11.33\scriptsize{$\pm$7.02}  & 9.33\scriptsize{$\pm$3.06}  & 12.67\scriptsize{$\pm$4.62} \\
& \textbf{\pname{}, $\Delta$ $\leq$ 5}  & \bluehl{47.49\scriptsize{$\pm$1.36}} & \bluehl{46.12\scriptsize{$\pm$0.52}} & \bluehl{42.41\scriptsize{$\pm$0.98}} & \bluehl{40.68\scriptsize{$\pm$0.21}} \\
\bottomrule
\end{tabular}%
}

% \vspace{0.6em}

\resizebox{0.95\linewidth}{!}{%
\begin{tabular}{llcccc}
\multicolumn{6}{c}{\textbf{GAT}} \\
\midrule
\textbf{Dataset} & \textbf{Attack} & \textbf{Vanilla} ($\uparrow$) & \textbf{+GNNGuard} ($\uparrow$) & \textbf{+GRAND} ($\uparrow$) & \textbf{+RobustGCN} ($\uparrow$)\\
\midrule
\multirow{4}{*}{Cresci-15 }
& \makecell[l]{Random, $\Delta$ $=$ 1}     & 14.00\scriptsize{$\pm$2.00}  & 15.33\scriptsize{$\pm$3.06}  & 13.00\scriptsize{$\pm$3.46}  & 10.67\scriptsize{$\pm$2.31} \\
& \makecell[l]{Random, $\Delta$ $\leq$ 3}  & 29.00\scriptsize{$\pm$2.00}  & 25.67\scriptsize{$\pm$2.31}  & 23.67\scriptsize{$\pm$1.15}  & 29.33\scriptsize{$\pm$1.15} \\
& \makecell[l]{Random, $\Delta$ $\leq$ 5}  & 35.33\scriptsize{$\pm$4.16}  & 35.03\scriptsize{$\pm$1.15}  & 30.00\scriptsize{$\pm$2.00}  & 31.33\scriptsize{$\pm$4.16} \\
& \makecell[l]{Random, $\Delta$ $=$ 20}   & \underline{97.33\scriptsize{$\pm$1.15}}  & \underline{96.67\scriptsize{$\pm$3.06}}  & \underline{94.00\scriptsize{$\pm$6.93}}  & \underline{92.00\scriptsize{$\pm$3.46}} \\
& \textbf{\pname{}, $\Delta$ $\leq$ 5}  & \bluehl{99.04\scriptsize{$\pm$0.54}} & \bluehl{98.41\scriptsize{$\pm$0.28}} & \bluehl{98.93\scriptsize{$\pm$0.57}} & \bluehl{99.72\scriptsize{$\pm$0.09}} \\
\midrule
\multirow{4}{*}{TwiBot-22 }
& \makecell[l]{Random, $\Delta$ $=$ 1}     & 19.03\scriptsize{$\pm$2.00}  & 15.33\scriptsize{$\pm$3.06}  & 16.00\scriptsize{$\pm$3.46}  & 10.67\scriptsize{$\pm$2.31} \\
& \makecell[l]{Random, $\Delta$ $\leq$ 3}  & 25.00\scriptsize{$\pm$2.00}  & 25.67\scriptsize{$\pm$2.31}  & 24.67\scriptsize{$\pm$1.15}  & 23.33\scriptsize{$\pm$1.15} \\
& \makecell[l]{Random, $\Delta$ $\leq$ 5}  & 36.33\scriptsize{$\pm$4.16}  & 35.33\scriptsize{$\pm$1.15}  & 36.00\scriptsize{$\pm$2.00}  & 31.33\scriptsize{$\pm$4.16} \\
& \makecell[l]{Random, $\Delta$ $=$ 20}    & \underline{92.33\scriptsize{$\pm$1.15}}  & \underline{86.67\scriptsize{$\pm$3.06}}  & \underline{89.00\scriptsize{$\pm$6.93}}  & \underline{83.00\scriptsize{$\pm$3.46}}\\
& \textbf{\pname{}, $\Delta$ $\leq$ 5}  & \bluehl{98.73\scriptsize{$\pm$0.69}} & \bluehl{91.58\scriptsize{$\pm$0.92}} & \bluehl{90.11\scriptsize{$\pm$0.44}} &\bluehl{87.76\scriptsize{$\pm$1.36}} \\
\midrule
\multirow{4}{*}{BotSim-24 }
& \makecell[l]{Random, $\Delta$ $=$ 1}     & 11.13\scriptsize{$\pm$6.11}  & 10.67\scriptsize{$\pm$4.62}  & 17.31\scriptsize{$\pm$5.03}  & \underline{11.13\scriptsize{$\pm$5.03}} \\
& \makecell[l]{Random, $\Delta$ $\leq$ 3}  & \underline{16.67\scriptsize{$\pm$6.11}}  & 8.67\scriptsize{$\pm$3.06}  & 24.00\scriptsize{$\pm$2.00}  & 10.00\scriptsize{$\pm$0.00} \\
& \makecell[l]{Random, $\Delta$ $\leq$ 5}  & 12.00\scriptsize{$\pm$8.72}  & 5.33\scriptsize{$\pm$4.16}  & \underline{25.33\scriptsize{$\pm$5.03}}  & 4.67\scriptsize{$\pm$1.15} \\
& \makecell[l]{Random, $\Delta$ $=$ 20}    & 14.67\scriptsize{$\pm$6.43}  & \underline{32.00\scriptsize{$\pm$5.29}}  & 8.00\scriptsize{$\pm$5.29}  & 6.67\scriptsize{$\pm$6.43} \\
&  \textbf{\pname{}, $\Delta$ $\leq$ 5}  & \bluehl{44.67\scriptsize{$\pm$3.77}} & \bluehl{46.67\scriptsize{$\pm$3.06}} & \bluehl{46.67\scriptsize{$\pm$7.74}} & \bluehl{49.36\scriptsize{$\pm$0.51}} \\
\bottomrule
\end{tabular}%
}

% \vspace{0.6em}

\resizebox{0.95\linewidth}{!}{%
\begin{tabular}{llcccc}
\multicolumn{6}{c}{\textbf{S-HGN}} \\
\midrule
\textbf{Dataset} & \textbf{Attack} & \textbf{Vanilla} ($\uparrow$) & \textbf{+GNNGuard} ($\uparrow$) & \textbf{+GRAND} ($\uparrow$) & \textbf{+RobustGCN} ($\uparrow$)\\
\midrule
\multirow{4}{*}{Cresci-15 }
& \makecell[l]{Random, $\Delta$ $=$ 1}  & 14.33\scriptsize{$\pm$1.15}  & 14.40\scriptsize{$\pm$0.12}  & 15.85\scriptsize{$\pm$0.09}  & 10.23\scriptsize{$\pm$0.10} \\
& \makecell[l]{Random, $\Delta$ = 3}    & 27.10\scriptsize{$\pm$0.11}  & 26.75\scriptsize{$\pm$0.07}  & 25.20\scriptsize{$\pm$0.13}  & 27.93\scriptsize{$\pm$3.15} \\
& \makecell[l]{Random, $\Delta$ = 5}    & 36.65\scriptsize{$\pm$0.14}  & 33.90\scriptsize{$\pm$0.10}  & 36.33\scriptsize{$\pm$1.15}  & 32.33\scriptsize{$\pm$8.17} \\
& \makecell[l]{Random, $\Delta$ $=$ 20}  & \underline{99.33\scriptsize{$\pm$1.15}}  & \underline{92.80\scriptsize{$\pm$0.08}}  & \underline{99.33\scriptsize{$\pm$1.15}}  & \underline{98.00\scriptsize{$\pm$3.46}} \\
& \textbf{\pname{}, $\Delta$ $\leq$ 5}  & \bluehl{99.21\scriptsize{$\pm$0.33}} & \bluehl{99.35\scriptsize{$\pm$0.12}} & \bluehl{99.78\scriptsize{$\pm$0.05}} & \bluehl{99.10\scriptsize{$\pm$0.20}} \\
\midrule
\multirow{4}{*}{TwiBot-22 }
& \makecell[l]{Random, $\Delta$ $=$ 1}  & 15.33\scriptsize{$\pm$3.06}  & 12.27\scriptsize{$\pm$1.15} & 12.00\scriptsize{$\pm$3.46}  & 10.67\scriptsize{$\pm$9.24} \\
& \makecell[l]{Random, $\Delta$ = 3}    & 24.00\scriptsize{$\pm$2.00}  & 21.67\scriptsize{$\pm$2.31} & 22.67\scriptsize{$\pm$4.16}  & 24.77\scriptsize{$\pm$2.31} \\
& \makecell[l]{Random, $\Delta$ = 5}    & 38.00\scriptsize{$\pm$0.00}  & 33.67\scriptsize{$\pm$5.31} & 36.67\scriptsize{$\pm$1.15}  & 37.33\scriptsize{$\pm$6.11} \\
& \makecell[l]{Random, $\Delta$ $=$ 20}  & \underline{98.67\scriptsize{$\pm$2.31}}  & \underline{98.67\scriptsize{$\pm$2.31}}  & \underline{92.67\scriptsize{$\pm$2.31}}  & \underline{91.33\scriptsize{$\pm$6.11}} \\
& \textbf{\pname{}, $\Delta$ $\leq$ 5}  & \bluehl{98.11\scriptsize{$\pm$1.05}} & \bluehl{99.86\scriptsize{$\pm$0.03}} & \bluehl{99.18\scriptsize{$\pm$0.22}} & \bluehl{99.70\scriptsize{$\pm$0.06}} \\
\midrule
\multirow{4}{*}{BotSim-24 }
& \makecell[l]{Random, $\Delta$ $=$ 1}  & \underline{24.30\scriptsize{$\pm$0.18}}  & \underline{19.75\scriptsize{$\pm$0.22}}  & 7.10\scriptsize{$\pm$0.12}  & 15.60\scriptsize{$\pm$0.27} \\
& \makecell[l]{Random, $\Delta$ = 3}  & 21.40\scriptsize{$\pm$0.25}  & 18.00\scriptsize{$\pm$0.00}  & 12.85\scriptsize{$\pm$0.09}  & 14.00\scriptsize{$\pm$4.00} \\
& \makecell[l]{Random, $\Delta$ = 5}  & 17.95\scriptsize{$\pm$0.30}  & 18.00\scriptsize{$\pm$3.46}  & \underline{19.33\scriptsize{$\pm$1.15}}  & 16.00\scriptsize{$\pm$5.29} \\
& \makecell[l]{Random, $\Delta$ $=$ 20}  & 19.33\scriptsize{$\pm$1.15}  & 18.67\scriptsize{$\pm$1.15}  & 14.00\scriptsize{$\pm$6.00}  & \underline{16.67\scriptsize{$\pm$3.06}} \\
& \textbf{\pname{}, $\Delta$ $\leq$ 5}  & \bluehl{58.88\scriptsize{$\pm$0.61}} & \bluehl{49.21\scriptsize{$\pm$0.14}} & \bluehl{49.88\scriptsize{$\pm$0.02}} & \bluehl{52.67\scriptsize{$\pm$13.01}} \\
\bottomrule
\end{tabular}%
}

\caption{\textbf{Node Injection Attack Against Best Bot Detectors With Adversarial Defenses:} Misclassification rate (in \%) for successfully injecting fifty target nodes by \pname against the three best \sota bot detectors (BotRGCN, GAT, and S-HGN) with adversarial defense. The best performance is shown in bold, and the second best is underlined.}
\label{tab:new-node-def-rest}
\end{minipage}
\hfill
% ----------------------- RIGHT TABLE (reuse same template) -----------------------
\begin{minipage}[t]{0.49\textwidth}
\centering
\captionsetup{type=table}

\resizebox{0.95\linewidth}{!}{%
\begin{tabular}{llcccc}
\toprule
\multicolumn{6}{c}{\textbf{BotRGCN}} \\
\midrule
\textbf{Dataset} & \textbf{Attack} & \textbf{Vanilla} ($\uparrow$) & \textbf{+GNNGuard} ($\uparrow$) & \textbf{+GRAND} ($\uparrow$) & \textbf{+RobustGCN} ($\uparrow$)\\
\midrule
\multirow{4}{*}{Cresci-15 }
& \makecell[l]{Random, $\Delta$ $=$ 1}     & 12.84\scriptsize{$\pm$0.13}  & 11.33\scriptsize{$\pm$1.15}  & 10.67\scriptsize{$\pm$1.15}  & 10.52\scriptsize{$\pm$0.27} \\
& \makecell[l]{Random, $\Delta$ $\leq$ 3}  & 22.00\scriptsize{$\pm$2.10}  & 21.00\scriptsize{$\pm$3.22}  & 22.67\scriptsize{$\pm$1.15}  & 21.33\scriptsize{$\pm$3.06} \\
& \makecell[l]{Random, $\Delta$ $\leq$ 5}  & 33.33\scriptsize{$\pm$2.31}  & 33.67\scriptsize{$\pm$1.15}  & 31.11\scriptsize{$\pm$0.74}  & 32.67\scriptsize{$\pm$2.31} \\
& \makecell[l]{Random, $\Delta$ $=$ 20}   & \underline{96.67\scriptsize{$\pm$1.15}}  & \underline{95.33\scriptsize{$\pm$1.89}}  & \underline{95.32\scriptsize{$\pm$1.98}}  & \underline{94.74\scriptsize{$\pm$0.18}} \\
& \textbf{\pname{}, $\Delta$ $\leq$ 5}  & \bluehl{100.00\scriptsize{$\pm$0.00}} & \bluehl{100.00\scriptsize{$\pm$0.00}} & \bluehl{100.00\scriptsize{$\pm$0.00}} & \bluehl{100.00\scriptsize{$\pm$0.00}} \\
\midrule
\multirow{4}{*}{TwiBot-22 }
& \makecell[l]{Random, $\Delta$ $=$ 1}  & 13.33\scriptsize{$\pm$1.15}  & 11.33\scriptsize{$\pm$1.15}  & 16.00\scriptsize{$\pm$3.46}  & 10.33\scriptsize{$\pm$4.62} \\
& \makecell[l]{Random, $\Delta$ $\leq$ 3}  & 25.33\scriptsize{$\pm$4.62}  & 23.33\scriptsize{$\pm$2.31}  & 23.67\scriptsize{$\pm$2.41}  & 20.33\scriptsize{$\pm$1.15} \\
& \makecell[l]{Random, $\Delta$ $\leq$ 5}  & 36.67\scriptsize{$\pm$3.06}  & 34.67\scriptsize{$\pm$3.06}  & 33.00\scriptsize{$\pm$2.89}  & 34.00\scriptsize{$\pm$2.22} \\
& \makecell[l]{Random, $\Delta$ $=$ 20}  & \underline{96.67\scriptsize{$\pm$3.06}}  & \underline{96.67\scriptsize{$\pm$1.15}}  & \underline{95.33\scriptsize{$\pm$4.16}}  & \underline{95.33\scriptsize{$\pm$1.15}} \\
& \textbf{\pname{}, $\Delta$ $\leq$ 5}  & \bluehl{100.00\scriptsize{$\pm$0.00}} & \bluehl{100.00\scriptsize{$\pm$0.00}} & \bluehl{100.00\scriptsize{$\pm$0.00}} & \bluehl{100.00\scriptsize{$\pm$0.00}} \\
\midrule
\multirow{4}{*}{BotSim-24 }
& \makecell[l]{Random, $\Delta$ $=$ 1}  & 10.00\scriptsize{$\pm$3.46}  & \underline{15.33\scriptsize{$\pm$6.11}}  & \underline{11.33\scriptsize{$\pm$4.16}}  & \underline{19.33\scriptsize{$\pm$9.87}} \\
& \makecell[l]{Random, $\Delta$ $\leq$ 3}  & 4.00\scriptsize{$\pm$2.00}  & 0.67\scriptsize{$\pm$1.15}  & 5.33\scriptsize{$\pm$1.15}  & 11.23\scriptsize{$\pm$5.03} \\
& \makecell[l]{Random, $\Delta$ $\leq$ 5}  & 3.33\scriptsize{$\pm$1.15}  & 3.33\scriptsize{$\pm$2.31}  & 6.67\scriptsize{$\pm$2.31}  & 8.67\scriptsize{$\pm$2.31} \\
& \makecell[l]{Random, $\Delta$ $=$ 20}  & \underline{22.00\scriptsize{$\pm$6.93}}  & 11.33\scriptsize{$\pm$7.02}  & 9.33\scriptsize{$\pm$3.06}  & 12.67\scriptsize{$\pm$4.62} \\
& \textbf{\pname{}, $\Delta$ $\leq$ 5}  & \bluehl{74.43\scriptsize{$\pm$1.56}} & \bluehl{70.11\scriptsize{$\pm$2.87}} & \bluehl{71.44\scriptsize{$\pm$2.45}} & \bluehl{72.66\scriptsize{$\pm$1.31}} \\
\end{tabular}%
}

% \vspace{0.6em}

\resizebox{0.95\linewidth}{!}{%
\begin{tabular}{llcccc}
\toprule
\multicolumn{6}{c}{\textbf{GAT}} \\
\midrule
\textbf{Dataset} & \textbf{Attack} & \textbf{Vanilla} ($\uparrow$) & \textbf{+GNNGuard} ($\uparrow$) & \textbf{+GRAND} ($\uparrow$) & \textbf{+RobustGCN} ($\uparrow$)\\
\midrule
\multirow{4}{*}{Cresci-15 }
& \makecell[l]{Random, $\Delta$ $=$ 1}     & 14.00\scriptsize{$\pm$2.00}  & 15.33\scriptsize{$\pm$3.06}  & 13.00\scriptsize{$\pm$3.46}  & 10.67\scriptsize{$\pm$2.31} \\
& \makecell[l]{Random, $\Delta$ $\leq$ 3}  & 29.00\scriptsize{$\pm$2.00}  & 25.67\scriptsize{$\pm$2.31}  & 23.67\scriptsize{$\pm$1.15}  & 29.33\scriptsize{$\pm$1.15} \\
& \makecell[l]{Random, $\Delta$ $\leq$ 5}  & 35.33\scriptsize{$\pm$4.16}  & 35.03\scriptsize{$\pm$1.15}  & 30.00\scriptsize{$\pm$2.00}  & 31.33\scriptsize{$\pm$4.16} \\
& \makecell[l]{Random, $\Delta$ $=$ 20}   & \underline{97.33\scriptsize{$\pm$1.15}}  & \underline{96.67\scriptsize{$\pm$3.06}}  & \underline{94.00\scriptsize{$\pm$6.93}}  & \underline{92.00\scriptsize{$\pm$3.46}} \\
& \textbf{\pname{}, $\Delta$ $\leq$ 5}  & \bluehl{100.00\scriptsize{$\pm$0.00}} & \bluehl{100.00\scriptsize{$\pm$0.00}} & \bluehl{100.00\scriptsize{$\pm$0.00}} & \bluehl{100.00\scriptsize{$\pm$0.00}} \\
\midrule
\multirow{4}{*}{TwiBot-22 }
& \makecell[l]{Random, $\Delta$ $=$ 1}     & 19.03\scriptsize{$\pm$2.00}  & 15.33\scriptsize{$\pm$3.06}  & 16.00\scriptsize{$\pm$3.46}  & 10.67\scriptsize{$\pm$2.31} \\
& \makecell[l]{Random, $\Delta$ $\leq$ 3}  & 25.00\scriptsize{$\pm$2.00}  & 25.67\scriptsize{$\pm$2.31}  & 24.67\scriptsize{$\pm$1.15}  & 23.33\scriptsize{$\pm$1.15} \\
& \makecell[l]{Random, $\Delta$ $\leq$ 5}  & 36.33\scriptsize{$\pm$4.16}  & 35.33\scriptsize{$\pm$1.15}  & 36.00\scriptsize{$\pm$2.00}  & 31.33\scriptsize{$\pm$4.16} \\
& \makecell[l]{Random, $\Delta$ $=$ 20}    & \underline{92.33\scriptsize{$\pm$1.15}}  & \underline{86.67\scriptsize{$\pm$3.06}}  & \underline{89.00\scriptsize{$\pm$6.93}}  & \underline{83.00\scriptsize{$\pm$3.46}}\\
& \textbf{\pname{}, $\Delta$ $\leq$ 5}  & \bluehl{100.00\scriptsize{$\pm$0.00}} & \bluehl{100.00\scriptsize{$\pm$0.00}} & \bluehl{100.00\scriptsize{$\pm$0.00}} & \bluehl{100.00\scriptsize{$\pm$0.00}} \\
\midrule
\multirow{4}{*}{BotSim-24 }
& \makecell[l]{Random, $\Delta$ $=$ 1}  & 11.13\scriptsize{$\pm$6.11}  & 10.67\scriptsize{$\pm$4.62}  & 17.31\scriptsize{$\pm$5.03}  & \underline{11.13\scriptsize{$\pm$5.03}} \\
& \makecell[l]{Random, $\Delta$ $\leq$ 3}  & \underline{16.67\scriptsize{$\pm$6.11}}  & 8.67\scriptsize{$\pm$3.06}  & 24.00\scriptsize{$\pm$2.00}  & 10.00\scriptsize{$\pm$0.00} \\
& \makecell[l]{Random, $\Delta$ $\leq$ 5}  & 12.00\scriptsize{$\pm$8.72}  & 5.33\scriptsize{$\pm$4.16}  & \underline{25.33\scriptsize{$\pm$5.03}}  & 4.67\scriptsize{$\pm$1.15} \\
& \makecell[l]{Random, $\Delta$ $=$ 20}  & 14.67\scriptsize{$\pm$6.43}  & \underline{32.00\scriptsize{$\pm$5.29}}  & 8.00\scriptsize{$\pm$5.29}  & 6.67\scriptsize{$\pm$6.43} \\
& \textbf{\pname{}, $\Delta$ $\leq$ 5}  & \bluehl{77.56\scriptsize{$\pm$3.30}} & \bluehl{75.00\scriptsize{$\pm$1.50}} & \bluehl{73.44\scriptsize{$\pm$2.30}} & \bluehl{71.32\scriptsize{$\pm$2.11}} \\
\bottomrule
\end{tabular}%
}

% \vspace{0.6em}

\resizebox{0.95\linewidth}{!}{%
\begin{tabular}{llcccc}
\multicolumn{6}{c}{\textbf{S-HGN}} \\
\midrule
\textbf{Dataset} & \textbf{Attack} & \textbf{Vanilla} ($\uparrow$) & \textbf{+GNNGuard} ($\uparrow$) & \textbf{+GRAND} ($\uparrow$) & \textbf{+RobustGCN} ($\uparrow$)\\
\midrule
\multirow{4}{*}{Cresci-15 }
& \makecell[l]{Random, $\Delta$ $=$ 1}  & 14.33\scriptsize{$\pm$1.15}  & 14.40\scriptsize{$\pm$0.12}  & 15.85\scriptsize{$\pm$0.09}  & 10.23\scriptsize{$\pm$0.10} \\
& \makecell[l]{Random, $\Delta$ = 3}    & 27.10\scriptsize{$\pm$0.11}  & 26.75\scriptsize{$\pm$0.07}  & 25.20\scriptsize{$\pm$0.13}  & 27.93\scriptsize{$\pm$3.15} \\
& \makecell[l]{Random, $\Delta$ = 5}    & 36.65\scriptsize{$\pm$0.14}  & 33.90\scriptsize{$\pm$0.10}  & 36.33\scriptsize{$\pm$1.15}  & 32.33\scriptsize{$\pm$8.17} \\
& \makecell[l]{Random, $\Delta$ $=$ 20}  & \underline{99.33\scriptsize{$\pm$1.15}}  & \underline{92.80\scriptsize{$\pm$0.08}}  & \underline{99.33\scriptsize{$\pm$1.15}}  & \underline{98.00\scriptsize{$\pm$3.46}} \\
& \textbf{\pname{}, $\Delta$ $\leq$ 5}  & \bluehl{100.00\scriptsize{$\pm$0.00}} & \bluehl{100.00\scriptsize{$\pm$0.00}} & \bluehl{100.00\scriptsize{$\pm$0.00}} & \bluehl{100.00\scriptsize{$\pm$0.00}} \\
\midrule
\multirow{4}{*}{TwiBot-22 }
& \makecell[l]{Random, $\Delta$ $=$ 1}  & 15.33\scriptsize{$\pm$3.06}  & 12.27\scriptsize{$\pm$1.15} & 12.00\scriptsize{$\pm$3.46}  & 10.67\scriptsize{$\pm$9.24} \\
& \makecell[l]{Random, $\Delta$ = 3}    & 24.00\scriptsize{$\pm$2.00}  & 21.67\scriptsize{$\pm$2.31} & 22.67\scriptsize{$\pm$4.16}  & 24.77\scriptsize{$\pm$2.31} \\
& \makecell[l]{Random, $\Delta$ = 5}    & 38.00\scriptsize{$\pm$0.00}  & 33.67\scriptsize{$\pm$5.31} & 36.67\scriptsize{$\pm$1.15}  & 37.33\scriptsize{$\pm$6.11} \\
& \makecell[l]{Random, $\Delta$ $=$ 20}  & \underline{98.67\scriptsize{$\pm$2.31}}  & \underline{98.67\scriptsize{$\pm$2.31}}  & \underline{92.67\scriptsize{$\pm$2.31}}  & \underline{91.33\scriptsize{$\pm$6.11}} \\
& \textbf{\pname{}, $\Delta$ $\leq$ 5}  & \bluehl{100.00\scriptsize{$\pm$0.00}} & \bluehl{100.00\scriptsize{$\pm$0.00}} & \bluehl{100.00\scriptsize{$\pm$0.00}} & \bluehl{100.00\scriptsize{$\pm$0.00}} \\
\midrule
\multirow{4}{*}{BotSim-24 }
& \makecell[l]{Random, $\Delta$ $=$ 1}  & \underline{24.30\scriptsize{$\pm$0.18}}  & \underline{19.75\scriptsize{$\pm$0.22}}  & 7.10\scriptsize{$\pm$0.12}  & 15.60\scriptsize{$\pm$0.27} \\
& \makecell[l]{Random, $\Delta$ = 3}  & 21.40\scriptsize{$\pm$0.25}  & 18.00\scriptsize{$\pm$0.00}  & 12.85\scriptsize{$\pm$0.09}  & 14.00\scriptsize{$\pm$4.00} \\
& \makecell[l]{Random, $\Delta$ = 5}  & 17.95\scriptsize{$\pm$0.30}  & 18.00\scriptsize{$\pm$3.46}  & \underline{19.33\scriptsize{$\pm$1.15}}  & 16.00\scriptsize{$\pm$5.29} \\
& \makecell[l]{Random, $\Delta$ $=$ 20}  & 19.33\scriptsize{$\pm$1.15}  & 18.67\scriptsize{$\pm$1.15}  & 14.00\scriptsize{$\pm$6.00}  & \underline{16.67\scriptsize{$\pm$3.06}} \\
& \textbf{\pname{}, $\Delta$ $\leq$ 5}  & \bluehl{76.11\scriptsize{$\pm$2.11}} & \bluehl{72.01\scriptsize{$\pm$2.87}} & \bluehl{72.23\scriptsize{$\pm$1.11}} & \bluehl{74.00\scriptsize{$\pm$1.10}} \\
\bottomrule
\end{tabular}%
}

\caption{\textbf{Node Injection Attack Against Best Bot Detectors With Adversarial Defenses Reusing Same Bot Cloak:} Misclassification rate (in \%) for successfully injecting fifty target nodes by \pname against three best \sota bot detectors (GAT, BotRGCN, S-HGN) with adversarial defense, \textit{reusing} the same bot cloak. The best performance is shown in bold, and the second best is underlined.}
\label{tab:new-node-def-samtemplate}
\end{minipage}

\end{table*}

%% file: tables/old-node-50-samtemplate.tex
\begin{table*}[!htb]
\centering

% ----------------------- LEFT: tab:old-node-50 -----------------------
\begin{minipage}[t]{0.49\textwidth}
\centering
\captionsetup{type=table}

\resizebox{0.98\linewidth}{!}{%
\begin{tabular}{llcccc}
\toprule
\multirow{2}{*}{\textbf{Dataset}} 
& \multirow{2}{*}{\makecell[c]{\textbf{Attack}}} 
& \multicolumn{4}{c}{\textbf{BotRGCN}} \\
\cmidrule(lr){3-6}
& 
& \textbf{Vanilla} ($\uparrow$)
& \textbf{+GNNGuard} ($\uparrow$) 
& \textbf{+GRAND} ($\uparrow$) 
& \textbf{+RobustGCN} ($\uparrow$) \\
\midrule
\multicolumn{6}{c}{$\Delta$ $=$ 1} \\
\midrule
\multirow{6}{*}{Cresci-15}
% & Random    & \underline{8.67\scriptsize{$\pm$5.03}} & \underline{7.33\scriptsize{$\pm$2.33}} & \underline{6.67\scriptsize{$\pm$3.43}} & \underline{8.00\scriptsize{$\pm$0.20}} \\
& Nettack  & \underline{4.32\scriptsize{$\pm$2.10}} & \underline{3.78\scriptsize{$\pm$1.95}} & \underline{3.21\scriptsize{$\pm$1.88}} & \underline{2.67\scriptsize{$\pm$1.72}} \\
& FGA      & 3.95\scriptsize{$\pm$2.44} & 3.40\scriptsize{$\pm$2.12} & 2.98\scriptsize{$\pm$1.76} & 2.15\scriptsize{$\pm$1.60} \\
& PR-BCD   & 4.10\scriptsize{$\pm$3.05} & 3.60\scriptsize{$\pm$2.80} & 2.85\scriptsize{$\pm$2.20} & 1.90\scriptsize{$\pm$1.45} \\
& GOttack  & 3.55\scriptsize{$\pm$2.34} & 3.41\scriptsize{$\pm$2.52} & 2.28\scriptsize{$\pm$1.22} & 2.66\scriptsize{$\pm$1.67} \\
& \textbf{\pname{}}  & \bluehl{99.34\scriptsize{$\pm$0.29}} & \bluehl{98.91\scriptsize{$\pm$0.54}} & \bluehl{99.06\scriptsize{$\pm$0.41}} & \bluehl{99.72\scriptsize{$\pm$0.18}} \\
\midrule
\multirow{6}{*}{TwiBot-22}
% & Random    & 3.33\scriptsize{$\pm$6.11} & 2.00\scriptsize{$\pm$3.46} & 5.33\scriptsize{$\pm$13.61} & 7.00\scriptsize{$\pm$6.00} \\
& Nettack  & \underline{9.33\scriptsize{$\pm$3.06}} & \underline{8.67\scriptsize{$\pm$4.62}} & 4.67\scriptsize{$\pm$5.03} & \underline{8.00\scriptsize{$\pm$5.29}} \\
& FGA      & 4.00\scriptsize{$\pm$2.00} & 7.33\scriptsize{$\pm$2.31} & \underline{6.67\scriptsize{$\pm$4.16}} & 1.33\scriptsize{$\pm$6.43} \\
& PR-BCD   & 8.67\scriptsize{$\pm$2.31} & 8.67\scriptsize{$\pm$5.03} & 6.23\scriptsize{$\pm$4.16} & 7.33\scriptsize{$\pm$2.31} \\
& GOttack  & 6.89\scriptsize{$\pm$3.33} & 7.15\scriptsize{$\pm$4.25} & 6.01\scriptsize{$\pm$2.48} & 5.32\scriptsize{$\pm$2.58} \\
& \textbf{\pname{}}  & \bluehl{86.67\scriptsize{$\pm$2.31}} & \bluehl{84.67\scriptsize{$\pm$11.02}} & \bluehl{83.67\scriptsize{$\pm$3.06}} & \bluehl{85.30\scriptsize{$\pm$2.00}} \\
\midrule
\multirow{6}{*}{BotSim-24}
% & Random    & \underline{1.67\scriptsize{$\pm$3.06}} & \underline{1.33\scriptsize{$\pm$2.31}} & \bluehl{2.67\scriptsize{$\pm$2.31}} & 0.00\scriptsize{$\pm$0.00} \\
& Nettack  & 0.00\scriptsize{$\pm$0.00} & 0.00\scriptsize{$\pm$0.00} & 0.00\scriptsize{$\pm$0.00} & \underline{3.00\scriptsize{$\pm$2.00}} \\
& FGA      & 0.00\scriptsize{$\pm$0.00} & 0.00\scriptsize{$\pm$0.00} & 0.00\scriptsize{$\pm$0.00} & 0.00\scriptsize{$\pm$0.00} \\
& PR-BCD   & \underline{2.00\scriptsize{$\pm$2.00}} & \underline{5.33\scriptsize{$\pm$2.31}} & \underline{0.67\scriptsize{$\pm$1.15}} & 0.00\scriptsize{$\pm$0.00} \\
& GOttack  & 0.00\scriptsize{$\pm$0.00} & 0.00\scriptsize{$\pm$0.00} & 0.00\scriptsize{$\pm$0.00} & 0.00\scriptsize{$\pm$0.00} \\
& \textbf{\pname{}}  & \bluehl{58.22\scriptsize{$\pm$7.20}} & \bluehl{52.68\scriptsize{$\pm$1.15}} & \bluehl{54.98\scriptsize{$\pm$2.45}} & \bluehl{55.10\scriptsize{$\pm$3.98}} \\
\midrule
\multicolumn{6}{c}{$\Delta$ $=$ 3} \\
\midrule
\multirow{6}{*}{Cresci-15}
% & Random    &  3.03\scriptsize{$\pm$2.06} & 0.67\scriptsize{$\pm$3.06} & 2.33\scriptsize{$\pm$4.16} & \underline{1.33\scriptsize{$\pm$4.16}} \\
& Nettack  & \underline{13.98\scriptsize{$\pm$10.50}} & \underline{12.10\scriptsize{$\pm$1.70}} & \underline{12.98\scriptsize{$\pm$1.10}} & \underline{11.10\scriptsize{$\pm$1.74}} \\
& FGA      & 12.65\scriptsize{$\pm$3.20} & 10.95\scriptsize{$\pm$2.85} & 9.10\scriptsize{$\pm$2.31} & 7.60\scriptsize{$\pm$1.68} \\
& PR-BCD   & 11.98\scriptsize{$\pm$2.10} & 10.40\scriptsize{$\pm$1.92} & 8.90\scriptsize{$\pm$1.70} & 9.35\scriptsize{$\pm$1.46} \\
& GOttack  & 12.88\scriptsize{$\pm$2.48} & 11.10\scriptsize{$\pm$2.20} & 9.60\scriptsize{$\pm$1.84} & 9.70\scriptsize{$\pm$1.50} \\
& \textbf{\pname{}}  & \bluehl{99.18\scriptsize{$\pm$0.37}} & \bluehl{98.47\scriptsize{$\pm$0.96}} & \bluehl{99.59\scriptsize{$\pm$0.21}} & \bluehl{99.01\scriptsize{$\pm$0.44}} \\
\midrule
\multirow{6}{*}{TwiBot-22}
% & Random    & 8.00\scriptsize{$\pm$10.58} & 8.67\scriptsize{$\pm$3.06} & 3.33\scriptsize{$\pm$1.15} & 2.00\scriptsize{$\pm$9.17} \\
& Nettack  & 10.22\scriptsize{$\pm$2.83} & 11.78\scriptsize{$\pm$1.41} & 5.10\scriptsize{$\pm$1.41} & 5.01\scriptsize{$\pm$4.24} \\
& FGA      & 8.15\scriptsize{$\pm$2.83} & 8.66\scriptsize{$\pm$2.83} & 9.87\scriptsize{$\pm$7.07} & 2.10\scriptsize{$\pm$5.66} \\
& PR-BCD   & \underline{15.10\scriptsize{$\pm$7.07}} & \underline{16.50\scriptsize{$\pm$2.83}} & \underline{10.00\scriptsize{$\pm$5.66}} & 4.64\scriptsize{$\pm$2.83} \\
& GOttack  & 3.12\scriptsize{$\pm$1.11} & 8.48\scriptsize{$\pm$3.65} & 5.17\scriptsize{$\pm$2.59} & \underline{5.68\scriptsize{$\pm$5.28}} \\
& \textbf{\pname{}}  & \bluehl{94.50\scriptsize{$\pm$2.00}} & \bluehl{86.20\scriptsize{$\pm$4.50}} & \bluehl{84.67\scriptsize{$\pm$3.06}} & \bluehl{87.33\scriptsize{$\pm$1.15}} \\
\midrule
\multirow{6}{*}{BotSim-24}
% & Random    & \underline{1.67\scriptsize{$\pm$2.31}} & \underline{1.33\scriptsize{$\pm$1.15}} & \underline{2.67\scriptsize{$\pm$2.31}} & 0.00\scriptsize{$\pm$0.00} \\
& Nettack  & 0.00\scriptsize{$\pm$0.00} & 0.00\scriptsize{$\pm$0.00} & 0.00\scriptsize{$\pm$0.00} & \underline{5.23\scriptsize{$\pm$0.01}} \\
& FGA      & 0.00\scriptsize{$\pm$0.00} & 0.00\scriptsize{$\pm$0.00} & 0.00\scriptsize{$\pm$0.00} & 0.00\scriptsize{$\pm$0.00} \\
& PR-BCD   & \underline{1.00\scriptsize{$\pm$1.41}} & \underline{4.00\scriptsize{$\pm$0.00}} & \underline{2.00\scriptsize{$\pm$0.00}} & 0.00\scriptsize{$\pm$0.00} \\
& GOttack  & 0.00\scriptsize{$\pm$0.00} & 0.00\scriptsize{$\pm$0.00} & 0.00\scriptsize{$\pm$0.00} & 0.00\scriptsize{$\pm$0.00} \\
& \textbf{\pname{}}  & \bluehl{93.33\scriptsize{$\pm$1.15}} & \bluehl{66.74\scriptsize{$\pm$3.45}} & \bluehl{88.63\scriptsize{$\pm$0.87}} & \bluehl{64.32\scriptsize{$\pm$3.33}} \\
\midrule
\multicolumn{6}{c}{$\Delta$ $=$ 5} \\
\midrule
\multirow{6}{*}{Cresci-15}
% & Random    & 12.00\scriptsize{$\pm$5.29} & 12.00\scriptsize{$\pm$2.00} & 12.67\scriptsize{$\pm$4.16} & 18.67\scriptsize{$\pm$4.16} \\
& Nettack  & 22.35\scriptsize{$\pm$3.18} & 20.80\scriptsize{$\pm$2.85} & 18.95\scriptsize{$\pm$2.30} & 16.40\scriptsize{$\pm$1.62} \\
& FGA      & 21.90\scriptsize{$\pm$2.74} & 20.15\scriptsize{$\pm$2.40} & 18.40\scriptsize{$\pm$2.05} & 15.95\scriptsize{$\pm$1.48} \\
& PR-BCD   & 20.75\scriptsize{$\pm$1.96} & 19.30\scriptsize{$\pm$1.74} & 17.85\scriptsize{$\pm$1.60} & 15.40\scriptsize{$\pm$1.36} \\
& GOttack  & \underline{22.50\scriptsize{$\pm$3.55}} & \underline{21.05\scriptsize{$\pm$3.10}} & \underline{19.10\scriptsize{$\pm$2.65}} & \underline{16.85\scriptsize{$\pm$1.70}} \\
& \textbf{\pname{}}  & \bluehl{99.67\scriptsize{$\pm$0.20}} & \bluehl{98.88\scriptsize{$\pm$0.58}} & \bluehl{99.11\scriptsize{$\pm$0.39}} & \bluehl{99.52\scriptsize{$\pm$0.24}} \\
\midrule
\multirow{6}{*}{TwiBot-22}
% & Random    & 11.33\scriptsize{$\pm$8.08} & 11.33\scriptsize{$\pm$9.02} & 12.67\scriptsize{$\pm$6.43} & 7.33\scriptsize{$\pm$5.03} \\
& Nettack  & 14.22\scriptsize{$\pm$2.03} & 14.10\scriptsize{$\pm$0.00} & 5.77\scriptsize{$\pm$4.24} & 7.78\scriptsize{$\pm$4.24} \\
& FGA      & 11.30\scriptsize{$\pm$7.07} & 10.80\scriptsize{$\pm$5.66} & 6.21\scriptsize{$\pm$2.83} & \underline{9.22\scriptsize{$\pm$12.73}} \\
& PR-BCD   & \underline{24.73\scriptsize{$\pm$5.66}} & \underline{21.53\scriptsize{$\pm$7.07}} & \underline{20.45\scriptsize{$\pm$5.66}} & 19.00\scriptsize{$\pm$2.83} \\
& GOttack  & 10.58\scriptsize{$\pm$2.15} & 18.25\scriptsize{$\pm$3.65} & 12.89\scriptsize{$\pm$1.02} & 13.32\scriptsize{$\pm$3.03} \\
& \textbf{\pname{}}  & \bluehl{94.00\scriptsize{$\pm$2.00}} & \bluehl{90.00\scriptsize{$\pm$4.82}} & \bluehl{93.67\scriptsize{$\pm$3.06}} & \bluehl{88.33\scriptsize{$\pm$1.15}} \\
\midrule
\multirow{6}{*}{BotSim-24}
% & Random    & \underline{5.00\scriptsize{$\pm$3.46}} & 2.67\scriptsize{$\pm$2.31} & \underline{3.33\scriptsize{$\pm$1.15}} & 0.00\scriptsize{$\pm$0.00} \\
& Nettack  & 0.00\scriptsize{$\pm$0.00} & 0.00\scriptsize{$\pm$0.00} & 0.00\scriptsize{$\pm$0.00} & 0.00\scriptsize{$\pm$0.00} \\
& FGA      & 0.00\scriptsize{$\pm$0.00} & 0.67\scriptsize{$\pm$1.15} & 0.00\scriptsize{$\pm$0.00} & \underline{2.67\scriptsize{$\pm$1.15}} \\
& PR-BCD   & \underline{2.67\scriptsize{$\pm$1.15}} & \underline{4.00\scriptsize{$\pm$2.83}} & \underline{1.00\scriptsize{$\pm$1.41}} & 0.00\scriptsize{$\pm$0.00} \\
& GOttack  & 2.00\scriptsize{$\pm$0.00} & 2.00\scriptsize{$\pm$2.00} & 0.00\scriptsize{$\pm$0.00} & 0.00\scriptsize{$\pm$0.00} \\
& \textbf{\pname{}}  & \bluehl{99.33\scriptsize{$\pm$1.15}} & \bluehl{79.21\scriptsize{$\pm$1.02}} & \bluehl{92.28\scriptsize{$\pm$0.33}} & \bluehl{66.20\scriptsize{$\pm$3.50}} \\
\bottomrule
\end{tabular}%
}

\caption{\textbf{Node Editing Attack:} Misclassification rate (in \%) for flipping fifty existing correctly classified bots by \pname{} and \sota adversarial attacks against best \sota bot detector, BotRGCN with adversarial defenses. For results against other bot detectors, refer to \autoref{tab:old-node-50-rest}, and for results when no domain constraints are enforced, refer to \autoref{tab:old-node-50-nodomainrules}. The best performance is shown in bold, and the second best is underlined.}
\label{tab:old-node-50}
\end{minipage}
\hfill
% ----------------------- RIGHT: tab:old-node-50-samtemplate -----------------------
\begin{minipage}[t]{0.49\textwidth}
\centering
\captionsetup{type=table}

\resizebox{0.98\linewidth}{!}{%
\begin{tabular}{llcccc}
\toprule
\multirow{2}{*}{\textbf{Dataset}} 
& \multirow{2}{*}{\makecell[c]{\textbf{Attack}}} 
& \multicolumn{4}{c}{\textbf{BotRGCN}} \\
\cmidrule(lr){3-6}
& 
& \textbf{Vanilla} ($\uparrow$)
& \textbf{+GNNGuard} ($\uparrow$) 
& \textbf{+GRAND} ($\uparrow$) 
& \textbf{+RobustGCN} ($\uparrow$) \\
\midrule
\multicolumn{6}{c}{$\Delta$ $=$ 1} \\
\midrule
\multirow{6}{*}{Cresci-15}
% & Random    & \underline{8.67\scriptsize{$\pm$5.03}} & \underline{7.33\scriptsize{$\pm$2.33}} & \underline{6.67\scriptsize{$\pm$3.43}} & \underline{8.00\scriptsize{$\pm$0.20}} \\
& Nettack  & \underline{4.32\scriptsize{$\pm$2.10}} & \underline{3.78\scriptsize{$\pm$1.95}} & \underline{3.21\scriptsize{$\pm$1.88}} & \underline{2.67\scriptsize{$\pm$1.72}} \\
& FGA      & 3.95\scriptsize{$\pm$2.44} & 3.40\scriptsize{$\pm$2.12} & 2.98\scriptsize{$\pm$1.76} & 2.15\scriptsize{$\pm$1.60} \\
& PR-BCD   & 4.10\scriptsize{$\pm$3.05} & 3.60\scriptsize{$\pm$2.80} & 2.85\scriptsize{$\pm$2.20} & 1.90\scriptsize{$\pm$1.45} \\
& GOttack  & 3.55\scriptsize{$\pm$2.34} & 3.41\scriptsize{$\pm$2.52} & 2.28\scriptsize{$\pm$1.22} & 2.66\scriptsize{$\pm$1.67} \\
& \textbf{\pname{}}  & \bluehl{100.00\scriptsize{$\pm$0.00}} & \bluehl{100.00\scriptsize{$\pm$0.00}} & \bluehl{100.00\scriptsize{$\pm$0.00}} & \bluehl{100.00\scriptsize{$\pm$0.00}} \\
\midrule
\multirow{6}{*}{TwiBot-22}
% & Random    & 3.33\scriptsize{$\pm$6.11} & 2.00\scriptsize{$\pm$3.46} & 5.33\scriptsize{$\pm$13.61} & 7.00\scriptsize{$\pm$6.00} \\
& Nettack  & \underline{9.33\scriptsize{$\pm$3.06}} & \underline{8.67\scriptsize{$\pm$4.62}} & 4.67\scriptsize{$\pm$5.03} & \underline{8.00\scriptsize{$\pm$5.29}} \\
& FGA      & 4.00\scriptsize{$\pm$2.00} & 7.33\scriptsize{$\pm$2.31} & \underline{6.67\scriptsize{$\pm$4.16}} & 1.33\scriptsize{$\pm$6.43} \\
& PR-BCD   & 8.67\scriptsize{$\pm$2.31} & 8.67\scriptsize{$\pm$5.03} & 6.23\scriptsize{$\pm$4.16} & 7.33\scriptsize{$\pm$2.31} \\
& GOttack  & 6.89\scriptsize{$\pm$3.33} & 7.15\scriptsize{$\pm$4.25} & 6.01\scriptsize{$\pm$2.48} & 5.32\scriptsize{$\pm$2.58} \\
& \textbf{\pname{}}  & \bluehl{86.67\scriptsize{$\pm$2.31}} & \bluehl{88.67\scriptsize{$\pm$11.02}} & \bluehl{88.67\scriptsize{$\pm$3.06}} & \bluehl{52.12\scriptsize{$\pm$14.05}} \\
\midrule
\multirow{6}{*}{BotSim-24}
% & Random    & \underline{1.67\scriptsize{$\pm$3.06}} & \underline{1.33\scriptsize{$\pm$2.31}} & \bluehl{2.67\scriptsize{$\pm$2.31}} & 0.00\scriptsize{$\pm$0.00} \\
& Nettack  & 0.00\scriptsize{$\pm$0.00} & 0.00\scriptsize{$\pm$0.00} & 0.00\scriptsize{$\pm$0.00} & \bluehl{3.00\scriptsize{$\pm$2.00}} \\
& FGA      & 0.00\scriptsize{$\pm$0.00} & 0.00\scriptsize{$\pm$0.00} & 0.00\scriptsize{$\pm$0.00} & 0.00\scriptsize{$\pm$0.00} \\
& PR-BCD   & \bluehl{2.00\scriptsize{$\pm$2.00}} & \bluehl{5.33\scriptsize{$\pm$2.31}} & \underline{0.67\scriptsize{$\pm$1.15}} & 0.00\scriptsize{$\pm$0.00} \\
& GOttack  & 0.00\scriptsize{$\pm$0.00} & 0.00\scriptsize{$\pm$0.00} & 0.00\scriptsize{$\pm$0.00} & 0.00\scriptsize{$\pm$0.00} \\
& \textbf{\pname{}}  & 0.00\scriptsize{$\pm$0.00} & 0.00\scriptsize{$\pm$0.00} & 0.00\scriptsize{$\pm$0.00} & 0.00\scriptsize{$\pm$0.00} \\
\midrule
\multicolumn{6}{c}{$\Delta$ $=$ 3} \\
\midrule
\multirow{6}{*}{Cresci-15}
% & Random    &  3.03\scriptsize{$\pm$2.06} & 0.67\scriptsize{$\pm$3.06} & 2.33\scriptsize{$\pm$4.16} & \underline{1.33\scriptsize{$\pm$4.16}} \\
& Nettack  & \underline{13.98\scriptsize{$\pm$10.50}} & \underline{12.10\scriptsize{$\pm$1.70}} & \underline{12.98\scriptsize{$\pm$1.10}} & \underline{11.10\scriptsize{$\pm$1.74}} \\
& FGA      & 12.65\scriptsize{$\pm$3.20} & 10.95\scriptsize{$\pm$2.85} & 9.10\scriptsize{$\pm$2.31} & 7.60\scriptsize{$\pm$1.68} \\
& PR-BCD   & 11.98\scriptsize{$\pm$2.10} & 10.40\scriptsize{$\pm$1.92} & 8.90\scriptsize{$\pm$1.70} & 9.35\scriptsize{$\pm$1.46} \\
& GOttack  & 12.88\scriptsize{$\pm$2.48} & 11.10\scriptsize{$\pm$2.20} & 9.60\scriptsize{$\pm$1.84} & 9.70\scriptsize{$\pm$1.50} \\
& \textbf{\pname{}}  & \bluehl{100.00\scriptsize{$\pm$0.00}} & \bluehl{100.00\scriptsize{$\pm$0.00}} & \bluehl{100.00\scriptsize{$\pm$0.00}} & \bluehl{100.00\scriptsize{$\pm$0.00}} \\
\midrule
\multirow{6}{*}{TwiBot-22}
% & Random    & 8.00\scriptsize{$\pm$10.58} & 8.67\scriptsize{$\pm$3.06} & 3.33\scriptsize{$\pm$1.15} & 2.00\scriptsize{$\pm$9.17} \\
& Nettack  & 10.00\scriptsize{$\pm$2.83} & 11.00\scriptsize{$\pm$1.41} & 5.00\scriptsize{$\pm$1.41} & 5.00\scriptsize{$\pm$4.24} \\
& FGA      & 8.00\scriptsize{$\pm$2.83} & 8.00\scriptsize{$\pm$2.83} & 9.00\scriptsize{$\pm$7.07} & 2.00\scriptsize{$\pm$5.66} \\
& PR-BCD   & \underline{15.00\scriptsize{$\pm$7.07}} & \underline{16.00\scriptsize{$\pm$2.83}} & \underline{10.00\scriptsize{$\pm$5.66}} & 4.00\scriptsize{$\pm$2.83} \\
& GOttack  & 3.12\scriptsize{$\pm$1.11} & 8.48\scriptsize{$\pm$3.65} & 5.17\scriptsize{$\pm$2.59} & \underline{5.68\scriptsize{$\pm$5.28}} \\
& \textbf{\pname{}}  & \bluehl{94.00\scriptsize{$\pm$2.00}} & \bluehl{96.00\scriptsize{$\pm$4.00}} & \bluehl{94.67\scriptsize{$\pm$3.06}} & \bluehl{99.33\scriptsize{$\pm$1.15}} \\
\midrule
\multirow{6}{*}{BotSim-24}
% & Random    & \underline{1.67\scriptsize{$\pm$2.31}} & \underline{1.33\scriptsize{$\pm$1.15}} & \underline{2.67\scriptsize{$\pm$2.31}} & 0.00\scriptsize{$\pm$0.00} \\
& Nettack  & 0.00\scriptsize{$\pm$0.00} & 0.00\scriptsize{$\pm$0.00} & 0.00\scriptsize{$\pm$0.00} & \underline{5.23\scriptsize{$\pm$0.01}} \\
& FGA      & 0.00\scriptsize{$\pm$0.00} & 0.00\scriptsize{$\pm$0.00} & 0.00\scriptsize{$\pm$0.00} & 0.00\scriptsize{$\pm$0.00} \\
& PR-BCD   & 1.00\scriptsize{$\pm$1.41} & \underline{4.00\scriptsize{$\pm$0.00}} & \underline{2.00\scriptsize{$\pm$0.00}} & 0.00\scriptsize{$\pm$0.00} \\
& GOttack  & 0.00\scriptsize{$\pm$0.00} & 0.00\scriptsize{$\pm$0.00} & 0.00\scriptsize{$\pm$0.00} & 0.00\scriptsize{$\pm$0.00} \\
& \textbf{\pname{}}  & \bluehl{99.33\scriptsize{$\pm$1.15}} & \bluehl{74.50\scriptsize{$\pm$2.50}} & \bluehl{100.00\scriptsize{$\pm$0.00}} & \bluehl{74.12\scriptsize{$\pm$1.50}} \\
\midrule
\multicolumn{6}{c}{$\Delta$ $=$ 5} \\
\midrule
\multirow{6}{*}{Cresci-15}
% & Random    & 12.00\scriptsize{$\pm$5.29} & 12.00\scriptsize{$\pm$2.00} & 12.67\scriptsize{$\pm$4.16} & 18.67\scriptsize{$\pm$4.16} \\
& Nettack  & 22.35\scriptsize{$\pm$3.18} & 20.80\scriptsize{$\pm$2.85} & 18.95\scriptsize{$\pm$2.30} & 16.40\scriptsize{$\pm$1.62} \\
& FGA      & 21.90\scriptsize{$\pm$2.74} & 20.15\scriptsize{$\pm$2.40} & 18.40\scriptsize{$\pm$2.05} & 15.95\scriptsize{$\pm$1.48} \\
& PR-BCD   & 20.75\scriptsize{$\pm$1.96} & 19.30\scriptsize{$\pm$1.74} & 17.85\scriptsize{$\pm$1.60} & 15.40\scriptsize{$\pm$1.36} \\
& GOttack  & \underline{22.50\scriptsize{$\pm$3.55}} & \underline{21.05\scriptsize{$\pm$3.10}} & \underline{19.10\scriptsize{$\pm$2.65}} & \underline{16.85\scriptsize{$\pm$1.70}} \\
& \textbf{\pname{}}  & \bluehl{100.00\scriptsize{$\pm$0.00}} & \bluehl{100.00\scriptsize{$\pm$0.00}} & \bluehl{100.00\scriptsize{$\pm$0.00}} & \bluehl{100.00\scriptsize{$\pm$0.00}} \\
\midrule
\multirow{6}{*}{TwiBot-22}
% & Random    & 11.33\scriptsize{$\pm$8.08} & 11.33\scriptsize{$\pm$9.02} & 12.67\scriptsize{$\pm$6.43} & 7.33\scriptsize{$\pm$5.03} \\
& Nettack  & 14.22\scriptsize{$\pm$2.03} & 14.10\scriptsize{$\pm$0.00} & 5.77\scriptsize{$\pm$4.24} & 7.78\scriptsize{$\pm$4.24} \\
& FGA      & 11.30\scriptsize{$\pm$7.07} & 10.80\scriptsize{$\pm$5.66} & 6.21\scriptsize{$\pm$2.83} & \underline{9.22\scriptsize{$\pm$12.73}} \\
& PR-BCD   & \underline{24.73\scriptsize{$\pm$5.66}} & \underline{21.53\scriptsize{$\pm$7.07}} & \underline{20.45\scriptsize{$\pm$5.66}} & 19.00\scriptsize{$\pm$2.83} \\
& GOttack  & 10.58\scriptsize{$\pm$2.15} & 18.25\scriptsize{$\pm$3.65} & 12.89\scriptsize{$\pm$1.02} & 13.32\scriptsize{$\pm$3.03} \\
& \textbf{\pname{}}  & \bluehl{94.00\scriptsize{$\pm$2.00}} & \bluehl{96.00\scriptsize{$\pm$4.00}} & \bluehl{94.67\scriptsize{$\pm$3.06}} & \bluehl{99.33\scriptsize{$\pm$1.15}} \\
\midrule
\multirow{6}{*}{BotSim-24}
% & Random    & \underline{5.00\scriptsize{$\pm$3.46}} & 2.67\scriptsize{$\pm$2.31} & \underline{3.33\scriptsize{$\pm$1.15}} & 0.00\scriptsize{$\pm$0.00} \\
& Nettack  & 0.00\scriptsize{$\pm$0.00} & 0.00\scriptsize{$\pm$0.00} & 0.00\scriptsize{$\pm$0.00} & 0.00\scriptsize{$\pm$0.00} \\
& FGA      & 0.00\scriptsize{$\pm$0.00} & 0.67\scriptsize{$\pm$1.15} & 0.00\scriptsize{$\pm$0.00} & \underline{2.67\scriptsize{$\pm$1.15}} \\
& PR-BCD   & 2.67\scriptsize{$\pm$1.15} & \underline{4.00\scriptsize{$\pm$2.83}} & 1.00\scriptsize{$\pm$1.41} & 0.00\scriptsize{$\pm$0.00} \\
& GOttack  & 2.00\scriptsize{$\pm$0.00} & 2.00\scriptsize{$\pm$2.00} & 0.00\scriptsize{$\pm$0.00} & 0.00\scriptsize{$\pm$0.00} \\
& \textbf{\pname{}}  & \bluehl{99.33\scriptsize{$\pm$1.15}} & \bluehl{95.21\scriptsize{$\pm$5.02}} & \bluehl{100.00\scriptsize{$\pm$0.00}} & \bluehl{74.00\scriptsize{$\pm$1.50}} \\
\bottomrule
\end{tabular}%
}

\caption{\textbf{Node Editing Attack Reusing Same Bot Cloak:} Misclassification rate (in \%) for flipping fifty existing correctly classified bots by \pname{} and \sota adversarial attacks against best \sota bot detector, BotRGCN with adversarial defenses and \textit{reusing} the same bot cloak. The best performance is shown in bold, and the second best is underlined.}
\label{tab:old-node-50-samtemplate}
\end{minipage}

\end{table*}

%% file: tables/old-node-50-nodomainrules.tex
\begin{table*}[!tb]
\centering
\resizebox{0.6\linewidth}{!}{%
\begin{tabular}{llcccc}
\toprule
\multirow{2}{*}{\textbf{Dataset}} 
& \multirow{2}{*}{\makecell[c]{\textbf{Attack}}} 
& \multicolumn{4}{c}{\textbf{BotRGCN}} \\
\cmidrule(lr){3-6}
& 
& \textbf{Vanilla} ($\uparrow$)
& \textbf{+GNNGuard} ($\uparrow$) 
& \textbf{+GRAND} ($\uparrow$) 
& \textbf{+RobustGCN} ($\uparrow$) \\
\midrule
\multicolumn{6}{c}{$\Delta$ $=$ 1} \\
\midrule
\multirow{6}{*}{Cresci-15}
& Nettack  & 94.67\scriptsize{$\pm$6.55} & 90.10\scriptsize{$\pm$9.22} & 90.30\scriptsize{$\pm$3.75} & 92.58\scriptsize{$\pm$1.05} \\
& FGA      & 96.89\scriptsize{$\pm$4.81} & 92.21\scriptsize{$\pm$3.48} & 93.75\scriptsize{$\pm$5.64} & 91.44\scriptsize{$\pm$10.00} \\
& PR-BCD   & 96.36\scriptsize{$\pm$8.26} & 93.15\scriptsize{$\pm$8.06} & 92.91\scriptsize{$\pm$1.59} & 93.16\scriptsize{$\pm$0.69} \\
& GOttack  & \underline{98.54\scriptsize{$\pm$12.40}} & \underline{96.57\scriptsize{$\pm$8.98}} & \underline{94.04\scriptsize{$\pm$1.68}} & \underline{93.79\scriptsize{$\pm$1.67}} \\
& \textbf{\pname{}} (ours) & \bluehl{99.34\scriptsize{$\pm$0.29}} & \bluehl{98.91\scriptsize{$\pm$0.54}} & \bluehl{99.06\scriptsize{$\pm$0.41}} & \bluehl{99.72\scriptsize{$\pm$0.18}} \\
\midrule
\multirow{6}{*}{TwiBot-22}
& Nettack  & 79.77\scriptsize{$\pm$6.77} & 41.72\scriptsize{$\pm$3.23} & 68.68\scriptsize{$\pm$18.75} & \underline{84.54\scriptsize{$\pm$22.74}} \\
& FGA      & 71.90\scriptsize{$\pm$4.23} & 47.03\scriptsize{$\pm$4.80} & 80.60\scriptsize{$\pm$4.92} & 74.40\scriptsize{$\pm$10.08} \\
& PR-BCD   & 75.77\scriptsize{$\pm$6.05} & 49.69\scriptsize{$\pm$19.23} & 73.99\scriptsize{$\pm$14.54} & 61.81\scriptsize{$\pm$7.41} \\
& GOttack  & \underline{84.29\scriptsize{$\pm$11.63}} & \underline{54.90\scriptsize{$\pm$10.57}} & \underline{80.73\scriptsize{$\pm$8.34}} & 78.42\scriptsize{$\pm$13.55} \\
& \textbf{\pname{}} (ours) & \bluehl{86.67\scriptsize{$\pm$2.31}} & \bluehl{84.67\scriptsize{$\pm$11.02}} & \bluehl{83.67\scriptsize{$\pm$3.06}} & \bluehl{85.30\scriptsize{$\pm$2.00}} \\
\midrule
\multirow{6}{*}{BotSim-24}
& Nettack  & 56.57\scriptsize{$\pm$6.63} & 51.22\scriptsize{$\pm$7.08} & 55.36\scriptsize{$\pm$4.80} & 52.57\scriptsize{$\pm$1.70} \\
& FGA      & 55.67\scriptsize{$\pm$6.96} & 52.35\scriptsize{$\pm$13.91} & \underline{55.95\scriptsize{$\pm$3.67}} & \bluehl{58.04\scriptsize{$\pm$25.09}} \\
& PR-BCD   & 50.62\scriptsize{$\pm$8.04} & \bluehl{53.19\scriptsize{$\pm$8.01}} & 53.53\scriptsize{$\pm$4.46} & 52.71\scriptsize{$\pm$3.69} \\
& GOttack  & \underline{57.18\scriptsize{$\pm$2.32}} & 50.48\scriptsize{$\pm$12.18} & \bluehl{58.43\scriptsize{$\pm$5.58}} & \underline{56.50\scriptsize{$\pm$10.64}} \\
& \textbf{\pname{}} (ours) & \bluehl{58.22\scriptsize{$\pm$7.20}} & \underline{52.68\scriptsize{$\pm$1.15}} & 54.98\scriptsize{$\pm$2.45} & 55.10\scriptsize{$\pm$3.98} \\
\midrule
\multicolumn{6}{c}{$\Delta$ $=$ 5} \\
\midrule
\multirow{6}{*}{Cresci-15}
& Nettack  & \underline{99.45\scriptsize{$\pm$9.40}} & 98.02\scriptsize{$\pm$12.41} & \underline{95.78\scriptsize{$\pm$7.20}} & \underline{95.45\scriptsize{$\pm$17.53}} \\
& FGA      & 97.56\scriptsize{$\pm$12.43} & 80.11\scriptsize{$\pm$9.65} & 94.65\scriptsize{$\pm$3.63} & 94.88\scriptsize{$\pm$12.32} \\
& PR-BCD   & 98.65\scriptsize{$\pm$5.25} & 96.12\scriptsize{$\pm$15.73} & 94.55\scriptsize{$\pm$14.93} & 92.54\scriptsize{$\pm$6.27} \\
& GOttack  & 99.33\scriptsize{$\pm$17.24} & \bluehl{99.58\scriptsize{$\pm$3.58}} & 90.98\scriptsize{$\pm$5.49} & 90.36\scriptsize{$\pm$7.36} \\
& \textbf{\pname{}} (ours) & \bluehl{99.67\scriptsize{$\pm$0.20}} & \underline{98.88\scriptsize{$\pm$0.58}} & \bluehl{99.11\scriptsize{$\pm$0.39}} & \bluehl{99.52\scriptsize{$\pm$0.24}} \\
\midrule
\multirow{6}{*}{TwiBot-22}
& Nettack  & 94.15\scriptsize{$\pm$1.05} & 58.54\scriptsize{$\pm$8.34} & 86.73\scriptsize{$\pm$21.07} & \bluehl{96.78\scriptsize{$\pm$19.05}} \\
& FGA      & \underline{96.13\scriptsize{$\pm$7.15}} & 60.52\scriptsize{$\pm$8.20} & 92.42\scriptsize{$\pm$1.48} & 88.58\scriptsize{$\pm$5.33} \\
& PR-BCD   & 94.08\scriptsize{$\pm$2.68} & 66.14\scriptsize{$\pm$21.72} & 89.85\scriptsize{$\pm$12.04} & 87.45\scriptsize{$\pm$4.14} \\
& GOttack  & \bluehl{99.55\scriptsize{$\pm$6.16}} & \underline{75.96\scriptsize{$\pm$7.77}} & \underline{93.53\scriptsize{$\pm$4.10}} & \underline{92.52\scriptsize{$\pm$18.37}} \\
& \textbf{\pname{}} (ours) & 94.00\scriptsize{$\pm$2.00} & \bluehl{90.00\scriptsize{$\pm$4.00}} & \bluehl{93.67\scriptsize{$\pm$3.06}} & 88.33\scriptsize{$\pm$1.15} \\
\midrule
\multirow{6}{*}{BotSim-24}
& Nettack  & 97.78\scriptsize{$\pm$8.96} & 74.79\scriptsize{$\pm$2.64} & 78.58\scriptsize{$\pm$7.48} & \underline{89.08\scriptsize{$\pm$4.04}} \\
& FGA      & \underline{98.23\scriptsize{$\pm$2.98}} & 82.77\scriptsize{$\pm$8.82} & 88.12\scriptsize{$\pm$7.35} & 84.02\scriptsize{$\pm$20.30} \\
& PR-BCD   & 92.45\scriptsize{$\pm$3.58} & \underline{85.63\scriptsize{$\pm$10.46}} & 82.61\scriptsize{$\pm$1.42} & 85.12\scriptsize{$\pm$6.51} \\
& GOttack  & 96.86\scriptsize{$\pm$7.70} & 82.87\scriptsize{$\pm$15.75} & \underline{89.54\scriptsize{$\pm$3.56}} & \bluehl{90.72\scriptsize{$\pm$4.93}} \\
& \textbf{\pname{}} (ours) & \bluehl{99.33\scriptsize{$\pm$1.15}} & \bluehl{89.21\scriptsize{$\pm$1.02}} & \bluehl{92.28\scriptsize{$\pm$0.33}} & 44.00\scriptsize{$\pm$19.16} \\
\bottomrule
\end{tabular}%
}
\caption{\textbf{Node Editing Attack Without Domain Constraints:} Misclassification rate (in \%) for flipping fifty existing correctly classified bots by \pname{} and \sota adversarial attacks with budget $\Delta$ of five against best \sota bot detector, BotRGCN with adversarial defenses and \textit{without} enforcing domain constraints on attacks. For results when domain constraints are enforced, refer to \autoref{tab:old-node-50} and for results when the same bot cloak is reused, refer to \autoref{tab:old-node-50-samtemplate}. The best performance is shown in bold, and the second best is underlined.}
\label{tab:old-node-50-nodomainrules}
\end{table*}

%% file: tables/constrained-baseline-controls.tex
\begin{table}[!t]
\centering
\scriptsize
\setlength{\tabcolsep}{3pt}
\renewcommand{\arraystretch}{1.05}
\begin{tabular}{lcc}
\toprule
\textbf{Attack} & \textbf{Unconstrained} & \textbf{Constrained} \\
\midrule
FGA & 71.90$\pm$4.23 & 4.00$\pm$2.00 \\
FGA-mod & 70.44$\pm$1.34 & 7.11$\pm$2.23 \\
PR-BCD & 75.77$\pm$6.05 & 8.67$\pm$2.31 \\
PR-BCD-mod & 74.93$\pm$5.90 & 13.21$\pm$1.45 \\
LR-BCD & 82.33$\pm$2.13 & 9.86$\pm$4.65 \\
Homophily & 60.11$\pm$3.69 & 13.12$\pm$1.01 \\
\textbf{\pname{} (ours)} & \bluehl{86.67$\pm$2.31} & \bluehl{86.67$\pm$2.31} \\
\bottomrule
\end{tabular}
\caption{\textbf{Control study on TwiBot-22/BotRGCN for node editing with $\Delta=1$}: FGA-mod and PR-BCD-mod add in-loop projection onto the admissible bot$\rightarrow$human edit set; Homophily replaces OT-guided coupling with embedding/homophily matching while keeping the same feature set and local context.}
\label{tab:constraint-controls}
\end{table}

%% file: tables/ablation.tex
\begin{table*}[!htb]
\centering
\resizebox{0.5\linewidth}{!}{%
\begin{tabular}{lcccc}
\toprule

\multirow{2}{*}{\makecell[c]{\textbf{Loss Component}}} 
& \multicolumn{4}{c}{\textbf{BotRGCN}} \\
\cmidrule(lr){2-5}
& \textbf{Vanilla} ($\uparrow$)
& \textbf{+GNNGuard} ($\uparrow$) 
& \textbf{+GRAND} ($\uparrow$) 
& \textbf{+RobustGCN} ($\uparrow$) \\
\midrule

\multicolumn{5}{c}{Cresci-15} \\
\midrule
$-\lambda_{\mathrm{BCE}}$ & 38.21\scriptsize{$\pm$5.12} & 32.98\scriptsize{$\pm$0.22} & 33.34\scriptsize{$\pm$2.21} & 35.76\scriptsize{$\pm$1.80} \\
$-\lambda_{\mathrm{sp}}$  & 96.88\scriptsize{$\pm$5.89} & 95.12\scriptsize{$\pm$5.41} & 95.34\scriptsize{$\pm$7.13} & 94.23\scriptsize{$\pm$0.71} \\
$-\lambda_{\mathrm{pl}}$ & 57.99\scriptsize{$\pm$3.38} &52.67\scriptsize{$\pm$0.26} &50.71\scriptsize{$\pm$1.76}& 55.45\scriptsize{$\pm$4.05} \\
$-\alpha_{\deg}$          & 73.23\scriptsize{$\pm$0.23} & 70.23\scriptsize{$\pm$1.60} & 71.11\scriptsize{$\pm$5.20} & 68.54\scriptsize{$\pm$4.36}\\
$-\alpha_{\mathrm{age}}$  & \underline{87.23\scriptsize{$\pm$1.78}} & \underline{85.77\scriptsize{$\pm$4.72}} & \underline{84.82\scriptsize{$\pm$6.47}} & \underline{86.76\scriptsize{$\pm$0.07}}\\
\textbf{\pname{}}   & \bluehl{99.67\scriptsize{$\pm$0.20}} & \bluehl{98.88\scriptsize{$\pm$0.58}} & \bluehl{99.11\scriptsize{$\pm$0.39}} & \bluehl{99.52\scriptsize{$\pm$0.24}} \\
\midrule

\multicolumn{5}{c}{TwiBot-22} \\
\midrule
$-\lambda_{\mathrm{BCE}}$ & 22.45\scriptsize{$\pm$6.44} & 20.12\scriptsize{$\pm$5.58} & 17.23\scriptsize{$\pm$2.73} & 19.87\scriptsize{$\pm$1.26} \\
$-\lambda_{\mathrm{sp}}$  & 92.12\scriptsize{$\pm$7.65} & 87.34\scriptsize{$\pm$2.70} & 91.22\scriptsize{$\pm$0.76} & 84.12\scriptsize{$\pm$0.79} \\
$-\lambda_{\mathrm{pl}}$ & 55.23\scriptsize{$\pm$6.77} &50.12\scriptsize{$\pm$4.83} &52.55\scriptsize{$\pm$6.45} &52.98\scriptsize{$\pm$5.84} \\
$-\alpha_{\deg}$          & 70.98\scriptsize{$\pm$4.29} & 68.12\scriptsize{$\pm$7.78} & 65.22\scriptsize{$\pm$3.04} & 68.79\scriptsize{$\pm$4.42}\\
$-\alpha_{\mathrm{age}}$  & \underline{83.88\scriptsize{$\pm$6.63}} & \underline{81.54\scriptsize{$\pm$4.95}} & \underline{80.02\scriptsize{$\pm$6.89}} & \underline{78.35\scriptsize{$\pm$4.62}}\\
\textbf{\pname{}}   & \bluehl{94.00\scriptsize{$\pm$2.00}} & \bluehl{90.00\scriptsize{$\pm$4.00}} & \bluehl{93.67\scriptsize{$\pm$3.06}} & \bluehl{88.33\scriptsize{$\pm$1.15}} \\
\midrule

\multicolumn{5}{c}{BotSim-24} \\
\midrule
$-\lambda_{\mathrm{BCE}}$ & 32.45\scriptsize{$\pm$5.64} & 25.34\scriptsize{$\pm$0.39} & 29.43\scriptsize{$\pm$1.84} & 10.11\scriptsize{$\pm$2.33} \\
$-\lambda_{\mathrm{sp}}$  & 96.55\scriptsize{$\pm$0.66} & 85.73\scriptsize{$\pm$1.88} & 90.21\scriptsize{$\pm$0.82} & 38.45\scriptsize{$\pm$2.24} \\
$-\lambda_{\mathrm{pl}}$  & 60.34\scriptsize{$\pm$5.09} & 58.17\scriptsize{$\pm$2.93} & 53.13\scriptsize{$\pm$2.97} & 25.33\scriptsize{$\pm$1.69} \\
$-\alpha_{\deg}$          & 69.22\scriptsize{$\pm$2.15} & 65.23\scriptsize{$\pm$7.49} & 64.67\scriptsize{$\pm$5.18} & 32.32\scriptsize{$\pm$4.87}\\
$-\alpha_{\mathrm{age}}$  & \underline{85.34\scriptsize{$\pm$1.38}} & \underline{83.78\scriptsize{$\pm$5.83}} & \underline{82.44\scriptsize{$\pm$1.32}} & \underline{37.23\scriptsize{$\pm$3.04}}\\
\textbf{\pname{}} & \bluehl{99.33\scriptsize{$\pm$1.15}} & \bluehl{79.21\scriptsize{$\pm$1.02}} & \bluehl{92.28\scriptsize{$\pm$0.33}} & \bluehl{66.20\scriptsize{$\pm$3.50}} \\
\bottomrule
\end{tabular}%
}
\caption{\textbf{Ablation Study:} Ablation results for misclassification rate (in \%) for node editing. \pname{} results against BotRGCN with standard adversarial defenses with budget $\Delta$ of 5. The best performance is shown in bold, and the second best is underlined.}
\label{tab:ablation}
\end{table*}
 

%% file: tables/sensitivity.tex
\begin{table*}[!htb]
\centering
\setlength{\tabcolsep}{6pt}
\renewcommand{\arraystretch}{1.05}
\resizebox{0.50\linewidth}{!}{%
\begin{tabular}{lccc}
\toprule
\textbf{Hyperparameter} & \textbf{Cresci-15} & \textbf{TwiBot-22} & \textbf{BotSim-24} \\
\midrule

\multicolumn{4}{c}{\textbf{OT Regularizer($\varepsilon$)}} \\
\midrule
$\varepsilon=0.01$ & 98.11\scriptsize{$\pm$4.23} & 84.00\scriptsize{$\pm$6.00} & 84.00\scriptsize{$\pm$1.56} \\
$\varepsilon=0.1$  & \underline{99.30\scriptsize{$\pm$2.36}} & 90.67\scriptsize{$\pm$3.06} & \underline{98.00\scriptsize{$\pm$3.88}} \\
$\varepsilon=1$    & 53.33\scriptsize{$\pm$17.67} & \underline{92.67\scriptsize{$\pm$2.31}} & 18.00\scriptsize{$\pm$3.03} \\
$\varepsilon=10$   & 33.33\scriptsize{$\pm$5.66} & 32.00\scriptsize{$\pm$2.00} & 22.65\scriptsize{$\pm$2.25} \\
$\varepsilon=100$  & 13.83\scriptsize{$\pm$5.57} & 22.00\scriptsize{$\pm$1.03} & 18.23\scriptsize{$\pm$1.20} \\
\textbf{\pname{}} ($\varepsilon=0.2$) & \bluehl{99.67\scriptsize{$\pm$0.20}} & \bluehl{94.00\scriptsize{$\pm$2.00}} & \bluehl{99.33\scriptsize{$\pm$1.15}} \\

\midrule
\multicolumn{4}{c}{\textbf{BCE Threshold ($\tau_{\text{bdry}}$)}} \\
\midrule
$\tau_{\text{bdry}}=0.00$ & 92.50\scriptsize{$\pm$2.50} & 85.33\scriptsize{$\pm$1.55} & 83.05\scriptsize{$\pm$1.33} \\
$\tau_{\text{bdry}}=0.05$ & 97.00\scriptsize{$\pm$2.00} & 91.50\scriptsize{$\pm$1.00} & 97.20\scriptsize{$\pm$1.15} \\
$\tau_{\text{bdry}}=0.15$ & \underline{99.05\scriptsize{$\pm$1.50}} & 93.25\scriptsize{$\pm$1.25} & \underline{97.33\scriptsize{$\pm$1.50}} \\
$\tau_{\text{bdry}}=0.30$ & 98.89\scriptsize{$\pm$1.07} & \underline{93.67\scriptsize{$\pm$1.76}} & 95.15\scriptsize{$\pm$1.33} \\
$\tau_{\text{bdry}}=0.50$ & 98.15\scriptsize{$\pm$1.67} & 90.11\scriptsize{$\pm$2.50} & 90.67\scriptsize{$\pm$1.33} \\
\textbf{\pname{}} ($\tau_{\text{bdry}}=0.10$) & \bluehl{99.67\scriptsize{$\pm$0.20}} & \bluehl{94.00\scriptsize{$\pm$2.00}} & \bluehl{99.33\scriptsize{$\pm$1.15}} \\

\bottomrule
\end{tabular}%
}
\caption{\textbf{Sensitivity Study:} Misclassification rate (in \%) for flipping fifty existing correctly classified bots as the OT regularizer ($\varepsilon$) and  BCE Threshold ($\tau_{\text{bdry}}$) is varied, impacting the human-bot decision boundary and and the strength of the OT margin constraint required for a successful cloak. \pname{} results against vanilla BotRGCN with budget $\Delta$ of 5.  Best is highlighted and second best is underlined within each block.}
\label{tab:sensitivity}
\end{table*}

%% file: tables/time-gpu.tex
\begin{table*}[!htb]
\centering
\resizebox{0.5\linewidth}{!}{%
\begin{tabular}{llccc}
\toprule
\textbf{System Resource} ($\downarrow$)
& \textbf{Attack}
& \textbf{Cresci-15} 
& \textbf{TwiBot-22} 
& \textbf{BotSim-24} \\
\midrule

\multirow{6}{*}{\makecell[c]{Time\\ (sec)}}
& Random   & 0.01\scriptsize{$\pm$0.01} & 0.18\scriptsize{$\pm$0.02} & 0.03\scriptsize{$\pm$0.01} \\
& Nettack  & 20.3\scriptsize{$\pm$0.22}          & 1.4\scriptsize{$\pm$1.33}            & 19.7\scriptsize{$\pm$1.65} \\
& FGA      & \underline{0.3\scriptsize{$\pm$0.14}} & \underline{1.1\scriptsize{$\pm$0.06}} & \underline{0.3\scriptsize{$\pm$1.57}} \\
& PR-BCD   & 1.4\scriptsize{$\pm$0.13}           & 2.9\scriptsize{$\pm$0.47}            & 2.1\scriptsize{$\pm$0.75} \\
& GOttack  & 1.9\scriptsize{$\pm$3.56}           & 853.9\scriptsize{$\pm$19.95}        & 7.5\scriptsize{$\pm$1.75} \\
% \cmidrule(lr){2-5}
& \textbf{\pname{}} (ours) & \bluehl{0.1\scriptsize{$\pm$0.13}} & \bluehl{0.5\scriptsize{$\pm$0.14}} & \bluehl{0.1\scriptsize{$\pm$0.13}} \\
\midrule

\multirow{6}{*}{\makecell[c]{RAM\\ (MB)}}
& Random   & 10\scriptsize{$\pm$2}      & 8237\scriptsize{$\pm$11} & 1991\scriptsize{$\pm$10} \\
& Nettack  & 2496\scriptsize{$\pm$43}            & 21102\scriptsize{$\pm$12133}         & 2973\scriptsize{$\pm$628} \\
& FGA      & \underline{2307\scriptsize{$\pm$0}} & \underline{9036\scriptsize{$\pm$72}} & \underline{1720\scriptsize{$\pm$13}} \\
& PR-BCD   & 2307\scriptsize{$\pm$0} & 9037\scriptsize{$\pm$72}             & 1857\scriptsize{$\pm$21} \\
& GOttack  & 2430\scriptsize{$\pm$13}            & 10121\scriptsize{$\pm$214}           & 2069\scriptsize{$\pm$623} \\
% \cmidrule(lr){2-5}
& \textbf{\pname{}} (ours) & \bluehl{1455\scriptsize{$\pm$8}} & \bluehl{7974\scriptsize{$\pm$25}} & \bluehl{1450\scriptsize{$\pm$2}} \\
\midrule

\multirow{6}{*}{\makecell[c]{GPU\\ (MB)}}
& Random   & 116\scriptsize{$\pm$11}  & 20812\scriptsize{$\pm$2}            & 146\scriptsize{$\pm$6} \\
& Nettack  & \underline{288\scriptsize{$\pm$58}}  & \underline{5982\scriptsize{$\pm$4}} & \underline{156\scriptsize{$\pm$40}} \\
& FGA      & 812\scriptsize{$\pm$26}              & 9222\scriptsize{$\pm$28}            & 286\scriptsize{$\pm$2} \\
& PR-BCD   & 812\scriptsize{$\pm$26}              & 6178\scriptsize{$\pm$4}             & 286\scriptsize{$\pm$1} \\
& GOttack  & 324\scriptsize{$\pm$2}               & \underline{5982\scriptsize{$\pm$15}} & 240\scriptsize{$\pm$2} \\
% \cmidrule(lr){2-5}
& \textbf{\pname{}} (ours) & \bluehl{9\scriptsize{$\pm$1}} & \bluehl{13\scriptsize{$\pm$2}} & \bluehl{11\scriptsize{$\pm$1}} \\
\bottomrule
\end{tabular}%
}
\caption{\textbf{System Overhead Study:} System resource usage by \pname{} and \sota adversarial attacks against vanilla BotRGCN with budget $\Delta$ of 1. The best performance is shown in bold, and the second best is underlined.}
\label{tab:time-gpu}
\end{table*}

%% file: tables/struc-category.tex
\begin{table}[t]
\centering
\resizebox{0.5\linewidth}{!}{
\begin{tabular}{ll}
\toprule
\textbf{Outgoing follows (Target Bot $\rightarrow$)} & \textbf{Incoming follows ($\rightarrow$ Target Bot)} \\
\midrule
Target Bot $\rightarrow$ Human & Human $\rightarrow$ Target Bot \\
Target Bot $\rightarrow$ Human & Bots $\rightarrow$ Target Bot \\
Target Bot $\rightarrow$ Human & Both $\rightarrow$ Target Bot \\
Target Bot $\rightarrow$ Human & Nobody $\rightarrow$ Target Bot \\
\addlinespace
Target Bot $\rightarrow$ Bots & Human $\rightarrow$ Target Bot \\
Target Bot $\rightarrow$ Bots & Bots $\rightarrow$ Target Bot \\
Target Bot $\rightarrow$ Bots & Both $\rightarrow$ Target Bot \\
Target Bot $\rightarrow$ Bots & Nobody $\rightarrow$ Target Bot \\
\addlinespace
Target Bot $\rightarrow$ Both & Human $\rightarrow$ Target Bot \\
Target Bot $\rightarrow$ Both & Bots $\rightarrow$ Target Bot \\
Target Bot $\rightarrow$ Both & Both $\rightarrow$ Target Bot \\
Target Bot $\rightarrow$ Both & Nobody $\rightarrow$ Target Bot \\
\addlinespace
Target Bot $\rightarrow$ Nobody & Human $\rightarrow$ Target Bot \\
Target Bot $\rightarrow$ Nobody & Bots $\rightarrow$ Target Bot \\
Target Bot $\rightarrow$ Nobody & Both $\rightarrow$ Target Bot \\
Target Bot $\rightarrow$ Nobody & Nobody $\rightarrow$ Target Bot \\
\bottomrule
\end{tabular}
}
\caption{\textbf{Structural Categories.} Sixteen structural categories used to characterize target bot's neighborhood using 1-hop ego-graph structure around the target bot. Rows enumerate the \emph{outgoing} follow type (target Bot account following Humans/Bots/Both/Nobody) and the \emph{incoming} follow type (followers of the target Bot are Humans/Bots/Both/Nobody).}
\label{tab:struc-category}
\end{table}

%% file: figs/test-fig.tex
% ===================== ICML-friendly preamble bits =====================
% In your preamble (ICML template already loads many of these, but it's safe):
% \usepackage{graphicx}    % for \resizebox
% \usepackage{tikz}
% \usetikzlibrary{arrows.meta}
% \usepackage{subcaption}  % ICML generally allows this; if not, use \captionof from capt-of.

% ===================== Figure =====================
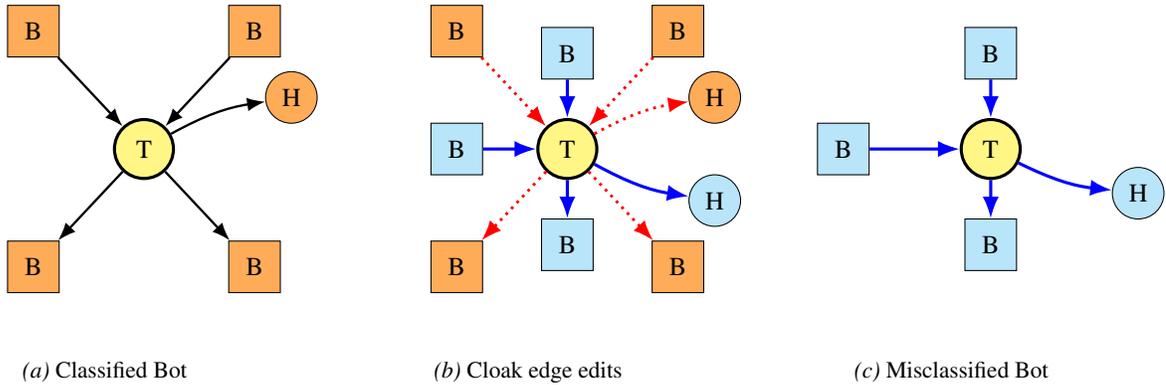
\begin{figure}[t]
\centering

\tikzset{
  >=Latex,
  target/.style={circle, draw=black, very thick, fill=yellow!60, minimum size=8mm, inner sep=0pt},
  oldnode/.style={draw=black, fill=orange!65, minimum size=7mm, inner sep=0pt},
  oldbot/.style={oldnode, rectangle},
  oldhuman/.style={oldnode, circle},
  newnode/.style={draw=black, fill=cyan!25, minimum size=7mm, inner sep=0pt},
  newbot/.style={newnode, rectangle},
  newhuman/.style={newnode, circle},
  kept/.style={-Latex, line width=0.9pt, black},
  removed/.style={-Latex, line width=1.1pt, red, dotted},
  added/.style={-Latex, line width=1.2pt, blue},
}

% ICML: use \columnwidth in 2-col mode (same as \linewidth inside the column).
% The 0.98 factor gives a tiny safety margin.
\resizebox{0.98\columnwidth}{!}{%
\begin{minipage}{\columnwidth}
\centering

% -------------------- (a) BEFORE --------------------
\begin{subfigure}[b]{0.33\columnwidth}
\centering
\begin{tikzpicture}
  % Tight, consistent bounding box to prevent "phantom" whitespace overflow
  \path[use as bounding box] (-3.35,-2.65) rectangle (3.65,2.65);

  % Nodes
  \node[target] (T) at (0,0) {T};

  % Old in-neighbors (bots)
  \node[oldbot] (Bin1) at (-1.5, 1.6) {B};
  \node[oldbot] (Bin2) at ( 1.5, 1.6) {B};

  % Old out-neighbors (2 bots, 1 human)
  \node[oldbot]   (Bout1) at (-1.5,-1.6) {B};
  \node[oldbot]   (Bout2) at ( 1.5,-1.6) {B};
  \node[oldhuman] (Hout)  at ( 2.0, 0.7) {H};

  % Edges (deg=5: in=2, out=3)
  \draw[kept] (Bin1) -- (T);
  \draw[kept] (Bin2) -- (T);

  \draw[kept] (T) -- (Bout1);
  \draw[kept] (T) -- (Bout2);
  \draw[kept, bend left=10] (T) to (Hout);
\end{tikzpicture}
\subcaption{Classified Bot}
\end{subfigure}\hfill%
% -------------------- (b) EDITS (DEL/ADD) --------------------
\begin{subfigure}[b]{0.33\columnwidth}
\centering
\begin{tikzpicture}
  \path[use as bounding box] (-3.35,-2.65) rectangle (3.65,2.65);

  % Nodes
  \node[target] (T) at (0,0) {T};

  % Old nodes (same positions)
  \node[oldbot]   (Bin1)  at (-1.5, 1.6) {B};
  \node[oldbot]   (Bin2)  at ( 1.5, 1.6) {B};
  \node[oldbot]   (Bout1) at (-1.5,-1.6) {B};
  \node[oldbot]   (Bout2) at ( 1.5,-1.6) {B};
  \node[oldhuman] (Hout)  at ( 2.0, 0.7) {H};

  % New nodes introduced by cloak
  \node[newbot]   (BnewIn1)  at (0.0, 1.3) {B};
  \node[newbot]   (BnewIn2)  at (-1.5, 0.0) {B};
  \node[newbot]   (BnewOut)  at (0.0,-1.3) {B};
  \node[newhuman] (HnewOut)  at (2.0,-0.7) {H};

  % Kept edges (thin black) -- only the two incoming bots remain kept here
  \draw[removed] (Bin1) -- (T);
  \draw[removed] (Bin2) -- (T);

  % Deleted edges (red dotted): old outgoing to bots + old outgoing to human
  \draw[removed] (T) -- (Bout1);
  \draw[removed] (T) -- (Bout2);
  \draw[removed, bend left=10] (T) to (Hout);

  % Added edges (blue)
  \draw[added] (BnewIn1) -- (T);
  \draw[added] (BnewIn2) -- (T);
  \draw[added] (T) -- (BnewOut);
  \draw[added, bend right=10] (T) to (HnewOut);
\end{tikzpicture}
\subcaption{Cloak edge edits}
\end{subfigure}\hfill%
% -------------------- (c) AFTER --------------------
\begin{subfigure}[b]{0.33\columnwidth}
\centering
\begin{tikzpicture}
  \path[use as bounding box] (-3.35,-2.65) rectangle (3.65,2.65);

  % Nodes
  \node[target] (T) at (0,0) {T};

  % Old nodes kept
  % \node[oldbot] (Bin1)  at (-1.5, 1.6) {B};
  % \node[oldbot] (Bin2)  at ( 1.5, 1.6) {B};

  % New nodes (now part of 1-hop)
  \node[newbot]   (BnewIn1) at (0.0, 1.3) {B};
  \node[newbot]   (BnewIn2) at (-2.0, 0.0) {B};
  \node[newbot]   (BnewOut) at (0.0,-1.3) {B};
  \node[newhuman] (HnewOut) at (2.0,-0.6) {H};

  % Edges present after cloak
  % \draw[kept] (Bin1) -- (T);
  % \draw[kept] (Bin2) -- (T);

  \draw[added] (BnewIn1) -- (T);
  \draw[added] (BnewIn2) -- (T);
  \draw[added] (T) -- (BnewOut);
  \draw[added, bend right=10] (T) to (HnewOut);
\end{tikzpicture}
\subcaption{Misclassified Bot}
\end{subfigure}

\end{minipage}%
} % end resizebox

\caption{\textbf{Cloaking target bot (node id=81779) with misclassified boundary bot cloak (node id=41575) in TwiBot-22 dataset.}
Target bot (T) is yellow; pre-existing neighbors are orange; new neighbors introduced by \pname are light blue. Red dotted arrows denote deleted incident edges; blue arrows denote newly added edges.}
\label{fig:case-study}
\end{figure}